\documentclass{article} 
\usepackage[preprint]{colm2025_conference}

\usepackage{microtype}
\usepackage{hyperref}
\usepackage{url}
\usepackage{booktabs}

\usepackage{lineno}

\definecolor{darkblue}{rgb}{0, 0, 0.5}
\hypersetup{colorlinks=true, citecolor=darkblue, linkcolor=darkblue, urlcolor=darkblue}

\usepackage{soul}
\usepackage[utf8]{inputenc}
\usepackage{graphicx}
\usepackage{algorithm}
\usepackage[noend]{algpseudocode}

\usepackage[english]{babel}
\usepackage{csquotes}

\usepackage{natbib}
\usepackage{amsmath}
\usepackage{amssymb}
\usepackage{amsthm}
\usepackage{parskip}
\usepackage{amsfonts, bm, dsfont}
\usepackage{kantlipsum, lipsum}
\usepackage{threeparttable}
\usepackage{pifont}
\usepackage{array}
\usepackage{multirow}
\usepackage{colortbl}
\usepackage{enumitem}
\usepackage[figuresright]{rotating}
\usepackage{stmaryrd}
\usepackage{subcaption}

\usepackage{tabularx}

\usepackage{amsthm}
\newtheorem{definition}{Definition}

\title{Where Paths Collide: A Comprehensive Survey of Classic and Learning-Based Multi-Agent Pathfinding}

\author{Shiyue Wang \& Haozheng Xu\thanks{The first two authors contribute equally.} \\
East China Normal University\\
\texttt{\{wangshiyue,xuhaozheng\}@stu.ecnu.edu.cn} \\
\AND
Yuhan Zhang \& Jingran Lin \\
University of Electronic Science and Technology of China\\
\texttt{\{yuhan\_zhang@std.,jingranlin@\}uestc.edu.cn} \\
\AND
Changhong Lu \& Xiangfeng Wang\\
East China Normal University\\
\texttt{\{chlu@math,xfwang@cs\}.ecnu.edu.cn} \\
\And
Wenhao Li\thanks{Corresponding authors: Xiangfeng Wang and Wenhao Li.}\\
Tongji University\\
\texttt{whli@tongji.edu.cn}\\
}

\begin{document}

\ifcolmsubmission
\linenumbers
\fi

\maketitle




\begin{abstract}
Multi-Agent Path Finding (MAPF) is a fundamental problem in artificial intelligence and robotics, requiring the computation of collision-free paths for multiple agents navigating from their start locations to designated goals. 
As autonomous systems become increasingly prevalent in warehouses, urban transportation, and other complex environments, MAPF has evolved from a theoretical challenge to a critical enabler of real-world multi-robot coordination. 
This comprehensive survey bridges the long-standing divide between classical algorithmic approaches and emerging learning-based methods in MAPF research.
We present a unified framework that encompasses search-based methods (including Conflict-Based Search, Priority-Based Search, and Large Neighborhood Search), compilation-based approaches (SAT, SMT, CSP, ASP, and MIP formulations), and data-driven techniques (reinforcement learning, supervised learning, and hybrid strategies). 
Through systematic analysis of experimental practices across 200+ papers, we uncover significant disparities in evaluation methodologies, with classical methods typically tested on larger-scale instances (up to 200×200 grids with 1000+ agents) compared to learning-based approaches (predominantly 10-100 agents). 
We provide a comprehensive taxonomy of evaluation metrics, environment types, and baseline selections, highlighting the need for standardized benchmarking protocols. 
Finally, we outline promising future directions including mixed-motive MAPF with game-theoretic considerations, language-grounded planning with large language models, and neural solver architectures that combine the rigor of classical methods with the flexibility of deep learning. 
This survey serves as both a comprehensive reference for researchers and a practical guide for deploying MAPF solutions in increasingly complex real-world applications.
The project website is~\url{https://wangsh1yue.github.io/Where-Paths-Collide/}.
\end{abstract}

\section{Introduction}\label{sec:intro}



Multi-Agent Path Finding (MAPF)~\citep{stern2019multi} is a longstanding and foundational problem in robotics and artificial intelligence, focusing on planning collision-free paths for multiple agents moving in a shared environment. 
Traditionally, MAPF is formulated on an undirected graph ${\cal{G}} = ({\cal{V}}, {\cal{E}})$, in which vertices ${\cal{V}}$ represent possible locations and edges ${\cal{E}}$ denote valid transitions between locations~\citep{surynek2022problem}. 
A widely adopted discretization strategy places agents on a grid, enabling them to move to adjacent cells at discrete time steps. 
Nevertheless, continuous formulations of MAPF that allow more flexible trajectories in space and time have also been actively studied~\citep{andreychuk2022multi,yang2023path}. 
This broad problem statement has spawned numerous theoretical breakthroughs and a diverse range of algorithms over the decades.

\begin{figure}[htb!]
    \centering
    \begin{subfigure}[b]{0.48\textwidth}
        \centering
        \includegraphics[width=\textwidth]{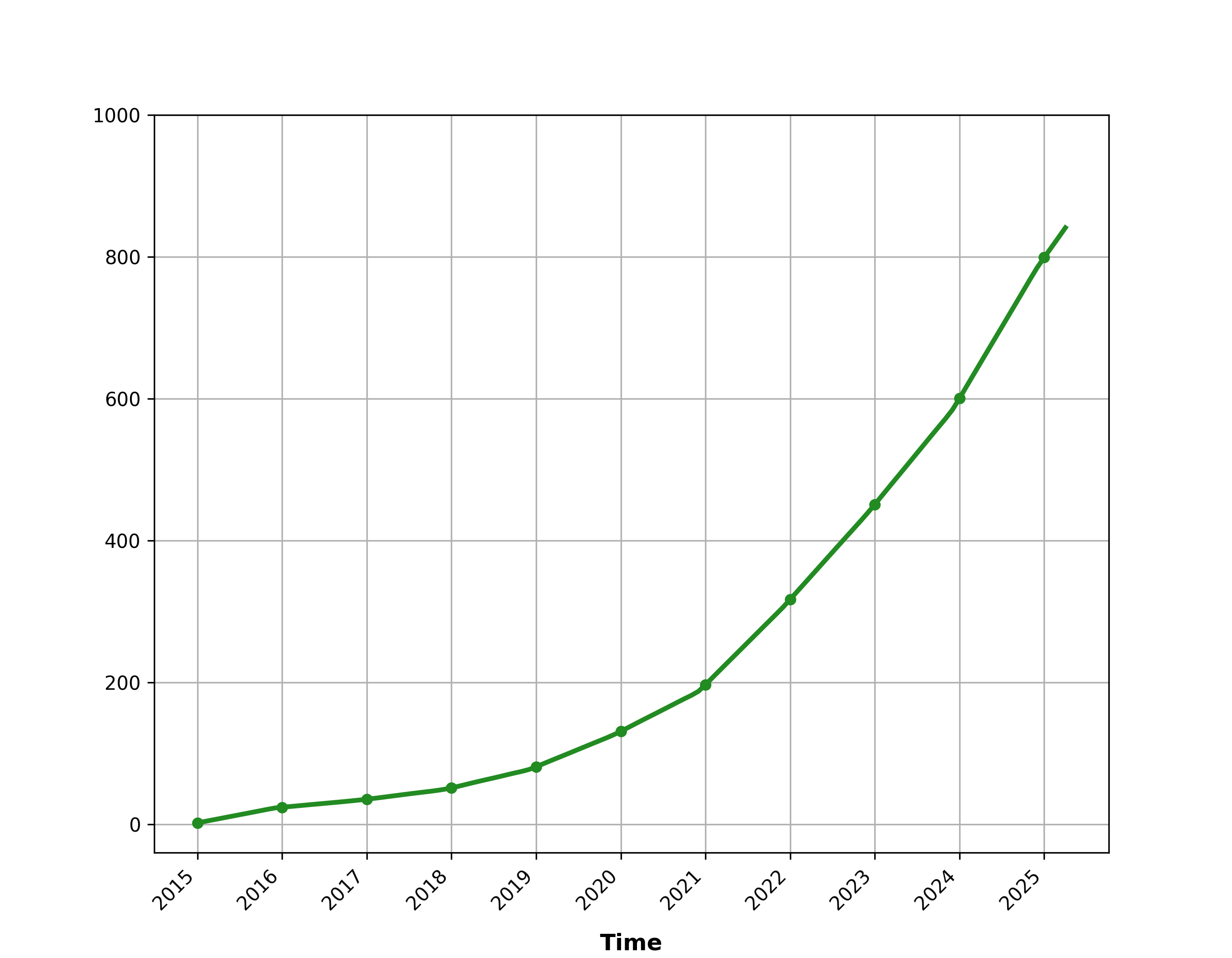}
        \caption{Query=``multi-agent pathfinding''}
        \label{fig:mapf_papers_1}
    \end{subfigure}
    \begin{subfigure}[b]{0.48\textwidth}
        \centering
        \includegraphics[width=\textwidth]{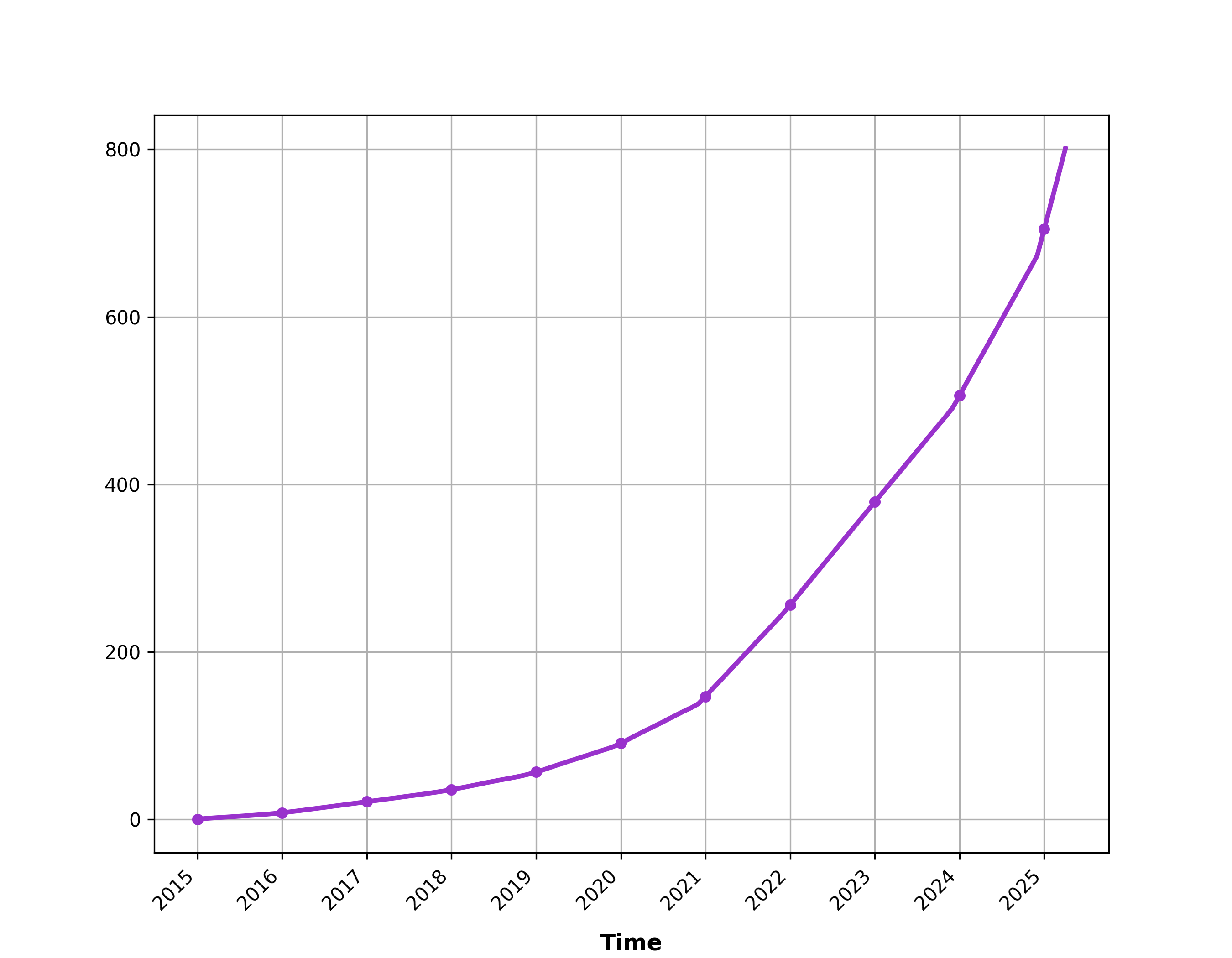}
        \caption{Query=``MAPF''}
        \label{fig:mapf_papers_2}
    \end{subfigure}
    \caption{The trends of the cumulative numbers of Google Scholar papers that contain the keyphrases ``multi-agent pathfinding'' and `` MAPF'' since January 2015, respectively. The statistics are calculated using exact match by queryingthe keyphrases in title or abstract by months. Both queries show a notable growth trend, particularly after 2020, indicating the increasing research interest and expansion.}
    \label{fig:mapf_papers}
\end{figure}

MAPF sees practical deployment in an extensive set of application domains, as shown in Figure~\ref{fig:mapf_papers}. 
In warehouse automation, fleets of automated guided vehicles coordinate to transport goods efficiently~\citep{varambally2022mapf,wang2024mapf}. 
In traffic management for autonomous driving, vehicles that share urban road networks must navigate safely and cooperatively~\citep{teng2017coordinating,ren2024multi}. 
In safety-critical scenarios such as air traffic control or railway scheduling, MAPF techniques contribute to conflict resolution and scheduling~\citep{ho2020decentralized,shrestha20216g,li2021scalable,chen2022multi}. 
Beyond these areas, MAPF is a key element in optimizing virtual character movements in video games~\citep{rahmani2020multi,ma2017feasibility}, designing efficient airport surface operations~\citep{morris2016planning,von2024towards}, coordinating automatic parking in vehicle infrastructure~\citep{okoso2019multi,okoso2022high}, enabling multi-robot exploration and coordination~\citep{tang2024large,almadhoun2019survey}, and orchestrating swarm drone fleets~\citep{pyke2021dynamic,tjiharjadi2022systematic}. 
As physical systems and robotic capabilities continue to advance, including the emergence of embodied intelligence~\citep{paoloposition} and the low-altitude economy~\citep{leet2024safe}, the practical deployment of MAPF has grown in both scale and complexity, reinforcing MAPF’s role as a crucial foundation for cooperative robotic and autonomous systems.

Despite the extensive research on MAPF, the field faces several challenges that stem from both theoretical and practical considerations. 
Classical approaches to MAPF, which mainly include search-based and compilation-based methods, have been the backbone of MAPF research for decades. 
Search-based methods often emphasize the optimality of solutions, employing traditional graph search, heuristics, and tailored techniques to minimize path collision and search overhead. 
Examples include the widely studied A*\nobreakdash-based and conflict-based search algorithms~\citep{sharon2015conflict,okumura2022priority}. 
Compilation-based methods, on the other hand, transform MAPF into other well-understood mathematical formulations (e.g., integer linear programs, satisfiability problems), thereby leveraging mature solvers~\citep{surynek2016efficient,surynek2022problem}. 
While such methods can find high-quality or even provably optimal solutions, they frequently struggle with large problem instances or dynamic environments. 
Real-time constraints, uncertainties in dynamics and perception, and non-stationarity can severely degrade performance when these classical methods are applied directly~\citep{sartoretti2019primal,alkazzi2024comprehensive}.

In recent years, there has been a noticeable growth in learning-based approaches to MAPF—ranging from imitation learning and reinforcement learning (RL) to evolutionary methods and even emerging paradigms involving foundation models~\citep{alkazzi2024comprehensive}. 
The motivation behind these data-driven approaches is twofold. 
First, learning-based methods can be more adaptable in complex or partially observable environments, where classical methods may become computationally infeasible or require excessive domain-specific heuristics. 
Second, they offer the potential to generalize from experience across different instances, reducing the design effort needed when facing new or changing environments. 
Nevertheless, these learning-based solutions typically scale to fewer agents (e.g., on the order of hundreds)~\citep{li2022multi,skrynnik2024learn} as compared to some of the most advanced classical methods that can handle thousands~\citep{friedrich2024scalable,okumura2024engineering}. 
This gap in scalability, along with various methodological differences, underscores the necessity of a more holistic view that integrates classical and learning insights.

A number of surveys have highlighted either the classical MAPF literature or the recent surge of learning-based approaches~\citep{stern2019multi,ma2022graph,surynek2022problem,alkazzi2024comprehensive,chung2024learning}. 
However, the field currently lacks a comprehensive examination that places both paradigms side by side, evaluates their respective advantages and shortcomings, and provides guidance for combining them. 
In particular, the community could benefit from a detailed analysis of 
(i) how learning-based solutions can draw inspiration from the theoretical properties and algorithmic designs of classical MAPF and 
(ii) the ways in which classical methods can leverage learning components for improved scalability and robustness in real-world applications. 
Such an integrated view can spur new innovations, especially in learning-based MAPF, as it brings into focus opportunities for synergy, such as replacing certain heuristic modules in classical solutions with learned policies, or fusing optimization-based back-ends with representation learning.

\begin{figure}[htb!]
    \centering
    \includegraphics[width=\linewidth]{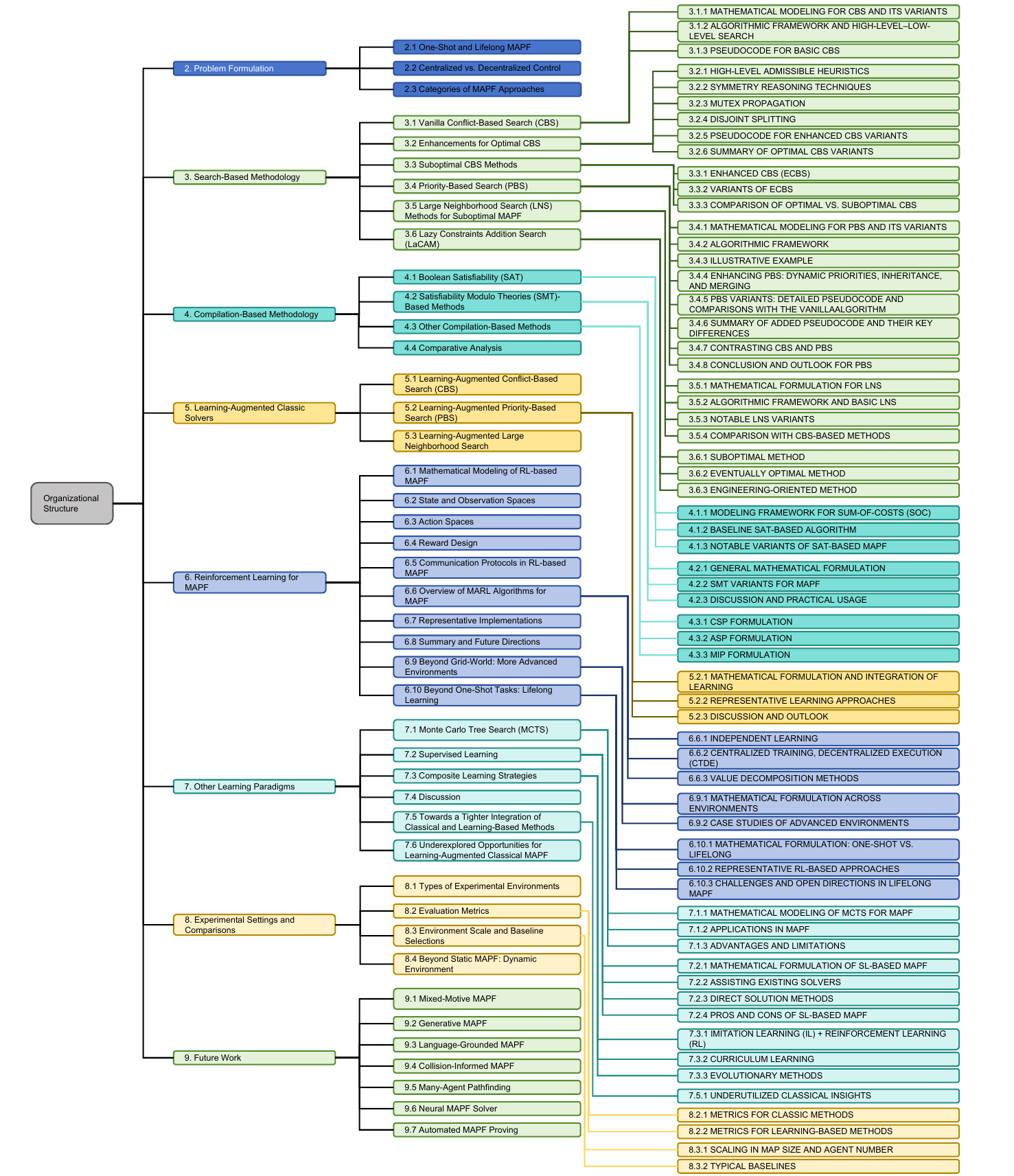}
    \caption{Structure diagram of the paper (Section 2 - Section 9).}
    \label{fig:overview}
\end{figure}

Furthermore, empirical evaluations of MAPF solutions differ widely across publications, making it difficult to assess their performance in a uniform manner. 
While many studies employ grid-based maps with static obstacles, others look at continuous or dynamic environments, using different metrics such as makespan, sum of costs, or success rate. 
Additional variability arises from the choice of baseline algorithms, map size, and the number of agents. 
This survey aims to address these inconsistencies by providing a systematic analysis of experimental settings and evaluation metrics employed by both classical and learning-based MAPF methods. 
Such a comparative study can foster best practices in benchmarking and draw attention to critical gaps in real-world deployability.

In this survey, we make several key contributions:

\begin{itemize}[leftmargin=*]
    \item We offer a unified viewpoint for understanding MAPF and its wide-ranging variants, clarifying how different methods formalize and solve the same underlying challenge.
    \item We provide a comprehensive review of both classical approaches---including search-based and compilation-based methods and learning-based approaches, highlighting recent developments in RL, imitation learning, evolutionary algorithms, and emerging large language model techniques.
    \item We present an in-depth comparison between classical and learning-based MAPF, focusing on scalability, robustness, constraints, and adaptability to dynamic environments.
    \item We conduct a detailed analysis of experimental design across existing MAPF literature, examining map types, map sizes, number of agents, evaluation criteria, and choice of baselines, thus providing insights into generalizability and real-world applicability.
    \item We outline promising directions for future research, including potential for foundation models in MAPF, dynamic environment handling, hybrid classical-learning approaches, and more practical deployments beyond controlled laboratory settings.
\end{itemize}


The remainder of this survey is organized as follows, as shown in Figure~\ref{fig:overview}. 
In Section~\ref{sec:formulation}, we introduce the classical definition of MAPF and discuss various formulations, noting that a single mathematical model cannot capture the breadth of MAPF approaches. 
We subsequently explore in Section~\ref{sec:search} and Section~\ref{sec:compilation} the main categories of classical methods: search-based and compilation-based solutions. 
Then, in Section~\ref{sec:augmenting},~\ref{sec:rl} and~\ref{sec:others}, we shift our focus to learning-based MAPF methods, beginning with hybrid frameworks that incorporate learned modules into otherwise classical pipelines, followed by fully data-driven RL solutions and, finally, other learning paradigms such as imitation learning, evolutionary algorithms, and emerging approaches using large language models. 
In Section~\ref{sec:exp}, we provide a comparative analysis of experimental setups employed in the literature, highlighting how variations in map settings, agent populations, performance metrics, and baseline choices can influence reported outcomes. 
We conclude in Section~\ref{sec:future} by identifying future directions for MAPF research, such as laying the groundwork for foundation models, extending MAPF to dynamic or partially observable domains, and demonstrating early experimental results that validate these directions.

By synthesizing these diverse perspectives, we aim for this survey to serve as a comprehensive guide to both classical and learning-based MAPF. 
We believe that bridging these two domains ultimately benefits the broader research and industrial communities, accelerating progress toward scalable, robust, and intelligent solutions in multi-agent coordination.

\section{Problem Formulation}\label{sec:formulation}

\begin{figure}[htb!]
    \centering
    \begin{subfigure}[b]{0.5\textwidth}
        \centering
        \includegraphics[width=\textwidth]{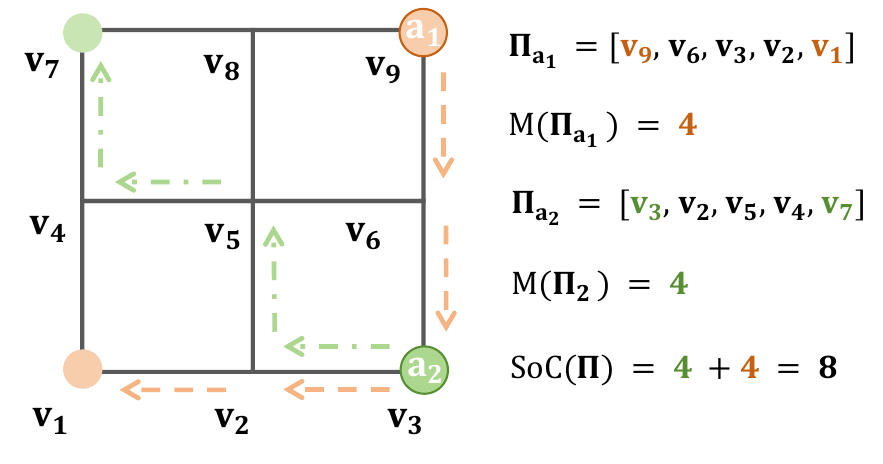}
        \caption{}
        \label{fig:mapf_1}
    \end{subfigure}
    \begin{subfigure}[b]{0.43\textwidth}
        \centering
        \includegraphics[width=\textwidth]{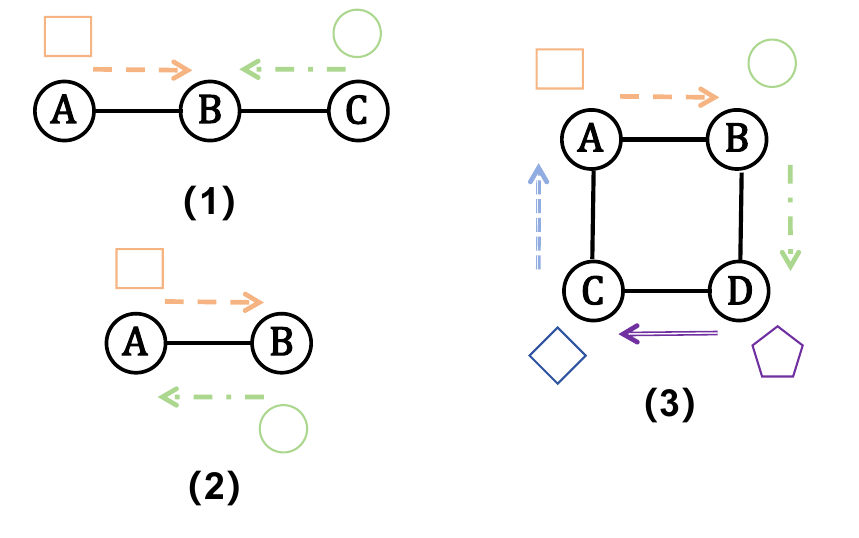}
        \caption{}
        \label{fig:mapf_2}
    \end{subfigure}
    \caption{Illustration of (a): a MAPF instance and (b): different types of conficts. (a): the individual paths of two agents, $a_1$ and $a_2$, along with a solution whose sum of costs (SoC) is 8. (b): (1) a
vertex confict, (2) a swapping confict, and (3) a cycle confict.}
    \label{fig:mapf}
\end{figure}

As discussed in Section~\ref{sec:intro}, Multi-Agent Path Finding (MAPF) broadly involves finding collision-free paths for a set of $n$ agents operating within a shared environment. 
The definitions in this section revolve around the mainstream MAPF formulations and several influential variants, which constitute the bulk of research addressed in this survey. 
Notably, many additional and more specialized MAPF variants also exist in the literature (see, e.g., \citep{stern2019multi} for a comprehensive overview). 
In the classic formulation, as shown in Figure~\ref{fig:mapf}, the environment is represented as an undirected graph $\mathcal{G} = (\mathcal{V}, \mathcal{E})$, where $\mathcal{V}$ is the set of vertices (possible locations), and $\mathcal{E}$ is the set of edges indicating valid deterministic transitions between locations~\citep{stern2019multi,surynek2022problem}. 
Each agent $i \in \{1, \dots, n\}$ is assigned a start vertex $s_i \in \mathcal{V}$ and, in many scenarios, a goal vertex $g_i \in \mathcal{V}$. 
Over discrete (or continuous) time steps, each agent must traverse the graph from $s_i$ to $g_i$ without colliding with other agents. 
A \emph{collision} occurs if two agents occupy the same vertex at the same time or if they traverse the same edge in opposite directions at the same time. 
The objective is to devise a set of paths that respect these collision-free constraints while optimizing one or more cost functions defined below.


\subsection{One-Shot and Lifelong MAPF}

\paragraph{One-Shot MAPF}
In a \emph{one-shot} MAPF problem, each agent has a fixed start position $s_i$ and a fixed goal position $g_i$. 
Once an agent reaches its goal, it has effectively completed its mission. 
Two primary optimization objectives typically arise in one-shot MAPF:
\begin{enumerate}[leftmargin=*]
    \item \textbf{Makespan Minimization.} 
    The makespan objective minimizes the maximum arrival time among all agents. 
    Formally, given a path $\pi_i$ for each agent $i$ and a cost function $\mathrm{Cost}(\pi_i)$ measuring the length (or time) of the path,\footnote{In time-discretized settings, $\mathrm{Cost}(\pi_i)$ often equals the number of time steps until agent $i$ reaches $g_i$.} the makespan is
    \begin{equation}\label{eq:makespan}
        \text{Makespan}(\{\pi_i\}_{i=1}^n) \;=\; \max_{1 \leq i \leq n} \mathrm{Cost}(\pi_i).
    \end{equation}
    Minimizing \eqref{eq:makespan} focuses on reducing the worst-case completion time across all agents.
    
    \item \textbf{Sum-of-Costs (SoC).}
    The SoC objective aims to minimize the total cost incurred by all agents. 
    If $\mathrm{Cost}(\pi_i)$ denotes the cost of agent $i$'s path, then
    \begin{equation}\label{eq:soc}
        \text{SoC}(\{\pi_i\}_{i=1}^n) \;=\; \sum_{i=1}^{n} \mathrm{Cost}(\pi_i).
    \end{equation}
    Equivalently, one may view this as minimizing the average cost per agent. 
    Methods optimized for SoC often strive to balance overall efficiency~\citep{surynek2022problem}.
\end{enumerate}

\paragraph{Lifelong MAPF}
In contrast, \emph{lifelong} MAPF involves a sequence of tasks or an ongoing flow of tasks where agents must complete multiple goals over an extended horizon~\citep{stern2019multi,tang2024large}. 
Rather than halting once a single goal is reached, each agent is either reassigned to new tasks or continues patrolling the environment. 
In this setting, optimization objectives typically concern maximizing the number of completed tasks in a limited time, minimizing idle time between tasks, or balancing similar operational metrics:
\begin{equation}\label{eq:lifelong}
    \max_{\{\pi_i\}_{i=1}^n} \; \text{TasksCompleted}(\{\pi_i\}_{i=1}^n),
\end{equation}
where \(\text{TasksCompleted}(\{\pi_i\}_{i=1}^n)\) indicates the cumulative total of tasks or subtasks that are successfully finished within a particular time horizon.

\paragraph{Remark 1}
In one-shot MAPF problems, beyond the most common optimization objectives of makespan and sum-of-costs, there exist other task-driven optimization goals. 
These include minimizing the total non-waiting steps required to complete all objectives, the total number of steps before all agents settle in their target positions, and various other metrics. 
For a more comprehensive overview of these alternative objective functions, readers can refer to \citet{stern2019multi}.

\paragraph{Remark 2} 
Beyond the two most prevalent problem settings, one-shot MAPF and lifelong MAPF, numerous problem variants exist in the literature. 
These include anonymous MAPF (where goal locations are interchangeable among agents), colored MAPF (where agents of the same color can be assigned to any goal of the matching color), and other specializations. 
For a more detailed discussion of these variants, we direct readers to \citet{stern2019multi}.

\paragraph{Remark 3}
The MAPF problem formulation presented in this paper implicitly assumes discrete time steps, uniform action durations (each action consumes exactly one time step), and that each agent occupies exactly one grid cell. 
In practical applications, these constraints are frequently relaxed. 
Examples include scenarios where different states and actions may consume varying amounts of time, grid cells may probabilistically contain multiple agents, and continuous-time extensions corresponding to motion planning problems. 
For a more thorough exploration of these generalizations, we refer readers to \citet{stern2019multi}.

\subsection{Centralized vs. Decentralized Control}

\paragraph{Centralized MAPF}
In a \emph{centralized} approach, a single planning entity has access to the global state of the environment, including the positions and goals of all agents. 
The central planner generates collision-free paths for every agent using either search-based methods \citep{sharon2015conflict,okumura2022priority} or compilation-based frameworks \citep{surynek2016efficient,surynek2022problem}, often yielding high-quality or even provably optimal solutions. 
However, centralized techniques can suffer from high computational overhead in large-scale or dynamically changing environments.

\paragraph{Decentralized MAPF}
Under \emph{decentralized} control, individual agents (or subsets of agents) plan their paths locally, possibly with partial observability or constrained communication among agents \citep{paoloposition,tjiharjadi2022systematic}. 
Decentralized frameworks must address the challenge of coordinating agents with limited global knowledge, frequently relying on consensus strategies or local collision avoidance rules. 
Formally, each agent $i$ has local information $\Omega_i$, representing its (potentially partial) view of the environment. 
The agent’s decision rule $\delta_i$ selects an action $a_i$ at each time step based on $\Omega_i$, that is,
\begin{equation}
    a_i(t) \;=\; \delta_i\bigl(\Omega_i(t)\bigr).
\end{equation}
These decentralized settings better reflect realistic constraints with limited sensing or communication but may require sophisticated algorithms to handle global conflict resolution.

\subsection{Categories of MAPF Approaches}

Drawing from our discussion in Section~\ref{sec:intro}, solutions to MAPF can be broadly categorized as:

\textbf{Search-Based Methods:} Classical graph search and tree-based algorithms that explicitly enumerate or prune the space of collision-free paths. They often guarantee completeness or optimality under certain assumptions but may struggle with large-scale instances.

\textbf{Compilation-Based Methods:} Formulate MAPF as an Integer Linear Program (ILP), a Satisfiability (SAT) problem, or other well-studied optimization frameworks \citep{surynek2016efficient}. These methods exploit powerful generic solvers but may also face scalability issues or long solve times.

\textbf{Learning-Based Methods:} Leverage diverse machine learning paradigms, such as RL, imitation learning, and evolutionary algorithms~\citep{alkazzi2024comprehensive,skrynnik2024learn}. While these can more readily adapt to uncertain or partially observable environments, they frequently handle fewer agents compared to large-scale classical approaches \citep{friedrich2024scalable}.

\textbf{Hybrid Methods:} Integrate learning components (e.g., learned heuristics or policies) into a classical MAPF pipeline to balance performance gains from learning with analytical guarantees from traditional solvers.

\begin{figure}[htb!]
    \centering
    \includegraphics[width=\linewidth]{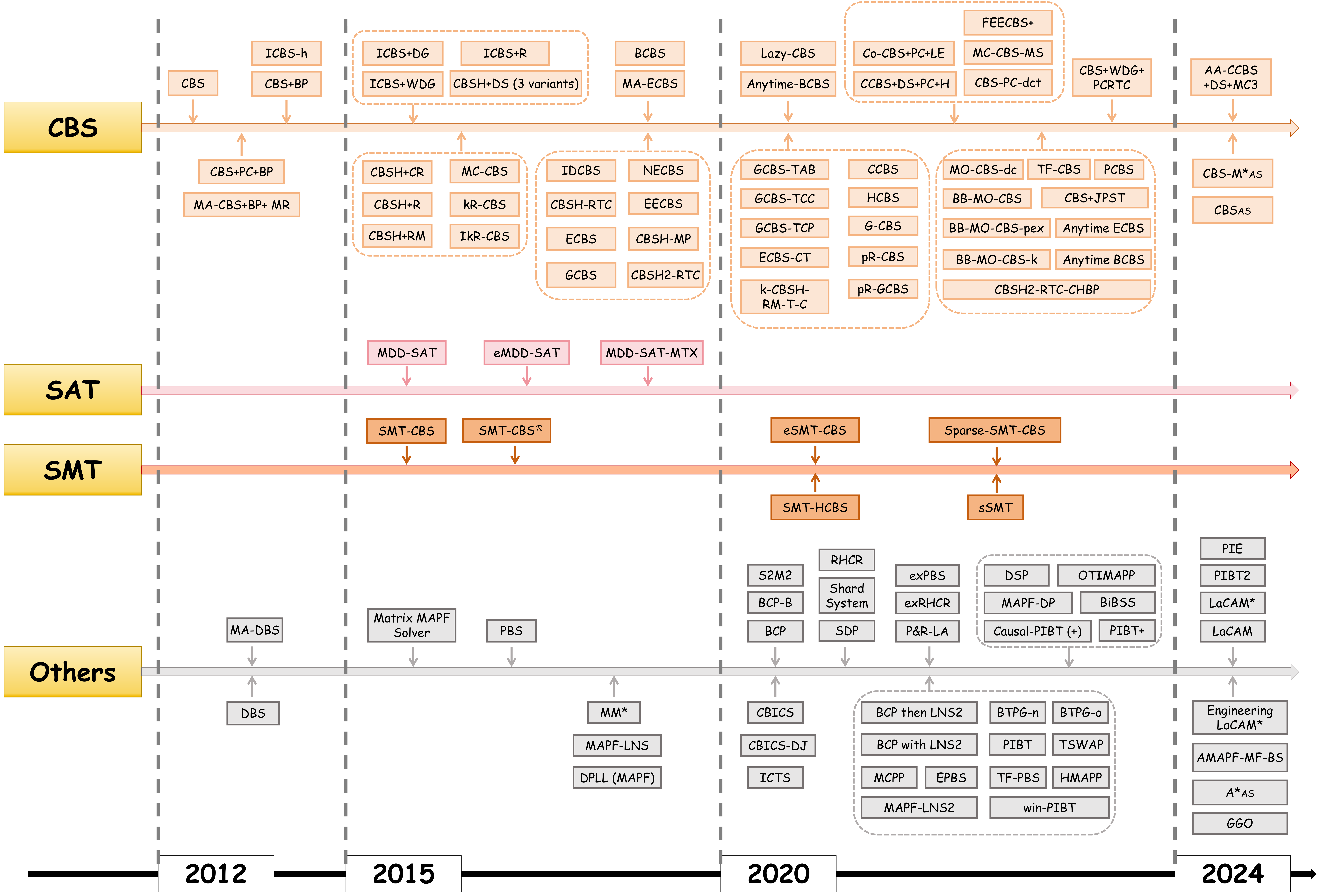}
    \caption{A chronological timeline of classical algorithms developed between 2012 and 2024, organized by core solving paradigms: CBS and its variants; SAT-based methods; SMT-based frameworks; and various other approaches.}
    \label{fig:classic_trend}
\end{figure}

\begin{figure}[htb!]
    \centering
    \includegraphics[width=\linewidth]{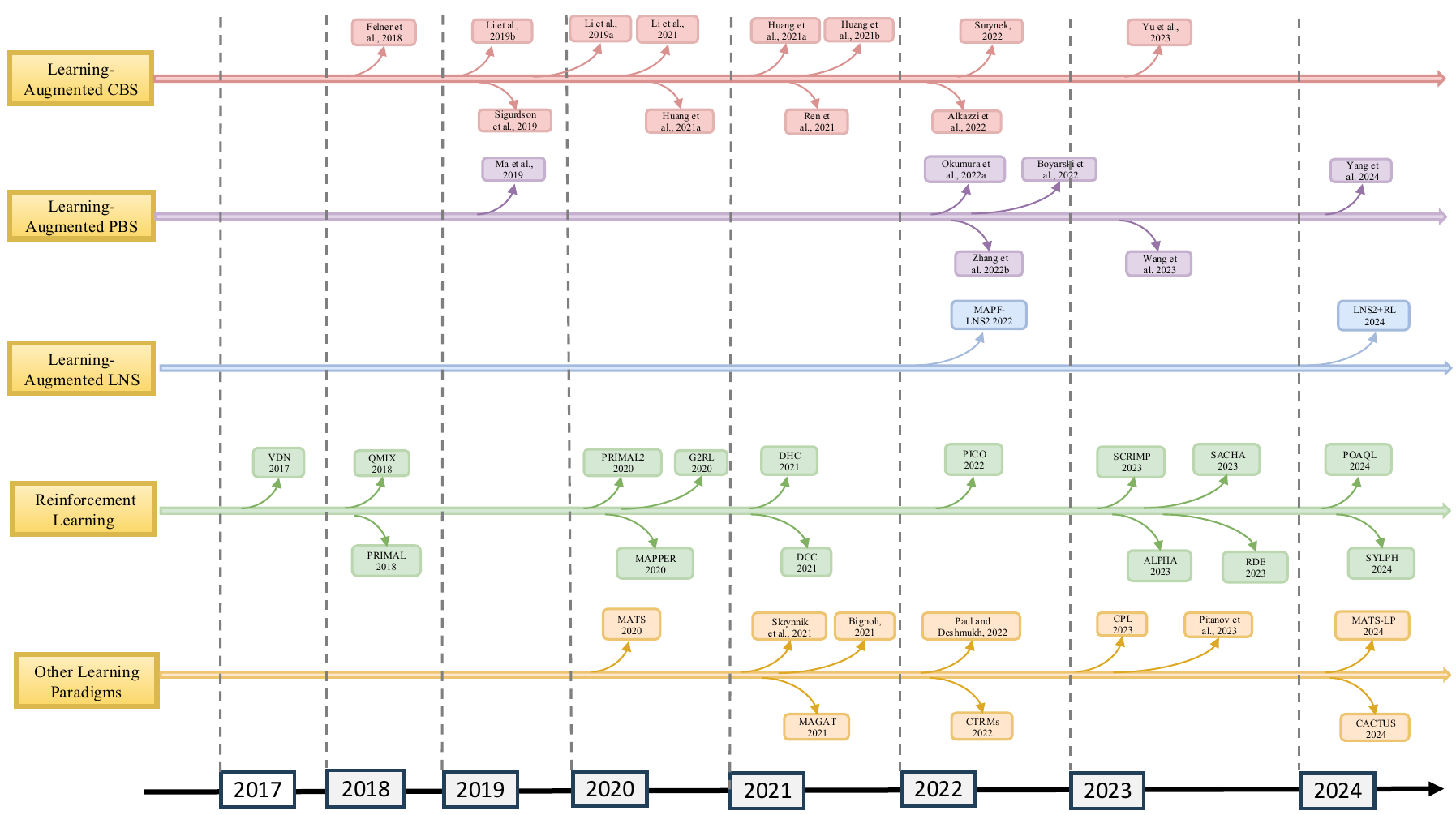}
    \caption{An evolution diagram of learning-based MAPF algorithms from 2017 to 2024, highlighting five main categories: enhanced CBS, enhanced PBS, enhanced LNS, reinforcement learning, and other learning paradigms.}
    \label{fig:learning_trend}
\end{figure}

The chronological timeline of representative classical and learning-based algorithms are shown in Figure~\ref{fig:classic_trend} and~\ref{fig:learning_trend}.

\section{Search-Based Methodology}\label{sec:search}

The growing breadth of MAPF research and its expanding real-world applications underscore the enduring importance of search-based strategies, even as learning-driven approaches gain momentum. 
Despite the recent successes of reinforcement learning and other data-centric methods in tackling this coordination challenge, search-based algorithms maintain a unique niche thanks to their algorithmic transparency, theoretical rigor, and strong performance in a wide range of problem instances. 
In particular, these algorithms excel at systematically uncovering collision-free solution paths in high-dimensional, combinatorial search spaces—a common scenario in large-scale MAPF setups. 
The following sections delve into the most influential and widely studied search-based methodologies, starting with the foundational Conflict-Based Search (CBS) and its extensive enhancements (both optimal and suboptimal), then moving to Priority-Based Search (PBS) as a complementary framework. 
We also explore Large Neighborhood Search (LNS) approaches, which have gained traction in suboptimal, time-sensitive, or resource-constrained MAPF settings. 
By examining the theoretical underpinnings and implementation details of these methods, we aim to highlight both their individual merits and the broader methodological interplay that fuels the advancement of MAPF solutions.



\subsection{Vanilla Conflict-Based Search (CBS)}\label{sec:cbs}

Prior to the advent of \emph{Conflict-Based Search} (CBS)~\citep{sharon2015conflict}, a variety of search-based algorithms were employed to tackle the MAPF problem. These methods not only addressed the problem effectively but also provided significant inspiration for CBS and its variants, including A*, \emph{Windowed Hierarchical Cooperative A*} (WHCA*)~\citep{silver2005cooperative}, \emph{Enhanced Partial
Expansion A*} (EPEA*)~\citep{felner2012partial}, M*~\citep{wagner2011m}, \emph{Operator
Decomposition} (OD)~\citep{standley2010finding}, ODrM*~\citep{ferner2013odrm} and \emph{Increasing Cost Tree Search} (ICTS)~\citep{sharon2011pruning}.

\begin{figure}[htb!]
    \centering
    \includegraphics[width=\linewidth]{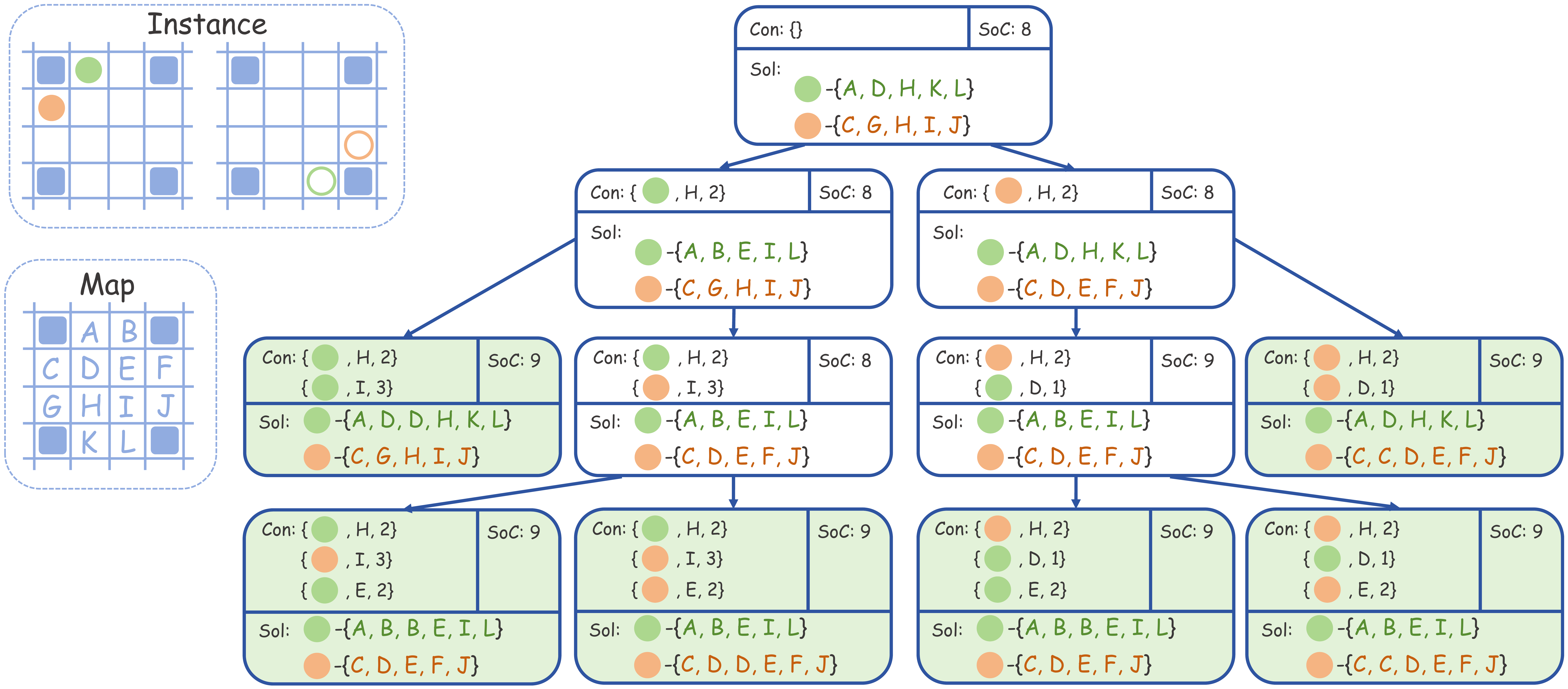}
    \caption{Illustration of the CBS process on a simple two-agent instance. Left: Two related source (solid circle) and target (hollow circle) configurations overlaid on a 5$\times$5 grid with obstacles (blue squares) and two agents (green and orange circles). The underlying graph labels traversable grids A–L. Right: The CT built from the root (Con = $\{ \}$, SoC = 8), where each node box reports: Con: the set of agent–location–time constraints (e.g. $\{$\textcolor[HTML]{ACD78E}{\ding{108}}, H, 2$\}$ forbids the green agent from H at timestep 2), SoC: the current sum-of-costs of the two individual shortest-path solutions, Sol: the ordered vertex sequences for the green and orange agents (shown in matching colors). At the root, the joint plan has a collision at (H,2) which represents the point in space-time, so CBS branches into two children that respectively prohibit either agent at that conflict. Subsequent collisions at (I,3) and (E,2) likewise generate further binary splits. Leaf nodes (shaded in light green) represent conflict-free plans whose SoC = 9, which is the optimal cost-minimal, collision-free solution found by CBS.}
    \label{fig:cbs}
\end{figure}

In this section, we introduce CBS, one of the most influential search-based methods for the MAPF problem. 
We first present a mathematical formulation of the MAPF problem tailored to CBS and its variants, followed by an overview of the CBS algorithmic framework and a pseudocode description of the basic CBS procedure. 
The illustration of the CBS process on a simple two-agent instance is shown in Figure~\ref{fig:cbs}.

\subsubsection{Mathematical Modeling for CBS and Its Variants}

Consistent with the notations in Section~\ref{sec:formulation}, consider an undirected graph $\mathcal{G} = (\mathcal{V}, \mathcal{E})$, a set of $n$ agents $\{1,\dots,n\}$, and a discrete time horizon $T$. 
For each agent $i$, we denote by $s_i \in \mathcal{V}$ its start vertex and by $g_i \in \mathcal{V}$ its goal vertex. 
A path (or trajectory) for agent~$i$ over the time horizon $T$ is a sequence of vertices
\[
   \pi_i \;=\; \bigl(v_{i}(0),\,v_{i}(1),\dots,v_{i}(T)\bigr), \quad v_i(t)\in \mathcal{V},
\]
such that $v_i(0)= s_i$ and $v_i(T)= g_i$. 
(If the agent is assumed to stop once reaching $g_i$, we may hold $v_i(t) = g_i$ for all remaining $t$.)

\vspace{1mm}\noindent
\textbf{Decision Variables.} 
From a constraint-based perspective, one introduces the binary variables 
\[
   x_{i,v,t} \;=\; \begin{cases}
   1, & \text{if agent $i$ is at vertex $v$ at time $t$,}\\
   0, & \text{otherwise}.
   \end{cases}
\]
Hence, $\pi_i$ is represented by setting $x_{i,v,t}=1$ exactly for $v = v_i(t)$ along agent~$i$’s path.

\vspace{1mm}\noindent
\textbf{Constraints.} 
The MAPF problem is characterized by twin collision-avoidance constraints:
\begin{subequations}\label{eq:cbs:constraints}
\begin{align}
   &x_{i,v,t} + x_{j,v,t} \;\le\; 1, \ \ \forall \, t, \,\forall \, v\in \mathcal{V}, \,\forall\, i < j,\ \ \text{(vertex collision avoidance)} \label{eq:collision-vertex}\\[4pt]
   &x_{i,u,t} \,+\, x_{i,v,t+1} \;+\; x_{j,v,t} \,+\, x_{j,u,t+1} \;\le\; 3, \ \ \forall \, (u,v)\in \mathcal{E},\forall \, t, \,\forall\, i < j, \ \ \text{(edge collision avoidance)}\label{eq:collision-edge}
\end{align}
\end{subequations}
where~\eqref{eq:collision-vertex} prevents multiple agents from occupying the same vertex at the same time, while~\eqref{eq:collision-edge} ensures that no two agents traverse the same edge in opposite directions at the same time.\footnote{One might equivalently model \eqref{eq:collision-edge} as constraints disallowing $x_{i,u,t} = x_{j,v,t}$ and $x_{i,v,t+1} = x_{j,u,t+1}$ for $(u,v)\in\mathcal{E}$. 
The form above simply encodes that at most three of these four binary variables can be $1$.} 
Additional constraints ensure that each agent maintains a consistent path from $s_i$ to $g_i$, typically:
\[
   \sum_{v\in \mathcal{V}} x_{i,v,t} \;=\; 1, \qquad\forall i,\;\forall 0\le t \le T,
\]
and $x_{i,v_i(0),0} = 1$, $x_{i,v_i(T),T} = 1$. 

\vspace{1mm}\noindent
\textbf{Objective.} 
CBS can be used to minimize different objectives.  Common choices include:
\begin{itemize}[leftmargin=*]
\item \textbf{Makespan}: $\displaystyle \min \max_{1 \leq i \leq n} \mathrm{Cost}(\pi_i)$, where $\mathrm{Cost}(\pi_i)$ typically represents the time at which agent~$i$ reaches $g_i$. 
\item \textbf{Sum of Costs (SoC)}: $\displaystyle \min \sum_{i=1}^n \mathrm{Cost}(\pi_i)$. 
\end{itemize}
The classical CBS framework is often presented for the SoC objective but can be adapted to the makespan objective. 

\subsubsection{Algorithmic Framework and High-Level--Low-Level Search}\label{sec:cbs:framework}

To solve the above MAPF formulation, CBS employs a two-level search:
\begin{itemize}[leftmargin=*]
    \item \emph{High-Level Search.} 
    Maintains a \emph{conflict tree} (CT), where each node contains (i) a set of constraints (i.e., forbidden vertex/time or edge/time tuples) for each agent and (ii) a set of complete paths, one per agent, that obey these constraints. 
    Upon finding a conflict between any two paths, the high-level search \emph{branches} into two new CT nodes, each adding a constraint to exactly one agent’s path. 
    \item \emph{Low-Level Search.} 
    Given the constraints for a single agent, the low-level search finds an optimal path for that agent respecting those constraints. 
    Typically, one uses standard single-agent pathfinding algorithms (e.g., A* or Dijkstra) in a time-expanded graph to compute the new path. 
\end{itemize}

The high-level search terminates when it reaches a node in the CT whose paths are all conflict-free. 
The set of paths at that node is an optimal joint solution under the given objective (makespan or SoC), provided the branching and low-level searches use consistent admissible heuristics or cost functions.

To make the two-level CBS search scheme more concrete, we illustrate a small example on a $2\times 3$ grid (see Figure~\ref{fig:cbs-example-grid}). 
The grid is indexed by \((r,c)\) with \(r\in\{0,1\}\) for rows and \(c\in\{0,1,2\}\) for columns. 
We consider two agents:

\begin{itemize}[leftmargin=*]
\item Agent~1 starts at \(s_1 = (0,0)\) and has goal \(g_1 = (1,2)\). 
\item Agent~2 starts at \(s_2 = (1,0)\) and has goal \(g_2 = (0,2)\).
\end{itemize}

\noindent
At each time step, the agents can either move to an adjacent cell (up, down, left, or right if valid) or remain in place if desired (cf.\ Section~\ref{sec:formulation}). 
For simplicity of exposition, we assume both agents seek to minimize their individual path lengths (Sum of Costs).

\begin{figure}[ht]
\centering
\begin{tabular}{c@{\hspace{1.5em}}c@{\hspace{1.5em}}c}
\(\mathbf{(0,0)}\) & \(\mathbf{(0,1)}\) & \(\mathbf{(0,2)}\)\\
\(\mathbf{(1,0)}\) & \(\mathbf{(1,1)}\) & \(\mathbf{(1,2)}\)
\end{tabular}
\caption{A $2\times 3$ grid environment for the CBS example.  
Agent~1 aims to move from $(0,0)$ to $(1,2)$; 
Agent~2 aims to move from $(1,0)$ to $(0,2)$.}
\label{fig:cbs-example-grid}
\end{figure}

We now provide a step-by-step example of how CBS resolves conflicts and arrives at a collision-free solution, using the small $2\times3$ grid from Figure~\ref{fig:cbs-example-grid}. 
We assume both agents can move one cell per discrete time step (up, down, left, or right) or remain idle. The objective is the Sum of Costs (SoC), where each agent’s cost is the earliest time step at which it reaches its goal.

\begin{enumerate}[leftmargin=1.5em]
\item \textbf{Root Node (No Constraints).}  
  \begin{enumerate}
  \item Each agent first plans a shortest path \emph{independently}, ignoring collisions.
    \[
      \pi_1^0: (0,0)\ \to\ (0,1)\ \to\ (0,2)\ \to\ (1,2),
    \]
    \[
      \pi_2^0: (1,0)\ \to\ (1,1)\ \to\ (1,2)\ \to\ (0,2).
    \]
  \item The root node $N_0$ of the Conflict Tree (CT) stores: 
    \begin{itemize}
      \item \emph{Constraint Sets:} All empty.
      \item \emph{Paths:} $\pi_1^0,\,\pi_2^0$.
      \item \emph{Cost}: $\mathrm{SoC} = \mathrm{Cost}(\pi_1^0)+\mathrm{Cost}(\pi_2^0)$. 
        (Each path has length $3$ in this example, so $\mathrm{SoC}=6$.)
    \end{itemize}
  \item We place $N_0$ into the Open List, which is typically a priority queue ordered by SoC (or by makespan, if optimizing that objective).
  \end{enumerate}

\item \textbf{Detecting a Conflict.}  
  \begin{enumerate}
  \item We pop $N_0$ from the Open List and check the paths $\pi_1^0,\,\pi_2^0$ for collisions:
    \begin{itemize}
      \item At $t=0$: Agent~1 is at \((0,0)\) and Agent~2 is at \((1,0)\). No conflict.
      \item At $t=1$: Agent~1 is at \((0,1)\) and Agent~2 is at \((1,1)\). No conflict.
      \item At $t=2$: Agent~1 is at \((0,2)\) and Agent~2 is at \((1,2)\). They occupy different cells, hence no vertex conflict yet. 
        However, suppose we look ahead to $t=3$: 
        Agent~1 would move to \((1,2)\), and Agent~2 would move to \((0,2)\). 
        They attempt to \emph{swap} positions via the edge \((0,2)\leftrightarrow(1,2)\) simultaneously. 
        This is a classic \emph{edge conflict} in MAPF, as they traverse the same edge in opposite directions at the same time.\footnote{A conflict could also arise if both agents tried to occupy the same cell. The resolution mechanism is similar.}
    \end{itemize}
  \item Because we have found a conflict between Agents~1 and~2 at (or around) time $t=3$, we must \emph{branch} in the Conflict Tree.
  \end{enumerate}

\item \textbf{Branching into Two Child Nodes.}  
  \begin{enumerate}
  \item At node $N_0$, we have one conflict: \(\mathrm{Conflict}(1,2,t=3)\). The algorithm creates two child nodes $N_A$ and $N_B$:
    \[
      N_A:\ \text{Add constraint forbidding Agent~1’s conflicting move, replan for Agent~1 only};
    \]
    \[
      N_B:\ \text{Add constraint forbidding Agent~2’s conflicting move, replan for Agent~2 only}.
    \]
  \item \emph{Constraint Sets in Each Child Node.}
    \begin{itemize}
      \item In $N_A$, Agent~1 receives an additional constraint:
      \[
        \text{``Agent~1 may not occupy (or use edge to) }(1,2)\text{ at }t=3\text{.''}
      \] 
      \item In $N_B$, Agent~2 receives a similar constraint forbidding the move to \((0,2)\) at $t=3$.
    \end{itemize}
  \end{enumerate}

\item \textbf{Low-Level Replanning for a Single Agent.}  
  \paragraph{Child Node $N_A$.} 
  \begin{enumerate}
  \item Agent~2’s path \(\pi_2^0\) remains unchanged.
  \item Agent~1 runs a low-level path search (e.g., A*) with the new constraint:
    \[
      \text{Cannot use edge or cell }(0,2)\leftrightarrow(1,2)\text{ at }t=3.
    \]
  \item A feasible updated path might be:
    \[
       \pi_1^A:\quad
       (0,0)\xrightarrow{t=0\to1}
       (0,1)\xrightarrow{t=1\to2}
       (1,1)\xrightarrow{t=2\to3}
       (1,2).
    \]
  \item In time-expanded terms:
    \begin{itemize}
      \item $t=0$: Agent~1 at (0,0).
      \item $t=1$: Move to (0,1).
      \item $t=2$: Move to (1,1).
      \item $t=3$: Move to (1,2) (Goal).
    \end{itemize}
  \item Check if $\pi_2^0$ and $\pi_1^A$ conflict:
    \[
      \pi_2^0: \ (1,0)\,\to\,(1,1)\,\to\,(1,2)\,\to\,(0,2).
    \]
    \begin{itemize}
      \item $t=1$: Agent~1 at (0,1), Agent~2 at (1,1) (no conflict).
      \item $t=2$: Agent~1 at (1,1), Agent~2 at (1,2). Different cells, no conflict.
      \item $t=3$: Agent~1 at (1,2), Agent~2 at (0,2). Different cells, no conflict.
    \end{itemize}
  \item No further conflict is found. Thus $N_A$ is conflict-free.
  \item The new SoC for $N_A$ is 
    \(\mathrm{Cost}(\pi_1^A)+\mathrm{Cost}(\pi_2^0) = 3 + 3 = 6\).\footnote{Strictly speaking, each agent took 3 moves to reach the goal. Another counting convention might yield a sum of $(3 + 3)=6$. The main point is that the total cost remains feasible.}
  \end{enumerate}

  \paragraph{Child Node $N_B$.}
  \begin{enumerate}
  \item Agent~1’s path remains \(\pi_1^0\).
  \item Agent~2 runs a low-level search forbidding \((0,2)\) at $t=3$. 
  \item A possible updated path for Agent~2 might be:
    \[
      \pi_2^B:\quad
      (1,0)\,\to\,(1,1)\,\to\,(0,1)\,\to\,(0,2).
    \]
    Technically, Agent~2 detours or adjusts its timing to avoid reaching \((0,2)\) at $t=3$. It might arrive at \((0,2)\) at $t=4$ or pass through \((0,1)\) if timing constraints so require.
  \item If $\pi_2^B$ and $\pi_1^0$ remain in conflict, CBS would branch again. But suppose $\pi_2^B$ yields no further collision. Then $N_B$ is also conflict-free, providing a second valid solution.
  \end{enumerate}

\item \textbf{Selecting the Final Solution.}  
  \begin{enumerate}
  \item In practice, as soon as CBS finds \emph{any} conflict-free node in the CT, it \emph{terminates} with that solution if the algorithm is searching in best-first order (optimal search).
  \item In our example, both $N_A$ and $N_B$ are feasible child nodes with no further conflicts. They each yield a total SoC of 6. 
  \item CBS can return either solution: 
    \begin{itemize}
    \item \textbf{Solution\,1 (from $N_A$):}\\
      Agent~1’s path: 
      \[
       (0,0)\,\to\,(0,1)\,\to\,(1,1)\,\to\,(1,2),
      \]
      Agent~2’s path: 
      \[
       (1,0)\,\to\,(1,1)\,\to\,(1,2)\,\to\,(0,2).
      \]
    \item \textbf{Solution\,2 (from $N_B$):}\\
      Agent~1’s path:
      \[
       (0,0)\,\to\,(0,1)\,\to\,(0,2)\,\to\,(1,2),
      \]
      Agent~2’s path:
      \[
       (1,0)\,\to\,(1,1)\,\to\,(0,1)\,\to\,(0,2).
      \]
    \end{itemize}
  \item Both solutions are valid and yield the same cost under typical shortest-path metrics.
  \end{enumerate}
\end{enumerate}

\noindent
\textbf{Key Takeaways.}  
\emph{(i)}~CBS only constrains \emph{one} agent per conflict, creating minimal constraint sets.  
\emph{(ii)}~Each child node replans for a single agent at the low level, keeping other agents’ paths fixed.  
\emph{(iii)}~The search terminates upon encountering the first conflict-free node in best-first order, guaranteeing an optimal solution for the chosen cost metric (SoC or makespan).  
\emph{(iv)}~In the example, once a child node’s constraints eliminate conflicts, no further branching is needed.  
Real-world cases with more agents may generate deeper conflict trees and more branching nodes, but the same principle applies: each conflict is incrementally resolved by splitting into two child nodes, each adding a constraint to exactly one agent.

\subsubsection{Pseudocode for Basic CBS}\label{sec:cbs:algorithm}

Algorithm~\ref{alg:cbs} shows a simplified pseudocode for the basic CBS procedure. 
Subsequent CBS variants (e.g., \emph{Enhanced CBS}, \emph{Meta-agent CBS}, \emph{ICBS}~\citep{felner2018adding}) enrich or modify this basic structure by altering the conflict prioritization strategy, introducing more advanced heuristics, or refining how constraints are defined and propagated. 
Nevertheless, the core two-level architecture remains the same: a high-level conflict resolution guided by branching constraints and a low-level single-agent path search.

\begin{algorithm}[t]
\footnotesize
\caption{Basic Conflict-Based Search (CBS)\label{alg:cbs}}
\begin{algorithmic}[1]
\Require Graph $\mathcal{G}=(\mathcal{V},\mathcal{E})$, agents $\{1,\dots,n\}$, start vertices $\{s_i\}$, goal vertices $\{g_i\}$, objective (\emph{SoC} or \emph{Makespan}).
\State Initialize the root node $N_0$ of the conflict tree (CT):
\begin{itemize}
   \item For each agent $i$, compute an individual path $\pi_i$ from $s_i$ to $g_i$ ignoring collisions (e.g., by A*). 
   \item Set the \emph{constraint set} for each agent to be empty.
\end{itemize}
\State Insert $N_0$ into a \emph{priority queue} or \emph{open list} (e.g., sorted by $\sum_i \mathrm{Cost}(\pi_i)$).
\While{the open list is not empty}
  \State $N \gets \text{pop front of open list}$  \Comment{Choose node with smallest cost}
  \State Check for conflicts among the paths $\{\pi_i\}$ stored in $N$.
  \If{there is \emph{no conflict}}
    \State \textbf{return} $\{\pi_i\}$ as the optimal solution.
  \Else 
    \State Let $(i,j,t)$ be the first conflict detected at time $t$ between agents $i$ and $j$. 
    \State \(\!\!\!\!\!\!\) \Comment{Two new child nodes, each adding one extra constraint}
    \For{each agent $a \in \{i, j\}$}
       \State Create a new node $N'$ by copying $N$.
       \State Add a new constraint to agent $a$’s constraint set to \emph{forbid} the conflicting vertex or edge at time $t$.
       \State Recompute an optimal path $\pi_a$ for agent $a$ under the updated constraint set in $N'$ (low-level A*).
       \If{a path is found}
          \State Insert $N'$ into the open list, with updated total cost. 
       \EndIf
       \Comment{If no path is found for agent $a$, discard $N'$.}
    \EndFor
  \EndIf
\EndWhile
\State \textbf{return} \textit{No feasible solution} \Comment{In rare cases, if no valid paths exist}
\end{algorithmic}
\end{algorithm}

\noindent
\textbf{Algorithm Explanation.} 
\begin{itemize}[leftmargin=*]
    \item \emph{Initialization (Lines 1--2).} 
    The root node \emph{ignores} collisions at first. Each agent finds a shortest path $\pi_i$ from $s_i$ to $g_i$, and all constraint sets are empty. The initial cost of the root node is computed (sum of costs or makespan) over $\{\pi_i\}$.
    \item \emph{Conflict Checking (Lines 4--5).} 
    At each CT node, CBS searches for collisions among the paths. If no collision is found, a globally valid solution is reached.
    \item \emph{Branching (Lines 8--18).} 
    If a collision is detected at time $t$ between agents $i$ and $j$, the algorithm creates two new child nodes. Each child node introduces a new constraint \emph{for only one} of the two agents, disallowing the conflict at time $t$. 
    After adding the constraint, the algorithm re-plans \emph{only} for that agent. 
    If a feasible path is found, the child node is appended to the priority queue; otherwise, it is discarded.
\end{itemize}

By resolving conflicts incrementally rather than imposing all constraints upfront, CBS often explores fewer joint states than a naive multi-agent search would. 
The method is complete for grid-based or graph-based MAPF and, under suitable cost functions (e.g., path length), is also optimal. 
Subsequent sections will discuss how suboptimal or bounded-suboptimal variants of CBS trade off optimality guarantees for computational efficiency, often by imposing additional heuristics on conflict detection or node expansion order. 
Thus, CBS forms a core building block for the broader class of search-based MAPF algorithms, serving as a reference point for numerous enhancements and hybridizations.

Algorithm~\ref{alg:cbs} provides the high-level pseudocode for the canonical CBS procedure. We augment it below with a step-by-step illustration of how conflicts are resolved in the example from Section~\ref{sec:cbs:framework}.

\vspace{1em}
\noindent
\textbf{Example (continued).}  
In the $2\times3$ grid scenario (Figure~\ref{fig:cbs-example-grid}):
\begin{itemize}[leftmargin=*]
\item \emph{Initialization (Lines\,1--2):} 
  - CBS first obtains two unconstrained shortest paths for Agent~1 and Agent~2. 
  - These paths conflict at time $t=3$ in the example scenario (both trying to reach \((0,2)\) or crossing $(0,2)\leftrightarrow(1,2)$).
\item \emph{Branching (Lines\,8--19):} 
  - Suppose CBS labels the conflict as $(1,2,t=3)$ (i.e., Agents~1 and~2 are in conflict at time 3). 
  - Child node 1 adds a constraint to Agent~1, forbidding it from occupying the conflicting vertex at time 3. Agent~1 replans a new path, e.g., by detouring via \((1,1)\). 
  - Child node 2 adds a constraint to Agent~2 instead, causing Agent~2 to find a different route or timing that avoids the collision.
\item The conflict tree grows with each detected conflict, introducing constraints and verifying whether each replan is feasible. 
\item When CBS encounters a node in which \emph{no} conflict is detected, it terminates and returns the corresponding paths as an optimal joint solution.
\end{itemize}

This interplay of \emph{high-level conflict resolution} and \emph{low-level single-agent replanning} underlies all CBS-based methods. The branching ensures that only one agent at a time receives an additional constraint, preserving the minimal set of constraints necessary to eliminate each conflict. As a result, the algorithm often converges faster than joint multi-agent A* in practice, while still guaranteeing completeness and optimality under the chosen objective.

\subsection{Enhancements for Optimal CBS}
\label{sec:cbs:optimal}

\begin{figure}[htb!]
    \centering
    \includegraphics[width=\linewidth]{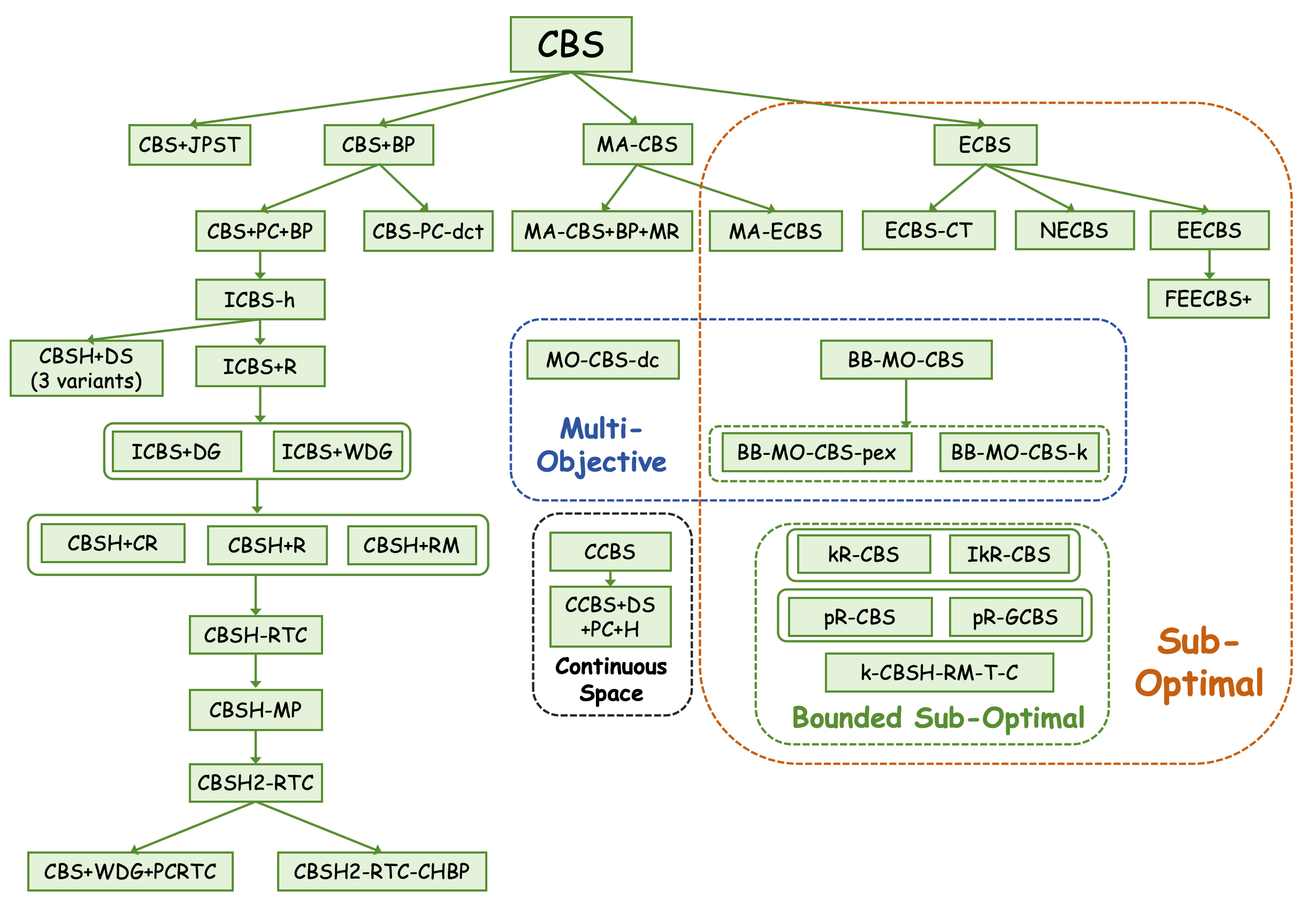}
    \caption{An evolutionary graph of the research work conducted on CBS. Starting from the vanilla CBS framework at the top, the diagram shows main development axes: optimal CBS variants (all nodes and branches outside the orange-dashed region); sub-optimal CBS variants (orange-dashed region); multi-objective CBS variants (blue-dashed region) and CBS variants in continuous space (black-dashed region).}
    \label{fig:cbs_variants}
\end{figure}

Building upon the baseline CBS algorithm described in Section~\ref{sec:cbs}, numerous enhancements have been proposed to achieve optimal (or bounded-optimal) solutions more efficiently \citep{sharon2015conflict,boyarski2015icbs,felner2018adding,li2019improved,li2019symmetry,li2020new,li2021pairwise,zhang2022multi}. 
These enhancements mainly address two interwoven aspects of the CBS framework:

\begin{enumerate}[leftmargin=*, itemsep=1pt, parsep=1pt]
    \item \emph{The high-level (HL) search}: introducing admissible heuristics, advanced conflict prioritization, and refined conflict splitting rules.
    \item \emph{The low-level (LL) search}: augmenting single-agent pathfinding with techniques such as multi-valued decision diagrams (MDDs) and mutex propagation to prune infeasible or redundant paths quickly. An evolutionary graph of the research work conducted on CBS is shown in Figure~\ref{fig:cbs_variants}.
\end{enumerate}

\paragraph{Multi-Valued Decision Diagrams (MDDs.)}
A multi-valued decision diagram, $\mathcal{MDD}_i^{\ell}$, for agent $i$, is a directed acyclic graph with $\ell+1$ levels, enumerating \emph{all} feasible single-agent paths of length $\ell$ that satisfy the agent-specific constraints in the current CT node.  
Each node in $\mathcal{MDD}_i^{\ell}$ represents a vertex~(location) and time-step pair $(v,t)$, and each edge encodes a valid transition (or wait action).  
MDDs are central to identifying whether two agents are \emph{dependent}---if their joint MDD is empty, then they cannot be simultaneously conflict-free without additional constraints (see \S\ref{sec:cbs:hl-heuristics}).

Collectively, these refinements have drastically improved CBS performance, sometimes reducing the number of conflict-tree (CT) node expansions by several orders of magnitude and rendering formerly intractable MAPF instances solvable in practice.

In this subsection, we highlight four core enhancement modules for \emph{optimal} CBS:
\begin{itemize}[leftmargin=*]
    \item \textit{High-level admissible heuristic} (\S\ref{sec:cbs:hl-heuristics}): Exploit conflict graphs~\citep{felner2018adding}, pairwise dependency graphs (DGs)~\citep{li2019improved}, and their weighted variants (WDGs) to guide the high-level search.
    \item \textit{Symmetry reasoning techniques} (\S\ref{sec:cbs:symmetry}): Introduce additional constraints (rectangle, corridor, target constraints, \emph{etc}.) to break symmetrical collisions early.
    \item \textit{Mutex propagation} (\S\ref{sec:cbs:mutex}): Employ MDD-based mutual-exclusion checks to identify unreachable or conflicting states in the low-level search.
    \item \textit{Disjoint splitting} (\S\ref{sec:cbs:disjoint}): Alter the branching scheme so that each conflict splits into two child nodes that enforce complementary constraints, ensuring their candidate plans remain mutually exclusive.
\end{itemize}
We first provide a brief overview of these modules, then show how to integrate each into the CBS pseudocode with minimal modifications highlighted.

\subsubsection{High-Level Admissible Heuristics}
\label{sec:cbs:hl-heuristics}

Standard CBS expands nodes in the conflict tree (CT) by using the sum of paths' costs (or makespan) as the priority.  
\emph{High-level admissible heuristics} augment this with more informed estimates of the minimal number of conflicts to resolve.  
Key representative heuristics include:
\begin{itemize}[itemsep=1pt, parsep=1pt, leftmargin=*]
    \item \textbf{Conflict Graph (CG)~\citep{felner2018adding}:} 
    Construct a graph whose vertices correspond to agents, adding an edge whenever two agents admit a \emph{cardinal conflict}.  
    Taking the minimum vertex cover (MVC) of this graph yields a heuristic value.  
    \item \textbf{Pairwise Dependency Graph (DG)~\citep{li2019improved}:} 
    Incorporates not only cardinal conflicts but also semi-cardinal or non-cardinal ones.  
    MDDs help identify whether two agents are \emph{dependent}. 
    The MVC of the DG then provides a refined heuristic estimate.
    \item \textbf{Weighted Pairwise Dependency Graph (WDG)~\citep{li2019improved}:} 
    Assign each edge a weight $\Delta_{i,j}$ measuring the difference in the pairwise SoCs for relevant conflicts.  
    The \emph{edge-weighted minimum vertex cover} (EWMVC) of the WDG further tightens the HL heuristic.
\end{itemize}

\subsubsection{Symmetry Reasoning Techniques}
\label{sec:cbs:symmetry}

\emph{Symmetry reasoning} adds specialized constraints whenever two agents repeatedly collide under symmetrical conditions:
\begin{itemize}[itemsep=1pt, parsep=1pt, leftmargin=*]
    \item \textbf{Rectangle symmetries~\citep{li2019symmetry}:} Two agents create a \emph{rectangle conflict} if each uses only Manhattan-optimal paths and collides in an axis-aligned sub-grid.  
    Extended forms (\emph{generalized rectangle symmetry}~\citep{li2021pairwise}) exploit intersecting MDDs to identify larger conflicting sub-lattices.
    \item \textbf{Target symmetries~\citep{li2020new}:} Where an agent waits indefinitely at its goal, forcing frequent collisions with other agents passing through that cell.
    \item \textbf{Corridor symmetries~\citep{li2020new}:} If two agents attempt to enter a narrow corridor (one-dimensional sub-grid) from opposite ends, they may repeatedly block each other.
\end{itemize}
By detecting these patterns, the HL search can introduce specialized constraints that eliminate \emph{all} symmetric permutations of the same conflict at once, dramatically reducing expansions.

\subsubsection{Mutex Propagation}
\label{sec:cbs:mutex}

\emph{Mutex propagation}~\citep{zhang2022multi} prunes infeasible or redundant portions of the MDD.  
Two MDD nodes $n_i$ and $n_j$ at time step $t$ are \emph{mutex} (mutually exclusive) if no valid collision-free sub-paths can place agents $i$ and $j$ at $n_i.\text{loc}$ and $n_j.\text{loc}$ respectively at time~$t$.  
Identifying and removing such \emph{mutex} pairs at each level yields a more accurate approximation of feasible states for the LL search, thus guiding CBS to detect and handle collisions more effectively.

\subsubsection{Disjoint Splitting}
\label{sec:cbs:disjoint}

In standard CBS, a conflict $(i, j, v, t)$ splits the CT node into two children: one forbids agent $i$ from being at $(v,t)$, the other forbids agent $j$.  
\emph{Disjoint splitting}~\citep{li2019disjoint} alters this branching so that one child \emph{verifies} agent $i$ must occupy $(v,t)$ (an \emph{include} constraint), while the other forbids it.  
A similar scheme applies to $(j, v, t)$ in the second child node.  
These complementary constraints ensure that the sets of candidate solutions for each child node are \emph{disjoint}, preventing repeated exploration of the same partial paths in multiple branches.

\subsubsection{Pseudocode for Enhanced CBS Variants}

We now present pseudocode snippets illustrating how each of these enhancement modules can be integrated into the baseline CBS of Algorithm~\ref{alg:cbs} (Section~\ref{sec:cbs:algorithm}).  
In each snippet, \textbf{additions} or \textbf{changes} compared to the baseline are enclosed in \fbox{\textbf{boxes}} for clarity.  
Although each module can be combined with the others in practice, we isolate them below for clarity.

\begin{algorithm}[ht]
\footnotesize
\caption{CBS with High-Level Admissible Heuristic (CBSH)}
\label{alg:cbs:hl-heuristic}
\begin{algorithmic}[1]
\Require Same inputs as in Algorithm~\ref{alg:cbs} (Sec.~\ref{sec:cbs:algorithm}), plus \fbox{\emph{ComputeHLHeuristic()}} function.
\State Initialize the root node $N_0$: 
   For each agent $i$, compute path $\pi_i$ ignoring collisions; all constraint sets are empty.
\State \fbox{\emph{Compute } $h(N_0) \!\!=\! \text{ComputeHLHeuristic}(\{\pi_i\});$}
\State Insert $N_0$ into open list, keyed by $f(N_0) = \mathrm{Cost}(\{\pi_i\}) + h(N_0)$.
\While{open list not empty}
  \State $N \gets$ pop node with smallest $f(N) = \mathrm{Cost}(N) + h(N)$
  \State Check for conflicts among $\{\pi_i\}$ in $N$.
  \If{no conflict}
    \State \Return $\{\pi_i\}$ (optimal solution)
  \Else
    \State Let $(i, j, t)$ be the first conflict.
    \For{each $a \in \{i, j\}$}
       \State Create $N'$ by copying $N$.
       \State Add constraint forbidding agent $a$ from the conflict at time $t$.
       \State Recompute $\pi_a$ for agent $a$ via low-level A*.
       \If{path found}
          \State \fbox{\emph{Compute } $h(N') \!\!=\! \text{ComputeHLHeuristic}(\{\pi_k\})$;}
          \State Insert $N'$ into open list with $f(N')=\mathrm{Cost}(N') + h(N')$.
       \EndIf
    \EndFor
  \EndIf
\EndWhile
\State \Return \textit{No feasible solution.}
\end{algorithmic}
\end{algorithm}

\noindent
\textbf{Explanation (CBSH).}  
The main change is in lines~2 and~16, which compute and maintain an \emph{admissible} HL heuristic, such as the MVC on a conflict graph (CG) or a pairwise dependency graph (DG/WDG).  
All other steps follow the baseline CBS logic.

\vspace{1em}

\begin{algorithm}[ht]
\footnotesize
\caption{CBS with Symmetry Reasoning (CBS-SR)}
\label{alg:cbs:symmetry}
\begin{algorithmic}[1]
\Require Same inputs as in Algorithm~\ref{alg:cbs}, plus \fbox{\emph{AddSymConstraint()}} function.
\State Initialize root node $N_0$ (same as baseline).
\State Insert $N_0$ into open list.
\While{open list not empty}
  \State $N \gets$ \emph{pop front of open list}
  \State Check for conflicts among $\{\pi_i\}$ in $N$.
  \If{no conflict}
    \State \Return $\{\pi_i\}$ (optimal solution)
  \Else
    \State Let $(i, j, t)$ be the first conflict.
    \State \textbf{Detect} \fbox{\emph{detectSymmetry}($i,j,t,\{\pi_i\}$)} 
    \If{a \emph{symmetry} is found (rectangle, corridor, \emph{etc.})}
       \State \fbox{\emph{AddSymConstraint}($i,j,t, N$)} \Comment{add specialized symmetry-breaking constraint}
       \State Recompute $\{\pi_i\}$ for the relevant agents
       \State Insert $N$ back into open list (updated constraints)
    \Else
       \For{each $a \in \{i, j\}$}
         \State Create $N'$ by copying $N$.
         \State Add standard CBS conflict constraint for agent $a$.
         \State Recompute $\pi_a$ for agent $a$.
         \If{path found}
            \State Insert $N'$ into open list
         \EndIf
       \EndFor
    \EndIf
  \EndIf
\EndWhile
\State \Return \textit{No feasible solution.}
\end{algorithmic}
\end{algorithm}

\noindent
\textbf{Explanation (CBS-SR).}  
Compared to Algorithm~\ref{alg:cbs}, lines~10--13 \textbf{(\emph{enclosed in boxes})} introduce new logic to detect symmetry conflicts and add \emph{symmetry-breaking constraints} (e.g., forbidding all permutations of a rectangle conflict).  
Only if symmetry is \emph{not} detected does the algorithm revert to the standard CBS branching in lines~15--24.

\vspace{1em}

\begin{algorithm}[ht]
\footnotesize
\caption{CBS with Mutex Propagation (CBS-Mutex)}
\label{alg:cbs:mutex}
\begin{algorithmic}[1]
\Require Same inputs as in Algorithm~\ref{alg:cbs}, plus \fbox{\emph{mutexCheckMDDs()}} subroutine.
\State Initialize root node $N_0$ with no constraints and unconstrained paths $\{\pi_i\}$.
\State \fbox{\emph{mutexCheckMDDs}($\{\pi_i\}$)} \Comment{remove mutually exclusive nodes from each agent's MDD}
\State Insert $N_0$ into open list.
\While{open list not empty}
  \State $N \gets$ \emph{pop front of open list}
  \State Check for conflicts among $\{\pi_i\}$ in $N$.
  \If{no conflict}
    \State \Return $\{\pi_i\}$ (optimal solution)
  \Else
    \State Let $(i,j,t)$ be the first conflict.
    \For{each $a \in \{i, j\}$}
       \State Create $N'$ by copying $N$.
       \State Add constraint forbidding agent $a$ from conflict.
       \State Recompute $\pi_a$ under updated constraints.
       \State \fbox{\emph{mutexCheckMDDs}($\{\pi_i\}$)} \Comment{prune infeasible MDD nodes again}
       \If{path found}
          \State Insert $N'$ into open list
       \EndIf
    \EndFor
  \EndIf
\EndWhile
\State \Return \textit{No feasible solution.}
\end{algorithmic}
\end{algorithm}

\noindent
\textbf{Explanation (CBS-Mutex).}  
Lines~2--3 and~16 \textbf{(\emph{enclosed in boxes})} call a \texttt{mutexCheckMDDs} routine that removes from each agent’s MDD any node that is mutually exclusive with another agent’s feasible states.  
This routine can be invoked after \emph{any} constraint update, typically in the low-level step.

\vspace{1em}

\begin{algorithm}[ht]
\footnotesize
\caption{CBS with Disjoint Splitting (CBS-DS)}
\label{alg:cbs:disjoint-splitting}
\begin{algorithmic}[1]
\Require Same inputs as in Algorithm~\ref{alg:cbs}.
\State Initialize $N_0$ with unconstrained paths
\State Insert $N_0$ into open list
\While{open list not empty}
  \State $N \gets \emph{pop front of open list}$
  \State Check for conflicts among $\{\pi_i\}$ in $N$
  \If{no conflict}
    \State \Return $\{\pi_i\}$ (optimal solution)
  \Else
    \State Let $(i, j, v, t)$ be the first conflict
    \For{each $a \in \{i, j\}$}
       \State Create $N'$ by copying $N$
       \State \fbox{\emph{(Disjoint Split)}:
       if \emph{Child 1} then add \emph{include} constraint $\langle a, v, t\rangle$else add \emph{exclude} constraint $\langle a, v, t\rangle$}
       \State Recompute $\pi_a$ for agent $a$
       \If{path found}
          \State Insert $N'$ into open list
       \EndIf
    \EndFor
  \EndIf
\EndWhile
\State \Return \textit{No feasible solution.}
\end{algorithmic}
\end{algorithm}

\noindent
\textbf{Explanation (CBS-DS).}  
Compared to the baseline branching (which forbids the conflict for either agent), lines~11--14 \textbf{(\emph{enclosed in boxes})} show a refined \emph{disjoint} approach: one child \emph{requires} agent $i$ (or $j$) at $(v,t)$, the other \emph{forbids} it. 
Each approach leads to distinct candidate solution sets without overlap.

\subsubsection{Summary of Optimal CBS Variants}

Each variant in Algorithms~\ref{alg:cbs:hl-heuristic}--\ref{alg:cbs:disjoint-splitting} modifies the high-level or low-level CBS operations to reduce redundant conflict exploration, prune infeasible states via multi-agent mutex detection, or unify symmetrical collision patterns. 
Crucially, these modules are largely orthogonal and can be combined for further performance gains (e.g., \emph{CBSH} plus \emph{Symmetry Reasoning} plus \emph{Disjoint Splitting}). 
Numerous implementations~\citep{boyarski2015icbs,felner2018adding,li2019improved,li2020new,li2021pairwise,zhang2022multi} confirm that such augmented CBS frameworks can optimally handle significantly larger MAPF instances than the plain baseline version, thus exemplifying how theoretical insights and well-designed heuristics can scale classical search-based methods in multi-agent pathfinding.

\begin{table*}[htb!]
\centering
\tiny
\renewcommand{\arraystretch}{1.2} 
\begin{threeparttable}
\resizebox{\textwidth}{!}{%
\begin{tabular}{lcccccccccccc}
\toprule
\multicolumn{1}{c}{\textbf{Papers}} & 
\multicolumn{1}{c}{\rotatebox{90}{\textbf{\begin{tabular}[c]{@{}l@{}}Prioritizing\\Conflicts\end{tabular}}}} &
\multicolumn{1}{c}{\rotatebox{90}{\textbf{\begin{tabular}[c]{@{}l@{}}Bypassing\\Conflicts\end{tabular}}}} & 
\multicolumn{1}{c}{\rotatebox{90}{\textbf{\begin{tabular}[c]{@{}l@{}}Rectangle\\Reasoning\end{tabular}}}} &
\multicolumn{1}{c}{\rotatebox{90}{\textbf{\begin{tabular}[c]{@{}l@{}}Generalized\\Rectangle\\Reasoning\end{tabular}}}} &
\multicolumn{1}{c}{\rotatebox{90}{\textbf{\begin{tabular}[c]{@{}l@{}}Target\\Reasoning\end{tabular}}}} &
\multicolumn{1}{c}{\rotatebox{90}{\textbf{\begin{tabular}[c]{@{}l@{}}Corridor\\Reasoning\end{tabular}}}} &
\multicolumn{1}{c}{\rotatebox{90}{\textbf{\begin{tabular}[c]{@{}l@{}}Corridor\\Target\\Reasoning\end{tabular}}}} &
\multicolumn{1}{c}{\rotatebox{90}{\textbf{\begin{tabular}[c]{@{}l@{}}Mutex\\Propagation\end{tabular}}}} &
\multicolumn{1}{c}{\rotatebox{90}{\textbf{\begin{tabular}[c]{@{}l@{}}Disjoint\\Splitting\end{tabular}}}} &
\multicolumn{1}{c}{\rotatebox{90}{\textbf{CG\tnote{1}}}} &
\multicolumn{1}{c}{\rotatebox{90}{\textbf{DG\tnote{2}}}} &
\multicolumn{1}{c}{\rotatebox{90}{\textbf{WDG\tnote{3}}}} \\
\midrule
\citet{boyarski2015icbs} & \ding{51} & & & & & & & & & & & \\
\citet{boyarski2015don} & & \ding{51} & & & & & & & & & & \\
\citet{felner2018adding} & & & & & & & & & & \ding{51} & & \\
\citet{li2019improved} & & & & & & & & & & & \ding{51} & \ding{51} \\
\citet{li2019symmetry} & & & \ding{51} & & & & & & & & & \\
\citet{li2019disjoint} & & & & & & & & & \ding{51} & & & \\
\citet{li2020new} & & & & & \ding{51} & \ding{51} & & & & & & \\
\citet{li2021pairwise} & & & & \ding{51} & & & \ding{51} & & & & & \\
\citet{zhang2022multi} & & & & & & & & \ding{51} & & & & \\
\bottomrule
\end{tabular}%
}
\begin{tablenotes}
\item[] 1. Conflict Graph; 2. Pairwise Dependency Graph; 3. Weighted Pairwise Dependency Graph.
\end{tablenotes}
\end{threeparttable}
\caption{Overview of pivotal variants of CBS algorithm. Checkmark \ding{51} indicates that the corresponding technique is employed in the given paper.}
\label{tab:cbs}
\end{table*}

\subsection{Suboptimal CBS Methods}\label{sec:cbs:suboptimal}

While optimal Conflict-Based Search (CBS) variants (Section~\ref{sec:cbs:optimal}) can guarantee the minimum possible solution cost, their computational requirements can escalate in large-scale or time-critical scenarios. 
To address these limitations, \textit{suboptimal} CBS extensions relax optimality in exchange for significantly improved runtime. 
In what follows, we first outline the archetypal suboptimal CBS framework---\textbf{Enhanced CBS} (ECBS)---as a reference point, providing pseudocode that highlights its differences from classical CBS. 
We then briefly survey several notable ECBS-based extensions, including EECBS~\citep{li2021eecbs}, FECBS~\citep{chan2022flex}, and ITA-ECBS~\citep{tang2024ita}. 
Each offers distinct strategies for distributing suboptimality factors (\(w\)), estimating costs in the high-level search, and coordinating how multiple agents share resources.

\subsubsection{Enhanced CBS (ECBS)}\label{sec:ecbs}

\paragraph{Overview and Key Ideas}
ECBS~\citep{barer2014suboptimal} introduces a suboptimality factor \(w \ge 1\) into both the high-level (HL) and low-level (LL) searches of classical CBS (Algorithm~\ref{alg:cbs}). 
Specifically,
\begin{itemize}[leftmargin=*]
    \item \emph{Low-Level Search Adjustment.} Each agent’s single-agent path planner (e.g., A*) is replaced or modified with a \emph{bounded-suboptimal} version (often referred to as a \emph{focal} search). For each agent~\(i\), the planner constructs an individual path \(\pi_i\) satisfying 
    \[
       \mathrm{Cost}(\pi_i) \;\le\; w \;\cdot\; \mathrm{Cost}(\pi_i^{\text{opt}}),
    \]
    where \(\mathrm{Cost}(\pi_i^{\text{opt}})\) is the agent’s true single-agent optimal cost. 

    \item \emph{High-Level Focal Search.} At the HL level, ECBS maintains two priority queues: 
    \begin{enumerate}[leftmargin=*]
        \item \textbf{OPEN}, sorted by the sum of \emph{lower-bound} path costs (often computed via an admissible heuristic on the single-agent path length). 
        \item \textbf{FOCAL}, containing all nodes from \textbf{OPEN} whose sum of lower-bound costs is within a multiplicative factor \(w\) of the minimum in \textbf{OPEN}. The queue \textbf{FOCAL} is then sorted by a user-defined \emph{focal heuristic} (e.g., the \emph{sum of actual costs} or some conflict-based metric).
    \end{enumerate}
    The node popped from \textbf{FOCAL} (rather than \textbf{OPEN}) is the one expanded at each iteration. If a solution node \(N_{\text{sol}}\) is returned, its final cost is guaranteed to be at most \(w\) times the optimal cost.
\end{itemize}

In this way, ECBS systematically explores a \emph{relaxed} search space guided by a suboptimality bound, often achieving significant speed-ups compared to optimal CBS. 
Algorithm~\ref{alg:ecbs} highlights the changes from the baseline CBS pseudocode (Algorithm~\ref{alg:cbs} in Section~\ref{sec:cbs:algorithm}). 
Lines and logic added or modified for ECBS are \textbf{enclosed in \fbox{boxes}}.

\begin{algorithm}[ht]
\footnotesize
\caption{ECBS for Suboptimal MAPF}
\label{alg:ecbs}
\begin{algorithmic}[1]
\Require Graph $\mathcal{G}=(\mathcal{V},\mathcal{E})$, agents $\{1,\dots,n\}$, start vertices $\{s_i\}$, goal vertices $\{g_i\}$, suboptimality bound $w \ge 1$.
\State Initialize the root node $N_0$:
\begin{itemize}
  \item \textbf{For each agent $i$:} run a \fbox{\emph{bounded-suboptimal A*}} (or a focal search) to compute a path $\pi_i$ with cost at most $w \cdot \mathrm{Cost}(\pi_i^{\text{opt}})$.  
  \item Set constraint sets to empty for all agents.
\end{itemize}
\State \fbox{Compute a lower-bound cost $L(N_0)$ as the sum of per-agent lower bounds ($L_i$).}
\State \fbox{Set OPEN = \{\}, FOCAL = \{\}. Insert $N_0$ into OPEN with key $\text{key}_\text{OPEN}(N_0) = L(N_0)$.}
\State \fbox{Let $L_{\min} = L(N_0)$ and define the \emph{focal threshold} $f_\text{th} = w \cdot L_{\min}$.}
\State \fbox{Populate FOCAL with all nodes $N$ in OPEN satisfying $L(N)\le f_\text{th}$.}
\State \fbox{Sorting them by a secondary heuristic (e.g., actual sum of costs).}
\While{\textbf{FOCAL} not empty}
  \State \fbox{$N \gets$ pop the node in FOCAL with the smallest focal-heuristic.}
  \State Check for conflicts among the paths $\{\pi_i\}$ in $N$.
  \If{no conflict}
    \State \Return $\{\pi_i\}$ \quad \Comment{Bounded-suboptimal solution.}
  \Else
    \State Let $(i, j, t)$ be the first detected conflict.
    \For{each $a \in \{i, j\}$}
       \State Create $N'$ by copying $N$.
       \State Add a conflict constraint for agent $a$ at time $t$.
       \State \fbox{Replan $\pi_a$ with \emph{bounded-suboptimal A*} to satisfy the new constraint.}
       \If{a new path is found}
          \State \fbox{$L(N') \gets \sum_i L_i(N')$ \quad (sum of agent-level lower-bound costs).}
          \State Insert $N'$ into OPEN with key $\text{key}_\text{OPEN}(N') = L(N')$.
       \EndIf
    \EndFor
    \State \fbox{Update $L_{\min}$ as $\min_{N\in \text{OPEN}} L(N)$ and recompute $f_\text{th} = w \cdot L_{\min}$.}
    \State \fbox{FOCAL $\gets \{\,N \in \text{OPEN}\,\mid\, L(N)\le f_\text{th}\}$, sorted by focal-heuristic.}
  \EndIf
\EndWhile
\State \Return \textit{No feasible solution under suboptimal CBS.}
\end{algorithmic}
\end{algorithm}

\noindent\textbf{Algorithm Explanation.}  
\begin{itemize}[leftmargin=*]
    \item \emph{Line\,1--3.} Agents compute initial \emph{bounded-optimal} (or \emph{focal}) single-agent paths. A lower bound \(L_i\) on each path cost is stored, summing to \(L(N_0)\). 
    \item \emph{Lines\,4--6.} ECBS manages two queues: \textbf{OPEN} (sorted by \emph{lower-bound} sum of costs) and \textbf{FOCAL} (containing nodes whose lower-bound sum is within a factor \(w\) of the smallest in \textbf{OPEN}). 
    \item \emph{Line\,10--22.} Conflict handling proceeds similar to classical CBS, except that \emph{bounded-suboptimal} single-agent search is used for replanning agent \(a\). 
    \item \emph{Lines\,23--25.} After generating new children, ECBS recalculates \(L_{\min}\) from \textbf{OPEN}, updates the \emph{focal threshold} \(f_\text{th}\), and refills \textbf{FOCAL} accordingly.
\end{itemize}

\subsubsection{Variants of ECBS}

\paragraph{Explicit Estimation CBS (EECBS).}  
EECBS~\citep{li2021eecbs} incorporates additional heuristics in the high-level search to estimate inadmissible (but more informed) path costs. Specifically, it leverages an \emph{online learning} approach to refine these estimates during node expansion. In practice, EECBS often reduces the search-tree size compared to vanilla ECBS, because it can more effectively prioritize nodes likely to yield feasible solutions.

\paragraph{Flexible ECBS (FECBS).}  
FECBS~\citep{chan2022flex} relaxes the assumption that \emph{one global} factor \(w\) must govern suboptimality uniformly across all agents. Instead, it allows individual agents to have distinct suboptimality margins based on environmental constraints, agent importance, or resource availability. This \emph{asymmetric suboptimality} broadens the set of solvable instances within tight time budgets, making FECBS particularly relevant to real-world settings where tasks exhibit varying priorities.

\paragraph{ITA-ECBS.}  
ITA-ECBS~\citep{tang2024ita} integrates \emph{target-assignment optimization} into the suboptimal CBS framework. Beyond searching for collision-free paths, it simultaneously refines how \emph{goal vertices} are distributed among agents, governed by a suboptimal factor \(w\). Experimental results suggest that ITA-ECBS can outperform standard ECBS-TA~\citep{barer2014suboptimal} across a broad range of metrics, highlighting the benefits of coupling suboptimal MAPF with target allocation in a single search procedure.

\subsubsection{Comparison of Optimal vs.\ Suboptimal CBS}

Table~\ref{tab:cbs-opt-subopt} summarizes the relationship between classical \emph{optimal} CBS algorithms (Section~\ref{sec:cbs:optimal}) and \emph{suboptimal} CBS extensions. While optimal variants can be significantly improved with heuristics (e.g., conflict prioritization, symmetry breaking, mutex propagation), they still must expand or branch whenever needed to guarantee the minimal possible cost. By contrast, suboptimal approaches (ECBS, EECBS, FECBS, ITA-ECBS) permit bounding the solution cost by \(w\) times the optimum, gaining substantial computational savings in many instances.

\begin{table}[ht]
\centering
\footnotesize
\caption{High-Level Comparison of Optimal and Suboptimal CBS Varieties}
\label{tab:cbs-opt-subopt}
\begin{tabular}{p{0.26\linewidth}p{0.29\linewidth}p{0.34\linewidth}}
\toprule
\bf Method & \bf Search Strategy & \bf Characteristics \\ 
\midrule
\multicolumn{3}{l}{\emph{Optimal CBS Variants (Sec.~\ref{sec:cbs:optimal})}} \\
\midrule
Plain CBS \citep{sharon2015conflict} & Standard HL + LL recursion & No suboptimal factor; guaranteed minimal cost. \\
CBS with HL heuristics & Admits conflict-based or pairwise dependency heuristics & Reduces expansions by focusing on potential cardinal conflicts first. \\ 
CBS with Symmetry \& Mutex reasoning & Adds specialized constraints and mutual-exclusion checks & Aggressively prunes search space to handle repeated collisions.\\
\midrule
\multicolumn{3}{l}{\emph{Suboptimal CBS Variants (Sec.~\ref{sec:cbs:suboptimal})}} \\
\midrule
ECBS \citep{barer2014suboptimal} & Focal search at both HL and LL with factor $w$ & Solutions cost at most $w \cdot \mathrm{Cost}(\text{opt})$, typically much faster in practice. \\
EECBS \citep{li2021eecbs} & ECBS + explicit cost estimation at HL & Learns better heuristics online, often fewer expansions.\\
FECBS \citep{chan2022flex} & ECBS + flexible per-agent $w$ & Adapts suboptimal factors individually, more practical under diverse agent constraints. \\
ITA-ECBS \citep{tang2024ita} & ECBS with integrated target assignment & Jointly solves target assignment + suboptimal path planning, often outperforms ECBS-TA. \\
\bottomrule
\end{tabular}
\end{table}

\noindent
From a theoretical standpoint, suboptimal CBS solutions retain \emph{completeness} and offer bounded suboptimality, yet forego the strict cost guarantees of optimal methods. Empirically, they can tackle significantly larger or more complex MAPF scenarios within limited compute budgets, thus bridging the gap between purely optimal solutions and real-time industrial deployment.

\subsection{Priority-Based Search (PBS)}\label{sec:pbs}

\begin{figure}[htb!]
    \centering
    \includegraphics[width=\linewidth]{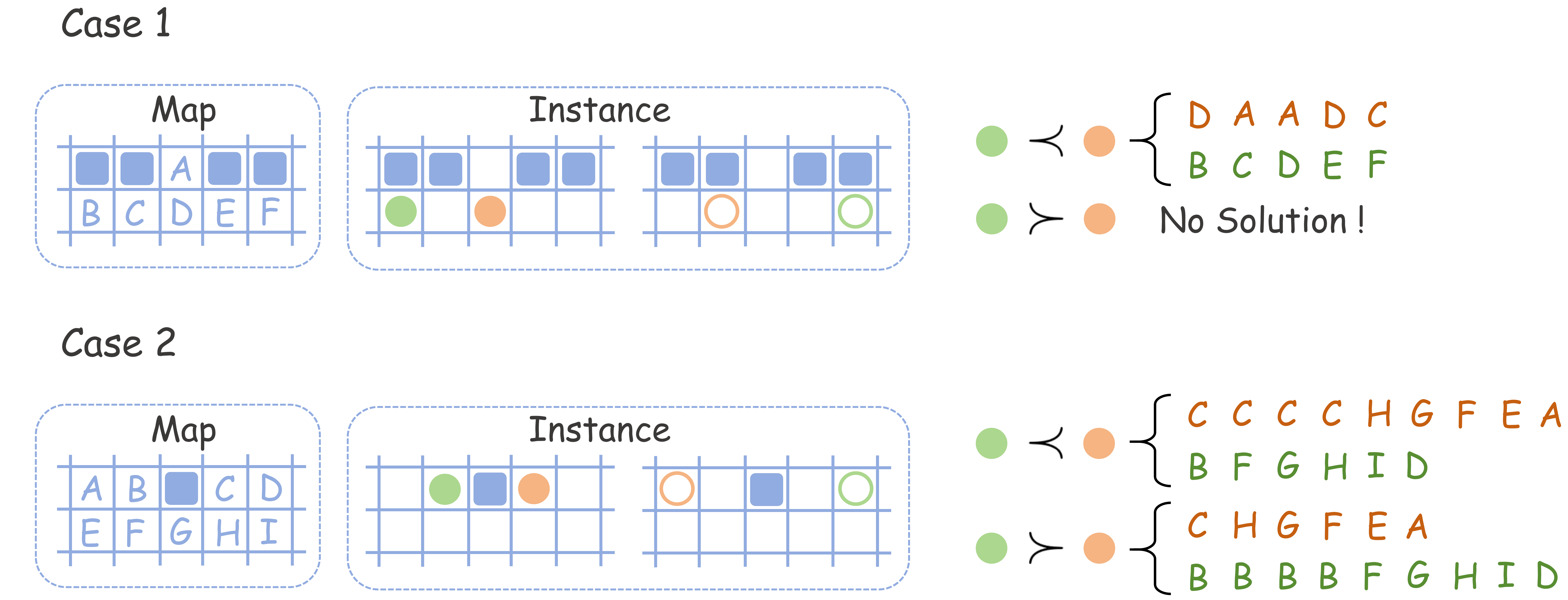}
    \caption{Illustration of how agent‐priority ordering affects the outcome of prioritized planning in PBS. Case 1 (top row): Map \& Instance (left): a 2 $\times$ 5 grid with obstacles (blue squares), nodes labeled A–F, and two agents (green and orange) whose sources are shown by solid circles and whose targets are shown by hollow circles. Results (right): we compare two priority orders for planning—green $\prec$ orange (upper branch) versus orange $\succ$ green (lower branch). Under the green‐first ordering, we obtain conflict‐free paths shown in the figure, whereas the orange‐first ordering yields no feasible solution, demonstrating that a "wrong" priority can break completeness. Case 2 (bottom row): Map \& Instance (left): a different 2 $\times$ 5 grid (nodes A–I), with sources and targets similarly annotated. Results (right): here both priority orders succeed, but produce markedly different path sequences.}
    \label{fig:pbs}
\end{figure}

Following our in-depth review of Conflict-Based Search (CBS) in Section~\ref{sec:cbs}, we now turn to an alternative yet closely related search-based paradigm for MAPF known as \emph{Priority-Based Search} (PBS). While CBS pursues a multi-agent search guided by conflicts and constraints, PBS tackles the collision-avoidance challenge by imposing a (partial) \emph{priority ordering} on agents, such that each agent plans its path while \emph{respecting} the fixed  (i.e. \emph{Prioritized Planning}, PP~\citep{erdmann1987multiple}; \emph{First Come First Served}, FCFS~\citep{dresner2004multiagent}) or dynamically  (i.e. \emph{Safe Interval Path Planning}, SIPP~\citep{phillips2011sipp}) assigned higher-priority agents’ paths.
The illustration of how agent-priority ordering affects the outcome of prioritized planning in
PBS is shown in~\ref{fig:pbs}.

This subsection mirrors the structure of Section~\ref{sec:cbs}: we begin by introducing the mathematical modeling that underlies PBS, then discuss the fundamental PBS algorithmic framework (comparable to “vanilla CBS”), examine notable enhancements (including dynamic priority inheritance and merging), and finally contrast PBS with CBS through a comparative lens.

\subsubsection{Mathematical Modeling for PBS and Its Variants}
\label{sec:pbs:modeling}

As in the MAPF definitions of Section~\ref{sec:formulation} and the constraint-based formulation for CBS (Section~\ref{sec:cbs}), we let:
\[
   \mathcal{G} = (\mathcal{V}, \mathcal{E}), 
   \quad
   n\text{ agents } \{1,\dots,n\},
   \quad
   s_i, g_i \in \mathcal{V},
\]
represent the environment and agent setup. We also adopt a discrete time horizon $T$, with agent~$i$’s path $\,\pi_i = \bigl(v_i(0), v_i(1), \dots, v_i(T)\bigr)$ or, equivalently, the binary indicators $x_{i,v,t}\in\{0,1\}$ marking whether agent~$i$ occupies vertex $v$ at time $t$. 

Unlike CBS, which relies on conflict-tree branching when collisions are found, PBS systematically assigns each agent a \emph{priority level}:
\[
   \mathsf{P}:\{1,\dots,n\}\;\rightarrow\;\{1,\dots,n\},
\]
such that a \emph{lower numeric value} of $\mathsf{P}(i)$ indicates \emph{higher} priority. The core principle then becomes:
\[
   \text{When agent $i$ plans or updates its path, it must remain collision-free w.r.t.\ all agents $j$ for which } \mathsf{P}(j) < \mathsf{P}(i).
\]
Mathematically, we incorporate constraints:
\begin{subequations}\label{eq:pbs:constraint}
\begin{align}
   &x_{i,v,t} + x_{j,v,t} \;\le\; 1 
   \quad \text{for all }t, \,\forall v, \text{ only if } \mathsf{P}(j)<\mathsf{P}(i), \label{eq:pbs:vertex}\\[4pt]
   &x_{i,u,t} + x_{i,v,t+1} + x_{j,v,t} + x_{j,u,t+1}
     \;\le\; 3
   \quad \text{for all }(u,v), \text{ only if } \mathsf{P}(j)<\mathsf{P}(i), \label{eq:pbs:edge}
\end{align}
\end{subequations}
meaning that agent~$i$ must avoid vertex and edge collisions with higher-priority agents but need not explicitly enforce constraints against lower-priority agents (those with $\mathsf{P}(j)>\mathsf{P}(i)$). 

Consistent with Section~\ref{sec:cbs}, we can adopt various objectives:
\begin{itemize}[leftmargin=*]
   \item \textbf{Sum of Costs (SoC).}  Minimize $\sum_{i=1}^n \mathrm{Cost}(\pi_i)$, adding only the collision constraints for higher-priority agents. 
   \item \textbf{Makespan.}  Minimize $\max_{1\le i\le n}\mathrm{Cost}(\pi_i)$ under the same one-sided collision-avoidance scheme.
\end{itemize}
Alternatively, in \emph{lifelong} settings, each agent repeatedly receives new tasks while \emph{still} respecting the paths of currently higher-priority neighbors.

\subsubsection{Algorithmic Framework}\label{sec:pbs:framework}

\paragraph{Overview.}  
In \emph{vanilla PBS}, one fixes a total ordering
\[
   \mathsf{P}(1) < \mathsf{P}(2) < \dots < \mathsf{P}(n),
\]
and iterates through agents in ascending priority. Each agent’s path is found via standard single-agent pathfinding (e.g., A*) on a time-expanded graph that marks all space-time cells occupied by \emph{higher-priority} agents as forbidden. Because the lower-priority agents have not yet planned, they impose no constraints on the current agent’s route. Hence, any collisions that might arise with same- or lower-priority agents are simply left to be resolved when those agents plan their paths (i.e., they must themselves detour around the higher-priority path). Algorithm~\ref{alg:pbs:vanilla} outlines the process for a single pass from highest- to lowest-priority agent.

\begin{algorithm}[ht]
\caption{Vanilla Priority-Based Search (PBS)}
\label{alg:pbs:vanilla}
\footnotesize
\begin{algorithmic}[1]
\Require Graph $\mathcal{G}=(\mathcal{V},\mathcal{E})$, $n$ agents each with $(s_i, g_i)$, total priority order $\mathsf{P}(1)<\dots<\mathsf{P}(n)$.
\State For $i=1$ to $n$ (in ascending priority):
   \begin{itemize}
   \item Identify higher-priority agents $\{j \mid \mathsf{P}(j)<\mathsf{P}(i)\}$.
   \item Treat their chosen paths $\{\pi_j\}$ as static obstacles in time-space.
   \item Compute $\pi_i$ via single-agent pathfinding subject to these obstacles.
   \item If no path is found, \textbf{report} \emph{infeasibility} under the current priority order.
   \end{itemize}
\State \textbf{Return} all paths $\{\pi_1,\dots,\pi_n\}$ once computed.
\end{algorithmic}
\end{algorithm}

\noindent
\textbf{Comparison with CBS.}  In CBS (Algorithm~\ref{alg:cbs}), \emph{all} agents have partial paths that are refined by \emph{conflict constraints} whenever collisions occur. By contrast, \emph{vanilla PBS} never replans a \emph{higher}-priority agent’s path. Once assigned, it remains fixed; collisions \emph{must} be handled by lower-priority agents. This approach can be implemented quickly and scales to large $n$ in many practical settings. Its main drawback is that a poor priority ordering may drastically inflate total path costs or lead to unsolvable constraints for the lower-priority agents.

\paragraph{Dynamic or Partial Orderings.} 
To improve beyond the naive fixed ordering, one can adopt a conflict-driven \emph{branching on partial orders}~\citep{ma2019searching}. Analogous to the branching in CBS, each time a collision arises between agents $i$ and $j$ of \emph{undetermined} relative priority, the search forks into two CT nodes: one sets $i \prec j$, the other $j \prec i$. Each child node then plans or replans the relevant agent’s path accordingly. The search thus explores only partial orderings that are necessary to resolve collisions, often reducing overhead compared to enumerating all $n!$ permutations.

\subsubsection{Illustrative Example}

Consider a $2\times 3$ grid (as in Figure~\ref{fig:cbs-example-grid} for CBS), with two agents:
\[
   \text{Agent 1: }s_1=(0,0)\to g_1=(1,2), 
   \quad
   \text{Agent 2: }s_2=(1,0)\to g_2=(0,2).
\]
Assume $\mathsf{P}(1)<\mathsf{P}(2)$, i.e., Agent~1 has higher priority. Then:
\begin{enumerate}[leftmargin=*, itemsep=1pt, parsep=1pt]
   \item \textbf{Plan $\pi_1$:} Agent~1 finds a direct route $(0,0)\!\to\!(0,1)\!\to\!(0,2)\!\to\!(1,2)$. 
   \item \textbf{Plan $\pi_2$:} With $\pi_1$ fixed, Agent~2 sees that $(0,2)$ and $(1,2)$ are occupied or transitioning at times $t=2,3,\dots$. Hence, it must detour, e.g., $(1,0)\!\to\!(1,1)\!\to\!(1,1)\!\to\!(1,2)\!\dots$ or wait further. 
\end{enumerate}
If $\mathsf{P}(1)$ was chosen poorly, i.e., the direct path for Agent~1 might block Agent~2 for a long time. Still, by design, Agent~2 shoulders the entire collision-avoidance effort. This can be either advantageous (e.g., if Agent~1 is truly more critical) or suboptimal for system-level objectives.

\subsubsection{Enhancing PBS: Dynamic Priorities, Inheritance, and Merging}

Over the years, a variety of PBS enhancements have emerged, analogous to the “optimal” and “suboptimal” CBS variants in Sections~\ref{sec:cbs:optimal} and \ref{sec:cbs:suboptimal}. Here, we focus on three main families of techniques:

\vspace{2mm}
\noindent
\textbf{(1) Priority Inheritance with Backtracking (PIBT / winPIBT).}
Originally proposed by \citet{okumura2022priority}, \emph{PIBT} operates \emph{one time-step at a time}. It dynamically reassigns priorities whenever a higher-priority agent becomes blocked by a lower-priority agent in the next step. The blocked low-priority agent \emph{inherits} the higher priority and attempts to move out of the way, allowing for local deadlock resolution without globally changing the entire priority order.

\begin{itemize}[leftmargin=*]
   \item \textbf{winPIBT}~\citep{okumura2019winpibt} generalizes this to a \emph{$w$-step windowed} approach, planning a short horizon of $w>1$ steps for each agent at each iteration. The added foresight helps mitigate repeated short-sighted collisions.
   \item Both PIBT and winPIBT guarantee \emph{eventual} goal reachability on certain classes of graphs (e.g., biconnected or cycle-rich networks), even though local re-planning can lead to suboptimal global cost.
\end{itemize}

\vspace{1mm}
\noindent
\textbf{(2) Conflict-Driven PBS with Merging.}
\citet{ma2019searching} unify the ideas of partial-order branching and \emph{merging} proposed in the CBS context~\citep{boyarski2022merging}. When the search detects repeated collisions between agents (or groups of agents), it merges them into a single \emph{meta-agent}, which is then planned jointly on the time-expanded graph. This can drastically reduce repeated collisions among those agents, but it \emph{increases} the complexity of their internal pathfinding. As in CBS merging, a careful choice of which agents to merge can yield a significant speed-up in practice.

\vspace{1mm}
\noindent
\textbf{(3) Game-Theoretic Incentive (Mechanism Design).}
\citet{friedrich2024scalable} propose a \emph{mechanism-design} approach, ensuring that agents have no incentive to defect or misrepresent their costs. They fix the priority order (or partial order) in a manner independent of agent reports, then apply approximate MAPF algorithms (like a suboptimal PBS) to assign collision-free paths. Agents pay a \emph{VCG-like} tax that reflects the externality they impose on others. Such strategyproof PBS frameworks can handle thousands of agents while retaining game-theoretic guarantees.

\vspace{1mm}
\begin{algorithm}[ht]
\footnotesize
\caption{PBS with Dynamic Priority Inheritance (Step-by-Step Per Timestep)}
\label{alg:pbs:dynamic}
\begin{algorithmic}[1]
\Require Graph $\mathcal{G}$, $n$ agents, \emph{initial} priority order $\mathsf{P}_0$, time horizon $T$.
\State \textbf{Initialize:} For each agent $i$, set $\pi_i(0) = s_i$. 
\For{$t=0 \to T-1$}
   \State Let $\Pi_{<t+1} = \{\pi_i(\tau)\mid \tau\le t\}$ \Comment{All prior commitments}
   \State Sort agents by current priorities, from highest to lowest.
   \For{each agent $i$ in descending priority}
       \State $\textbf{PlanStep}(\pi_i[t+1]\mid \Pi_{<t+1};\ \mathsf{P}, \mathcal{G})$ 
             \Comment{Pick next location avoiding collisions with 
               higher-priority paths.}
       \If{blocked by a lower-priority agent $j$}
          \State \textbf{priority-inherit: } $\mathsf{P}(j)\gets \mathsf{P}(i)-\epsilon$ 
                \Comment{agent $j$ temporarily inherits higher priority}
          \State $\textbf{ReplanStep}(j)$ 
                \Comment{agent $j$ tries to move first, unblocking $i$}
          \If{still no feasible step}
             \State \emph{backtrack $(i,j)$} 
                   \Comment{both try an alternate local step}
          \EndIf
       \EndIf
   \EndFor
   \If{all agents at $g_i$ or no valid moves possible}
      \State \textbf{break}
   \EndIf
\EndFor
\State \Return $\{\pi_i(\tau)\} \quad \text{(one-step-at-a-time solution)}$
\end{algorithmic}
\end{algorithm}

\noindent
\textbf{Algorithm~\ref{alg:pbs:dynamic}:}  
Illustrates a possible stepwise planning routine used by PIBT-like methods. In each time-step iteration, agents update their next move in priority order. If a higher-priority agent is blocked, it triggers \emph{priority inheritance}, letting the lower-priority agent move first. Such a scheme ensures local progress in many congested environments. Notably, the final solution is not guaranteed to be \emph{globally} cost-optimal.

\subsubsection{PBS Variants: Detailed Pseudocode and Comparisons with the Vanilla Algorithm}

In Section~\ref{sec:pbs:framework}, we presented the \emph{vanilla} PBS procedure in Algorithm~\ref{alg:pbs:vanilla}, illustrating how agents plan in ascending priority order.  We also discussed three major enhanced PBS families: (i) \emph{Priority Inheritance with Backtracking (PIBT-like)}, (ii) \emph{Conflict-Driven PBS with Merging}, and (iii) \emph{Mechanism-Design-based PBS}.  Previously, Algorithm~\ref{alg:pbs:dynamic} provided a stepwise \emph{PIBT-like} approach.  We now complement that with pseudocode for the other two methods, highlighting their key differences from Algorithm~\ref{alg:pbs:vanilla} (the base version).  In each pseudocode, we mark the *modified or additional lines* in a \fbox{\textbf{box}} and provide line-by-line explanations.

\paragraph{(1) Recap: Base PBS Algorithm (Algorithm~\ref{alg:pbs:vanilla}).}
For ease of reference, we restate the base pseudocode:

\begin{algorithm}[ht]
\caption{\textbf{Base PBS}: Vanilla Priority-Based Search (review)}
\label{alg:pbs:vanilla-recap}
\footnotesize
\begin{algorithmic}[1]
\Require Graph $\mathcal{G}=(\mathcal{V},\mathcal{E})$, $n$ agents each with $(s_i, g_i)$, total priority order $\mathsf{P}(1)<\cdots<\mathsf{P}(n)$.
\Statex \underline{\emph{Single pass in ascending priority:}}
\For{$i \gets 1$ to $n$}
  \State Let $\mathcal{X} \gets \{\pi_j \mid \mathsf{P}(j) < \mathsf{P}(i)\}$ 
    \Comment{\emph{Paths of higher-priority agents}}
  \State $\pi_i \gets \textsc{SingleAgentPathPlanner}(s_i,\ g_i,\ \mathcal{X})$ 
    \Comment{\emph{Avoid collisions with $\mathcal{X}$.}}
  \If{\textsc{NoPathFound}($i$)}
    \State \textbf{return} ``\emph{Infeasible under current ordering}''
  \EndIf
\EndFor
\State \textbf{return} $\{\pi_1,\ldots,\pi_n\}$ 
\Comment{\emph{Collision-free w.r.t.\ higher-priority paths}}
\end{algorithmic}
\end{algorithm}

\noindent
\textbf{Algorithm~\ref{alg:pbs:vanilla-recap} Explanation (Lines 1--9):}
1) We fix a total ordering, iterating agents from highest priority ($i=1$) to lowest ($i=n$).  
2) When planning agent $i$’s path, the paths for all $j$ with $\mathsf{P}(j) < \mathsf{P}(i)$ are treated as \emph{spatio-temporal obstacles}.  
3) If agent $i$ cannot find any path that avoids higher-priority obstacles, the algorithm declares failure under that ordering.  
4) Otherwise, we produce a set of final paths in which each lower-priority agent yields to all higher-priority ones.

\paragraph{(2) Conflict-Driven PBS with Merging.}
In \emph{conflict-driven} PBS, the priority order is \emph{not} fully fixed a priori. Instead, we discover necessary order constraints \emph{on the fly}: whenever two agents $i$ and $j$ collide, we branch by forcing either $i \prec j$ or $j \prec i$.  This procedure can be enriched with \emph{merging}~\citep{boyarski2022merging,ma2019searching}: after repeated collisions between the same pair (or set) of agents, we merge them into a single \emph{meta-agent}, which is then planned collectively.  Algorithm~\ref{alg:pbs:conflict-merge} shows the high-level approach; lines with a bold box \fbox{\textbf{...}} are \emph{modified/additional steps} relative to the \emph{base PBS} logic in Algorithm~\ref{alg:pbs:vanilla-recap}.

\begin{algorithm}[ht]
\caption{Conflict-Driven Partial-Order PBS with Merging}
\label{alg:pbs:conflict-merge}
\footnotesize
\begin{algorithmic}[1]
\Require Graph $\mathcal{G}=(\mathcal{V},\mathcal{E})$, $n$ agents each with $(s_i,g_i)$, \fbox{\textbf{no final total priority given initially}}.
\State Initialize a node $N_0$. 
\State \fbox{\textbf{For each agent $i$, compute an unconstrained path $\pi_i$, ignoring collisions.}}
\State \fbox{\textbf{Set partial-order $\prec_{N_0}$ to empty (no constraints among agents).}}
\State Insert $N_0$ into an \textsc{OpenList} or \textsc{Queue}.
\vspace{1mm}
\While{OpenList is not empty}
  \State $N \gets$ pop front of OpenList
  \State Check collisions among $\{\pi_i\}$ in $N$.
  \If{no collision}
    \State \textbf{return} $\{\pi_i\}$ as a valid plan \Comment(\emph{partial order is fully consistent})
  \EndIf
  \State Let $(i,j)$ be a collision in $N$.
  \If{\fbox{\textbf{exceeds merge threshold (e.g.\ repeated collisions)}}}
    \State \fbox{\textbf{Merge}($i,j$) \textbf{into meta-agent} $M_{ij}$} 
           \Comment(\emph{\textbf{Key difference from base PBS}})
    \State Recompute a joint path $\pi_{M_{ij}}$ for $M_{ij}$ avoiding collisions with other agents
    \State \textbf{Insert} $N$ back into OpenList with updated $\{\pi_{M_{ij}}, \pi_k\}$ 
  \Else
    \For{each ordering choice in $\{\,i \prec j,\ j \prec i\}$}
      \State $N' \gets$ clone of $N$
      \State \fbox{\textbf{Augment partial-order}} $\prec_{N'}$ by adding $i \prec j$ or $j \prec i$
      \State Replan path(s) for whichever agent(s) must yield to the newly enforced order
      \State Insert $N'$ into OpenList
    \EndFor
  \EndIf
\EndWhile
\State \textbf{return} ``\emph{No feasible partial order found}''
\end{algorithmic}
\end{algorithm}

\paragraph{Explanation of Algorithm~\ref{alg:pbs:conflict-merge}:}
\begin{enumerate}[leftmargin=*]
  \item \textbf{Initialization (Lines 1--3).}  Compare to \emph{base PBS}, we do \emph{not} specify a total priority from the outset.  Each agent is given an unconstrained path, ignoring collisions.
  \item \textbf{Collision Check and Branching (Lines 7--10, 14--18).}  On detecting a collision between agents (or meta-agents) $i$ and $j$, the algorithm can \emph{branch} into two child nodes: one forcing $i\prec j$, the other $j\prec i$.  In each child, we replan the yield-to-$\prec$ agent(s).
  \item \textbf{Merging (Lines~11--13).}  If $i$ and $j$ have collided repeatedly or exceed some threshold, we \emph{merge} them into a new meta-agent $M_{ij}$.  We then re-find a single path for $M_{ij}$ (i.e., planning all agents in $M_{ij}$ simultaneously).  This “solves collisions inside $M_{ij}$” cheaply but yields a bigger subproblem for $M_{ij}$’s next collisions with other agents.
  \item \textbf{Loop until solution or exhausted (Lines~4--5,19).}  The partial order $\prec_N$ grows as collisions arise. Eventually, if we find a node with no collisions among the updated paths, that solution is a valid plan that respects the partial-order constraints.
\end{enumerate}

\noindent
\textbf{Differences from Base PBS.}  Instead of committing to a single pass in ascending order, we \emph{split} or \emph{merge} whenever collisions appear.  This approach can produce solutions that the fixed-order base PBS might fail to find easily, and can sometimes deliver better paths or prove certain cost bounds (though the latter is still typically suboptimal relative to “optimal CBS”).

\vspace{2em}

\paragraph{(3) Mechanism-Design-Based PBS (Strategyproof Priority Ordering).}
In real-world applications involving distinct stakeholders or self-interested agents, a predetermined or conflict-driven priority might be \emph{manipulated} by agents lying about their costs or start/goal times to gain better routes.  \citet{friedrich2024scalable} propose a \emph{mechanism design} approach, ensuring that the priority assignment is \emph{independent of agents’ reported data}.  One can then apply an approximate (or even a simple single-pass) PBS routine to allocate collision-free paths, while charging each agent a \emph{payment} reflecting the externality they impose.  Algorithm~\ref{alg:pbs:mechanism} illustrates a \emph{high-level} procedure.

\begin{algorithm}[ht]
\caption{Mechanism-Design PBS (Strategyproof Priority Allocation)}
\label{alg:pbs:mechanism}
\footnotesize
\begin{algorithmic}[1]
\Require Graph $\mathcal{G}$, $n$ agents with $(s_i,g_i)$, each agent’s “(possibly reported) cost/time values” $(c_i,\dots)$, but \fbox{\textbf{a fixed random or external ordering}} $\mathsf{P}$ that does not depend on $c_i$.
\State \fbox{\textbf{Assign priorities} $\mathsf{P}(1)<\cdots<\mathsf{P}(n)$ \textbf{from a source independent of agents’ data}} 
       \Comment(e.g.\ random shuffle)
\State \textbf{Run PBS} with the chosen $\mathsf{P}$ to generate $\{\pi_i\}$ 
       \Comment(see Algorithm~\ref{alg:pbs:vanilla-recap})
\State Let $\mathrm{Cost}(\pi_i)$ be the (suboptimal) path cost for agent $i$, 
       and $v_i$ the agent’s reported \emph{value} for achieving $g_i$.
\State $W(\{\pi_i\}) \;\gets\;\sum_{i=1}^n \bigl[v_i - c_i\cdot \mathrm{Cost}(\pi_i)\bigr]^{+}$
       \Comment($[\cdot]^+$ means $\max\{0,\cdot\}$)
\vspace{1pt}
\For{each agent $i$}
  \State \fbox{\textbf{Compute externality}}: 
        $W_{-i}(\{\pi_i\}) = \sum_{j\neq i} [\,v_j - c_j\,\mathrm{Cost}(\pi_j)\,]^{+}.$
  \State \fbox{\textbf{Define agent $i$’s payment}}
       \[
         p_i \;\;=\;\;
         \max_{\,\mathsf{P}_\text{all}\,\text{valid}} 
            \Bigl\{\,W_{-i}\bigl(\text{PBS}(\mathsf{P}_\text{all})\bigr)\!\Bigr\}
         \;\;-\;\;
         W_{-i}\bigl(\{\pi_i\}\bigr)
         \,.
       \]
       \Comment(\emph{like a VCG tax, subject to the range of priorities})
\EndFor
\State \textbf{return} $(\{\pi_i\},\{p_i\})$ where $p_i$ is each agent’s charge.
\end{algorithmic}
\end{algorithm}

\paragraph{Explanation of Algorithm~\ref{alg:pbs:mechanism}:}
\begin{itemize}[leftmargin=*]
  \item 
  \underline{Lines 1--2: \emph{Exogenous Priority}}.  
  The priority order $\mathsf{P}$ is chosen \emph{without} referencing $c_i,v_i$, hence an agent cannot influence its priority by lying.  
  \item 
  \underline{Lines 3--4: \emph{Suboptimal PBS}}.  
  We run standard or partial-order PBS as a black-box subroutine.  The resulting paths can be suboptimal but feasible.  
  \item 
  \underline{Lines 6--7: \emph{Tax Payment}}.  
  Each agent $i$ pays a \emph{VCG-like} tax equal to the difference between the best possible “others’ welfare” if $i$ were absent and the “others’ welfare” in the presence of $i$.  The set of possible alternative priorities $\mathsf{P}_\text{all}$ is restricted to remain \emph{independent} of $i$’s own data.  
\end{itemize}

\noindent
\textbf{Differences from Base PBS.}  
Although the path planning step (Line~2) can be quite similar to Algorithm~\ref{alg:pbs:vanilla-recap}, this approach (i) enforces that $\mathsf{P}$ not be manipulated by agent $i$, and (ii) appends a \emph{payment} computation, ensuring the mechanism is \emph{strategyproof}.  This extra step has no direct analog in the classical (non-game-theoretic) PBS.

\subsubsection{Summary of Added Pseudocode and Their Key Differences}

Table~\ref{tab:pbs-compare-variants} summarizes these three PBS variants (PIBT-like, conflict-driven merging, mechanism-based) in relation to \emph{base PBS} (Algorithm~\ref{alg:pbs:vanilla-recap}).  Each extension modifies how priorities are assigned or updated, how collisions are resolved (via branching or merging), or how final solutions incorporate incentive-aligned payments.

\begin{table}[ht]
\centering
\footnotesize
\caption{Comparison of \emph{PIBT-like}, \emph{Conflict-Driven Merging}, and \emph{Mechanism-Design} PBS with respect to \emph{base PBS}.}
\label{tab:pbs-compare-variants}
\begin{tabular}{p{0.225\linewidth}p{0.25\linewidth}p{0.45\linewidth}}
\toprule
\textbf{Variant} & \textbf{Key Modification} & \textbf{Representative Pseudocode / Notable Differences} \\
\midrule
Base PBS & Single pass in ascending priority & Alg.~\ref{alg:pbs:vanilla-recap}.  Each agent yields to higher-priority paths.  No dynamic reorder.\\
\midrule
\emph{PIBT-like} & Plan \emph{one step at a time} in real time.  Apply \emph{priority inheritance} if blocked. & Alg.~\ref{alg:pbs:dynamic} replaces the single pass with a \textbf{for t=0..T} loop, re-evaluating next-step collisions.  No global solution guarantee.\\
\midrule
\emph{Conflict-Driven \& Merging} & \emph{Partial-order branching} for collisions. Once collisions are repeated, \emph{merge} colliding agents. & Alg.~\ref{alg:pbs:conflict-merge}.  Instead of a single ascending order, the partial order grows with collision constraints.  Merging forms meta-agents.\\
\midrule
\emph{Mechanism-Design} & \emph{Random or exogenous} priority to ensure strategyproofness.  Compute \emph{VCG-like} tax for each agent. & Alg.~\ref{alg:pbs:mechanism}.  Similar to base PBS for path planning, but no agent can influence $\mathsf{P}$ by lying.  Payment stage appended.\\
\bottomrule
\end{tabular}
\end{table}

Collectively, these variants illustrate the versatility of PBS in diverse settings: from \emph{online stepwise} planning (PIBT) to \emph{conflict-driven merges} in offline MAPF, and even \emph{game-theoretic} multi-agent allocations.  

Thus, while the \emph{base} (vanilla) PBS procedure (Algorithm~\ref{alg:pbs:vanilla-recap}) remains a minimal blueprint for priority-based collision avoidance, actual deployments often rely on these advanced extensions to handle dynamic priorities, repeated collisions, or incentive alignment. Together, they display how PBS can be tailored to balance simplicity, scalability, real-time adaptability, and strategic considerations.

\subsubsection{Contrasting CBS and PBS}\label{sec:pbs:vscbs}

Both CBS and PBS revolve around \emph{managing collisions} among agents. Table~\ref{tab:cbs-vs-pbs} distills their key similarities and differences:

\begin{table}[ht]
\centering
\footnotesize
\caption{Comparison of Conflict-Based Search (CBS) and Priority-Based Search (PBS)}
\label{tab:cbs-vs-pbs}
\begin{tabular}{p{0.18\linewidth}p{0.38\linewidth}p{0.34\linewidth}}
\toprule
 & \textbf{CBS} (Section~\ref{sec:cbs}) & \textbf{PBS} (Section~\ref{sec:pbs})\\
\midrule
Core Idea 
& Two-level search (HL: constraints, LL: single-agent replan).  Detect collisions among \emph{all} agent paths, branch with constraints to fix collisions.  
& Implicit constraints from a \emph{(partial) priority order.}  Each agent actively avoids collisions with higher-priority agents’ paths.  Little/no replan of the higher-priority routes.\\[5pt]

Collision Handling 
& Upon conflict, \emph{both} agents branch.  In each child node, exactly one agent is constrained to avoid the conflict. A single final node is conflict-free.  
& Lower-priority agents must always detour: a conflict is effectively “won” by the higher-priority agent’s path.  Dynamic partial orders or inheritance can shift who is higher priority.\\[5pt]

Optimality 
& Baseline CBS is complete and optimal for SoC or makespan if the branching is exhaustive, with variants adding heuristics for speed.  
& Vanilla PBS is generally \emph{not} guaranteed to find an optimal solution unless one enumerates or adjusts all possible priority orders. Typically yields suboptimal or incomplete results in large grids.\\[5pt]

Algorithmic Enhancements 
& \textbf{Optimal:} Admissible HL heuristics, symmetry breaking, merge-and-replan, disjoint splitting  
\newline
\textbf{Suboptimal:} ECBS, EECBS, \emph{etc.} 
& \textbf{Static:} Single pass in total priority  
\newline
\textbf{Dynamic:} Conflict-driven partial orders; \emph{priority inheritance}; merges; windowed planning  
\newline
\textbf{Mechanism design:} Strategyproof priority or partial order \\
\bottomrule
\end{tabular}
\end{table}

\noindent
\textbf{High-Level Observations.}
\begin{itemize}[leftmargin=*]
   \item \emph{Complexity vs.\ Flexibility.}  CBS can systematically achieve optimal solutions given enough branching, but typically sees exponential blowups for large $n$. PBS is simpler but heavily depends on a good (partial) priority assignment to produce near-optimal results.
   \item \emph{Similar Enhancement Themes.}  Both incorporate merging, suboptimal bounding (via focal searches or simpler heuristics), and specialized constraints for symmetrical conflicts.
   \item \emph{Use Cases.}  PBS is often favored in \emph{online} or \emph{lifelong} MAPF, where stepwise or partial-order planning (as in PIBT/winPIBT) yields quick updates in dynamic environments. CBS or suboptimal CBS may still excel in medium-scale, \emph{offline} instances with strong demand for solution optimality or bounded suboptimality.
\end{itemize}

\subsubsection{Conclusion and Outlook for PBS}

Priority-Based Search (PBS) stands alongside CBS as one of the fundamental search-based paradigms in multi-agent pathfinding. Its hallmark decision rule—“lower-priority agents must defer to higher-priority paths”—offers a straightforward collision-avoidance mechanism that can scale to large agent sets under real-time or constrained computation. We have seen how a simple, \emph{static} total ordering can suffice in some applications but also how \emph{dynamic} partial orders, \emph{priority inheritance}, and \emph{merging} can substantially improve PBS’s completeness and solution quality.

Despite PBS’s relative simplicity, it generally lacks the robust global optimality framework present in classical CBS. Nonetheless, specialized PBS extensions have bridged this gap somewhat by enumerating partial orders or integrating additional constraints in a manner reminiscent of CBS’s conflict-resolution tree. Furthermore, the synergy with \emph{mechanism design} offers a powerful route to strategyproof path assignments in open multi-robot systems.

In the subsequent sections, we explore compilation-based MAPF solvers (e.g., ILP, SAT) and learning-based methods. Both directions can be integrated with or inspired by search strategies akin to CBS or PBS, underscoring the diversity of approaches in modern MAPF research.

\subsection{Large Neighborhood Search (LNS) Methods for Suboptimal MAPF}\label{sec:lns}

\begin{figure}[htb!]
    \centering
    \includegraphics[width=\linewidth]{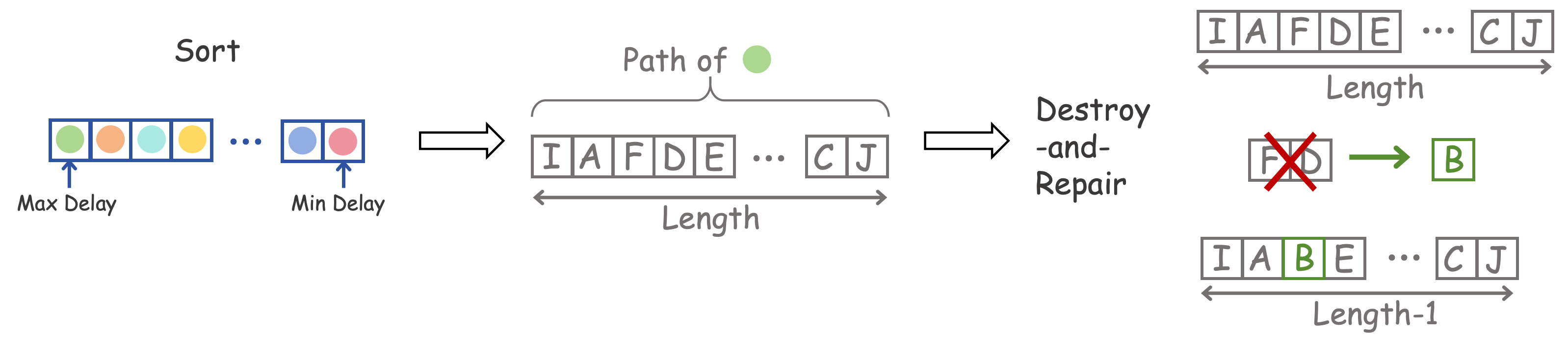}
    \caption{Illustration of the core workflow in LNS. Left: all agents (or path segments) are ranked by their ``delay'' scores—from highest (green) to lowest (pink). We select the target agent (green) and extract its current path (middle). Right: During the destroy phase (top right), a contiguous subsegment {F,D} is removed from the path. In the repair phase (bottom right), the best replacement vertex B (in green) is inserted at the removal point, producing the new path.}
    \label{fig:lns}
\end{figure}

Large Neighborhood Search (LNS) is a meta-heuristic optimization approach that has shown promise in solving suboptimal MAPF problems with potentially large numbers of agents and dynamic constraints~\citep{li2021anytime,li2022mapf,lam2023exact,tan2024benchmarking,phan2024adaptive}. 
Unlike Conflict-Based Search (CBS) methods that systematically resolve collisions through high-level and low-level searches (Section~\ref{sec:cbs}), LNS treats MAPF as a sequence of iterative refinements over a global solution, repeatedly ``destroying'' and ``repairing'' selected parts (or neighborhoods) of candidate solutions.
The illustration of the core workflow in LNS is shown in Figure~\ref{fig:lns}.

This section introduces the mathematical formulation of LNS in the MAPF setting, followed by a presentation of the baseline LNS algorithm and prominent LNS variants. 
We also compare how LNS complements or contrasts with both optimal and suboptimal CBS frameworks discussed in Section~\ref{sec:cbs:optimal} and Section~\ref{sec:cbs:suboptimal}.

\subsubsection{Mathematical Formulation for LNS}

We cast LNS-based MAPF using a similar problem definition as in Section~\ref{sec:formulation}, with discrete time steps (or a permissible continuous time horizon) and a shared environment modeled by an undirected graph $\mathcal{G} = (\mathcal{V}, \mathcal{E})$. 
Each agent $i \in \{1,\dots,n\}$ has a start vertex $s_i$ and a goal vertex $g_i$. 
A candidate \emph{global solution} is defined as a collection of single-agent paths $\{\pi_i\}_{i=1}^n$ where each $\pi_i$ is collision-free with respect to other paths.

\paragraph{Decision Variables and Constraints.}
As in Conflict-Based Search (Section~\ref{sec:cbs}), we can define binary variables $x_{i,v,t}$ indicating whether agent~$i$ is at vertex~$v$ at time~$t$, and impose collision-avoidance constraints:
\begin{align}
x_{i,v,t} + x_{j,v,t} &\;\le\; 1, 
&&\forall\,t, \,\forall\,v \in \mathcal{V}, \,\forall\,i < j, \label{eq:lns-vertex}\\
x_{i,u,t} + x_{i,v,t+1} + x_{j,v,t} + x_{j,u,t+1} &\;\le\; 3,
&&\forall\, (u,v)\in\mathcal{E}, \,\forall\, t, \,\forall\,i< j, \label{eq:lns-edge}
\end{align}
accompanied by single-agent path consistency and start/goal constraints. 

\paragraph{Objective.}
Typically, LNS methods target one or more of the following objectives:
\begin{itemize}[leftmargin=*]
    \item \textbf{Makespan Minimization:} 
    \[
       \min \max_{1 \leq i \leq n} \mathrm{Cost}(\pi_i), 
    \]
    where $\mathrm{Cost}(\pi_i)$ measures the time at which agent~$i$ reaches $g_i$.
    \item \textbf{Sum of Costs (SoC):}
    \[
       \min \sum_{i=1}^{n} \mathrm{Cost}(\pi_i).
    \]
    \item \textbf{Hybrid or Weighted Metrics (e.g., Weighted SoC):}
    \[
       \min \sum_{i=1}^{n} \alpha_i \,\mathrm{Cost}(\pi_i),
    \]
    where $\alpha_i$ sets per-agent weighting for more flexible optimization criteria.
\end{itemize}

\subsubsection{Algorithmic Framework and Basic LNS}\label{sec:lns:framework}

Large Neighborhood Search, originally popularized in vehicle routing and related combinatorial problems, proceeds by iteratively refining a \emph{current solution} (global set of paths) via two complementary actions:
\begin{enumerate}[leftmargin=*]
    \item \textbf{Destroy:} Select a subset of agents (or edges, time intervals, etc.) and temporarily remove or invalidate their paths, creating partial solutions with missing paths. 
    \item \textbf{Repair:} Replan the removed agents’ paths (using some constructive or heuristic method) in a way that hopefully reduces overall cost while maintaining collision-free feasibility.
\end{enumerate}

This meta-heuristic leverages the intuition that large-scale disruptions to the current solution can escape local minima more effectively than small, incremental updates. 
Algorithm~\ref{alg:lns} provides a simplified pseudocode for basic LNS in the MAPF context.

\begin{algorithm}[ht]
\footnotesize
\caption{Baseline LNS for Suboptimal MAPF\label{alg:lns}}
\begin{algorithmic}[1]
\Require Graph $\mathcal{G}$, agents $\{1,\dots,n\}$, start $\{s_i\}$, goal $\{g_i\}$, destroy ratio $\gamma$, max iterations $K$.
\State \textbf{Initialize} a feasible solution $\{\pi_i^{(0)}\}$ (e.g., by a greedy or search-based solver).
\State $S \gets \{\pi_i^{(0)}\}_{i=1}^n$ \Comment{current global solution}
\State $S_{\mathrm{best}} \gets S$ \Comment{best solution found so far}
\For{$k \gets 1$ \textbf{to} $K$}
   \State \textbf{Destroy Step:} 
   \Statex \hspace{1.5em}{\quad Select a subset of agents $\mathcal{D} \subseteq \{1,\dots,n\}$ (size $\approx \gamma n$). Remove their paths from $S$.}
   \State $S^{\mathrm{partial}} \gets S \setminus \{\pi_i : i \in \mathcal{D}\}$
   \State \textbf{Repair Step:}
   \Statex \hspace{1.5em}{\quad For each agent $i \in \mathcal{D}$, replan $\pi_i$ to minimize collisions and cost, given $S^{\mathrm{partial}}$.}
   \State Combine paths into new solution $S^{\prime}$.
   \If{$\mathrm{Cost}(S^{\prime}) < \mathrm{Cost}(S_{\mathrm{best}})$}
      \State $S_{\mathrm{best}} \gets S^{\prime}$ 
   \EndIf
   \State $S \gets$ \textsc{AcceptanceCriterion}$(S, S^{\prime})$ 
   \Comment{decide which solution to keep as current}
\EndFor
\State \Return $S_{\mathrm{best}}$
\end{algorithmic}
\end{algorithm}

\noindent
\textbf{Algorithm Explanation.}
\begin{itemize}[leftmargin=*]
    \item \emph{Initialization (Lines~1--3).} 
    A feasible solution is produced via any standard method (e.g., single-agent A* ignoring collisions, combined with a naive conflict-resolution pass). This serves as $S$, the current solution, and $S_{\mathrm{best}}$, the best solution found so far.
    \item \emph{Destroy Step (Lines~6--7).} 
    A fraction $\gamma$ of agents are randomly (or heuristically) selected, and their paths are removed from $S$. The remaining agents keep their existing paths, forming $S^{\mathrm{partial}}$.
    \item \emph{Repair Step (Lines~8--9).} 
    Each removed agent is re-inserted (planned) in a manner that aims to reduce collisions and overall cost. Any single-agent method, bounded-suboptimal pathfinding, or even an embedded MAPF solver can be used.
    \item \emph{Solution Acceptance (Lines~12--13).}
    After constructing $S^{\prime}$, LNS updates the global best $S_{\mathrm{best}}$ if $\mathrm{Cost}(S^{\prime})$ improves upon the previous best. 
    The routine \textsc{AcceptanceCriterion} decides whether to set $S^{\prime}$ or the old $S$ as the new \emph{current solution} for the next iteration. 
    Common strategies include always accept if $\mathrm{Cost}(S^{\prime}) \le \mathrm{Cost}(S)$ (greedy) or occasionally accept worse solutions to promote escaping local minima (similar to simulated annealing).
\end{itemize}

Compared to suboptimal CBS (Section~\ref{sec:cbs:suboptimal}), LNS explicitly leverages large-scale partial reoptimization rather than systematic conflict splitting. 
As a result, LNS can quickly explore diverse regions of the solution space, especially in large instances where enumerating conflict trees becomes costly.

\subsubsection{Notable LNS Variants}\label{sec:lns:variants}

A variety of specialized LNS strategies have been proposed to address efficiency, scalability, and dynamic adaptability in MAPF. 
In particular, we highlight five notable LNS-based methods below, each introducing unique refinements for suboptimal MAPF.  
For clarity, we provide pseudocode fragments that extend or modify the baseline LNS procedure (Algorithm~\ref{alg:lns}). 
All \fbox{\emph{boxed lines}} indicate \textbf{additions} or \textbf{changes} compared to the baseline, facilitating direct comparison across variants.

\vspace{1em}
\paragraph{Anytime LNS-based MAPF (ALNS).}  
\citet{li2021anytime} proposed an \emph{anytime} LNS approach that can produce a valid solution quickly, then iteratively refine it to higher quality if time permits. 
It incorporates adaptive destroy-repair strategies to prioritize collisions. 
Algorithm~\ref{alg:alns} highlights key modifications: an explicit \texttt{UpdateTimeBudget} routine to ensure the method returns the best solution before the time limit expires.

\begin{algorithm}[ht]
\footnotesize
\caption{Anytime LNS-based MAPF (ALNS)\label{alg:alns}}
\begin{algorithmic}[1]
\Require \fbox{Time budget $\tau$}, destroy ratio $\gamma$, max iterations $K$.
\State \textbf{Initialize} $S, S_{\mathrm{best}}$ as in Algorithm~\ref{alg:lns}. 
\For{$k \gets 1$ \textbf{to} $K$}
   \If{\fbox{\textsc{IsTimeExceeded}($\tau$)}}
      \State \textbf{return} \fbox{$S_{\mathrm{best}}$} \Comment{return best so far if time exceeded}
   \EndIf
   \State Perform \textbf{Destroy} and \textbf{Repair} steps as in Lines~6--9 of Algorithm~\ref{alg:lns}.
   \Statex \hspace{2.5em}{\quad \fbox{Possibly use a collision-based heuristic to select agent subset.}}
   \If{$\mathrm{Cost}(S^{\prime}) < \mathrm{Cost}(S_{\mathrm{best}})$}
      \State $S_{\mathrm{best}} \gets S^{\prime}$
   \EndIf
   \State $S \gets \textsc{AcceptanceCriterion}(S, S^{\prime})$
   \State \fbox{\textsc{UpdateTimeBudget}($\tau$)}
\EndFor
\State \Return $S_{\mathrm{best}}$
\end{algorithmic}
\end{algorithm}

\paragraph{MAPF-LNS2.}  
\citet{li2022mapf} introduced \emph{MAPF-LNS2}, which focuses on \emph{fast conflict resolution} by reconstructing agent paths within local neighborhoods. 
It often employs a more refined `destroy’ step that selectively removes only those agents heavily involved in collisions. 
While Algorithm~\ref{alg:lns2} is structurally similar to the baseline, the crucial distinction lies in the multi-layered repair strategy (Lines~10--12) that re-inserts agents with a specialized local solver.

\begin{algorithm}[ht]
\footnotesize
\caption{MAPF-LNS2: Focused Local Re-Optimization\label{alg:lns2}}
\begin{algorithmic}[1]
\Statex \textbf{Differences from Algorithm \ref{alg:lns} are marked with boxes.}
\Require Graph $\mathcal{G}$, destroy ratio $\gamma$, \fbox{focused conflict threshold $\eta$}, max iterations $K$.
\State Initialize solution $S, S_{\mathrm{best}}$ as before.
\For{$k\gets 1$ to $K$}
   \State \textbf{Destroy Step:} 
   \Statex \hspace{1.3em}{\quad \fbox{Identify agents with collision counts above $\eta$; remove their paths.}}
   \State $S^{\mathrm{partial}} \gets S\setminus\{\pi_i : i\in \mathcal{D}\}$ 
   \State \textbf{Repair Step:}
   \For{each $i\in \mathcal{D}$}
      \State \fbox{Use a \emph{local repair solver} to plan $\pi_i$ within a bounding region around $s_i$ and $g_i$.}
   \EndFor
   \State Combine new paths into $S^{\prime}$.
   \If{$\mathrm{Cost}(S^{\prime})< \mathrm{Cost}(S_{\mathrm{best}})$}
      \State $S_{\mathrm{best}} \gets S^{\prime}$
   \EndIf
   \State $S \gets \textsc{AcceptanceCriterion}(S, S^{\prime})$
\EndFor
\State \Return $S_{\mathrm{best}}$
\end{algorithmic}
\end{algorithm}

\paragraph{BCP-LNS for Exact Anytime Solutions.}  
\citet{lam2023exact} combined Branch-and-Cut-and-Price (BCP) methods with LNS to deliver \emph{exact} solutions in an anytime manner. 
Although BCP supports an \emph{optimal} solution guarantee, its worst-case runtime can be large. 
Hence, LNS heuristics are employed to prune the search space aggressively. 
Algorithm~\ref{alg:bcp-lns} shows the synergy, where a partial BCP-based cut is enforced in each \textbf{Repair Step} (Line~9).

\begin{algorithm}[ht]
\footnotesize
\caption{BCP-LNS: Integrating Branch-and-Cut-and-Price with LNS\label{alg:bcp-lns}}
\begin{algorithmic}[1]
\Statex \textbf{Key additions are marked with boxes.}.
\Require Graph $\mathcal{G}$, agents, max iterations $K$.
\State Initialize solution $S, S_{\mathrm{best}}$. 
\For{$k\gets 1$ to $K$}
   \State \textbf{Destroy Step:} same as baseline LNS (Algorithm~\ref{alg:lns}).
   \State \textbf{Repair Step:}
   \Statex \hspace{1.3em}{\quad \fbox{Invoke partial BCP routine to identify if any feasible path segment can be pruned by cutting planes.}}
   \Statex \hspace{1.3em}{\quad Recompute $\pi_i$ for $i\in\mathcal{D}$ with BCP constraints.}
   \State $S^{\prime} \gets S^{\mathrm{partial}} \cup \{\pi_i` \text{for} i\in\mathcal{D}\}$
   \If{$\mathrm{Cost}(S^{\prime})<\mathrm{Cost}(S_{\mathrm{best}})$}
      \State $S_{\mathrm{best}}\gets S^{\prime}$
   \EndIf
   \State $S\gets \textsc{AcceptanceCriterion}(S,S^{\prime})$
\EndFor
\State \Return $S_{\mathrm{best}}$
\end{algorithmic}
\end{algorithm}

\paragraph{Benchmarking LNS (B-LNS).}  
\citet{tan2024benchmarking} conducted a comprehensive evaluation of multiple LNS-based algorithms on standard MAPF benchmarks. 
Although there is no fundamentally new pseudocode, their approach systematically compares different \emph{destroy} heuristics (random, collision-based, region-based), \emph{repair} strategies (standard A*, suboptimal focus search), and acceptance criteria (greedy vs.\ simulated annealing). 
Their findings highlight how distinct LNS configurations trade off solution quality and runtime differently across problem scales.

\paragraph{Bandit-based Adaptive LNS (BA-LNS).}  
\citet{phan2024adaptive} introduced a \emph{multi-armed bandit} mechanism to adaptively select among multiple destroy-repair heuristics. 
Agents or conflict regions are assigned to ``arms,'' and the algorithm dynamically shifts selection probability toward heuristics producing the largest improvement over the last few iterations. 
Algorithm~\ref{alg:ba-lns} illustrates this scheme, where \textsc{BanditSelection} maintains a reward distribution for each destroy/repair pair.

\begin{algorithm}[ht]
\footnotesize
\caption{BA-LNS: Bandit-based Adaptive LNS\label{alg:ba-lns}}
\begin{algorithmic}[1]
\Statex \textbf{Key differences from Algorithm~\ref{alg:lns} are marked with boxes.}.
\Require Graph $\mathcal{G}$, destroy-heuristic set $\{\mathrm{DH}_m\}$, repair-heuristic set $\{\mathrm{RH}_n\}$, max iterations $K$.
\State Initialize $S, S_{\mathrm{best}}$, and \fbox{multi-armed bandit policy $\pi_{\mathrm{bandit}}$}.
\For{$k\gets 1$ to $K$}
   \State \fbox{$(\mathrm{DH}_{m^*}, \mathrm{RH}_{n^*}) \gets \textsc{BanditSelection}(\pi_{\mathrm{bandit}})$}
   \State \textbf{Destroy Step:} 
   \Statex \hspace{1.5em}{\quad \fbox{Use $\mathrm{DH}_{m^*}$ to remove paths for a subset of agents.}}
   \State \textbf{Repair Step:} 
   \Statex \hspace{1.5em}{\quad \fbox{Use $\mathrm{RH}_{n^*}$ to replan these agents’ paths.}}
   \State Evaluate $S^{\prime}$ and compute the improvement $\Delta = \mathrm{Cost}(S) - \mathrm{Cost}(S^{\prime})$.
   \If{$\mathrm{Cost}(S^{\prime}) < \mathrm{Cost}(S_{\mathrm{best}})$}
      \State $S_{\mathrm{best}} \gets S^{\prime}$
   \EndIf
   \State $S \gets \textsc{AcceptanceCriterion}(S,S^{\prime})$
   \State \fbox{\textsc{UpdateBanditReward}($\Delta,\,\mathrm{DH}_{m^*},\,\mathrm{RH}_{n^*},\,\pi_{\mathrm{bandit}}$)}
\EndFor
\State \Return $S_{\mathrm{best}}$
\end{algorithmic}
\end{algorithm}

\subsubsection{Comparison with CBS-based Methods}

Both LNS and CBS methods aim to solve MAPF in suboptimal (bounded or heuristic-driven) regimes, but they employ fundamentally different strategies:
\begin{itemize}[leftmargin=*]
    \item \textbf{Conflict Resolution vs.\ Destroy-Repair:}
    Suboptimal CBS (ECBS, EECBS, etc.) systematically resolves collisions via branching constraints and bounded-suboptimal single-agent replanning, while LNS performs large-scale partial reoptimizations on selected agents. 
    \item \textbf{Search Granularity:}
    CBS-based approaches track collision conflict trees, refining partial solutions at a conflict-by-conflict level. LNS manages solutions at a more \emph{global} scale (entire subsets of agents, or spatiotemporal neighborhoods). 
    \item \textbf{Performance and Scalability:}
    Empirical benchmarks~\citep{li2021anytime,li2022mapf,tan2024benchmarking,chan2022flex} indicate that LNS can handle larger MAPF instances more gracefully, converging to feasible solutions quickly, though with fewer theoretical guarantees than standard CBS. 
    On the other hand, suboptimal CBS methods can provide bounded suboptimality guarantees (e.g., solution cost $\leq w \cdot \mathrm{Cost}(\text{opt})$), which can be appealing in certain real-time applications.
\end{itemize}

LNS methods, particularly in the form of ALNS~\citep{li2021anytime} and MAPF-LNS2~\citep{li2022mapf}, demonstrate strong adaptability to dynamic environments and partial observations by re-optimizing large portions of the solution when new information arrives or constraints change. 
They can therefore be particularly beneficial in industrial scenarios (e.g., warehouse logistics or multi-drone coordination) where tasks evolve over time. 
Simultaneously, the line of BCP-LNS~\citep{lam2023exact} highlights how coupling LNS with rigorous optimization backends can bridge the gap between heuristic speed and exactness if computational resources permit. 
Lastly, adaptivity via bandit-based strategies~\citep{phan2024adaptive} offers a systematic means to tune LNS parameters online, thereby potentially outperforming static heuristics in diverse MAPF contexts.

Overall, LNS serves as a complementary approach to both optimal and suboptimal CBS, especially in large-scale or highly dynamic MAPF problems where partial reoptimization can effectively balance solution quality with rapid convergence. 
Future research may combine the conflict-tree perspective of CBS with the large-scale reoptimization advantages of LNS, further broadening the computational toolkit for MAPF researchers and practitioners alike.

\subsection{Lazy Constraints Addition Search (LaCAM)}
In recent years, the field of MAPF has continually grappled with the trade-off between computational complexity and real-time performance while striving for high-precision solutions. 
On the one hand, early time-independent and offline planning methods \citep{okumura2021time, okumura2021offline} enhanced solution accuracy by precomputing conflict-free paths. 
However, these approaches tend to incur high computational overhead, limiting their scalability and adaptability in dynamic environments. 
On the other hand, to meet real-time demands, researchers have introduced iterative optimization techniques \citep{okumura2021iterative}. 
Although this method offers clear advantages in terms of running time, its near real-time solution process may fall short of achieving optimal accuracy. 

\begin{figure}[htb!]
    \centering
    \includegraphics[width=\linewidth]{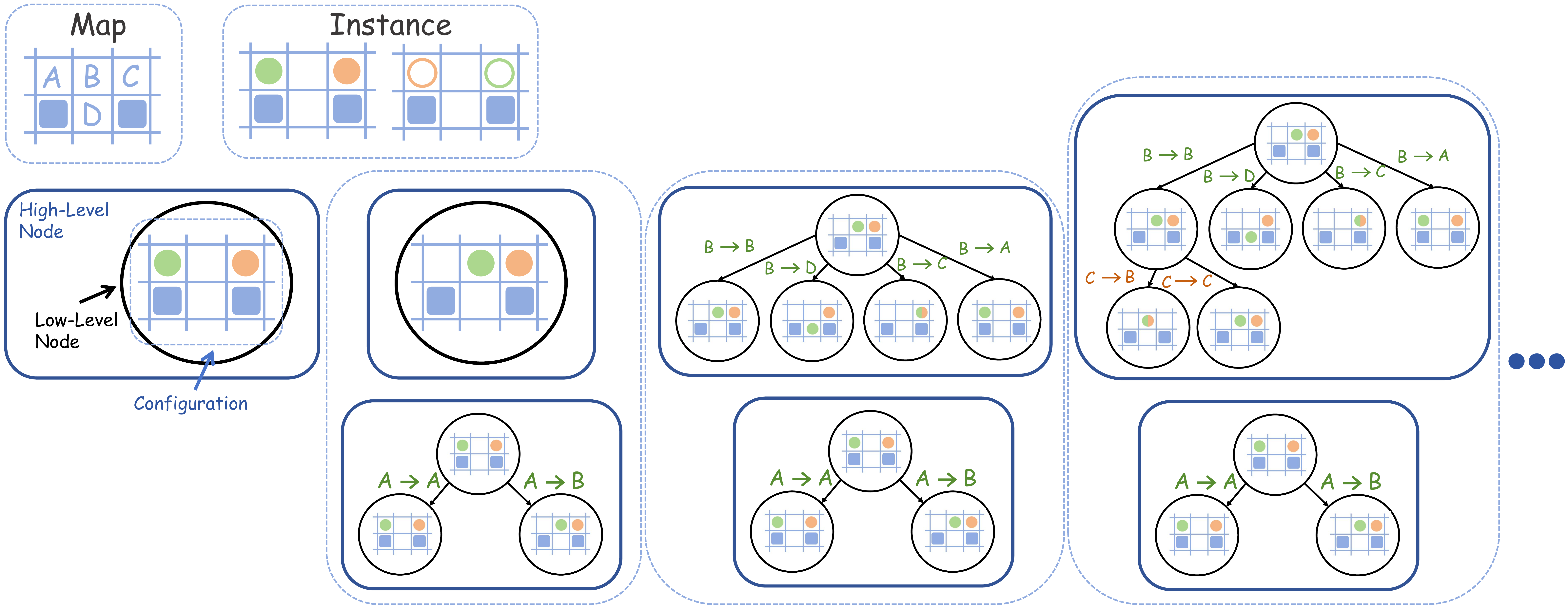}
    \caption{Illustration of LaCAM. Top left (Map \& Instance): A 3 $\times$ 3 grid (blue squares are represent obstacles) with nodes labeled A–D, and two agents (\textcolor[HTML]{ACD78E}{\ding{108}} and \textcolor[HTML]{F5B482}{\ding{108}}) whose sources (solid circles) and targets (hollow circles) are shown. A high‐level node (rounded rectangle) holding the current global configuration of all agents (black circle). Within each high‐level node, LaCAM invokes a lazy low‐level search for one agent cluster (shown by dashed gray boxes). Successor configurations are generated by cluster‐specific constraints, indicated by labeled arrows (e.g. A$\rightarrow$A, A$\rightarrow$B, B$\rightarrow$D, B$\rightarrow$C, C$\rightarrow$B, C$\rightarrow$C) that fix one cluster’s agent positions in the next configuration. Each low‐level invocation produces a minimal successor; once refined, it is merged back into the parent high‐level node and the next cluster is processed.}
    \label{fig:lacam}
\end{figure}

\subsubsection{Suboptimal Method}

More recently, research has focused on system robustness and scalability in large-scale environments \citep{okumura2023fault, okumura2023lacam}, leading to the development of an improved method known as Lazy Constraints Addition Search (LaCAM), as shown in Figure~\ref{fig:lacam}.
LaCAM is a step-by-step configuration search method inspired by the CBS framework. 
Similar to CBS, LaCAM employs a two-level architecture: the high-level search maintains a sequence of configurations, while the low-level search dynamically generates \emph{constraints}. The primary innovation of LaCAM (Algorithm \ref{alg:lacam}) is its lazy evaluation strategy at the low-level search, which incrementally creates minimal successors only when the corresponding high-level node is explored.

\begin{algorithm}[ht]
\footnotesize
\caption{LaCAM}
\label{alg:lacam}
\begin{algorithmic}[1]
\Require MAPF instance ($\mathcal{S}$: starts, $\mathcal{G}$: goals)
\State \textbf{Initialize:} Create root constraint $\mathcal{C}_{\text{init}} \leftarrow \langle~ parent: \bot, who: \bot, where: \bot~\rangle$;~$\mathcal{N}_{\text{goal}} \leftarrow \bot$
\State Initialize $Open$ and $Explored$
\State Create initial node:
$\mathcal{N}_{\text{init}} \leftarrow \langle~config: \mathcal{S},~tree: \llbracket~\mathcal{C}_{\text{init}}~\rrbracket,~neighbors: \emptyset,~parent: \bot
$
\State $Open.\textbf{push}(\mathcal{N}_{\text{init}})$; $Explored[\mathcal{S}] \leftarrow \mathcal{N}_{\text{init}}$
\While{$Open \neq \emptyset $}
    \State $\mathcal{N} \leftarrow Open.\textbf{top}()$
    \If{$\mathcal{N}.config=\mathcal{G}$}
        \State \Return $\textsc{backtrack}(\mathcal{N})$
    \EndIf
    \If{$\mathcal{N}.tree = \emptyset$}
        \State $Open.\textbf{pop}()$; \textbf{continue}
    \EndIf
    \State $\mathcal{C} \leftarrow \mathcal{N}.tree.\textbf{pop}()$
    \State $\textsc{LowLevelExpansion}(\mathcal{N}, \mathcal{C})$
    \State $Q_{\text{new}} \leftarrow \textsc{ConfigurationGenerator}(\mathcal{N}, \mathcal{C})$
    \If{$Q_{\text{new}} = \bot$}
        \State \textbf{continue}
    \EndIf
        \[
        \mathcal{N}_{\text{new}} \leftarrow \langle~config: Q_{\text{new}},~tree: \llbracket~\mathcal{C}_{\text{init}}~\rrbracket,~parent: \mathcal{N}~\rangle
        \]
        \State $Open.\textbf{push}(\mathcal{N}_{\text{new}})$
        \State $Explored[Q_{\text{new}}] \leftarrow \mathcal{N}_{\text{new}}$
\EndWhile
\State \Return \texttt{NO\_SOLUTION}
\end{algorithmic}
\end{algorithm}

\subsubsection{Eventually Optimal Method}

\citep{okumura2023improving} aimed to achieve a more balanced trade-off between accuracy and computational complexity by integrating pre-computation with online adjustment strategies. Specifically, their method differs from LaCAM in three main aspects (Algorithm \ref{alg:lacamstar}):
\begin{enumerate}
    \item It continues the search even after finding the goal configuration $\mathcal{G}$.
    \item It dynamically revises parent relationships among search nodes as needed.
    \item Unlike LaCAM, which is a sub-optimal algorithm, LaCAM* has been proven to be complete and optimal; see Theorem 1 in \citep{okumura2023improving} for further details.
\end{enumerate}

\begin{algorithm}[ht]
\footnotesize
\caption{LaCAM*}
\label{alg:lacamstar}
\begin{algorithmic}[1]
\Require MAPF instance ($\mathcal{S}$: starts, $\mathcal{G}$: goals), edge cost $\textbf{cost}_e$, admissible heuristic $\textbf{h}$
\State \textbf{Initialize:} Create root constraint $\mathcal{C}_{\text{init}} \leftarrow \langle~ parent: \bot, who: \bot, where: \bot~\rangle$;~$\mathcal{N}_{\text{goal}} \leftarrow \bot$
\State Initialize $Open$ (priority queue ordered by $f(\mathcal{N})=\mathcal{N}.g + h(\mathcal{N})$) and $Explored$
\State Create initial node:
$\mathcal{N}_{\text{init}} \leftarrow \langle~config: \mathcal{S},~tree: \llbracket~\mathcal{C}_{\text{init}}~\rrbracket,~neighbors: \emptyset,$$~parent: \bot,~neigh : \emptyset, ~g:0~\rangle
$
\State $Open.\textbf{push}(\mathcal{N}_{\text{init}})$; $Explored[\mathcal{S}] \leftarrow \mathcal{N}_{\text{init}}$
\While{$Open \neq \emptyset $$\land \lnot\textbf{interrupt}()$}
    \State $\mathcal{N} \leftarrow Open.\textbf{top}()$
    \If{$\mathcal{N}.config = \mathcal{G}$}
    \State $\mathcal{N}_{\text{goal}} \leftarrow \mathcal{N}$
    \EndIf
    \If{$(\mathcal{N}_{\text{goal}} \neq \bot) \land f(\mathcal{N}_{\text{goal}}) \leq f(\mathcal{N})$}
        \State $Open.\textbf{pop}()$; \textbf{continue}
    \EndIf
    \If{$\mathcal{N}.tree = \emptyset$}
        \State $Open.\textbf{pop}()$; \textbf{continue}
    \EndIf
    \State $\mathcal{C} \leftarrow \mathcal{N}.tree.\textbf{pop}()$
    \State $\textsc{LowLevelExpansion}(\mathcal{N}, \mathcal{C})$
    \State $Q_{\text{new}} \leftarrow \textsc{ConfigurationGenerator}(\mathcal{N}, \mathcal{C})$
    \If{$Q_{\text{new}} = \bot$}
        \State \textbf{continue}
    \EndIf
    \If{$\textsc{Explored}[Q_{\text{new}}] \neq \bot$}
        \State $\mathcal{N}.neighbors.\textbf{append}(Explored[Q_{\text{new}}])$
        \State $\textsc{DijkstraOpen} \leftarrow \llbracket \mathcal{N} \rrbracket$ \Comment{Priority queue ordered by $g$}
        \While{$\textsc{DijkstraOpen} \neq \emptyset$}
            \State $\mathcal{N}_{\text{from}} \leftarrow \textsc{DijkstraOpen}.\textbf{pop}()$
            \For{$\mathcal{N}_{\text{to}} \in \mathcal{N}_{\text{from}}.neighbors$}
                \State $g \leftarrow \mathcal{N}_{\text{from}}.g + \textbf{cost}_e(\mathcal{N}_{\text{from}}, \mathcal{N}_{\text{to}})$
                \If{$g < \mathcal{N}_{\text{to}}.g$}
                    \State $\mathcal{N}_{\text{to}}.g \leftarrow g$;~$\mathcal{N}_{\text{to}}.parent \leftarrow \mathcal{N}_{\text{from}}$;~$\textsc{DijkstraOpen}.\textbf{push}(\mathcal{N}_{\text{to}})$
                    \If{$(\mathcal{N}_{\text{goal}} \neq \bot) \land f(\mathcal{N}_{\text{to}}) < f(\mathcal{N}_{\text{goal}})$}
                        \State $Open.\textbf{push}(\mathcal{N}_{\text{to}})$
                    \EndIf
                \EndIf
            \EndFor
        \EndWhile
    \Else
        \State Create successor node:
        \[
        \mathcal{N}_{\text{new}} \leftarrow \langle~config: Q_{\text{new}},~tree: \llbracket~\mathcal{C}_{\text{init}}~\rrbracket,~neighbors: \emptyset,~parent: \mathcal{N},~g:\mathcal{N}.g + \textbf{cost}_e(\mathcal{N},Q_{\text{new}})~\rangle
        \]
        \State $Open.\textbf{push}(\mathcal{N}_{\text{new}})$;~$Explored[Q_{\text{new}}] \leftarrow \mathcal{N}_{\text{new}}$
        \State $\mathcal{N}.neighbors.\textbf{append}(\mathcal{N}_{\text{new}})$
    \EndIf
\EndWhile
\If{$(\mathcal{N}_{\text{goal}} \neq \bot) \land Open = \emptyset$}
    \State \Return $\textsc{Backtrack}(\mathcal{N}_{\text{goal}})$ \Comment{optimal}
\ElsIf{$\mathcal{N}_{\text{goal}} \neq \bot$}
    \State \Return $\textsc{Backtrack}(\mathcal{N}_{\text{goal}})$ \Comment{suboptimal}
\ElsIf{$\textsc{Open} = \emptyset$}
    \State \Return \texttt{NO\_SOLUTION}
\Else
    \State \Return \texttt{FAILURE}
\EndIf
\end{algorithmic}
\end{algorithm}

\subsubsection{Engineering-Oriented Method}

The latest engineering effort, \citep{okumura2024engineering} further validates this approach by systematically combining multiple strategies to effectively reduce computational complexity while maintaining high planning accuracy, thereby addressing the practical needs of large-scale, real-time applications. Overall, these works underscore the importance of dynamically balancing solution accuracy and complexity in the MAPF problem, and future research is likely to focus on further optimizing this trade-off using adaptive techniques.

\section{Compilation-Based Methodology}\label{sec:compilation}

Transitioning from the systematic graph-based approaches discussed in Section~\ref{sec:search}, we now turn to \emph{compilation-based} methodologies for MAPF. 
Unlike search-based techniques, which explicitly enumerate collision-free paths, compilation-based methods recast MAPF constraints into a target formalism (e.g., Boolean logic, linear programming) to leverage the power of general-purpose solvers. 
In doing so, they help researchers sidestep the complexities of state-space exploration by delegating path planning to mature frameworks such as SAT, SMT, CSP, ASP, or MIP. 
This shift not only allows for well-established theoretical guarantees—arising from decades of solver development—but also provides a flexible toolset for handling heterogeneous agent capabilities and specialized constraints. 
In the following subsections, we review several notable families of compilation-based approaches, highlight their encoding strategies through illustrative examples and pseudocode, and analyze how each balances scalability, ease of modeling, and solution quality.

\subsection{Boolean Satisfiability (SAT)}\label{sec:compilation-sat}

This section presents the family of \emph{Boolean satisfiability} (SAT) methods for Multi-Agent Path Finding (MAPF) under the sum-of-costs (SoC) objective. 
Building on a long line of work~\citep{surynek2016efficient,surynek2018sub,surynek2017integration,bartak2019sat,surynek2021mutex,surynek2022migrating,vcapek2021dpll}, SAT-based MAPF transforms the coordination of multiple agents---their collision avoidance, time discretization, and SoC minimization---into a propositional formula. 
A SAT solver is then invoked to find (or prove the non-existence of) collision-free paths within a specified time horizon and cost bound.

We first describe the \emph{modeling framework} for the SoC objective in detail, with a small \emph{toy example} illustrating how the formulation works in practice. 
We then present a \emph{baseline} SAT-based algorithm, expanding on its pseudocode and design. 
Finally, we discuss several \emph{notable variants} and show how each modifies the baseline procedure, highlighting their distinct pseudocode differences.

\subsubsection{Modeling Framework for Sum-of-Costs (SoC)}\label{sec:sat:modeling}

In the SAT-based paradigm, the MAPF problem is cast as a decision question: 
\[
  \textit{``Is there a set of collision-free paths for all agents such that the total SoC is at most } \xi \text{?''}
\]
To attempt different SoC bounds, we usually increment $\xi$ or an equivalent parameter until we find a feasible (i.e., satisfiable) assignment. Below, we introduce the key ingredients of this formulation.

\paragraph{Graph Layout and Time Discretization.}

Let $\mathcal{G}=(\mathcal{V},\mathcal{E})$ be an undirected graph. We have $n$ agents, each agent $i$ starting at $s_i \in \mathcal{V}$ and aiming to reach $g_i \in \mathcal{V}$. Time is discretized into integer steps $t=0,1,2,\dots$. A \emph{makespan} parameter $T$ will limit the maximum number of timesteps we consider; that is, no agent schedule can exceed $t=T$.

\paragraph{States and Moves.}

An agent’s route is described by where it is at each timestep. The classical SAT encoding introduces:

\begin{itemize}[leftmargin=*]
    \item \(\mathbf{X_{i,v,t}}\): a Boolean variable that is \texttt{true} if and only if agent~\(i\) is at vertex~\(v\in\mathcal{V}\) at time~\(t\).
    \item \(\mathbf{E_{i,(u\to v),t}}\): a Boolean variable that is \texttt{true} if and only if agent~\(i\) moves from vertex \(u\) at time~\(t\) to vertex~\(v\) at time~\(t+1\). Sometimes one allows \((u\to u)\) for a \emph{wait} edge.
\end{itemize}

\paragraph{Sum-of-Costs Tracking.}

In order to minimize \emph{sum-of-costs}, we count how many time steps each agent actually uses in its path. A standard approach (see~\citep{surynek2016efficient}) is:

\begin{itemize}[leftmargin=*]
    \item Compute each agent's shortest path length $d_i$ ignoring other agents (e.g., by BFS/Dijkstra). Let \(\xi_0 = \sum_{i=1}^n d_i\). This is a natural \emph{lower bound} on the total cost.
    \item Allow an extra \(\Delta\ge 0\) steps beyond this bound, so any feasible solution must satisfy:
      \[
         \text{SoC} \;\; \le\;\; \xi_0 + \Delta.
      \]
    \item For each agent $i$, define a Boolean variable \(C_{i,t}\) indicating that agent~$i$ has \emph{not} yet reached its goal at time $t$, or equivalently is ``actively using'' the time step $t$. By summing these across all $i,t$, we get the total SoC, and we constrain it to be $\le \xi_0 + \Delta$.
\end{itemize}

\paragraph{Collision Avoidance.}

For a collision-free solution:

\begin{itemize}[leftmargin=*]
    \item \(\sum_{i=1}^n X_{i,v,t} \;\le\;1\) for every $v\in\mathcal{V}$ and $t=0,\dots,T$, ensuring no two agents occupy the same vertex at the same time;
    \item Agents are similarly forbidden to swap edges in opposite directions simultaneously.
\end{itemize}

\paragraph{Flow Consistency.}

If $X_{i,u,t}$ is \texttt{true}, then agent~$i$ must choose exactly one valid move to a neighbor (or stay put) at time~$t$:
\[
  X_{i,u,t} \;\Longrightarrow\;
  \sum_{v : (u,v)\in\mathcal{E}\cup\{(u,u)\}} E_{i,(u\to v),t} = 1,
\]
with $E_{i,(u\to v),t}$ implying $X_{i,u,t}$ and $X_{i,v,t+1}$. These constraints ensure each agent transitions consistently from one vertex to the next.

\subparagraph{Toy Example.}

\begin{figure}[ht]
\centering
\small
\begin{minipage}{0.65\linewidth}
\begin{center}
\setlength{\unitlength}{1.5pt}
\begin{picture}(100,45)
\put(5,30){\circle*{3}}   \put(5,13){\circle*{3}}
\put(40,30){\circle*{3}} \put(40,13){\circle*{3}}
\put(75,30){\circle*{3}} \put(75,13){\circle*{3}}
\put(5,30) {\line(1,0){35}} \put(5,13){\line(1,0){35}}
\put(40,30){\line(1,0){35}} \put(40,13){\line(1,0){35}}
\put(5,30)  {\line(0,-1){17}} \put(40,30){\line(0,-1){17}}
\put(75,30){\line(0,-1){17}}

\put(2,35){\makebox(0,0){$A_0$}}
\put(37,35){\makebox(0,0){$B$}}
\put(72,35){\makebox(0,0){$C$}}
\put(2,8){\makebox(0,0){$D$}}
\put(37,8){\makebox(0,0){$E$}}
\put(72,8){\makebox(0,0){$F_0$}}

\end{picture}
\end{center}
\end{minipage}
\hfill
\begin{minipage}{0.3\linewidth}
\footnotesize
\textbf{Two Agents}:  
\begin{itemize}
    \item Agent~1: Start $A_0$, Goal $F_0$  
    \item Agent~2: Start $F_0$, Goal $A_0$
\end{itemize}
All edges can be traveled in 1 time step, or an agent can wait in its current vertex.
\end{minipage}
\vspace{-2mm}
\caption{\footnotesize A toy 2D grid snippet (6 vertices). Agent~1 must go from $A_0$ to $F_0$, while Agent~2 does exactly the reverse. We illustrate $(A_0\leftrightarrow B \leftrightarrow C)$ on top row and $(D\leftrightarrow E \leftrightarrow F_0)$ below. Agents can also move vertically between the top and bottom rows. Note that $A_0$ and $F_0$ are effectively diagonally across.}
\label{fig:toy-mapf}
\end{figure}

Consider two agents on the small environment of Figure~\ref{fig:toy-mapf}. 
Agent~1 tries to go from $A_0$ to $F_0$, and agent~2 from $F_0$ to $A_0$. Let $T=4$. We introduce Boolean variables:

\[
  X_{1,A_0,0}, X_{1,B,0}, \dots, X_{2,E,3}, \dots
\]
covering all reachable $(v,t)$ in up to 4 time steps. Then $E_{1,(A_0 \to B),0}$, $E_{2,(F_0 \to E),0}$, etc., record moves. The collision-avoidance constraints forbid $X_{1,B,t}$ and $X_{2,B,t}$ from both being \texttt{true} at the same time $t$, among others. If we also choose $\Delta=1$ on top of $\xi_0 = 4$ (assuming each agent’s single-agent shortest path is 2 steps, so $\xi_0=4$), then the constraint 
\[
 \sum_{i=1}^2 \sum_{t=0}^{3} C_{i,t} \;\le\;1
\]
ensures the total cost is $\le 5$. If the solver returns SAT, we decode $X_{i,v,t}$ to see the actual routes (e.g.\ perhaps each agent waits one step to avoid collisions). If it is UNSAT, we escalate $\Delta$ or $T$ and try again.

This toy example, though small, shows the gist: each agent’s presence at each location/time is a Boolean variable, and constraints enforce legality and cost.

\subsubsection{Baseline SAT-Based Algorithm}
\label{sec:baseline}

Algorithm~\ref{alg:sat-baseline-detailed} outlines a \emph{baseline} method, adapted from \citep{surynek2016efficient}. The approach systematically increments $\Delta$ from 0 upwards, setting $\xi = \sum_{i}d_i + \Delta$ and $T = \max_i d_i + \Delta$. For each candidate, we construct a formula $\Phi(\Delta)$ capturing:

\begin{itemize}[leftmargin=*]
    \item \textbf{Agent variables}: $X_{i,v,t}$, $E_{i,(u\to v),t}$, $C_{i,t}$
    \item \textbf{Flow/collision constraints}: ensuring valid single-agent movements and no pair collisions
    \item \textbf{SoC bound}: $\sum_{i,t} C_{i,t} \le \Delta$
\end{itemize}

The first $\Delta$ for which $\Phi(\Delta)$ is satisfiable yields an SoC-optimal solution.

\begin{algorithm}[ht]
\small
\caption{Baseline SAT-based SoC Algorithm}\label{alg:sat-baseline-detailed}
\begin{algorithmic}[1]
\Require $\mathcal{G}=(\mathcal{V},\mathcal{E})$; $n$ agents with $(s_i,g_i)$; integer $\Delta_{\max}$ (optional).
\State \textbf{Compute} $d_i = \text{shortestPathLength}(s_i,g_i)$ for each agent~$i$.
\State $\xi_{0} \gets \sum_{i=1}^n d_i$;  \quad $\mu_{0} \gets \max_{i} d_i$.
\State $\Delta \gets 0$.  \quad \textit{// Extra cost budget}
\While{ \texttt{true} } 
   \State $\xi \;\gets\; \xi_0 + \Delta$;\quad $T \;\gets\; \mu_0 + \Delta$.
   \State $\Phi(\Delta)\;\gets\; \emptyset$ \quad \textit{// Start building formula}
   \For{\textbf{each agent} $i=1,\dots,n$}
       \State Create variables $X_{i,v,t}$ for $t = 0,\dots,T$, $v\in\mathcal{V}$ (or pruned by reachability).
       \State Create variables $E_{i,(u\to v),t}$ for valid edges $(u,v)\in \mathcal{E}\cup\{(u,u)\}$ and $0\le t < T$.
       \State Create cost-flag variables $C_{i,t}$ for $0\le t < T$.
       \State Add \textbf{flow constraints} ensuring consistent motion from $t$ to $t+1$.
   \EndFor
   \State Add \textbf{collision-avoidance constraints} for all pairs $(i,j)$ at each time $t$.
   \State Add \textbf{SoC bound:} \(\sum_{i=1}^n \sum_{t=0}^{T-1} C_{i,t} \;\;\le\;\Delta\).
   \State \textbf{Run SAT solver} on $\Phi(\Delta)$.
   \If{\text{solver returns \textsc{SAT}}}
       \State \textbf{Extract} assignment and decode agent paths from $X_{i,v,t}=\text{true}$.
       \State \Return \text{Collision-free solution with SoC}~$\le \xi$.
   \Else
       \State $\Delta \gets \Delta + 1$.
       \If{$\Delta > \Delta_{\max}$ (if any limit is set)}
         \State \Return \textsc{NoSolutionFound} \quad // or continue indefinitely
       \EndIf
   \EndIf
\EndWhile
\end{algorithmic}
\end{algorithm}

\paragraph{Implementation Notes.}

\begin{itemize}[leftmargin=*]
    \item \emph{Incremental SAT solving}: Instead of building $\Phi(\Delta)$ from scratch each time, one can reuse constraints from previous $\Delta$, adding only the new bounding or incremental changes. 
    \item \emph{Pruning unreachable states}: Typically, one prunes spatiotemporal states $(v,t)$ that are obviously unreachable given the agent’s start and goal (the MDD idea).
    \item \emph{Extraction of solution}: A \textsc{SAT} assignment sets certain $X_{i,v,t}$ to \texttt{true}. Reconstructing each agent’s path is straightforward by following $E_{i,(u\to v),t}$ from $t=0$ forward.
\end{itemize}

\subsubsection{Notable Variants of SAT-Based MAPF}\label{sec:sat:variants}

While the baseline captures the essence of SAT-based MAPF, various refinements yield superior performance or functionality. We summarize five major directions here. Each variant can be thought of as branching from Algorithm~\ref{alg:sat-baseline-detailed} with \emph{additions or modifications}, which we highlight in pseudocode.

\paragraph{(1) MDD-SAT with Independence Detection} 
(\citealp{surynek2017integration}).

The idea is to \emph{(i) build MDDs} for each agent rather than blindly enumerating all vertices at all $T$ layers and \emph{(ii) detect large sets of agents that cannot collide}, solving them in separate, smaller SAT formulas. If collisions appear across subgroups, they either replan or merge those subgroups. Algorithm~\ref{alg:mdd-id} sketches the changes from the baseline, with new/modified lines enclosed in \fbox{boxes}.

\begin{algorithm}[ht]
\small
\caption{MDD-SAT with Independence Detection (ID)}
\label{alg:mdd-id}
\begin{algorithmic}[1]
\Require $\mathcal{G}=(\mathcal{V},\mathcal{E})$, $n$ agents, start/goal $(s_i,g_i)$
\State \textbf{Initialize groups}: $\text{Groups}\gets\{\{1\},\dots,\{n\}\}$.
\For{$\Delta = 0,1,\dots$} 
   \State $\xi \leftarrow \sum_i d_i + \Delta$; \quad $T \leftarrow \max_i d_i + \Delta$;
   \For{\fbox{each group } $G_j \in \text{Groups}$}
      \State \fbox{\textbf{Build MDDs} for each agent in $G_j$. (Pruned state-space)}
      \State \fbox{Construct partial formula $\Phi_{G_j}(\Delta)$; solve via SAT.}
      \If{\texttt{UNSAT}}
         \State \textbf{break} (increase $\Delta$ and retry)
      \EndIf
   \EndFor
   \State \fbox{Combine solutions for each $G_j$. Check cross-group collisions.}
   \If{\fbox{no collision occurs across groups}}
      \State \textbf{return} solution with SoC~$\le\xi$.
   \Else
      \State \fbox{Re-plan or \emph{merge} conflicting groups and rerun.}
   \EndIf
\EndFor
\end{algorithmic}
\end{algorithm}

\paragraph{(2) Suboptimal MDD-SAT Variants}
(\citealp{surynek2018sub}).

One can trade off solution quality vs.\ runtime by allowing $\mathrm{SoC} \le w\cdot \mathrm{OPT}$, where $w>1$. 
The pseudocode mostly resembles the baseline, but we prematurely accept a solution once an SoC within $w\times \xi_0$ is found, or skip certain $\Delta$ steps. 
Algorithm~\ref{alg:suboptimal-mdd-sat} highlights differences in \fbox{boxes}.

\begin{algorithm}[ht]
\small
\caption{Suboptimal MDD-SAT (modified from baseline)}
\label{alg:suboptimal-mdd-sat}
\begin{algorithmic}[1]
\Require $\mathcal{G}=(\mathcal{V},\mathcal{E})$, $n$ agents; factor $w\ge 1$.
\State \textbf{Compute} $d_i$ for each agent. Let $\xi_0 = \sum_i d_i$.
\State \textbf{Set} $\xi_{\text{max}} \leftarrow w\cdot \xi_0$
\For{\fbox{$\Delta = 0,\dots,\lfloor (w-1)\cdot \xi_0\rfloor$}}
   \State $\xi \leftarrow \xi_0 + \Delta$; \quad $T \leftarrow \max_i d_i + \Delta$
   \State Build (MDD-based) formula $\Phi(\Delta)$; solve via SAT
   \If{SAT}
      \State \textbf{Decode solution} with SoC $\le \xi$
      \State \fbox{\textbf{return} suboptimal solution with factor $\le w$.}
   \EndIf
\EndFor
\end{algorithmic}
\end{algorithm}

\noindent
In practice, these suboptimal methods often solve large instances much faster, at the cost of a looser cost guarantee.

\paragraph{(3) Mutex Propagation} 
(\citealp{surynek2021mutex}).

Besides collision-avoidance, advanced \emph{mutex} constraints discover deeper conflicts among partial states, forbidding them at the formula level. 
We incorporate a \texttt{mutexCheck} routine (enclosed in \fbox{a box}), which after building the baseline constraints, enumerates or propagates \emph{mutually exclusive} states. 
An example: if $X_{i,u,t}$ and $X_{j,v,t}$ cannot ever appear in the same valid solution due to more intricate reachability conflicts, we add a clause $\neg X_{i,u,t}\lor \neg X_{j,v,t}$ to prune that partial assignment from the solver.

\begin{algorithm}[ht]
\small
\caption{Baseline $+$ \textbf{Mutex Propagation}}
\label{alg:mutex-prop}
\begin{algorithmic}[1]
\State \textbf{Build baseline} formula $\Phi(\Delta)$ as in Algorithm~\ref{alg:sat-baseline-detailed}.
\State \fbox{\textbf{mutexCheck}$(\Phi,\Delta)$:} 
   \begin{enumerate}
     \item \fbox{Identify conflicting pairs $(X_{i,u,t}, X_{j,v,t})$ from deeper analysis.}
     \item \fbox{For each pair, add clause $(\lnot X_{i,u,t} \lor \lnot X_{j,v,t})$ to $\Phi$. }
   \end{enumerate}

\State \textbf{Solve} the enriched formula $\Phi(\Delta)$ with a SAT solver.
\end{algorithmic}
\end{algorithm}

Experiments show these mutex clauses can dramatically reduce solver overhead.

\paragraph{(4) Coupling SoC and Makespan Bounds} 
(\citealp{bartak2019sat}).

Instead of enumerating $\Delta$ alone, \citeauthor{bartak2019sat} treat $(T, \xi)$ jointly. 
They prove that if a solution with $\mathrm{SoC}< \xi$ exists, it can be scheduled within $T \approx \mu_0 + (\xi-\xi_0)$. 
Hence, they systematically vary $T,\xi$ together. 
The pseudocode is essentially the baseline, but with a double parameter $(T,C)$ updated in tandem:

\begin{algorithm}[ht]
\small
\caption{Coupled SoC and Makespan \textbf{(Bartak \& \v{S}vancara, 2019)}}
\label{alg:bartak}
\begin{algorithmic}[1]
\State $D_{\mathrm{sum}}\gets \sum_i d_i$, \quad $D_{\mathrm{max}}\gets \max_i d_i$
\State $\delta \gets 0$
\While{\texttt{true}}
   \State $T \gets D_{\mathrm{max}} + \delta$; \quad $C \gets D_{\mathrm{sum}} + \delta$
   \State Build formula $\Phi(T,C)$ (ensuring each agent ends by time $T$; total cost $\le C$).
   \State \textbf{SAT-solve} $\Phi(T,C)$
   \If{SAT is true}
     \State \textbf{Return} solution with SoC~$\le C$
   \Else
     \State $\delta \gets \delta + 1$
   \EndIf
\EndWhile
\end{algorithmic}
\end{algorithm}

\paragraph{(5) DPLL(MAPF)} 
(\citealp{vcapek2021dpll}).

Finally, a more radical variant integrates MAPF conflict checks \emph{inside} the solver’s DPLL/CDCL loop, discovering collisions on partial assignments and learning conflict clauses on-the-fly. 
The high-level loop is still akin to the baseline, but the SAT engine’s internal “\emph{partial MAPF check}” step dynamically prunes collisions. 
We highlight the changed lines in \fbox{boxes}, focusing on real-time clause additions:

\begin{algorithm}[ht]
\small
\caption{DPLL(MAPF) with On-the-Fly Conflict Checks}
\label{alg:dpll-mapf}
\begin{algorithmic}[1]
\While{\texttt{true}}
   \State Build initial formula $\Phi(\Delta)$ \textit{(not necessarily all collision constraints upfront)}
   \State \textbf{Run DPLL} with partial assignments:
   \While{\text{partial assignment not complete}}
      \State \fbox{\textbf{Check} if current partial assignment forces a collision.}
      \If{\fbox{collision found}}
         \State \fbox{Create conflict clause forbidding that partial combination.}
         \State \textbf{Backtrack}
      \Else
         \State \textbf{Assign next literal} in $X_{i,v,t}$ or $E_{i,(u\to v),t}$.
      \EndIf
   \EndWhile
   \If{solution found} 
      \State \Return solution
   \Else 
      \State $\Delta \gets \Delta + 1$
   \EndIf
\EndWhile
\end{algorithmic}
\end{algorithm}

\paragraph{Comparison.}

\begin{enumerate}[leftmargin=*]
    \item \emph{MDD vs.\ naive expansion}: MDD-based approaches can drastically reduce the formula size, especially for large $T$, by cutting unreachable states.
    \item \emph{Suboptimal vs.\ optimal:} Permitting suboptimal solutions (\(w>1\)) can slash runtime at the expense of cost fidelity.
    \item \emph{Mutex propagation \& ID:} Aggressively prunes obviously incompatible states/agent groups, often crucial for large instances.
    \item \emph{Coupled bounding:} Linking $T$ and $\xi$ ensures that once satisfiable, the discovered plan is truly SoC-optimal.
    \item \emph{On-the-fly conflict checks (DPLL(MAPF))}: Potentially reduces the solver’s search depth by immediately learning collision clauses, but also requires deeper integration with the solver.
\end{enumerate}

Each variant may excel in different scenarios. 
In dense conflict domains, \emph{independence detection} helps isolate small colliding subsets. 
In large open grids, bounding $T$ and $\xi$ together can yield early detection of feasible solutions. 
For extremely large-scale or real-time needs, \emph{suboptimal} strategies may be more practical.

\subsection{Satisfiability Modulo Theories (SMT)-Based Methods}
\label{sec:smt}

Satisfiability Modulo Theories (SMT) methods constitute a second major family of \emph{compilation-based} approaches to MAPF, complementing the search-based techniques (Section~\ref{sec:search}). 
As discussed in our \emph{Problem Formulation} (Section~\ref{sec:formulation}), MAPF demands planning collision-free paths for multiple agents on either discrete or continuous representations of the environment. 
SMT-based methods leverage logical satisfiability enhanced by specialized \emph{theory} solvers (e.g., linear arithmetic, geometry) to handle key constraints such as agent motion, collision avoidance, and objective optimization \citep{surynek2019multi,surynek2019multi2,surynek2020continuous,surynek2020multi,surynek2021sum,surynek2019conflict}.

In essence, an SMT-based MAPF solver encodes agent paths, collision constraints, timing requirements, and objective functions as a set of logical clauses augmented with theory-specific formulas. 
A general SMT solver then searches for a \emph{model} (i.e., an assignment satisfying all clauses) to produce a set of collision-free paths. 
If the solver finds a collision, or if a solution violates cost bounds, additional ``nogood'' constraints are iteratively added. 
These constraints refine the solution space until either a feasible MAPF plan emerges or the problem is shown to be unsatisfiable under the imposed conditions.

\subsubsection{General Mathematical Formulation.}

Following our standard MAPF notation in Section~\ref{sec:formulation}, let $n$ be the number of agents, and let $\mathcal{G}=(\mathcal{V},\mathcal{E})$ be an undirected graph (or a continuous embedding with discretized waypoints). 
For each agent $i\in\{1,\ldots,n\}$, we define:
\begin{itemize}[leftmargin=*]
    \item A path variable \(\pi_i\) representing a sequence of (vertex, time) or (configuration, time) pairs.
    \item \(\mathrm{Cost}(\pi_i)\) as the travel cost (makespan component if one-shot, or incremental cost if we are optimizing \emph{sum of costs}).
    \item Additional Boolean or real-valued decision variables, such as
    \[
    \begin{aligned}
        X_{t,v}(i) & = \begin{cases}
        \text{true}, & \text{agent $i$ occupies vertex $v$ at time $t$}, \\
        \text{false}, & \text{otherwise};
        \end{cases}\\
        E_{t,u,v}(i) & = \begin{cases}
        \text{true}, & \text{agent $i$ starts traversing edge $(u,v)$ at time $t$},\\
        \text{false}, & \text{otherwise}.
        \end{cases}
    \end{aligned}
    \]
\end{itemize}

\noindent
To accommodate the \emph{continuous-time} setting \citep{surynek2019multi}, some SMT formulations introduce real-valued variables for time indices (rather than enumerating integer steps). 
The collision-avoidance and objective constraints then become theory clauses in, e.g., linear arithmetic or geometry. 
In discrete-time or grid-based MAPF, simpler integer arithmetic often suffices \citep{surynek2022problem}.

\begin{definition}[SMT-based MAPF Encoding (Generic)]
\label{def:smt-mapf}
Let $\Xi$ represent the set of all decision variables encapsulating agent positions, motions, and times. 
We define $F(\Xi)$ as the collection of logical clauses that enforce:
\begin{enumerate}[leftmargin=*]
    \item \emph{Initial \& Goal Feasibility.} 
    Each agent $i$ starts at $s_i$ and must eventually reach $g_i$.
    \item \emph{Path Consistency.} 
    An agent can only move to adjacent vertices (or valid continuous positions) with correct timing.
    \item \emph{Collision Avoidance.} 
    No two agents occupy or traverse overlapping regions at the same continuous or discrete time.
    \item \emph{Cost (Objective) Constraints.} 
    Either the overall makespan or sum of costs is bounded by a target value $\Lambda$. 
    Often enforced through a lazy branch-and-bound style: if a solution contradicts known or guessed $\Lambda$, it is disallowed by extra clauses.
\end{enumerate}
A \emph{model} of $F(\Xi)$ that satisfies all clauses (possibly under iterative refinements) corresponds to a set of collision-free agent paths achieving or improving the optimization goal.
\end{definition}

\noindent
\textbf{Core SMT Algorithm.} 
Algorithm~\ref{alg:smt-cbs-framework} summarizes a prototypical \emph{SMT-based} MAPF solver, which loosely follows the “conflict-based” or “lazy” refinement paradigm proposed by \citep{surynek2019multi,surynek2019multi2} and extended in subsequent works \citep{surynek2020continuous,surynek2020multi,surynek2021sum}. 
In the pseudocode, we assume an objective such as makespan or sum-of-costs is constrained by \(\Lambda\), which is progressively tightened or relaxed during the search.

\begin{algorithm}[ht]
\caption{Generic SMT-based MAPF Solver}
\label{alg:smt-cbs-framework}
\begin{algorithmic}[1]
\Require $\Sigma = (\mathcal{G}, n, \{s_i\}, \{g_i\})$, desired cost bound or optimization target $\Lambda$
\Ensure A collision-free solution with cost $\leq \Lambda$, or report \textsc{Unsatisfiable}
\vspace{3pt}
\State $\mathit{Constraints} \gets \varnothing \quad \triangleright \text{Initialize extra collision constraints}$
\State $F(\Xi,\Lambda) \gets \textsc{EncodeBaseFormulation}(\Sigma,\Lambda)$ 
\Comment{E.g., Def.~\ref{def:smt-mapf}}
\While{\textbf{true}}
    \State $\textit{result} \gets \textsc{SMTSolver}(F(\Xi,\Lambda) \cup \mathit{Constraints})$
    \If{$\textit{result} = \textsc{UNSAT}$}
        \State \Return \textsc{Unsatisfiable}
    \Else
        \State $\Pi \gets \textsc{extractSolution}(\textit{result})$ 
        \Comment{Build agent paths from solver’s model}
        \State $\mathit{collisions} \gets \textsc{validatePaths}(\Pi)$
        \If{ $\mathit{collisions} = \varnothing$}
            \State \Return $\Pi$ \Comment{Collision-free plan found}
        \Else
            \For{Each \textit{collision} $\in \mathit{collisions}$}
                \State \textit{AddDisjunctiveConstraint}($\mathit{collision}$, $\mathit{Constraints}$)
                \Comment{Forbid this collision in future solutions}
            \EndFor
        \EndIf
    \EndIf
\EndWhile
\end{algorithmic}
\end{algorithm}

\noindent
\textbf{A Simple Toy Example.}
Consider a small graph $\mathcal{G}=(\{v_1,v_2,v_3,v_4\}, \{(v_1,v_2),(v_2,v_3),(v_2,v_4)\})$ where two agents must swap positions:
\[
\begin{aligned}
    &\text{Agent~1: } s_1 = v_1, \; g_1 = v_4,\\
    &\text{Agent~2: } s_2 = v_3, \; g_2 = v_2.\\
\end{aligned}
\]
In a step-based model, we introduce Boolean variables such as $X_{t,v}(1)$ and $X_{t,v}(2)$ indicating which vertex each agent occupies at each time $t$, plus potential edge-traversal variables $E_{t,u,v}(i)$. 
Initially, we only encode constraints ensuring each agent can follow a path from start to goal. 
If the solver returns a plan where both agents collide (e.g., if they try to pass through vertex $v_2$ at the same time), we detect that collision through a validation check, and add a \emph{disjunction}:
\[
\neg X_{t,v_2}(1) \;\vee\;\neg X_{t,v_2}(2),
\]
forcing the solver to avoid that specific simultaneous occupation in future attempts. 
Repeating this refinement leads to a collision-free schedule.

\subsubsection{SMT Variants for MAPF}
While the above description outlines the fundamental use of SMT in MAPF, multiple variants have emerged to address specific objectives, environments, and agent capabilities:

\begin{description}
    \item[\textbf{Continuous-Time \& Geometric Agents.}] 
    Early works, such as \citep{surynek2019multi,surynek2019multi2}, consider continuous trajectories and geometric overlap (e.g., circular agents in 2D). 
    The solver lazily introduces real-valued time variables and constraints from geometry-based collision checks. 
    These approaches are powerful in capturing uncountably many possible collision instants, although they can face high complexity in large or sparse domains.
    
    \item[\textbf{Makespan-Optimal vs. Sum-of-Costs.}]
    SMT encodings also differ by objective. 
    \citep{surynek2019multi,surynek2020continuous} focus on minimizing completion time (makespan), introducing incremental bounding on $\Lambda$. 
    In contrast, \citep{surynek2021sum} extends these encodings to the more challenging \emph{sum-of-costs} objective by adding designations on how waiting times or arrival times accumulate across agents. 
    A solver iteratively refines an upper bound on $\sum_i \mathrm{Cost}(\pi_i)$ until no collision remains or the problem proves infeasible.

    \item[\textbf{Hybrid Conflict-Resolution Schemes.}]
    Works such as \citep{surynek2020multi} integrate \emph{Conflict-Based Search (CBS)} with SMT solvers. 
    Instead of classical CBS branching at collisions, the collisions are turned into disjunctive constraints fed back into an SMT loop.
    The approach capitalizes on the clause-learning capabilities of SMT to prune large swaths of invalid solution space.

    \item[\textbf{Scalability and Performance.}] 
    As reported by \citep{surynek2019conflict}, current SMT methods excel on small- to medium-scale problems, particularly in complex or continuous environments that break typical discrete search heuristics. 
    However, they may struggle on extremely large and sparse domains or with thousands of agents, where simpler decentralized or priority-based methods (Section~\ref{sec:search}) can be more scalable.
\end{description}

\noindent
\textbf{Illustrative Pseudocode for SMT Variants.}
Algorithm~\ref{alg:smt-sum-costs} outlines a \emph{sum-of-costs} oriented version (motivated by \citep{surynek2021sum}) that highlights how cost bounding and collision resolution intertwine. 
New or modified lines from the basic template (Algorithm~\ref{alg:smt-cbs-framework}) are labeled \emph{(*)}.

\begin{algorithm}[ht]
\caption{SMT Solver for MAPF Under Sum-of-Costs (\emph{Key Differences Highlighted})}
\label{alg:smt-sum-costs}
\begin{algorithmic}[1]
\Require Problem $\Sigma$, initial upper bound on sum-of-costs $\Lambda$, collision constraints $\mathit{Constraints}$
\State $F(\Xi,\Lambda) \gets \textsc{EncodeBaseFormulation}(\Sigma,\Lambda)$
\While{\textbf{true}}
    \State $\textit{assignment} \gets \textsc{SMTSolver}(F(\Xi,\Lambda)\,\cup\,\mathit{Constraints})$
    \If{$\textit{assignment} = \textsc{UNSAT}$}
        \State $\Lambda \gets \Lambda + 1$ \quad \textit{(*) Increase sum-of-costs bound}
        \State \textbf{continue}
    \EndIf
    \State $\Pi \gets \textsc{extractSolution}(\textit{assignment})$
    \If{$\textsc{validatePaths}(\Pi)\text{ has collisions}$}
        \State $\mathit{collisions} \gets \textsc{detectCollisions}(\Pi)$
        \For{Each c in collisions}
            \State \textsc{AddDisjunctiveConstraint}(c,$\mathit{Constraints}$)
        \EndFor
    \Else
        \If{$\textsc{Cost}(\Pi) \le \Lambda$} 
            \State \Return $\Pi$ \quad \textit{(Found feasible solution under current bound)}
        \Else
            \State \textsc{AddCostNogood}(\(\Pi\),$\mathit{Constraints}$) \quad \textit{(*) Forbid this high-cost plan}
        \EndIf
    \EndIf
\EndWhile
\end{algorithmic}
\end{algorithm}

\subsubsection*{Unified Pseudocode Across SMT Variants}

While Algorithm~\ref{alg:smt-sum-costs} focuses on the sum-of-costs objective, a variety of SMT-based MAPF methods share a common “collisions-as-constraints” logic. 
Below, Algorithm~\ref{alg:unified-smt-var} illustrates a single compact pseudocode unifying major differences among:
1) \textbf{Makespan vs. Sum-of-Costs Objectives}, and  
2) \textbf{Discrete vs. Continuous Settings}.  
We encode these differences through parameterized procedures or \emph{conditional lines}, where each variant simply activates the relevant module.

\begin{algorithm}[htb!]
\caption{Unified SMT-Based MAPF Solver Covering Multiple Variants}
\label{alg:unified-smt-var}
\begin{algorithmic}[1]
\Require 
  Problem $\Sigma = (\mathcal{G},\,n,\,\{s_i\},\,\{g_i\})$, 
  ObjectiveType $\in\{\textsc{Makespan}, \textsc{SumOfCosts}\}$,
  SettingType $\in\{\textsc{Discrete}, \textsc{Continuous}\}$,
  initial cost/makespan bound $\Lambda$,
  collision set $\mathit{Colls} \leftarrow \emptyset$
\State $F(\Xi,\Lambda) \gets \textsc{EncodeBase}(\Sigma,\text{ObjectiveType}, \text{SettingType}, \Lambda)$
\Comment{Build the base SMT formula, e.g. discrete or continuous constraints}
\While{\textbf{true}}
    \State $\textit{assignment} \gets \textsc{SMTSolver}(F(\Xi,\Lambda) \,\cup\, \textsc{CollisionConstraints}(\mathit{Colls}))$
    \If{$\textit{assignment} = \textsc{UNSAT}$}
        \State \textsc{AdjustBound}$(\Lambda)$ \Comment{Increment or otherwise modify $\Lambda$}
        \State \textbf{continue}
    \EndIf
    \State $\Pi \gets \textsc{ExtractPaths}(\textit{assignment}, \text{SettingType})$
    \State $\mathit{newColls} \gets \textsc{CheckCollisions}(\Pi, \text{SettingType})$
    \If{$\mathit{newColls}=\varnothing$}
        \If{($\textsc{Cost}(\Pi)\le \Lambda$ \textbf{and} ObjectiveType = \textsc{SumOfCosts}) 
         \textbf{or} 
        ($\textsc{Makespan}(\Pi)\le \Lambda$ \textbf{and} ObjectiveType = \textsc{Makespan})}
            \State \Return $\Pi$ \Comment{Feasible \emph{and} optimal under bound $\Lambda$}
        \Else
            \State \textsc{AddCostNogood}$(\Pi,F(\Xi,\Lambda))$ 
            \Comment{If cost or makespan still too high}
        \EndIf
    \Else
        \State \textsc{ForbidCollisions}$(\mathit{newColls}, F(\Xi,\Lambda))$
        \Comment{Disjunctive constraints for each collision}
        \State $\mathit{Colls} \gets \mathit{Colls} \,\cup\, \mathit{newColls}$
    \EndIf
\EndWhile
\end{algorithmic}
\end{algorithm}

\paragraph{Explanation and Comparison.}  
\begin{itemize}[leftmargin=*]
    \item \textbf{EncodeBase}~(\(\Sigma,\text{ObjectiveType},\text{SettingType}, \Lambda\)):  
    In a \textsc{Discrete} setting, this subroutine creates Boolean variables $X_{t,v}(i)$ for integer time $t$.  
    In a \textsc{Continuous} setting, it introduces real-valued times or piecewise-linear motion segments.  
    The objective or constraint on $\Lambda$ is likewise chosen according to \textsc{Makespan} or \textsc{SumOfCosts}.
    
    \item \textbf{CheckCollisions}~(\(\Pi,\text{SettingType}\)):  
    If \textsc{Discrete}, collisions are time-step conflicts.  
    If \textsc{Continuous}, we must detect geometric overlap in real-time.  
    Either yields a set of “collision tuples” to be forbidden next iteration.

    \item \textbf{AdjustBound}~(\(\Lambda\)):  
    For \textsc{Makespan}, we might do $\Lambda \leftarrow \Lambda + 1$.  
    For \textsc{SumOfCosts}, incrementing can be more nuanced (e.g., bounding total arrival time).  

    \item \textbf{AddCostNogood}~(\(\Pi,F\)):  
    Only relevant for sum-of-costs.  If $\sum_i \mathrm{Cost}(\pi_i)$ still exceeds $\Lambda$, we forbid that specific combination of time assignments in the SMT formula.  In a makespan context, no such line is needed (or it might just forbid solutions longer than $\Lambda$).
\end{itemize}

Through these modular conditions, a single loop can accommodate different SMT-based MAPF variants. 
For instance, \citep{surynek2019multi} and \citep{surynek2020continuous} use near-identical logic but focus on \textsc{Makespan} in continuous space; \citep{surynek2021sum} uses \textsc{SumOfCosts}; \citep{surynek2019multi2} and \citep{surynek2020multi} embed a Conflict-Based Search style branching into the loop, essentially refining collisions as disjunctions.  
Hence, Algorithm~\ref{alg:unified-smt-var} encapsulates their shared pattern: obtaining a candidate plan from an SMT solver, validating it for collisions, and either finalizing the plan or refining constraints to exclude collisions or reduce cost/makespan.

\subsubsection{Discussion and Practical Usage.}
SMT-based MAPF methods exemplify a powerful synergy between high-level collision refinement (similar to conflict-based search) and the clause-learning of SAT/SMT solvers. 
They are particularly appealing when:
\begin{itemize}[leftmargin=*]
    \item \emph{Continuous time and geometry} play a central role, as direct discretization quickly becomes intractable.
    \item The problem dimension is moderate, allowing the solver to effectively prune the space of collisions via advanced theory constraints.
    \item One desires optimal or near-optimal MAPF solutions under complex constraints (e.g., sum-of-costs, multi-criteria routing).
\end{itemize}
However, these methods often face scalability issues at large agent counts or when the environment is highly sparse and can be tackled more efficiently by specialized heuristics. 
Still, the continuous-time and geometric formalisms that SMT can handle natively make them an invaluable tool in scenarios unsuited to purely discrete or purely search-based approaches.

In summary, SMT-based MAPF formulations unify classical path feasibility with advanced collision detection and cost-based refinement in a single constraint-solving framework. 
Subsequent developments have shown that each variant—whether aimed at makespan, sum-of-costs, or continuous-space coverage—substantially broadens the range of solvable MAPF challenges, addressing aspects that classical search-based or purely ILP/SAT-based methods struggle to handle alone.



\subsection{Other Compilation-Based Methods}
\label{sec:compilation-other}

In addition to SAT (\S\ref{sec:compilation-sat}) and SMT (\S\ref{sec:smt}) formulations, several other \emph{compilation-based} approaches for Multi-Agent Path Finding (MAPF) have also emerged. 
This section presents three major classes of such methods: (i) \textbf{Constraint Satisfaction Problem (CSP)} encodings, (ii) \textbf{Answer Set Programming (ASP)} representations, and (iii) \textbf{Mixed-Integer Programming (MIP)} formulations. 
Each line of work offers a unique perspective on how to encode the MAPF problem into well-established computational frameworks, providing additional flexibility or efficiency gains under certain conditions. 
We first detail the mathematical modeling (objective, decision variables, and constraints) for each class, then illustrate their core algorithmic pipelines with pseudocode and small toy examples. 
We conclude by comparing these methods along key dimensions such as scalability, solution quality, and ease of implementation.

\subsubsection{CSP Formulation}
\label{sec:csp}

\paragraph{Mathematical Modeling.}
In a typical CSP-based approach (see, e.g., \citep{wang2019new}), one treats each agent as a variable whose \emph{domain} is the set of possible (finite-horizon) paths from $s_i$ to $g_i$. Formally, let $n$ be the number of agents, and let $X_i$ be the CSP variable for agent~$i$:
\[
   \text{Var} \;=\; \{X_1, X_2, \dots, X_n\}, 
   \quad
   \text{Dom}(X_i) \;=\; \Bigl\{\pi_i \mid \pi_i \text{ is a path from } s_i \text{ to } g_i \text{ with length} \leq T\Bigr\},
\]
where $T$ is a time horizon. The CSP \emph{constraints} enforce collision-avoidance: namely, no two chosen paths conflict. If $\pi_i(t)$ denotes the position of agent~$i$ at time $t$ (along path $\pi_i$), then for any pair of agents $i,j$, the constraint
\[
   \bigl(\pi_i(t) \neq \pi_j(t)\bigr) \;\;\wedge\;\; \bigl( (\pi_i(t) \neq \pi_j(t+1)) \vee (\pi_i(t+1) \neq \pi_j(t)) \bigr)
\]
ensures that no agents share the same vertex simultaneously or pass each other on the same edge in opposite directions.

Cost objectives (e.g., makespan or sum-of-costs) can be incorporated by searching for a \emph{smallest} $T$ such that the CSP is satisfiable, or by encoding additional constraints/variables tracking how many time steps each agent actually uses before completing its path.

\paragraph{Core Algorithm and Pseudocode.}
Building on a matrix-based path-counting idea, \citep{wang2019new} propose a solver that repeatedly:
1) selects the most \emph{constrained} agent (i.e., the one with the fewest feasible paths left),
2) commits to one of its available paths,
3) prunes the path set of other agents to remove newly inflicted collisions,
4) restarts if it becomes impossible to assign further agents.

A high-level sketch follows.

\begin{algorithm}[ht]
\small
\caption{CSP-Based MAPF Solver (\citealp{wang2019new} style)}
\label{alg:csp}
\begin{algorithmic}[1]
\Require Graph $G=(V,E)$, agents $A=\{a_1,\dots,a_n\}$ with start/goal $(s_i,g_i)$, horizon $T$, and max restarts $R$.
\Ensure A conflict-free assignment of paths or report failure.
\For{$r \gets 1$ to $R$}
   \State $P \leftarrow \emptyset$ \Comment{Assigned paths so far}
   \State $\textsc{Domain}(a_i) \gets \{\text{all length-}\le T\text{ paths from }s_i \text{ to }g_i\}$ 
   \State $\mathit{conflictFree} \leftarrow \text{true}$
   \While{$\exists\, a_i \text{ unassigned}$ \textbf{and} $\mathit{conflictFree}$}
       \State $a^* \gets \operatorname{argmin}_{a_i\text{ unassigned}} \Bigl|\textsc{Domain}(a_i)\Bigr|$
       \If{$\Bigl|\textsc{Domain}(a^*)\Bigr| = 0$}
          \State $\mathit{conflictFree} \leftarrow \text{false}$; \quad \textbf{break}
       \EndIf
       \State $\pi^* \gets \textsc{PickOnePath}\bigl(\textsc{Domain}(a^*)\bigr)$
       \State $P \leftarrow P \cup \{(a^*, \pi^*)\}$
       \State $\textsc{UpdateDomains}\bigl(\{\textsc{Domain}(a_i)\}_{i}, \pi^*\bigr)$ \Comment{Prune collisions}
   \EndWhile
   \If{$\mathit{conflictFree}$}
      \State \Return $P$ \Comment{All agents assigned conflict-free paths}
   \EndIf
\EndFor
\State \Return \textsc{NoSolutionFound}
\end{algorithmic}
\end{algorithm}

\paragraph{Toy Example and Explanation.}
Consider a grid with 4 cells $\{A,B,C,D\}$ arranged in a square, and two agents $a_1$, $a_2$. Suppose $a_1$ must go from $A$ to $D$, and $a_2$ from $C$ to $B$. One enumerates all $T$-step routes for each agent (e.g., $T=3$), then forbids any assignment that yields a direct collision or an edge swap. When picking $a_1$'s path, the solver prunes $a_2$'s domain to remove collisions. This procedure continues until either a complete valid set is found or no feasible assignment remains. Such CSP views can be implemented with off-the-shelf constraint solvers, although they may require specialized heuristics to handle large instances efficiently.

\paragraph{Mathematical Note.}
While classical CSPs entail purely \emph{discrete} decision variables, one can also embed certain numeric constraints for more refined collision conditions (e.g., partial occupancy of an edge). However, most CSP-based MAPF treatments assume time-discrete steps and single-vertex occupancy constraints, preserving a purely discrete constraint satisfaction framework.

\subsubsection{ASP Formulation}
\label{sec:asp}

\paragraph{Modeling and Background.}
\emph{Answer Set Programming} (ASP) is a logic-based paradigm suited for complex combinatorial search problems. The MAPF domain is encoded via sets of logical rules specifying agent positions, movement actions, collision avoidance, and cost minimization. A specialized ASP solver then finds \emph{answer sets}—that is, consistent truth-value assignments that satisfy all rules under the stable-model semantics.

In the ASP-based MAPF approach of \citep{gomez2020solving,gomez2021compact}, the environment is discretized into a grid or directed graph, time steps $t \in \{0,\dots,T\}$, and Boolean predicates:
\[
   \texttt{at}(a,x,y,t), 
   \quad
   \texttt{exec}(a,m,t),
   \quad
   \texttt{cost}(a,t,1),
   \;\dots
\]
The sum-of-costs objective is encoded via integer optimization statements in ASP, and the solver incrementally explores increasing makespan values $T$ until a feasible plan emerges. Additional constraints impose that no two agents occupy or swap the same vertex.

\paragraph{Formal Encoding Highlights.}
Define:
\begin{itemize}[leftmargin=*]
    \item $\texttt{move}(a,m,t)$: true if agent $a$ executes move $m$ at time $t$. 
    \item $\texttt{reachable}(x,y,t)$: true if cell $(x,y)$ can be occupied at time $t$ by any legal path.
\end{itemize}
Collision constraints typically appear as linear rules:
\[
   \texttt{:- at}(a,x,y,t),\;\texttt{at}(b,x,y,t),\; a \neq b.
\]
(``\(\texttt{:-}\)'' is an ASP notation for stating that this combination is forbidden.)
Cost minimization can be expressed through weak constraints \(\sim\!\! \texttt{cost}(a,t,1)[1@priority]\), awarding a penalty whenever $\texttt{cost}(a,t,1)$ is true. The ASP solver aims to minimize these penalty sums, effectively capturing sum-of-costs.

\paragraph{Pseudocode Structure.}
Although ASP solutions are typically not written in procedural style, a top-level algorithm might look like:

\begin{algorithm}[ht]
\small
\caption{Asp-MAPF-Solver (Sum-of-Costs Based)}
\label{alg:asp}
\begin{algorithmic}[1]
\State $T_{\text{min}} \;\gets\max_i \bigl(d_i\bigr)$ \Comment{Longest single-agent shortest path}
\For{$T \gets T_{\text{min}}, T_{\text{min}}+1,\dots$}
   \State $\Pi(T) \;\gets\;\textsc{GenerateASPEncoding}(G,A,T)$ 
   \State \textit{AS} $\gets \textsc{SolveASP}(\Pi(T))$
   \If{\textit{AS} $\neq \text{\emph{UNSAT}}$}
      \State \textit{BestModel} $\gets$ \textsc{GetMinCostModel}(\textit{AS})
      \If{\textit{BestModel.cost} \text{ is globally minimal}} 
         \State \textbf{return} \textit{BestModel}
      \EndIf
   \EndIf
\EndFor
\end{algorithmic}
\end{algorithm}

\paragraph{Toy Example.}
Consider two agents in a $3\times2$ grid. A typical ASP snippet could define:
\begin{verbatim}
time(0..T).
agent(a1; a2).
pos(0..2, 0..1).  % x in {0,1,2}, y in {0,1}

at(A, X, Y, 0) :- initial(A, X, Y).
% Agent can move to next cell if adjacent and not blocked
at(A, X2, Y2, T+1) :- at(A, X1, Y1, T), exec(A, move(X1,Y1,X2,Y2), T).
% Collision avoidance
:- at(A1, X, Y, T), at(A2, X, Y, T), A1 != A2.
\end{verbatim}
plus rules for cost penalties and a final directive to \(\#minimize\) the total cost. The solver tries each $T$ in ascending order, stopping when a feasible (and cost-minimal) model emerges.

\paragraph{Advantages.}
ASP-based encodings can be quite concise and benefit from advanced conflict-driven learning and optimization features of modern ASP solvers like \texttt{clingo}. They handle densely constrained scenarios well and allow for linear-sized encodings in terms of agents, as shown in \citep{gomez2020solving,gomez2021compact}.

\subsubsection{MIP Formulation}
\label{sec:mip}

\paragraph{Modeling Rationale.}
\emph{Mixed-Integer Programming} (MIP) formulations encode MAPF by enumerating time-expanded flows or by representing each agent’s path as a sequence of discrete decisions. The solver then uses linear constraints plus an objective function to enforce conflict-free routing and minimal cost. Advanced branch-and-cut routines prune the search, and \emph{branch-and-price} or \emph{branch-and-cut-and-price (BCP)} can further improve scalability. Notable examples include \citep{lam2022branch,lam2023exact}.

\paragraph{Decision Variables.}
One canonical MIP approach introduces a large (though potentially implicit) set of path variables:
\[
  \lambda_{p} \in \{0,1\},
\]
indicating whether a particular path $p$ is used by an agent. Let $\mathcal{P}_a$ be all possible paths for agent~$a$. Then,

\begin{equation}
\begin{aligned}
  \min \qquad & \sum_{a} \sum_{p \in \mathcal{P}_a} c_p \,\lambda_{p}, \\
  \text{s.t.}\quad & \sum_{p \in \mathcal{P}_a} \lambda_{p} = 1, \quad \forall a, \\
                   & \sum_{a} \sum_{p \in \mathcal{P}_a} x^p_v \,\lambda_{p} \;\;\le\; 1, \quad \forall v\in V,\\
                   & \text{(edge conflict constraints)},\;\; \lambda_{p} \in \{0,1\}.
\end{aligned}
\end{equation}
Here $c_p$ is the length (or cost) of path $p$, and $x^p_v=1$ if path $p$ visits vertex $v$. 
Similar definitions exist for edges. 
An alternative is a direct flow-based MIP with binary $x_{i,v,t}$ variables representing agent $i$’s occupancy at vertex $v$ and time $t$.

\paragraph{Core Idea: Branch-and-Cut-and-Price.}
Following \citep{lam2022branch,lam2023exact}, one rarely enumerates all $\mathcal{P}_a$ at once. Instead:
\begin{enumerate}[leftmargin=*]
    \item \emph{Column generation (pricing)}: generate new path columns with negative reduced cost by solving single-agent subproblems guided by dual variables.
    \item \emph{Cut separation}: detect collisions in the fractional solution and add linear constraints (cuts) forbidding them in the next re-optimization step.
    \item \emph{Branching}: if the solution remains fractional, pick a branching rule (e.g., forcing an agent to use, or not use, a particular edge) and create child subproblems. 
\end{enumerate}
This yields a \emph{BCP} framework that is guaranteed to converge to an optimal integral solution while handling large path sets implicitly.

\paragraph{Illustrative Pseudocode.}
Algorithm~\ref{alg:mip} outlines a simplified branch-and-cut-and-price loop.

\begin{algorithm}[ht]
\small
\caption{High-Level BCP Algorithm for MAPF (\citealp{lam2022branch,lam2023exact})}
\label{alg:mip}
\begin{algorithmic}[1]
\State \textbf{Initialize Master Problem}: no columns, no collision cuts.
\State \textbf{Push}(Master Problem) into node queue.
\While{\text{node queue not empty}}
   \State $N \gets \textbf{Pop}(\text{node queue})$
   \Repeat
      \State \textbf{Pricing Step:} For each agent, solve single-agent shortest path with dual-sensitive costs; add any column $p$ with negative reduced cost.
      \State \textbf{Cut Separation:} Check collisions in the fractional solution; add constraints forbidding these collisions if discovered.
      \State Re-solve the LP relaxation.
   \Until{\text{no new columns/cuts found}}
   \If{\text{the current solution is integral}}
      \State \text{Update global best if cost improved; prune node.}
   \Else
      \State \textbf{Branch:} pick a fractional condition, create 2 child nodes
      \State \textbf{Push} children into node queue
   \EndIf
\EndWhile
\State \Return \text{global best solution found}
\end{algorithmic}
\end{algorithm}

\paragraph{Toy Example.}
Consider again a $2\times 2$ grid with two agents swapping diagonals. 
In a column-based MIP, each agent $a$ has a (potentially infinite) set of paths. 
The dual variables for vertices at each time step raise or lower path costs, guiding the solver to prefer paths that avoid collisions. 
Collision cuts appear if, for instance, the fractional solution attempts to use a vertex or edge simultaneously across multiple agents. 
While small grids are easily solved by simpler means, the power of BCP emerges in larger or more intricate domains.

\subsection{Comparative Analysis}\label{sec:comparative}

Compilation-based solvers for MAPF can be broadly organized according to their underlying logical or mathematical framework (e.g., CSP, SAT, SMT, ASP, MIP) and the strategies they employ for conflict resolution (e.g., complete expansion vs.\ incremental constraint addition). Although they share the common goal of translating MAPF into a declarative formalism handled by general-purpose solvers, subtle differences in modeling choices often lead to varying levels of scalability, solution quality, and ease of implementation. This section synthesizes insights from the preceding discussion, as well as from various recent works~\citep{surynek2019tour,surynek2021conceptual,acha2021new}, to provide a more unified view of the advantages and drawbacks of each approach.

From a \textit{modeling complexity} perspective, Boolean SAT and ASP formulations often require discretizing agent positions in both space and time, resulting in large but structurally uniform propositional encodings. 
Works such as \citep{acha2021new} illustrate how specialized Boolean variables (e.g., shift-based encodings) can streamline representing collision avoidance (notably swap and follow conflicts) and enable conflict-driven clause learning. 
These streamlined encodings also appear in \textit{SMT-based} approaches, which embed additional theory solvers for linear arithmetic or geometry~\citep{surynek2019tour}. 
By contrast, \textit{MIP-based} methods~\citep{surynek2021conceptual} rely on linear constraints and integer decision variables to handle collision avoidance and objective functions. 
While MIP natively captures sum-of-costs or makespan objectives with linear constraints, it requires non-trivial means (often branch-and-cut or branch-and-price) to handle collisions incrementally without enumerating a prohibitive number of path variables. 
\textit{CSP-based} methods occupy yet another niche, treating possible paths as the domain of agent-specific variables and pruning collisions through constraint propagation.

Regarding \textit{scalability}, purely SAT-based or ASP-based approaches are highly effective on small and medium-size instances, particularly if the environment is dense in agents or obstacles. 
Their conflict-driven clause-learning (CDCL) mechanisms can prune a combinatorial search space quickly when collisions frequently arise~\citep{acha2021new}. 
However, as the graph or number of agents grows larger and the state space becomes more sparse, MIP-based or hybrid approaches sometimes outperform pure SAT/ASP due to more flexible branching rules and efficient ways of generating only profitable path columns~\citep{surynek2021conceptual}. 
SMT-based methods tend to excel when continuous or geometric formulations of MAPF demand sophisticated theory solvers, making them advantageous in robotic applications featuring complex dynamics~\citep{surynek2019tour}. 
CSP-based designs typically scale well when agent domains (i.e., feasible paths) remain manageable, although they may require careful variable-ordering heuristics to compete with other frameworks.

A \textit{further dimension} of comparison concerns the \textit{degree of conflict resolution} embedded in the solver. 
In a one-shot encoding (as in classical SAT or MIP expansions), all potential collision constraints may be included upfront, risking very large formulas. 
By contrast, \textit{incremental or lazy methods} iteratively detect actual collisions in candidate solutions and add new constraints to forbid them~\citep{surynek2019tour,surynek2021conceptual}. 
This strategy parallels the logic of Conflict-Based Search (CBS). 
Indeed, \citep{surynek2019tour} demonstrates the promise of a DPLL(MAPF) paradigm that refines constraints within the solver loop, mirroring how CBS resolves collisions one by one. 
Hybrids like SAT/SMT-CBS effectively unify the conflict-resolution strengths of CBS with the clause-learning prowess of SAT/SMT solvers, often achieving competitive results on a wide range of MAPF instances~\citep{gange2019lazy}.

Finally, \textit{implementation complexity} and \textit{practical considerations} also play decisive roles. 
For example, ASP encodings allow concise expression of MAPF rules at the cost of adopting APS- or MaxSAT-specific toolchains~\citep{acha2021new}, whereas MIP-based solutions can leverage extensively developed commercial solvers but must handle potential fractional assignments with custom cuts. 
Similarly, pure SAT approaches enjoy mature, highly optimized CDCL engines but might lack direct means for integrating advanced numeric constraints unless extended to SMT. 
Each framework can thus be advantageous in different industrial or research contexts, influenced by the size of the instance, the dynamic or continuous nature of the environment, desired optimality guarantees (makespan vs.\ sum-of-costs), and available solver technologies.

\begin{sidewaystable}
\centering
\caption{A comparative overview of major compilation-based MAPF approaches.}
\label{tab:compilation-comparison}
\renewcommand{\arraystretch}{1.2}
\resizebox{.9\textwidth}{!}{%
\begin{tabular}{p{0.10\linewidth} | p{0.2\linewidth} | p{0.2\linewidth} | p{0.2\linewidth} | p{0.20\linewidth} | p{0.10\linewidth}}
\hline
\textbf{Approach} & \textbf{Modeling Core \& Representation} & \textbf{Collision Handling Mechanism} & \textbf{Objective Handling} 
& \textbf{Strengths / Scalability} & \textbf{References}\\
\hline

\textbf{CSP} 
& 
\begin{itemize}[leftmargin=*, noitemsep]
  \item Each agent as a CSP variable. 
  \item Domains: all feasible paths for a given time horizon.
\end{itemize}
& 
\begin{itemize}[leftmargin=*, noitemsep]
  \item Constraints forbid overlapping domains (occupation) at each timestep.
  \item Often updated incrementally when collisions are detected.
\end{itemize}
& 
\begin{itemize}[leftmargin=*, noitemsep]
  \item Typically done by bounding time and searching for a valid assignment.
  \item Extensions allow cost-based filtering or iterative deepening.
\end{itemize}
& 
\begin{itemize}[leftmargin=*, noitemsep]
  \item Straightforward to implement with generic solvers.
  \item Good for moderate agent populations if domain sizes remain tractable.
\end{itemize}
& 
\citep{wang2019new}\\

\hline
\textbf{SAT} 
& 
\begin{itemize}[leftmargin=*, noitemsep]
  \item Propositional encoding with Boolean variables (e.g., $X_{i,v,t}$).
  \item Often uses time expansion or shift-based encodings.
\end{itemize}
& 
\begin{itemize}[leftmargin=*, noitemsep]
  \item All or most collision constraints included as clauses (vertex/edge conflicts).
  \item Incremental or lazy additions if employing SMT-CBS style.
\end{itemize}
& 
\begin{itemize}[leftmargin=*, noitemsep]
  \item Iteratively increasing time / cost bound until satisfiable.
  \item Can add pseudo-Boolean constraints or weighted clauses for sum-of-costs.
\end{itemize}
& 
\begin{itemize}[leftmargin=*, noitemsep]
  \item CDCL solvers excel in dense conflict settings.
  \item Effective for small-to-medium graphs with frequent collisions.
\end{itemize}
& 
\citep{stern2019multi,surynek2016efficient,surynek2019tour}\\

\hline
\textbf{SMT} 
& 
\begin{itemize}[leftmargin=*, noitemsep]
  \item Similar to SAT, but enriched with theories (e.g.\ linear arithmetic).
  \item Can represent discrete or continuous-time positions.
\end{itemize}
& 
\begin{itemize}[leftmargin=*, noitemsep]
  \item Collision constraints lazily added if a partial assignment yields conflicts.
  \item Solver learns theory-specific lemmas for geometry / timing.
\end{itemize}
& 
\begin{itemize}[leftmargin=*, noitemsep]
  \item Provides flexible control over makespan/sum-of-costs via quantifiers or linear constraints.
  \item Iterative bounding or branch-and-bound mechanisms.
\end{itemize}
& 
\begin{itemize}[leftmargin=*, noitemsep]
  \item Strong for continuous/complex domains where standard SAT is less natural.
  \item Gains power from advanced theory solvers.
\end{itemize}
& 
\citep{surynek2019multi,surynek2020continuous}\\

\hline
\textbf{ASP / MaxSAT} 
& 
\begin{itemize}[leftmargin=*, noitemsep]
  \item Logic-based rules, stable-model semantics.
  \item ASP typically encodes states/time discretely; can incorporate shift-based ideas.
\end{itemize}
& 
\begin{itemize}[leftmargin=*, noitemsep]
  \item Conflict constraints as integrity constraints or cardinalities (no overlapping).
  \item Swap/follow rules captured by dedicated logical clauses.
\end{itemize}
& 
\begin{itemize}[leftmargin=*, noitemsep]
  \item Weighted rules used for sum-of-costs or makespan.
  \item MaxSAT or ASP built-in optimization features (weak constraints).
\end{itemize}
& 
\begin{itemize}[leftmargin=*, noitemsep]
  \item Very concise problem definitions; strong solver support (e.g.\ \texttt{clingo}).
  \item Often outperforms search-based methods in dense, small-to-medium maps.
\end{itemize}
& 
\citep{acha2021new,gomez2020solving}\\

\hline
\textbf{MIP} 
& 
\begin{itemize}[leftmargin=*, noitemsep]
  \item Formulate MAPF as integer linear programs or flow-based models.
  \item Variables may represent path selections or time-expanded flows.
\end{itemize}
& 
\begin{itemize}[leftmargin=*, noitemsep]
  \item Pairwise collision constraints linearized via big-M or cut generation.
  \item Branch-and-cut(-and-price) to handle large path sets dynamically.
\end{itemize}
& 
\begin{itemize}[leftmargin=*, noitemsep]
  \item Direct handling of sum-of-costs or makespan via linear objective.
  \item Typically solved by powerful commercial or open-source MIP solvers.
\end{itemize}
& 
\begin{itemize}[leftmargin=*, noitemsep]
  \item Scales well if advanced column generation or cuts are used.
  \item Integrates naturally with real-time scheduling or numeric constraints.
\end{itemize}
& 
\citep{lam2022branch,surynek2021conceptual}\\

\hline
\end{tabular}%
}
\end{sidewaystable}

In summary, CSP, SAT, SMT, ASP, and MIP formulations for MAPF embody distinct trade-offs in how they encode agent movement, collisions, and cost functions. 
Recent work~\citep{surynek2019tour,surynek2021conceptual,acha2021new} consolidates these approaches in an increasingly unified view, often adding conflict-refinement mechanisms that resemble CBS at a higher level (or inside the solver itself). 
While no single framework dominates under all conditions, these compilation-based families collectively showcase how general-purpose solvers—when armed with informed modeling choices—can handle an ever-expanding scope of MAPF problems, from classic grid-based tasks to continuous or high-dimensional coordination challenges.



\section{Learning-Augmented Classic Solvers}\label{sec:augmenting}





Building on the classical methods outlined in Sections~\ref{sec:search} and~\ref{sec:compilation}, a natural evolution is to incorporate learning-based components into well-established MAPF solvers. 
Rather than discarding decades of research on search-based and compilation-based methods, these approaches seek to \emph{augment} classical pipelines by replacing or supporting specific modules with learned policies or heuristics. 
The motivation stems from the complementary strengths of each paradigm: 
while classical solvers offer rigid theoretical guarantees and can often handle a large number of agents, they tend to degrade in performance under real-time constraints or dynamic environments. 
Learning-based methods, by contrast, excel at adapting to complex, evolving scenarios and can reduce the heavy reliance on human-engineered heuristics, yet often struggle to match the scalability of purely classical techniques when the agent count or environment size becomes large.

This hybrid perspective is consistent with the broader direction introduced in Section~\ref{sec:intro}, where data-driven methods offer the flexibility to adapt to uncertainties and partial observability, while classical solvers provide efficient frameworks for collision avoidance and optimality guarantees in static or moderately changing environments. 
From a theoretical standpoint, the interplay between applied optimization frameworks and learned predictors presents fertile ground for rigorous analyses: 
carefully designed learning modules can \emph{guide or prune} large search spaces, improving computational tractability without sacrificing solution quality.

In the following subsections, we discuss three prominent strands of search-based MAPF solvers that have been augmented by learning-based elements. 
First, conflict-based methods leverage data-driven conflict resolution strategies to expedite node expansions and reduce search overhead. 
Second, priority-based methods can benefit from learned priority assignments or adaptive ordering schemes that dynamically resolve agent conflicts. 
Finally, large neighborhood search techniques incorporate customized learning heuristics to guide neighborhood selection and accelerate solution refinement. 
These augmented approaches illustrate an emerging shift towards \emph{hybrid MAPF pipelines} that build on the best of both worlds, offering the promise of robust and scalable solutions in increasingly complex multi-agent settings.

\subsection{Learning-Augmented Conflict-Based Search (CBS)}\label{sec:cbs:learning}

Conflict-Based Search (CBS)~\citep{sharon2015conflict} (see Section~\ref{sec:cbs}) has long been a standard bearer for solving MAPF optimally under the constraints in \eqref{eq:cbs:constraints}.  
However, despite notable enhancements to the baseline approach (Section~\ref{sec:cbs:optimal}), CBS can still face significant computational bottlenecks when the environment is highly dynamic or the conflict set is large.  
Recent work has therefore explored \emph{learning-augmented} CBS, wherein \emph{select modules} of the classical method---for example, node expansion, conflict selection, or heuristic evaluation---are partially \emph{learned} rather than purely hand-coded.  
The central idea is to maintain CBS’s original integer linear constraints and guaranteed collision-avoidance rules, while injecting data-driven insights that accelerate key search operations.  

\paragraph{Mathematical Formulation.}
Much like classical CBS, learning-augmented CBS retains binary variables
\[
   x_{i,v,t} \in \{0,1\}, \quad \forall i,\,v,\,t,
\]
with collision-avoidance constraints~\eqref{eq:collision-vertex}--\eqref{eq:collision-edge}, and potentially dimensional constraints tailored to SoC or makespan objectives (see Section~\ref{sec:cbs:framework}).  
To incorporate learning, one augments the baseline CBS with \emph{auxiliary} parameters \(\Theta\) that influence search decisions without altering feasibility.  
Thus, the feasible solution space remains governed by the same collision-free constraints, yet the \emph{search process} can become more adaptive.

Formally, each high-level (HL) node \(N\) in the conflict tree (CT) is characterized by:
\[
   N \;=\; 
   (\{\pi_i\}_{i=1}^n,\; \{\mathrm{constraints}_i\}_{i=1}^n,\; \mathrm{cost}(N)),
\]
where \(\{\pi_i\}\) are single-agent paths obeying \eqref{eq:cbs:constraints}, and each \(\mathrm{constraints}_i\) is a set of time-indexed vertex or edge constraints for agent~\(i\).  
In classical CBS, the node expansion order or conflict selection is often driven by manually defined heuristics (e.g., smallest sum of costs, earliest conflict).  
In learning-augmented CBS, we introduce a learned mapping
\[
   \delta_\Theta: \; N \;\mapsto\; \text{(ranked node/ conflict scoring)},
\]
to prioritize expansions or conflict splits based on data-driven predictions (e.g., anticipated search depth, likely feasibility, or future cost).  
These predictions do \emph{not} relax any constraints but \emph{guide} which branch of the search to explore first.

\paragraph{Key Learning-Augmented Modules.}
Existing works typically introduce learning in one of the following CBS modules. Below, we highlight parallels to well-known classical refinements from Sections~\ref{sec:cbs:optimal}--\ref{sec:cbs:suboptimal}:  
\begin{enumerate}[leftmargin=*]
    \item \textbf{High-Level Node Selection.}  
    The conflict tree is explored by popping HL nodes from a priority queue. Classical CBS often targets a minimal SoC or uses an admissible heuristic (\emph{e.g.}, conflict-graph-based)~\citep{felner2018adding,li2019improved,li2021pairwise}.  
    Learning-augmented approaches (e.g.,~\citep{huang2021learning-aamas,yu2023accelerating,yao2024accelerating}) replace this manual heuristic with a policy~\(\delta_\Theta\) that rank-orders nodes by predicted downstream performance.  
    Agents still replan paths via A*\ at the low level (or MDD-based searches~\citep{boyarski2015icbs,li2019symmetry}), ensuring consistency with classical collision avoidance.  

    \item \textbf{Conflict Selection \& Splitting.}  
    When a node \(N\) contains multiple pairwise collisions, classical CBS typically branches on the \emph{first detected} conflict or a \emph{cardinal} conflict identified via conflict-graph checks~\citep{felner2018adding,li2019improved}.  
    A \emph{learning-based} approach may instead predict which collision is most “critical” to resolve, attempting to prune the search tree earlier~\citep{huang2021learning}.  
    Structurally, this is reminiscent of \emph{disjoint splitting}~\citep{li2019disjoint} or \emph{symmetry-breaking}~\citep{li2020new,li2021pairwise}, but with data-driven prioritization.  

    \item \textbf{Adaptive Algorithm Selection.}  
    A broader \emph{meta-learner} may dynamically choose between different CBS variants (\emph{e.g.},\ c.f.\ Sections~\ref{sec:cbs:optimal}--\ref{sec:cbs:suboptimal} on optimal vs.\ suboptimal methods) or even between CBS and other MAPF solvers (like SAT or ILP-based methods~\citep{surynek2022problem}).  
    For instance, \citep{sigurdson2019automatic,ren2021mapfast,alkazzi2022mapfaster} use classifiers to identify which solver is likely fastest for a given instance.  
    Once selected, the standard conflict-based constraints remain intact, and the learned module purely handles solver dispatch.
\end{enumerate}

\paragraph{Illustrative Pseudocode.} 
Algorithm~\ref{alg:learning-cbs} compares a minimal learning-augmented CBS (focusing on node selection) to the classical version in Algorithm~\ref{alg:cbs}. 
Lines enclosed in \fbox{boxes} highlight the data-driven components.

\begin{algorithm}[ht]
\footnotesize
\caption{CBS with Learned Node-Selection (\textbf{Compare to Algorithm~\ref{alg:cbs}})}
\label{alg:learning-cbs}
\begin{algorithmic}[1]
\Require 
  Graph $\mathcal{G}=(\mathcal{V},\mathcal{E})$, 
  Agents $\{1,\dots,n\}$, 
  Start/Goal vertices $\{(s_i,g_i)\}_{i=1}^n$,  
  \fbox{Learned node-scorer} $\delta_\Theta(\cdot)$.
\State Initialize root node $N_0$:
  \begin{itemize}
     \item For each agent $i$, compute $\pi_i$ \emph{ignoring collisions} (standard single-agent A*).
     \item Set empty constraint sets for all agents (no additional constraints).
  \end{itemize}
\State \fbox{Compute node score } $\mathit{score}(N_0) \gets \delta_\Theta(N_0)$
\State \fbox{Insert $N_0$ into a priority queue, keyed by $\mathit{score}(N_0)$ (descending order).}
\While{priority queue is \emph{not empty}}
  \State $N \gets \text{pop front of queue}$ 
  \State \textbf{Check for conflicts} among $\{\pi_i\}_{i=1}^n$ in $N$.
  \If{no conflict}
    \State \Return \(\{\pi_i\}\) \quad(\emph{solution found})
  \Else
    \State Let $(i, j, t)$ be the \fbox{\emph{highest-ranked conflict}} according to $\delta_\Theta(N)$
    \For{each agent $a \in \{i,j\}$}
       \State $N' \gets \text{copy of }N$
       \State Add new constraint \emph{forbidding} agent $a$ from \((v,t)\), the conflicting position.
       \State Replan $\pi_a$ with standard single-agent pathfinding (respecting new constraints).
       \If{$\pi_a$ is feasible}
          \State \fbox{$\mathit{score}(N') \gets \delta_\Theta(N')$}
          \State Insert $N'$ into the queue keyed by $\mathit{score}(N')$
       \EndIf
    \EndFor
  \EndIf
\EndWhile
\State \Return \(\emptyset\) \,(\emph{no feasible solution})
\end{algorithmic}
\end{algorithm}

\noindent
The crucial distinction is that whereas classical CBS typically expands the node with the \emph{lowest computed cost} or the fewest conflicts, here we use a \emph{learned} function $\delta_\Theta(\cdot)$ to guide expansions, conflict splitting, or both.  

\paragraph{Comparison to Classical CBS.}
Table~\ref{tab:cbs:compare} synthesizes the main differences across four dimensions: HL node expansion, conflict selection, heuristic or priority function, and solver selection. We also reference the enhancements in Section~\ref{sec:cbs:optimal} (for optimal CBS) and Section~\ref{sec:cbs:suboptimal} (for suboptimal CBS) to pinpoint how learning-driven modules map onto the older, hand-designed heuristics.

\begin{sidewaystable}
\centering
\footnotesize
\caption{Comparison of \textbf{Classical} vs.\ \textbf{Learning-Augmented} CBS Approaches}
\label{tab:cbs:compare}
\resizebox{\linewidth}{!}{%
\begin{tabular}{p{0.1\linewidth}|p{0.45\linewidth}p{0.45\linewidth}}
\toprule
\textbf{Aspect} 
    & \textbf{Classical CBS Approaches} 
    & \textbf{Learning-Augmented CBS} 
    \\ 
\midrule

\textbf{Node Expansion}
    & Typically prioritizes the node with:
      \begin{itemize}[leftmargin=1em]
          \item Lowest sum of costs (SoC) or smaller makespan~\citep{sharon2015conflict,li2019disjoint}
          \item Admissible heuristics (e.g., conflict graph, DG/WDG)~\citep{felner2018adding,li2019improved,li2021pairwise}
          \item Possibly suboptimal expansions (ECBS)~\citep{barer2014suboptimal}
      \end{itemize}
    & Employs a \emph{learned} function 
      \(\delta_\Theta(N)\) to score each node:
      \begin{itemize}[leftmargin=1em]
          \item Prediction of conflict severity, solution depth, etc.
          \item Often trained via supervised imitation~\citep{huang2021learning-aamas} or RL signals
      \end{itemize}
    \\

\textbf{Conflict Selection}
    & Branches on first or earliest conflict, or on a recognized cardinal conflict:
      \begin{itemize}[leftmargin=1em]
          \item Disjoint splitting~\citep{li2019disjoint}
          \item Symmetry reasoning~\citep{li2019symmetry,li2021pairwise}
          \item Rectangle/corridor detection~\citep{li2020new}
      \end{itemize}
    & \begin{itemize}[leftmargin=1em]
          \item Uses a predictor \(\delta_\Theta\) to rank conflicts by urgency
          \item Can mimic or supersede classical heuristics, e.g.\ “cardinality” detection~\citep{huang2021learning}
          \item Potentially adapts to dynamic or partially observable domains
      \end{itemize}
    \\

\textbf{Heuristic or Priority Function}
    & Fixed or combinational heuristics:
      \begin{itemize}[leftmargin=1em]
          \item Conflict count, cardinal conflicts, min vertex cover on conflict graph~\citep{felner2018adding}, etc.
          \item Mutex reasoning with MDD~\citep{boyarski2015icbs,zhang2022multi}
      \end{itemize}
    & Parametric scoring module:
      \begin{itemize}[leftmargin=1em]
          \item GNN-based or transformer-based node rankings~\citep{yu2023accelerating,yao2024accelerating}
          \item Linear or neural scoring of conflicts~\citep{huang2021learning}
          \item Learned synergy of multiple features not captured by a single classical heuristic
      \end{itemize}
    \\

\textbf{Solver Selection}
    & In classical CBS, typically one solver is used:
      \begin{itemize}[leftmargin=1em]
          \item Optimal CBS variants or
          \item Suboptimal CBS (ECBS, EECBS, etc.)~\citep{barer2014suboptimal,li2021eecbs,chan2022flex,tang2024ita}
      \end{itemize}
    & A meta-learner orchestrates (per-instance or online):
      \begin{itemize}[leftmargin=1em]
          \item Chooses among CBS variants (ICBS, ECBS, FECBS, etc.)~\citep{boyarski2015icbs,chan2022flex}
          \item Can opt for entirely different MAPF frameworks (e.g., ILP-based~\citep{surynek2022problem}, or SAT-based~\citep{surynek2016efficient})
          \item Predicts best solver via offline-trained classifier~\citep{sigurdson2019automatic,ren2021mapfast,alkazzi2022mapfaster}
      \end{itemize}
    \\
\bottomrule
\end{tabular}%
}
\end{sidewaystable}

\noindent
\textbf{Relation to CBS Heuristics.}  
As summarized in Table~\ref{tab:cbs:compare}, classical CBS enhancements (Section~\ref{sec:cbs:optimal}) use carefully engineered heuristics (e.g., conflict-graph cardinality~\citep{felner2018adding} or symmetry constraints~\citep{li2020new,li2021pairwise}) to reduce the branching factor.  
A learning-based approach can be viewed as “unifying” or “automating” these heuristic cues under a single parametric model that is \emph{trained} rather than manually designed.  
For instance, a neural network might implicitly learn to detect frequent corridor conflicts or rectangles, approximating and even extending classical symmetry-breaking.  

\noindent
\textbf{Suboptimal CBS vs.\ Learned Guidance.}  
Suboptimal CBS variants like ECBS~\citep{barer2014suboptimal}, EECBS~\citep{li2021eecbs}, and FECBS~\citep{chan2022flex} offer a guaranteed cost bound but prioritize runtime performance.  
Similarly, a learned node-scorer or conflict selector can be integrated into these frameworks to further accelerate expansions.  
For example, one can combine the suboptimal factor $w$ with a learned function $\delta_\Theta(\cdot)$, yielding a dual ranking scheme
\[
   f(N) 
   \;=\; 
   \alpha \,\mathrm{Cost}(N) \;+\; \beta\,\delta_\Theta(N),
\]
where \(\mathrm{Cost}(N)\) might be the focal search cost used by ECBS, and \(\delta_\Theta(N)\) is a data-driven priority.  
By adjusting the weights \(\alpha,\beta\), one can retain partial suboptimality guarantees and benefit from learned heuristics’ speedups.  

\paragraph{Discussion and Outlook.}
Overall, learning-augmented CBS strives to preserve the strong theoretical underpinnings of conflict-based pathfinding~\citep{sharon2015conflict,felner2018adding,li2019improved}, while giving the search process more data-driven adaptability.  
Empirical studies~\citep{huang2021learning-aamas,yu2023accelerating,yao2024accelerating} indicate that learning-augmented expansions can considerably reduce runtime in complex or large-scale instances.  
Moreover, these approaches naturally extend to partially observable or dynamic domains (via online retraining or domain adaptation), where purely hand-designed heuristics may fail to capture all nuance.  

Key open challenges include:  
\begin{enumerate}[leftmargin=*, itemsep=0.5em]
    \item \textbf{Scalability:}  
    Extending these learning-augmented methods to thousands of agents (see, e.g., \citep{friedrich2024scalable,okumura2024engineering}) requires highly efficient inference and robust generalization.  
    \item \textbf{Guarantees under Heavy Pruning:}  
    If the learned policy prunes nodes aggressively, suboptimal or even incomplete solutions can result.  
    Transparent ways of bounding the search error would improve reliability.  
    \item \textbf{Handling Non-Stationary Environments:}  
    Many real-world MAPF scenarios involve dynamic obstacle layouts or uncertain agent dynamics~\citep{ren2024multi}.  
    Designing learning-augmented CBS pipelines resilient to such changes is an active area of research.
\end{enumerate}

\noindent
Nevertheless, \emph{conflict-based modeling} remains central to many state-of-the-art MAPF solvers, and adding learned components only \emph{augments} rather than discards these classical constraints.  
We thus anticipate further hybrid developments that integrate machine learning with advanced optimal or suboptimal CBS variants (or compilation-based methods) for robust, large-scale multi-agent coordination.



\subsection{Learning-Augmented Priority-Based Search (PBS)}\label{sec:pbs:learning}

\paragraph{Motivation and Overview.}
Priority-Based Search (PBS)~(Section~\ref{sec:pbs}) is a fast and popular framework for MAPF, wherein each agent plans its path under the constraint that it must respect the spatial--temporal trajectories of all higher-priority agents.  
While PBS can be scaled to large numbers of agents with relatively low search overhead, its performance hinges critically on the choice of \emph{priority ordering}, as a poorly chosen order can lead to excessively long paths or even unsolvable collisions in the low-priority subset.  
Classical PBS relies on manual heuristics or random assignments to finalize this ordering, with no single method dominating in all scenarios.

To address this limitation, \emph{learning-augmented} PBS approaches replace (or guide) the classic \emph{priority assignment} module by leveraging data-driven models.  
These hybrid pipelines preserve the fundamental \emph{collision constraints} (Eq.~\eqref{eq:pbs:vertex}--\eqref{eq:pbs:edge}) and the general feasibility domain of PBS, but allow a learned predictor to rank or reorder agents in a manner that effectively reduces collisions or enhances solution quality.  
In addition, learning-based modules may adapt to real-time or dynamic environments where a purely static ordering would be insufficient.  
Conceptually, these techniques bring PBS closer to the spirit of \emph{learning-augmented CBS} (Section~\ref{sec:cbs:learning}), while retaining the simplicity and scalability that have made purely classical PBS appealing.

\subsubsection{Mathematical Formulation and Integration of Learning}

Recall from Section~\ref{sec:pbs:modeling} that a classical PBS solution is governed by:
\[
   x_{i,v,t}\in\{0,1\}, 
   \quad 
   \text{for all agents } i, \text{ vertices } v, \text{ and times } t,
\]
subject to collision-avoidance constraints~\eqref{eq:pbs:constraint}.  
What fundamentally defines a PBS instance is a \emph{function} 
\[
   \mathsf{P}: 
      \{1,\dots, n\} 
      \;\rightarrow\; 
      \{1,\dots,n\},
\]
which ranks agents from highest to lowest priority (or imposes a partial order in more advanced variants).  
In \emph{learning-augmented PBS}, we introduce a \emph{parametric} priority function
\[
   \mathsf{P}_\Theta:\;\{1,\dots,n\}\;\mapsto\;\{1,\dots,n\},
\]
where \(\Theta\) denotes learnable parameters (e.g., weights of a classifier or neural policy).  
The search and path-finding constraints (Eqs.~\ref{eq:pbs:vertex}--\ref{eq:pbs:edge}) remain unchanged, but instead of a fixed hand-designed \(\mathsf{P}\), we compute or update \(\mathsf{P}_\Theta\) from data or from online observations:
\[
   \text{When agent $i$ plans its path, it must avoid collisions with each agent $j$ satisfying 
        } \mathsf{P}_\Theta(j) < \mathsf{P}_\Theta(i).
\]
The \emph{objective} (e.g., a sum-of-costs or makespan criterion) is also unchanged from the classical setting; the learning module aims to select a priority ordering that typically yields fewer collisions, shorter paths, or a higher success rate.

\vspace{1em}
\begin{algorithm}[ht]
\footnotesize
\caption{Priority-Based Search with a Learned Priority Assignment}
\label{alg:learning-pbs}
\begin{algorithmic}[1]
\Require 
   Graph $\mathcal{G}=(\mathcal{V},\mathcal{E})$, 
   Agents $\{1,\dots,n\}$, 
   Start/goal pairs $\{(s_i,g_i)\}_{i=1}^n$, 
   \textbf{learned} priority function~$\mathsf{P}_\Theta$.
\State \textbf{Initialize:} 
   \begin{itemize}
     \item \fbox{Compute priorities for each agent} 
           $r_i \gets \mathsf{P}_\Theta(i)$.
     \item Sort agents by $r_i$ in ascending order 
           (lowest $r_i$ means highest priority).
   \end{itemize}
\For{each agent $i$ in ascending $\mathsf{P}_\Theta$-order}
  \State Let $H_i \gets \{\pi_j \mid \mathsf{P}_\Theta(j)<\mathsf{P}_\Theta(i)\}$ 
         \Comment paths of \emph{higher-priority} agents
  \State Plan \(\pi_i\) on a time-expanded graph where all space--time cells used by $H_i$ are forbidden.
  \If{no feasible \(\pi_i\)}
     \State \textbf{return} ``\emph{Infeasible under current learned ordering}.''
  \EndIf
\EndFor
\State \textbf{return} $\{\pi_1,\dots,\pi_n\}$ 
       \quad(\emph{valid solution under } $\mathsf{P}_\Theta$)
\end{algorithmic}
\end{algorithm}
\vspace{1em}

\noindent
\textbf{Algorithm~\ref{alg:learning-pbs}} outlines a minimal \emph{one-shot} integration of a learned priority function.  
Like the vanilla version in Algorithm~\ref{alg:pbs:vanilla-recap}, we perform a single pass from highest to lowest priority.  
However, the order itself now comes from a predictor \(\mathsf{P}_\Theta\).  
In principle:
\begin{enumerate}[leftmargin=*]
   \item \emph{Offline Training.}  
   One can train \(\Theta\) (e.g., via supervised learning or evolutionary search) on a corpus of MAPF instances, where the “label” or “reward” might measure how well an ordering reduces the sum of costs or yields fewer collisions.
   \item \emph{Online Updating.}  
   In dynamic environments or repeated tasks, \(\mathsf{P}_\Theta\) can be re-evaluated each time new start/goal locations arrive or environmental conditions change, thereby adapting priorities in real time.
\end{enumerate}

\subsubsection{Representative Learning Approaches}

\paragraph{SVM- and Evolutionary-Based Prioritization.}
\citet{zhang2022learning} propose a \emph{support vector machine} (SVM) model to predict a promising ordering of agents, aiming to reduce collisions and hence overall solution cost.  
They show performance gains by incorporating \emph{stochastic restarts} and \emph{random rank perturbations} during planning to avoid local optima under a single ordering.  
Similarly, \citet{wang2023synthesizing} adopt a \emph{genetic algorithm} (GA) to evolve priority functions, treating permutations or partial orders of agents as candidate “genomes.”  
Offspring that yield fewer collisions or lower sum-of-costs are selected, gradually improving the learned priority scheme over multiple generations.

\paragraph{Reinforcement Learning and Attention Mechanisms.}
\citet{yang2024attention} present a \emph{hybrid} approach that couples PBS with a reinforcement learning (RL) policy.  
Instead of a fixed one-shot assignment, their method uses a \emph{Synthetic Score-based Attention Network} to assign conflict-free priorities in a more dynamic fashion, learning from trial-and-error interactions in a multi-agent environment.  
The attention mechanism processes local agent states (positions, goals, conflicts) to generate situational \emph{priority scores} online, which effectively reorder planning so as to reduce bottlenecks.  

\paragraph{Integration with Dynamic PBS Variants.}
Beyond the \emph{vanilla} single-pass algorithm, advanced PBS frameworks (e.g., dynamic priority inheritance~\citep{okumura2022priority}, partial-order branching~\citep{ma2019searching}, or merging~\citep{boyarski2022merging}) can similarly embed learned components.  
In such cases, \(\mathsf{P}_\Theta\) might re-rank agents at each collision, or a \emph{learning-based policy} might decide which agents to merge.  
These expansions mirror the ideas seen in learning-augmented CBS (Section~\ref{sec:cbs:learning}), underscoring a broader strategy of \emph{augmenting classical submodules}---instead of discarding them entirely.

\subsubsection{Discussion and Outlook}

Learning-augmented PBS exemplifies the central theme of \emph{hybridizing} classical MAPF solvers with data-driven models:  
it preserves the fundamental constraints of priority-based planning (Eq.~\ref{eq:pbs:constraint}) while injecting a learned module \(\mathsf{P}_\Theta\) to \emph{strategically} shape the planning order.  
Experiments suggest that such combinations can reduce the reliance on ad-hoc heuristics, improve success rates in congested or dynamic environments, and more gracefully scale when domain parameters shift.

Important open challenges parallel those seen in learning-augmented CBS and other MAPF methods:
\begin{itemize}[leftmargin=*]
    \item \textbf{Generalization and Scalability.}  
    Learned priority assignments may fail to extrapolate well to agent populations much larger than those seen in training or to drastically different topologies.
    \item \textbf{Robustness in Real-World Deployments.}  
    In environments where sensors, communication, or agent dynamics are uncertain, how can \(\mathsf{P}_\Theta\) adapt priorities online without causing deadlocks or collisions?
    \item \textbf{Hybrid \emph{vs.}\ Fully Data-Driven.}  
    While partial learning preserves classical PBS properties, future work may explore fully RL-based multi-agent frameworks that derive both collision-avoidance rules and priorities from large dataset pretraining, aligning with \emph{foundation models}~\citep{alkazzi2024comprehensive} discussed in Section~\ref{sec:others}.
\end{itemize}

Overall, learning-augmented PBS forms a promising middle ground:  
it capitalizes on decades of heuristic wisdom in priority-based planning, while employing machine learning to handle the notoriously difficult \emph{priority ordering} problem.  
As multi-agent applications continue to grow in scale and complexity, such hybrid pipelines will likely become increasingly relevant, especially when classical PBS alone struggles to meet real-time or adaptive demands.



\subsection{Learning-Augmented Large Neighborhood Search}\label{sec:lns:learning}

Large Neighborhood Search (LNS) methods (Section~\ref{sec:lns}) provide a suboptimal yet scalable framework for MAPF by iteratively ``destroying'' and ``repairing'' subsets of a global solution.  
Whereas classical LNS approaches rely on handcrafted heuristics to select which agents to remove (\emph{destroy}) and how to replan these agents (\emph{repair}), \emph{learning-augmented LNS} seeks to replace or assist these modules with data-driven policies.  
Such hybrid designs exploit the original mathematical formulation and collision-avoidance constraints of classical LNS-based MAPF (Equations~\eqref{eq:lns-vertex}--\eqref{eq:lns-edge}), while introducing learned components to guide destructive and reparative decisions.  

\paragraph{Mathematical Model.}  
We recall that LNS solutions for MAPF maintain a global set of trajectories \(\{\pi_i\}_{i=1}^n\), one path per agent, and impose vertex- and edge-collision constraints:
\begin{align}
&\text{\emph{(vertex constraint)}} 
&&x_{i,v,t} + x_{j,v,t} \;\le\; 1,
\quad \forall\,t,\, \forall\,v\in \mathcal{V}, \,\forall\,i<j, \label{eq:lns-vertex-recap}\\
&\text{\emph{(edge constraint)}} 
&&x_{i,u,t} + x_{i,v,t+1} + x_{j,v,t} + x_{j,u,t+1} \;\le\; 3,\quad
\forall\,(u,v)\in \mathcal{E}, \,\forall\,t,\,\forall\,i<j,
\label{eq:lns-edge-recap}
\end{align}
where \(x_{i,v,t}\) is a binary decision variable indicating whether agent~\(i\) is at vertex \(v\) at time \(t\).  
The \emph{objective} (e.g., makespan~\eqref{eq:makespan} or SoC~\eqref{eq:soc}) and the fundamental set of collision-free feasibility constraints remain unchanged from the classical setup.  
In learning-augmented LNS, one introduces additional \emph{auxiliary} variables or a learned function \(\delta_\Theta\) that steers the \emph{destroy} and \emph{repair} steps externally to these collision-avoidance constraints:  
\[
    \delta_\Theta: (\{\pi_i\}, \mathcal{G}, \textit{other contextual data}) \;\;\mapsto\; 
   (\text{choice of agents/paths to destroy}, \text{replanning strategy}).
\]
Because \(\delta_\Theta\) is not part of the core feasibility constraints~\eqref{eq:lns-vertex-recap}--\eqref{eq:lns-edge-recap}, it neither invalidates collision-free guarantees nor changes the set of admissible solutions; rather, it \emph{guides} which partial solutions are explored and how.  

\paragraph{From Classical to Learning-Augmented LNS.}  
The classical LNS scheme (Algorithm~\ref{alg:lns}) iterates between (i) randomly or heuristically \emph{destroying} a subset of agent paths and (ii) \emph{repairing} them using a local solver.  
In a learning-augmented approach, these two steps are partially or wholly driven by data-driven policies.  
Concretely:
\begin{itemize}[leftmargin=*]
    \item \emph{Learning-Based Destroy.}  
    Instead of removing agents uniformly at random or solely by collision frequency, one estimates which agents or spatial regions offer the greatest potential improvement if re-optimized.  
    For instance, a learned policy \(\delta_\Theta^{\text{destroy}}\) might identify ``high-conflict'' agents or time intervals that frequently cause deadlocks, thereby focusing the LNS loop on eliminating these problematic collisions more efficiently.
    
    \item \emph{Learning-Based Repair.}  
    During the \emph{repair} step, a learned module \(\delta_\Theta^{\text{repair}}\) can provide either (i) cost-effective single-agent trajectories that consider a learned heuristic beyond classical A*, or (ii) partial multi-agent solutions via reinforcement learning.  
    The same global feasibility constraints (\ref{eq:lns-vertex-recap})--(\ref{eq:lns-edge-recap}) still apply; however, the planner can benefit from data-driven intuition regarding which paths are likely to reduce collisions or improve objectives.
\end{itemize}
As a result, learning-augmented LNS typically shares the \emph{same} collision-avoidance model and objective function as classical LNS, but replaces handcrafted heuristics in the destroy-repair loop with parametric (e.g., neural) policies that can adapt to instance-specific features.

\paragraph{Example: LNS2+RL.}  
Recent work by~\citet{wang2024lns2+} proposes an LNS-based solver that integrates a multi-agent reinforcement learning (MARL) module into classical LNS2~\citep{li2022mapf}.  
Initial iterations prioritize a MARL-based low-level \emph{repair} to explore diverse path allocations, as MARL can discover geometric or collision patterns that standard shortest-path routines overlook.  
Later in the LNS cycles, the method switches to a priority-based re-planner (similar to suboptimal MAPF solvers in Section~\ref{sec:compilation} and Section~\ref{sec:cbs:suboptimal}), accelerating convergence by leveraging simpler heuristics in conflict-free subregions.  
Hence, LNS2+RL employs the same collision-prevention constraints as classical LNS2, but adaptively invokes a learned policy to \emph{repair} agent paths early on, then falls back to more efficient classical updates once major conflicts are resolved.  
Empirical results indicate that this blended approach outperforms naive LNS2 or purely learning-based approaches in both solution quality and runtime, especially as the number of agents increases.

\begin{algorithm}[t]
\footnotesize
\caption{Learning-Augmented LNS (High-Level Sketch)}
\label{alg:lns-learning}
\begin{algorithmic}[1]
\Require Graph $\mathcal{G}=(\mathcal{V},\mathcal{E})$, agents $\{1,\dots,n\}$, start/goal vertices $\{(s_i,g_i)\}_{i=1}^n$, destroy ratio $\gamma$, max iterations $K$, learned policy $\delta_\Theta$ for (destroy, repair).  
\State \textbf{Initialize} a feasible solution $S_0 = \{\pi_i^{(0)}\}$ using classical single-agent heuristics or a separate solver.
\State $S_{\mathrm{current}} \gets S_0;$ \quad $S_{\mathrm{best}} \gets S_0$
\For{$k = 1 \to K$}
    \State \textbf{Destroy Step:} 
    \Statex \quad Identify a subset of agents $\mathcal{D}$ via \(\delta_\Theta^{\text{destroy}}\bigl(S_{\mathrm{current}}\bigr)\) (size $\approx \gamma n$).
    \Statex \quad Remove $\pi_i$ for each $i\!\in\!\mathcal{D}$, yielding partial solution $S^{\mathrm{partial}}$.
    \State \textbf{Repair Step:} 
    \Statex \quad For each $i\in \mathcal{D}$, compute a new path $\hat{\pi}_i$ with \(\delta_\Theta^{\text{repair}}\bigl(S^{\mathrm{partial}},\mathcal{G}\bigr)\) subject to collision constraints (\ref{eq:lns-vertex-recap})--(\ref{eq:lns-edge-recap}).
    \Statex \quad Merge these new paths into $S' = S^{\mathrm{partial}} \cup \{\hat{\pi}_i: i\in\mathcal{D}\}$.
    \If{$\mathrm{Cost}(S') < \mathrm{Cost}(S_{\mathrm{best}})$}
       \State $S_{\mathrm{best}} \gets S'$
    \EndIf
    \State $S_{\mathrm{current}} \gets \textsc{AcceptanceCriterion}\bigl(S_{\mathrm{current}}, S'\bigr)$
\EndFor
\State \Return $S_{\mathrm{best}}$
\end{algorithmic}
\end{algorithm}

\paragraph{Illustrative Algorithm.}  
Algorithm~\ref{alg:lns-learning} outlines a generic learning-augmented LNS procedure.  
Compared to the baseline procedure (Algorithm~\ref{alg:lns}), the difference lies in Lines~5--6 (learning-based destroy) and Lines~7--8 (learning-based repair).  
One may also allow \(\delta_\Theta\) to interact with an external classical solver, \emph{e.g.}, switching from reinforcement learning to priority-based re-planning after a certain number of iterations.

\paragraph{Discussion and Outlook.}  
Learning-augmented LNS underscores the broader synergy between \emph{classical MAPF constraints} (which guarantee collision-free doctrines) and \emph{data-driven search guidance} (which reduces reliance on hand-engineered heuristics).  
By leveraging modern machine learning, one gains the ability to:  
\begin{itemize}[leftmargin=*]
    \item \textbf{Adapt to Complex Environments or Partial Observability.}  
    Learned destroy-repair policies can generalize from past experience and re-optimize quickly in changing or uncertain scenarios.  
    \item \textbf{Reduce Search Overhead.}  
    Targeting high-impact agents or collisions can significantly prune the solution space.  
    \item \textbf{Maintain Transparency of Constraints.}  
    Since the core collision-avoidance model remains identical to that in classical LNS, correctness and feasiblity are preserved.
\end{itemize}
Nevertheless, learning-augmented LNS also faces practical challenges.  
Data collection and policy training can be nontrivial, especially for large numbers of agents~\citep{friedrich2024scalable,okumura2024engineering}, and guaranteeing suboptimality bounds may require careful integration of classical cost metrics~\citep{li2022mapf,lam2023exact}.  
Despite these open questions, the convergence of LNS meta-heuristics and data-driven modules represents a promising path forward, balancing adaptive intelligence with well-established optimization principles.

\section{Reinforcement Learning for MAPF}
\label{sec:rl}

\begin{figure}[htb!]
    \centering
    \includegraphics[width=\linewidth]{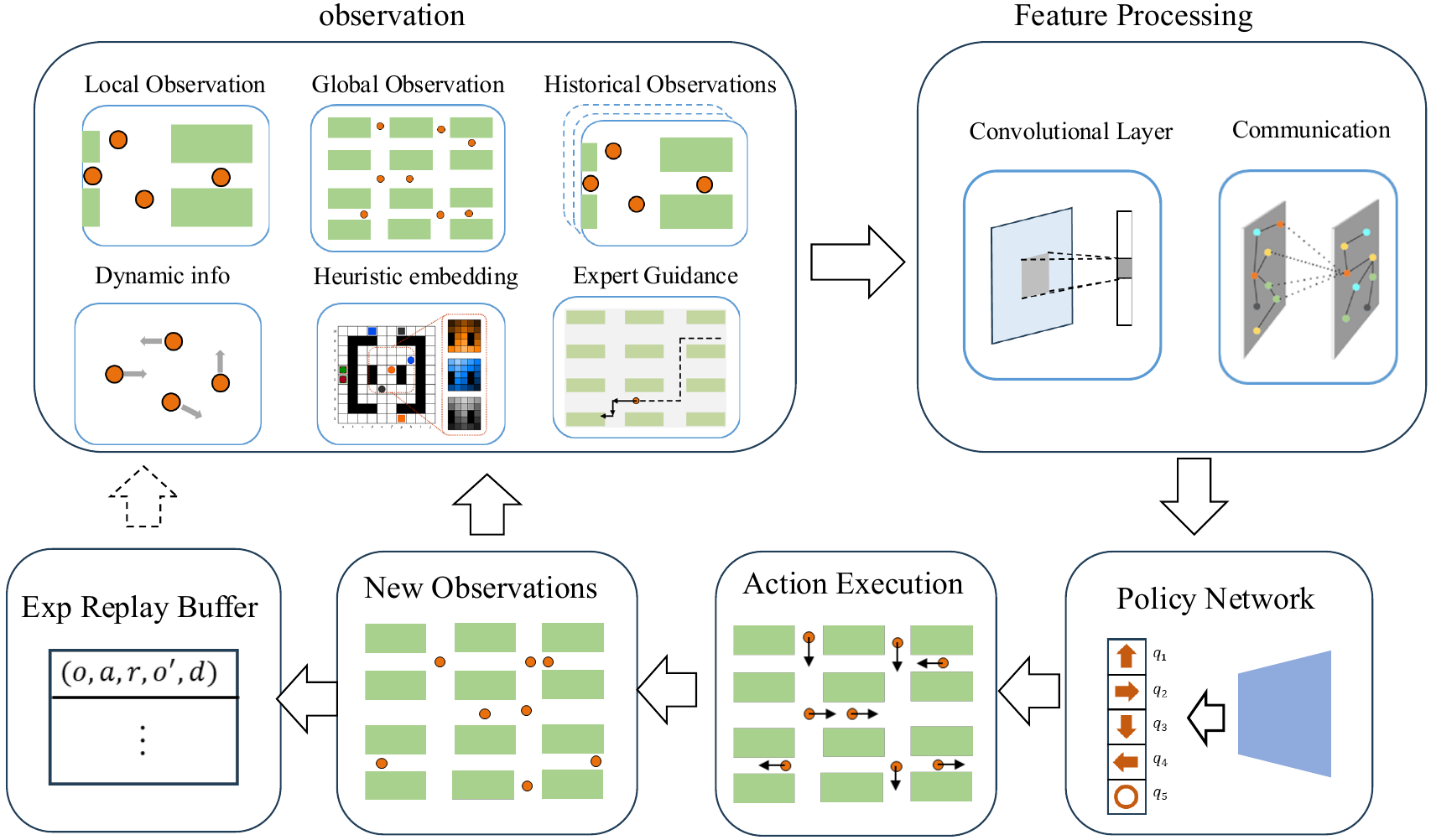}
    \caption{A general algorithmic flowchart for solving the MAPF problem using reinforcement learning. The process begins with various types of observations, including local, global, dynamic, historical, and expert-guided information. These observations are processed in the feature extraction stage, which involves convolutional layers and inter-agent communication. The extracted features are then passed to the policy network. The next steps include action execution and reward collection, followed by storing the experience in a replay buffer and generating new observations.}
    \label{fig:rl}
\end{figure}

Reinforcement learning (RL) offers a promising framework for tackling MAPF under decentralized decision-making and partial observability.
A general algorithmic flowchart for solving the MAPF problem using RL is shown in Figure~\ref{fig:rl}. 
In contrast to purely classical methods, RL-based approaches allow each agent to learn collision-free navigation and coordination policies in a data-driven manner, adapting to complex or dynamic environments. 
Nevertheless, bridging these learning ideas with MAPF requires carefully defined states, actions, rewards, collision-avoidance mechanisms, and inter-agent communication protocols. 
This section provides a comprehensive presentation of how RL can model and solve MAPF, offering mathematical formulations, explanations of popular multi-agent RL (MARL) paradigms, and a critical review of remaining challenges. 

\subsection{Mathematical Modeling of RL-based MAPF}
\label{subsec:rl_modeling}

\paragraph{MDP Definition.}
In an RL-based MAPF framework, agent $i$ (where $i\in\{1,\dots,n\}$) is typically governed by a Markov Decision Process (MDP) 
\[
\bigl\langle 
    \mathcal{S}^i,\,   
    \mathcal{A}^i,\,
    p(\cdot\mid \cdot),\,
    r^i,\,
    \gamma 
\bigr\rangle,
\]
where:
\begin{itemize}[leftmargin=*]
    \item $\mathcal{S}^i$ denotes the (local) state space for agent $i$, which may include partial environment observations, neighbor information, or heuristic guidance.
    \item $\mathcal{A}^i$ is the action space (e.g., $\{\textit{up}, \textit{down}, \textit{left}, \textit{right}, \textit{stay}\}$ in grid worlds).
    \item $p(\cdot\mid\cdot)$ is the transition probability function. In deterministic map-based settings, it might simplify to $s^i_{t+1}=f(s^i_t,a^i_t)$.
    \item $r^i:\mathcal{S}^i\times \mathcal{A}^i\rightarrow \mathbb{R}$ is an individual reward function (discussed in Section~\ref{subsec:reward_design}).
    \item $\gamma\in[0,1)$ is the discount factor for future rewards.
\end{itemize}
The goal of each agent is to learn a policy $\pi^i:\mathcal{S}^i\to \Delta(\mathcal{A}^i)$ (a probability distribution over actions) that maximizes the expected discounted return:
\begin{equation}\label{eq:rl-objective}
\pi^{i,*} 
~\in~
\arg \max_{\pi^i}
\mathbb{E}\biggl\{
\sum_{t=0}^\infty 
\gamma^t \, r^i\bigl(s_t^i,a_t^i\bigr)
\biggr\}.
\end{equation}

\paragraph{Collision-Avoidance Constraints.}
Since MAPF requires collision-free paths, collisions must be integrated into the RL formulation either as \emph{soft constraints} (negative rewards) or \emph{hard constraints} (invalid actions). 
For example, an agent may receive a large penalty $-\alpha \,(<0)$ upon attempting to occupy a vertex already claimed by another agent or upon traversing an edge in an opposite direction at the same time. 
In large-scale MAPF instances, these collision penalties or constraints can drastically affect how the agent explores the environment during training.

\subsection{State and Observation Spaces}
\label{subsec:rl_state_space}

In many MAPF setups, each agent can observe only part of the environment, such as its limited field of view (FOV). Consequently, the agent’s local state often contains only a subset of the global map. 
Table~\ref{tab:rl-state-spaces} details four common categories of observation. 
Each category can be used individually or in combination, depending on the complexity of the environment and the level of guidance needed.

\begin{table}[!ht]
\centering
\caption{Representative State (Observation) Partitioning in RL-based MAPF. 
Each category of state information contributes differently to collision avoidance and goal reaching.}
\label{tab:rl-state-spaces}
\begin{tabular}{p{2.8cm} p{2.9cm} p{6.3cm}}
\toprule
\textbf{Category} & 
\textbf{Information Examples} & 
\textbf{Remarks}\\
\midrule
\textit{Local static info} 
& 
Map obstacles, free cells in $W_{\text{FOV}}\times H_{\text{FOV}}$
& 
Provides a local occupancy grid around the agent, enabling immediate collision detection or obstacle avoidance~\citep{feng2024multi,chen2023transformer}. \\
\textit{Expert path guidance} 
& 
Recommended route from a classical solver (A*, D* Lite, etc.)
& 
Helps agents avoid getting lost in large or complex maps by providing explicit routing suggestions; drastically reduces RL sample complexity and training time~\citep{liu2020mapper,tang2024ensembling}.\\
\textit{Dynamic neighbor info} 
& 
Positions, velocities, or trajectories of nearby agents
& 
Enhances multi-agent collision avoidance by providing situational awareness of other agents in dense or dynamic settings~\citep{he2024alpha,wang2020mobile}.\\
\textit{Heuristic embedding} 
& 
Distance-to-goal, action feasibility, direction hints
& 
Gives agents a flexible sense of goal orientation (e.g., whether moving up brings them closer to the goal), combining classical heuristics with RL’s adaptive exploration~\citep{ma2021learning,song2023helsa,lin2023sacha}.\\
\bottomrule
\end{tabular}
\end{table}

\noindent
\textbf{Additional Discussion for Table~\ref{tab:rl-state-spaces}.} 
\begin{itemize}[leftmargin=*]
    \item \emph{Local static information.} This is often the minimum requirement in RL-based MAPF. Although it helps agents avoid immediate collisions, it may be insufficient for discovering long-horizon routes in large grids or environments with multiple bottlenecks. 
    \item \emph{Expert path guidance.} This allows each agent to receive suggestions from offline planners such as A* or D* Lite. By following or at least referencing these suggestions, agents reduce random exploration and can reach near-optimal routes more quickly.
    \item \emph{Dynamic neighbor information.} Sharing real-time positions or velocities of other agents is crucial in high-density scenarios; it permits emergent cooperative behavior in RL, such as implicit priority or yield policies at intersections.
    \item \emph{Heuristic embedding.} Instead of giving a single recommended path, some works prefer more general heuristic signals (e.g., Manhattan distance to goal). Agents then need to learn how to balance these heuristics with real-time local constraints and potential collisions.
\end{itemize}

\subsection{Action Spaces}
\label{subsec:action_space}

On a 2D grid, an agent typically has five actions $\{\textit{up}, \textit{down}, \textit{left}, \textit{right}, \textit{stay}\}$. 
In continuous domains (e.g., a nonholonomic robot in a 2D plane), the agent might select continuous velocities or steering commands, requiring continuous-action RL algorithms such as DDPG or TD3. 
When collisions are treated as hard constraints, an action that immediately causes a collision may be invalid. 
Alternatively, collisions can be penalized via large negative rewards (Section~\ref{subsec:reward_design}), disincentivizing agents from selecting dangerous maneuvers.

\subsection{Reward Design}
\label{subsec:reward_design}

Reward design is one of the most critical components of RL-based MAPF, balancing collision avoidance, efficiency, and multi-agent cooperation. 
Table~\ref{tab:rl-reward-structures} lists typical reward components, each focusing on different aspects of the MAPF objective. 
In practice, these rewards can be combined or weighted to form a single scalar signal.

\begin{table}[ht!]
    \centering
    \caption{Common Reward Structures in RL-based MAPF. Multiple components are often combined to guide agents effectively in large or dynamic environments.}
    \label{tab:rl-reward-structures}
    \begin{tabular}{m{3.0cm} m{5.2cm} m{5.2cm}}
    \toprule
    \textbf{Reward Type} & \textbf{Mathematical Formulation} & \textbf{Explanations/References}\\
    \midrule
    Goal-reaching 
    & 
    \(
    r^i_{\text{goal}}(s_t^i,a_t^i) 
    = 
    \begin{cases}
    +R_{G}, & \!\!\text{if }s_t^i=g_i \\
    -\varepsilon, & \!\!\text{otherwise}
    \end{cases}
    \)
    &
    Rewards the agent upon reaching its designated goal $g_i$. 
    $R_G>0$ is usually large to maximize success; 
    a small penalty $-\varepsilon$ encourages shorter paths~\citep{yin2024deep}.\\
    Cooperation  
    & 
    \(
    r^i_{\text{coop}}(s_t^i,a_t^i) 
    =
    \beta_1 \,r^i_{\text{indv}} + \beta_2\, r_{\text{team}}
    \)
    &
    Balances individual progress $r^i_{\text{indv}}$ with a shared team incentive $r_{\text{team}}$. 
    Encourages global success rather than local selfish behavior~\citep{zhao2023curriculum,tao2024poaql}.\\
    Expert-guided
    & 
    \(
    r^i_{\text{guide}}
    =
    \begin{cases}
    +R_{E}, & \!\!\text{if action \(\in\)\;expert path}\\
    0, & \!\!\text{otherwise}
    \end{cases}
    \)
    &
    Provides a positive reward $R_E>0$ for following a reference path (e.g., from A*) to reduce trial-and-error exploration~\citep{liu2020mapper,wang2020mobile,Pham2023OptimizingCM}.\\
    Collision penalty
    &
    \(
    r^i_{\text{collision}}(s_t^i,a_t^i)=-\alpha\,\mathbb{I}\{\text{collision at time }t\}
    \)
    &
    Strongly penalizes collisions ($-\alpha$), deterring agents from occupying the same cell or crossing edges simultaneously~\citep{qiu2020multi}. \\
    \bottomrule
    \end{tabular}
\end{table}

\noindent
\textbf{Additional Discussion for Table~\ref{tab:rl-reward-structures}.}
\begin{itemize}[leftmargin=*]
    \item \emph{Goal-reaching.} A common approach is to give a large positive reward only when the agent arrives at its target. Some variants also provide a small shaping reward for moving closer to the goal each step, improving learning speed but risking unintended local optima.
    \item \emph{Cooperation.} A purely local or per-agent reward may lead to greedy strategies. By adding a group-oriented term $r_{\text{team}}$, methods encourage agents to coordinate, reducing deadlocks or cycles.
    \item \emph{Expert-guided.} For difficult or sparse environments, referencing a path from classical planners (e.g., A*) significantly reduces RL training time. Approaches differ in how strictly they guide: some apply partial or decreasing weighting of the expert path.
    \item \emph{Collision penalty.} Typically, collisions incur a large negative reward to override other incentives. Alternatively, collisions may terminate the episode for the colliding agents, which also conveys a strong penalty signal.
\end{itemize}

\subsection{Communication Protocols in RL-based MAPF}
\label{subsec:comm}

Communication protocols determine how partial observations are shared among agents to improve collective decision-making. 
Table~\ref{tab:comm-rl} summarizes representative methods. 
Each approach modifies the local state $s^i_t$, integrating the messages or states received from neighbors.

\begin{table}[ht!]
\centering
\caption{Communication Mechanisms in Reinforcement Learning-based MAPF. 
Each approach modifies how local states $s^i_t$ are constructed, thereby affecting coordination quality.}
\label{tab:comm-rl}
\begin{tabular}{p{2.8cm} p{2.5cm} p{2.0cm} p{5.9cm}}
\toprule
\textbf{Type} & 
\textbf{Example Methods} &
\textbf{Scalability} &
\textbf{Further Details / References}\\
\midrule
\textit{Non-communication} 
& 
Local only
& 
High 
& 
Each agent relies entirely on its local information; collisions are avoided via local RL rules. Feasible if agents are sparse or environments are simple~\citep{chen2023transformer,qiu2020multi}.\\

\textit{Basic communication} 
& 
Message exchange
& 
Moderate
& 
Agents broadcast raw or summarized states (via GNNs) to neighbors, enabling more effective conflict resolution~\citep{Pham2023OptimizingCM}.\\

\textit{Priority-based} 
& 
Select top-$k$ neighbors
& 
Moderate
& 
Agents communicate only with the most relevant neighbors (e.g., highest conflict potential) to reduce bandwidth~\citep{ye2022multi,li2022multi}.\\

\textit{Attention-based} 
& 
Graph or Transformer 
& 
Moderate
& 
Agents weigh neighbors by attention mechanisms (e.g., BicNet, Graph Attention) to focus on the most critical relationships~\citep{guan2022ab,bignoli2021graph}.\\

\textit{Request-response} 
& 
On-demand triggers
& 
High 
& 
Agents broadcast or request info only when a neighbor’s action might influence their decision, limiting unnecessary messages~\citep{ma2021learning,zhou2024dhaa}.\\
\bottomrule
\end{tabular}
\end{table}

\noindent
\textbf{Additional Discussion for Table~\ref{tab:comm-rl}.}
\begin{itemize}[leftmargin=*]
    \item \emph{Non-communication.} Approaches with no inter-agent communication are typically easier to scale to many agents and ensure faster training, but they may lead to more collisions in dense areas. 
    \item \emph{Basic communication.} Agents may broadcast local states or partial observations to all neighbors within a certain radius. Social conventions (e.g., “move right if in conflict”) can emerge, but the overhead of repeated broadcasts can be high.
    \item \emph{Priority-based communication.} By allowing each agent to communicate only with those neighbors deemed “most critical,” networks avoid saturating communications. Various metrics (e.g., distance, possible collisions in the next steps) can establish these priorities.
    \item \emph{Attention-based communication.} Derived from modern deep learning architectures like Transformers, attention weighting helps each agent filter crucial messages. This is particularly helpful in scenarios with many neighbors.
    \item \emph{Request-response communication.} Agents solicit updates from others only when critical. This specialized approach can greatly reduce bandwidth consumption while preserving coordination, though it often requires more intricate logic at each agent.
\end{itemize}

\subsection{Overview of MARL Algorithms for MAPF}
\label{subsec:rl_algorithms}

In single-agent RL, each agent $i$ independently learns $\pi^i$ to maximize (\ref{eq:rl-objective}). 
However, the multi-agent setting introduces nonstationarity (since other agents are also learning) and partial observability. 
Below, we summarize popular multi-agent RL (MARL) paradigms and give short mathematical explanations so that MAPF practitioners can link them to classical MAPF solution concepts.

\subsubsection{Independent Learning}
Each agent $i$ runs a single-agent RL algorithm (e.g., DQN, PPO) treating all other agents as part of the environment. 
Though simple to implement, independent learners may converge slowly or fail to coordinate in dense MAPF scenarios. 
The agent’s Bellman update for action-value $Q^i$ is:
\[
Q^i_{t+1}(s_t^i,a_t^i)
\;\leftarrow\;
Q^i_t(s_t^i,a_t^i)
\;+\;
\eta \Bigl[
r^i_t 
  + \gamma\,\max_{a'}\,Q^i_t(s_{t+1}^i,a')
  - Q^i_t(s_t^i,a_t^i)
\Bigr].
\]

\subsubsection{Centralized Training, Decentralized Execution (CTDE)}
CTDE offers a more structured approach:
\[
Q_{\text{CT}}(s,a_1,\dots,a_n)
\;\; \text{with} \;\;
s = \bigl(s^1,\dots,s^n\bigr).
\]
During training, a \emph{centralized critic} has access to global states, actions, and possibly even agent IDs or goals. 
Once the critic $Q_{\text{CT}}$ is optimized, each agent executes a decentralized policy $\pi^i$ that conditions only on $s^i$. 
For instance, MADDPG uses a deterministic policy in continuous spaces:
\[
a^i = \mu^i_{\theta^i}(s^i), 
\]
while the critic $Q_{\Phi}$ is centralized:
\[
Q_{\Phi}(\mathbf{s},a_1,\dots,a_n)
= 
\mathbb{E}\Bigl[
r + \gamma \max_{a'} Q_{\Phi}(\mathbf{s}',a'_1,\dots,a'_n)
\Bigr].
\]
This improves coordination while preserving decentralized execution essential for MAPF solutions.

\subsubsection{Value Decomposition Methods}
For a team reward $r_{\text{team}}(\mathbf{s},\mathbf{a})$, value decomposition networks (VDN)~\citep{sunehag2018value} and QMIX~\citep{rashid2020monotonic} factorize the global action-value function into per-agent utilities, making the training feasible even if we have a single shared objective. For instance, VDN enforces:
\[
Q_{\text{VDN}}(\mathbf{s},\mathbf{a}) 
= 
\sum_i Q^i(s^i,a^i),
\]
while QMIX uses a monotonic mixing network:
\[
Q_{\text{QMIX}}(\mathbf{s},\mathbf{a})
=
f_{\text{mix}}\bigl(Q^1,\dots,Q^n;\,\mathbf{s}\bigr)\quad\text{with monotonic partial derivatives.}
\]
These methods are relevant when MAPF tasks optimize a global criterion (sum-of-costs or makespan). 
They can converge faster to coordinated solutions than purely independent methods.

\subsection{Representative Implementations}

Various MAPF studies integrate the above MARL techniques with specialized domain knowledge, such as neighbor-based collision checks or expert guidance. 
For example, \citep{Pham2023OptimizingCM} adopt a CTDE paradigm combined with graph neural networks (GNNs) for large warehouse pathfinding. 
\citep{ma2021learning} incorporate a QMIX-based approach where agents share a team reward, encouraging them to resolve conflicts cooperatively. 
These examples demonstrate how general MARL architectures (e.g., independent RL, CTDE, value decomposition) can be customized and scaled for multi-agent pathfinding.

\subsection{Summary and Future Directions}
\label{subsec:rl_summary_future}

Reinforcement learning has opened new avenues for MAPF, especially in decentralized control or partially observable, dynamic environments. 
However, its application is far from trivial. 
Table~\ref{tab:rl-limitations-solutions} summarizes some key limitations and highlights prospective solutions from the RL/MARL community.

\begin{table}[ht!]
\centering
\caption{Open Challenges of RL-based MAPF and Potential Research Avenues. Each challenge connects to broader RL/MARL directions, offering new opportunities for MAPF practitioners.}
\label{tab:rl-limitations-solutions}
\begin{tabular}{p{3.2cm}p{5.1cm}p{5.2cm}}
\toprule
\textbf{RL-based MAPF Challenge} 
& 
\textbf{Current Limitation} 
& 
\textbf{Possible RL/MARL Solutions}\\
\midrule
\textit{Scalability to large $n$} 
& 
Communication overhead and an exponential joint state space
& 
\begin{itemize}[leftmargin=1em]
\setlength\itemsep{0em}
\item Hierarchical RL for multi-level decisions.
\item Graph-based message passing with priority or attention.
\item Value decomposition for factorized learning.
\end{itemize}\\
\textit{Lack of theoretical guarantees} 
& 
No formal completeness or suboptimality bounds
& 
\begin{itemize}[leftmargin=1em]
\item Safe RL (e.g., barrier functions, constrained MDPs).
\item Hybrid frameworks with classical MAPF back-ends.
\end{itemize}\\
\textit{Reward shaping complexity} 
& 
Difficult to encode global MAPF objectives into local rewards
& 
\begin{itemize}[leftmargin=1em]
\item CTDE with a global critic that tracks collective performance.
\item Potential-based shaping to integrate classical heuristics.
\end{itemize}\\
\textit{Partial observability} 
& 
Agents are myopic, may cause collisions due to hidden areas
& 
\begin{itemize}[leftmargin=1em]
\item Recurrent policies or memory-based MARL.
\item On-demand or selective communication protocols.
\end{itemize}\\
\textit{Real-world complexity} 
& 
Uncertainties in dynamics, sensor noise, or kinematic constraints
& 
\begin{itemize}[leftmargin=1em]
\item Domain randomization or sim-to-real transfer to improve robustness.
\item Multi-fidelity simulation frameworks that gradually incorporate real-world constraints.
\end{itemize}\\
\bottomrule
\end{tabular}
\end{table}

\noindent
\textbf{Additional Discussion for Table~\ref{tab:rl-limitations-solutions}.} 
\begin{itemize}[leftmargin=*]
    \item \emph{Scalability.} Each agent’s learning complexity grows quickly as $n$ increases. Approaches that factorize value or adopt hierarchical structures (e.g., assigning subgoals or clusters of agents) can help manage complexity.
    \item \emph{Lack of theoretical guarantees.} Unlike classical MAPF, which can guarantee completeness or near-optimality under certain conditions, pure RL solutions lack formal proofs of collision-free motion. Integrating safe RL constraints or combining RL with classical verification tools remains an open area.
    \item \emph{Reward shaping complexity.} Designing rewards that encourage local progress but also ensure global success is nontrivial. Methods like potential-based shaping or difference rewards (which remove “baseline” contributions of other agents) can better align local and global goals.
    \item \emph{Partial observability.} Many realistic MAPF scenarios restrict an agent’s viewpoint. Memory-augmented RL (e.g., LSTM-based policies) or advanced communication schemes can mitigate some pitfalls of partial observations, but solutions remain highly domain-specific.
    \item \emph{Real-world complexity.} MAPF tasks in actual robotic deployments require robust solutions to handle agent dynamics, sensor noise, nonholonomic constraints, and cluttered or changing environments. Domain randomization and sim-to-real strategies can partially mitigate these challenges, but bridging the gap between simulation and reality is far from solved.
\end{itemize}

Overall, RL-based MAPF solutions capitalize on data-driven adaptivity, offering a unique advantage in dynamic or partially known settings. 
Future research can continue to refine these methods, balancing learning efficiency, communication strategies, and theoretical safety to achieve robust, scalable performance in real-world multi-agent coordination.

\subsection{Beyond Grid-World: More Advanced Environments}
\label{subsec:advanced_envs}

While much of the existing reinforcement learning (RL) research on MAPF (see Section~\ref{sec:rl}) focuses on two-dimensional grid worlds, real-world applications often depart markedly from this simplified model. 
These departures include embedding agents in 3D spaces (e.g., autonomous drones), allowing for continuous-valued states and actions, or imposing time-varying obstacles in dynamic environments. 
Moreover, certain real-world infrastructures---such as railway networks---require graph-based topology and additional motion constraints. 
This subsection reviews these non-traditional MAPF environments with an eye toward the mathematical modeling that aligns with Section~\ref{sec:formulation}’s general MAPF framework.

\subsubsection{Mathematical Formulation Across Environments}
To unify different environment types, we recall from Section~\ref{sec:formulation} that classical MAPF is modeled by a graph \(\mathcal{G}=(\mathcal{V},\mathcal{E})\). 
Below, we highlight the mathematical variations for each environment category. 
Table~\ref{tab:env-types} provides a concise comparison.

\begin{sidewaystable}
\centering
\caption{Representative Environment Types for MAPF, Illustrating Key Variables and Constraints.}
\label{tab:env-types}
\resizebox{\linewidth}{!}{%
\begin{tabular}{p{0.2\linewidth} p{0.4\linewidth} p{0.4\linewidth}}
\toprule
\textbf{Environment} & \textbf{Modeling Variables and Constraints} & \textbf{Representative References} \\
\midrule
\textbf{2D Grid} 
& 
\(\mathcal{G}\) is a 2D lattice; 
\(\mathcal{V} = \{(x,y)\}\), 
\(\mathcal{E}=\{\dots\}\). 
Actions are often discrete: \( \{\textit{up},\textit{down},\dots\} \). 
Constraints: adjacency in \(\mathcal{E}\). 
& 
\citep{skrynnik2024pogema} and many others rely on standard 2D grid benchmarks. \\
\textbf{3D Environment} 
& 
Extend grid or lattice to 3D cells: 
\(\mathcal{V} = \{(x,y,z)\}\). 
Actions typically \(\{\pm x,\pm y,\pm z,\textit{stay}\}\). 
Additional collision checks for 3D adjacency. 
& 
\citep{zhiyao2020deep} (PRIMALc for UAVs).\\
\textbf{Continuous Space} 
& 
Replace discrete graph with continuous domain \(\Omega \subseteq \mathbb{R}^d\). 
Agent $i$ has continuous state \(\mathbf{x}_i(t) \in \Omega\). 
Movement governed by \(\dot{\mathbf{x}}_i=f(\mathbf{x}_i,\mathbf{u}_i)\). 
High-dimensional collision constraints.  
& 
\citep{liu2024multi-agent,qiu2020multi,fan2020distributed}.\\
\textbf{Graph-based Infrastructure} 
& 
\(\mathcal{G}\) can be a strongly connected (directed) network or a specialized railway graph (Flatland). 
Agents restricted to follow tracks: 
\(\mathcal{V}=\{\text{junctions}\}, \mathcal{E}=\{\text{links}\}\). 
& 
\citep{mohanty2020flatland,van2021time}.\\
\textbf{Dynamic Environment} 
& 
Environment state evolves via a function \(\phi(t)\) that adds/deletes edges in \(\mathcal{E}\) or modifies obstacles in \(\mathcal{V}\). 
MAPF feasible sets change over time. 
& 
\citep{xie2024improved,ou2024reinforcement,kong2024multi}.\\
\bottomrule
\end{tabular}%
}
\end{sidewaystable}

\paragraph{State and Transition Functions.}
For grid or 3D lattice environments, agent positions typically remain on discrete vertices, with transitions governed by adjacency. 
By contrast, continuous-space formulations let each agent $i$ maintain a position \(\mathbf{x}_i(t)\in \Omega\subseteq \mathbb{R}^{d}\). 
The environment then defines transition dynamics:
\[
\mathbf{x}_i(t + 1)
= 
\mathbf{x}_i(t) 
\;+\; 
\Delta t \cdot \mathbf{u}_i(t), 
\]
or more complex non-linear motion constraints \(\dot{\mathbf{x}}_{i} = f_{i}(\mathbf{x}_i,\mathbf{u}_i)\). 
For dynamic variants, the set of permissible states or edges can evolve each timestep. 
In an RL setting, each agent’s local observation must capture these changes, either through dynamic occupancy grids~\citep{xie2024improved} or sensor measurements~\citep{qiu2020multi}.

\paragraph{Action Spaces.}
In discrete graphs (2D or 3D), each agent’s action space \(\mathcal{A}^i\) corresponds to stepping into an adjacent vertex if no collision occurs. 
For continuous domains, \(\mathcal{A}^i\) may be a continuous set (forces, velocities, turning angles), as reported in \citep{liu2024multi-agent,fan2020distributed}, or discretized approximations for safe flight corridors in drone applications~\citep{zhiyao2020deep}. 
Graph-based transit systems (e.g., railways) often require specialized “movement rules,” disallowing direct transitions outside the track network~\citep{mohanty2020flatland}.

\paragraph{Reward and Constraints.}
Regardless of dimensionality, RL-based methods frequently penalize collisions and reward reaching goals quickly. 
However, certain tasks (e.g., multi-train scheduling) incorporate \emph{hard safety constraints} or even higher-level scheduling costs (e.g., lateness penalties). 
In dynamic environments, time-varying elements factor into collisions:
\begin{equation}
\label{eq:dynamic-collision}
\mathbb{I}\bigl\{\mathbf{x}_i(t)\in \phi_{\text{obs}}(t)\bigr\},
\end{equation}
where \(\phi_{\text{obs}}(t)\) denotes the region of newly introduced obstacles at time $t$. 
Such expansions demand specialized RL reward structures (Table~\ref{tab:rl-reward-structures} in Section~\ref{subsec:reward_design}) and can be augmented by communication-based strategies (Table~\ref{tab:comm-rl} in Section~\ref{subsec:comm}).

\paragraph{Implications for RL.}
Moving beyond grid-worlds often increases state and action dimensionality, intensifies partial observability, and requires advanced exploration or communication protocols. 
Effective RL policies may integrate local sensor data (continuous domains) or specialized heuristics/communication to handle dynamic graphs. 
Approaches such as hierarchical RL or centralized training~\citep{Pham2023OptimizingCM} become especially relevant for environments with higher complexity.

\subsubsection{Case Studies of Advanced Environments}

\paragraph{3D UAV Coordination.}
In \citep{zhiyao2020deep}, the authors extend 2D PRIMAL to a 3D search space called \emph{PRIMALc}. 
Agents model future conflict states by exchanging predicted actions, effectively building a communication channel. 
From a mathematical viewpoint, each agent $i$’s state is \(\mathbf{x}_i(t)=(x_i,y_i,z_i)\), and collisions are enforced by $\|\mathbf{x}_i(t) - \mathbf{x}_j(t)\|_{2} \ge \delta$ for any $j \neq i$. 
Using careful reward shaping and gradient clipping stabilizes the imitation loss, yielding stable flight coordination among UAVs in a partially observable environment.

\paragraph{Continuous Robot Navigation.}
\citet{liu2024multi-agent} formulate each robot’s action as a continuous force vector, capturing underlying physical dynamics. 
Let \(\mathbf{u}_i(t)\in \mathbb{R}^2\) be the planar force, and agent $i$’s dynamics follow
\[
m \,\ddot{\mathbf{x}}_i(t) 
= 
\mathbf{u}_i(t),
\]
where $m$ is mass. 
Their RL training penalizes collisions and encourages minimal path length in continuous space. 
Similarly, \citet{qiu2020multi,fan2020distributed} map sensor observations to continuous velocity or steering actions and verify collision-avoidance in real or simulated prototypes.

\paragraph{Graph-Based Railway Scheduling.}
In the automated train scheduling context~\citep{mohanty2020flatland,van2021time}, trains follow track networks represented by a directed graph \( \mathcal{G}=(\mathcal{V}, \mathcal{E})\). 
Each vertex $v \in \mathcal{V}$ can hold at most one train (or a limited capacity of trains), and each edge imposes travel-time constraints. 
RL-based methods must coordinate schedules at each switch junction (vertex), using a multi-agent policy that respects collision-free track occupancy.

\paragraph{Dynamic Obstacle Handling.}
Dynamic MAPF (DMAPF) extends the standard setup by continuously altering obstacle configurations. 
\citet{xie2024improved} introduce a random process with probability $p_o$ adding or removing obstacles. 
In RL terms, the agent’s local state includes a dynamic occupancy grid, and the RL update must incorporate time-varying feasible sets and collision checks as in Eq.~\eqref{eq:dynamic-collision}. 
Similarly, \citet{ou2024reinforcement} propose GAR-CoNav, where graph attention networks process changing environmental features, guiding multi-agent navigation with minimal real-time collisions. 
Such dynamic setups highlight the demand for real-time, adaptive RL policies.

\vspace{1em}
\noindent
\textbf{Summary.} 
While discrete 2D grid worlds remain an invaluable starting point for RL-based MAPF, advanced environments---3D cells, continuous domains, specialized graphs, and dynamic obstacles---better reflect real-world scenarios but also pose higher computational complexity, stricter collision constraints, and more involved communication requirements. 
High-performing methods typically blend domain knowledge (e.g., sensor usage, specialized collisions checks) with advanced RL strategies (hierarchical or GNN-based).

\subsection{Beyond One-Shot Tasks: Lifelong Learning}
\label{subsec:lifelong}

In the classical MAPF setting, each agent is assigned a single start and goal location (one-shot task). 
Once an agent reaches its goal, it no longer partakes in further planning, effectively halting the problem. 
However, many real-world applications (e.g., large-scale warehouses, delivery fleets) follow an \emph{ongoing} or \emph{lifelong} mode, where agents must be re-tasked repeatedly with new objectives. 
This difference in problem setup impacts system-level metrics such as throughput, computation frequency, and scheduling. 
Here, we first distinguish mathematically between one-shot and lifelong MAPF and then discuss representative RL solutions that handle repeated re-planning.

\subsubsection{Mathematical Formulation: One-Shot vs. Lifelong}

Recall from Section~\ref{sec:formulation} that in a one-shot MAPF, each agent $i$ has a start vertex \(s_i\) and a goal vertex \(g_i\). 
The solution typically seeks to minimize makespan~\eqref{eq:makespan} or sum-of-costs~\eqref{eq:soc}, subject to collision-free constraints. 
In contrast, \emph{lifelong MAPF (LMAPF)} assigns agent $i$ a \emph{sequence} of goals, $\{g_{i}^1,g_{i}^2,\dots\}$, with possibly only the next target known at each step. 
Denote by $T_i$ the set of tasks for agent $i$, where task $j$ is $(s_{i}^j,g_{i}^j,\tau_i^j)$ specifying an origin, destination, and the time $\tau_i^j$ at which the goal is announced (if known in advance). 
The RL agent faces a repeated (or continual) planning scenario:
\[
\min_{\{\pi_i\}_{i=1}^n}
\; 
\sum_{j} \mathrm{Cost}(\pi_i^j) 
\quad
\text{subject to collision-free constraints across all tasks.}
\]
Moreover, some approaches aim to maximize throughput or the average number of completed tasks per unit time:
\begin{equation}\label{eq:lifelong-obj}
\max_{\{\pi_i\}} 
\;\;
\frac{\sum_{i=1}^n \text{TasksCompleted}_i(\{\pi_i\})}{T_{\text{horizon}}}.
\end{equation}
The environment can also be partially observable or dynamic (Section~\ref{subsec:advanced_envs}).
Table~\ref{tab:one-shot-lifelong} summarizes the conceptual and mathematical distinctions between one-shot and lifelong tasks.

\begin{table}[ht!]
\centering
\caption{Comparison of One-Shot vs. Lifelong MAPF in Terms of Key Variables and Incentives.}
\label{tab:one-shot-lifelong}
\begin{tabular}{p{2.5cm} p{4.3cm} p{5.5cm}}
\toprule
& \textbf{One-Shot MAPF} & \textbf{Lifelong MAPF} \\
\midrule
\textbf{Goal Set} & 
Static $\bigl(s_i \to g_i\bigr)$, single assignment 
& 
Sequential $\bigl(s_i \to g_i^1 \to g_i^2 \dots\bigr)$; possibly indefinite \\
\textbf{Objective} & 
Minimize makespan or sum-of-costs for one final solution 
& 
Minimize cumulative cost over repeated tasks or maximize throughput \\
\textbf{Policy} & 
Terminal once $g_i$ is reached 
& 
Continuous re-planning; new tasks appear dynamically \\
\textbf{RL Implication} &
Single-episode design; fewer exploration phases 
& 
Multi-episode or indefinite horizon; RL must adapt to new tasks \\
\textbf{Example Methods} &
Conflict-based or compilation-based solutions for fixed instance 
& 
PRIMAL2~\citep{damani2021primal}, 
Re-planning modules~\citep{chen2023towards}, 
Priority-based methods~\citep{gao2024pce} \\
\bottomrule
\end{tabular}
\end{table}

\subsubsection{Representative RL-Based Approaches}

\paragraph{PRIMAL2 for Dense Warehouses.}
\citet{damani2021primal} extend the PRIMAL approach to \emph{PRIMAL2}, targeting local, fully decentralized policies for agents that repeatedly receive new tasks in constrained and partially observable environments. 
The RL module is structured to handle real-time pathfinding as tasks arrive, with each agent $i$ maintaining a policy $\pi^i$, refined to adapt to dynamic reassignments $(s_i^j,g_i^j)$. 
Training includes reshaping local observations, so that each new subtask is seamlessly integrated into the agent’s partial view.

\paragraph{Re-planning Modules in Hybrid RL.}
\citet{chen2023towards} propose a \emph{Re-planning Module} that merges classical path planners with RL. 
Once an agent finishes a task, the module re-invokes a local or global RL-based solver to assign and plan for the next task. 
This pipeline ensures that partial solutions from past tasks can quickly adapt to new goals. 
From an MDP perspective, the state now tracks not only the agent’s current position but also the \emph{progress} (or idle time) since the last assignment.

\paragraph{Priority-Aware Communication \& Lifelong Scheduling.}
\citet{gao2024pce} present a \emph{Priority-aware Communication \& Experience replay} subroutine (PCE) suited for repeated tasks. 
In each re-planning phase, the system reassigns tasks to idle agents, and a priority-based communication ensures minimal collisions among active missions. 
This approach unifies ongoing scheduling decisions with standard RL updates, enabling robust performance over extended horizons.

\subsubsection{Challenges and Open Directions in Lifelong MAPF}
Adopting a lifelong viewpoint introduces additional complexities not found in single-shot tasks:
\begin{itemize}[leftmargin=1em]
    \item \emph{Frequent Re-planning.} Agents continually switch goals, requiring fast, incremental RL updates or real-time inference. 
    \item \emph{Scheduling Coupling.} At high agent densities (e.g., warehouses), the scheduling of new tasks can create bottlenecks, so RL must balance finishing old tasks quickly with preparing for new ones.
    \item \emph{State Explosion.} Each agent’s MDP must encode which task it is pursuing, local environment states, and partial observations. This might be mitigated by hierarchical RL or modular sub-polices (one per task type).
    \item \emph{Performance Metrics.} While sum-of-costs or makespan suffices for one-shot tasks, repeated assignment may require throughput (\ref{eq:lifelong-obj}) or time-averaged productivity measures akin to those used in queueing theory.
\end{itemize}

\vspace{1em}
\noindent
\textbf{Summary.} 
Switching from a one-shot perspective to lifelong MAPF captures the continuous nature of real-world operations, demanding greater adaptability from RL-based approaches. 
Agents must repeatedly plan while ensuring minimal collisions and high throughput. 
Recent works (e.g., PRIMAL2~\citep{damani2021primal}, \citep{chen2023towards,gao2024pce,matsui2023investigation,ma2017lifelong,skrynnik2024learn}) exemplify how multi-agent RL can handle indefinite sequences of tasks, but the field remains open for more sophisticated scheduling policies, dynamic communication schemes, and advanced reward shaping tailored to high-throughput, multi-round settings.

\begin{table*}[!htbp]
    \centering
    \resizebox{\textwidth}{!}{%
    \begin{tabular}{llllll}
    \toprule
    \multicolumn{1}{l}{}& \multicolumn{2}{l}{framework} & \multicolumn{3}{l}{categorization} \\
    & \multicolumn{2}{l}{\rule{4cm}{0.4pt}} &  \multicolumn{3}{l}{\rule{6cm}{0.4pt}} \\[0.5cm] 
    Model & information  & reward  & comm. & env & target \\
    \midrule
    PRIMAL~\citep{sartoretti2019primal} & heuristic & goal-reaching & non & grid & one-shot \\
    DHAA~\citep{zhou2024dhaa} & Local & goal-reaching & selective & grid & one-shot \\
    DHC~\citep{ma2021distributed} &heuristic& goal-reaching & attention-based & grid &one-shot  \\
    CPL~\citep{zhao2023curriculum} & heuristic & cooperation & non &grid&one-shot \\ 
    CACTUS~\citep{phan2024confidence} & heuristic & cooperation & non & grid & one-shot \\
    C3PIL~\citep{xie2024crowd} & heuristic & goal-reaching & basic & grid & one-shot \\ 
    MAPPER~\citep{liu2020mapper} & expert path+dynamic & expert-guidance & non & dynamic & one-shot \\
    AB-Mapper~\citep{guan2022ab} & expert path+dynamic & expert-guidance & attention-based & dynamic & one-shot \\
    CRAMP~\citep{Pham2023OptimizingCM} & heuristic & goal-reaching & attention-based & grid & one-shot \\
    ALPHA~\citep{he2024alpha} & global+dynamic & expert-guidance & non & grid & one-shot \\
    \citet{ye2022multi} & dynamic & goal-reaching & priority-based & grid & one-shot\\
    PICO~\citep{li2022multi} & heuristic & goal-reaching & priority-based & grid & one-shot \\
    DCC~\citep{ma2021learning} & heuristic & goal-reaching & selective & grid & one-shot \\
    EPH~\citep{tang2024ensembling} & expert path & goal-reaching & selective & grid & one-shot \\
    PRIMALc~\citep{zhiyao2020deep} & heuristic & goal-reaching & basic & 3D & one-shot \\
    \citet{liu2024multi-agent} & heuristic & goal-reaching & non & continuous & one-shot \\
    PRIMAL2~\citep{damani2021primal} & heuristic+dynamic & goal-reaching & non & grid & lifelong \\
    
    \end{tabular}%
    }
\end{table*}

\section{Other Learning Paradigms}\label{sec:others}

In addition to the reinforcement learning (RL) strategies described in Section~\ref{sec:rl}, a variety of other learning-based paradigms have emerged for multi-agent path finding (MAPF). 
These paradigms treat MAPF as a sequential decision-making or prediction problem, often integrating with the classical MAPF formulation in Section~\ref{sec:formulation} by re-defining objectives, constraints, or feasible sets. 
Unlike purely RL-based methods, which directly learn a policy from trial-and-error interactions, the approaches discussed here can leverage expert demonstrations, heuristic expansions, population-based search, or hybrid search-learning loops. 
This section provides a structured survey of four major categories of learning paradigms: 
(i) Monte Carlo Tree Search, 
(ii) Supervised Learning, 
(iii) Composite Learning Strategies, 
and 
(iv) Evolutionary Methods. 
We highlight how each method formally maps the standard MAPF constraints (e.g., collision-avoidance) onto distinct optimization variables and search processes.

\vspace{1em}
\noindent
\textbf{Roadmap for this Section.} 
Table~\ref{tab:paradigm-comparison} offers a broad comparison of the mathematical modeling and typical constraints across these learning paradigms. 
Subsequent subsections provide deeper discussion and references for each approach.

\begin{sidewaystable}
    \centering
    \caption{Comparison of Selected Non-RL Learning Paradigms in MAPF. 
    For each paradigm, we outline how MAPF is cast in terms of optimization variables, solution space, cost or reward objectives, and constraint enforcement.}
    \label{tab:paradigm-comparison}
    \resizebox{\linewidth}{!}{%
    \begin{tabular}{p{0.1\linewidth} p{0.3\linewidth} p{0.3\linewidth} p{0.3\linewidth}}
    \toprule
    \textbf{Paradigm} & \textbf{Optimization Variables} & \textbf{Cost / Reward Objective} & \textbf{Collision Avoidance Constraint} \\
    \midrule
    \textbf{MCTS} 
    & Node expansions at each decision step; selection policy over state-action trajectories 
    & 
    MCTS rollout reward, often combining:
    \begin{itemize}[leftmargin=*,noitemsep]
    \item Goal distance
    \item Collision penalties
    \end{itemize}
    & 
    \begin{itemize}[leftmargin=*,noitemsep]
    \item Pruning or prohibiting expansions leading to collisions
    \item Negative rollout returns if collisions occur
    \end{itemize}\\
    \textbf{Supervised Learning} 
    & Model (e.g., neural network) parameters that map environment states to next actions or path predictions 
    & 
    \begin{itemize}[leftmargin=*,noitemsep]
    \item Minimizing prediction error vs.\ expert labels
    \item Possibly domain-specific (makespan / SoC)
    \end{itemize}
    & 
    \begin{itemize}[leftmargin=*,noitemsep]
    \item Indirectly encoded via supervised data
    \item Network trained to predict collision-free steps
    \end{itemize}\\
    \textbf{Composite Learning (e.g., IL+RL, Curriculum)} 
    & Shared or modular policy parameters across multiple learning schemes 
    & 
    \begin{itemize}[leftmargin=*,noitemsep]
    \item Combination of RL rewards, IL losses, or staged curriculum objectives
    \item Often balancing short paths and conflict avoidance
    \end{itemize}
    & 
    \begin{itemize}[leftmargin=*,noitemsep]
    \item Collision penalty integrated into RL portion
    \item Imitation from collision-free expert trajectories
    \end{itemize}\\
    \textbf{Evolutionary Methods} 
    & Population of candidate solutions; genetic encoding of agent paths or policies 
    & 
    \begin{itemize}[leftmargin=*,noitemsep]
    \item Fitness function: $F(\{\pi_i\}) = -(\text{SoC}+\alpha\,\text{numCollisions})$
    \end{itemize}
    & 
    \begin{itemize}[leftmargin=*,noitemsep]
    \item Hard constraints in evolutionary search for feasible paths
    \item Mutations avoiding collision states 
    \end{itemize}\\
    \bottomrule
    \end{tabular}%
    }
\end{sidewaystable}

\subsection{Monte Carlo Tree Search (MCTS)}\label{subsec:mcts}

\begin{figure}[htb!]
    \centering
    \includegraphics[width=\linewidth]{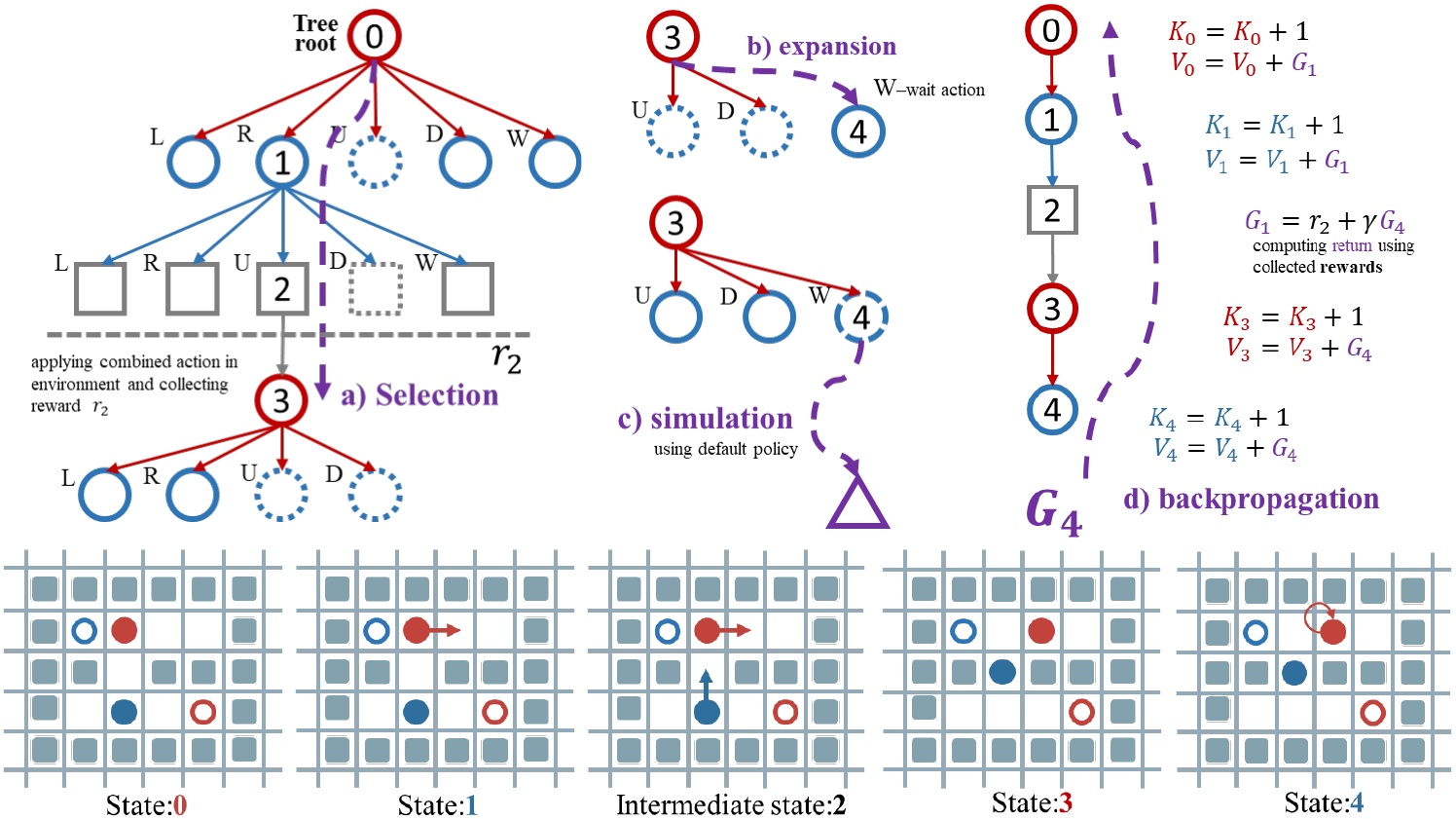}
    \caption{A flowchart of solving the MAPF problem using Monte Carlo Tree Search (MCTS). The tree structure represents each agent’s action as a separate node, reducing the branching factor in multi-agent settings. The diagram illustrates the four stages of MCTS: (a) Selection – choosing actions based on the current state; (b) Expansion – adding new nodes for possible actions; (c) Simulation – simulating outcomes using a default policy; and (d) Backpropagation – updating the tree with computed rewards. Agents and goals are shown on a grid as solid and hollow circles, respectively, with dashed-line actions excluded during selection.}
    \label{fig:mcts}
\end{figure}

Monte Carlo Tree Search (MCTS) is a general-purpose sampling-based planning algorithm widely used in sequential decision making.
Although it is often categorized under \emph{model-based} reinforcement learning, MCTS can be integrated into MAPF solvers in ways that differ from typical RL pipelines.
A flowchart of solving the MAPF problem using MCTS is shown in Figure~\ref{fig:mcts}.
This subsection provides a concise mathematical view of MCTS in the context of MAPF and highlights representative works that incorporate MCTS for collision avoidance.

\subsubsection{Mathematical Modeling of MCTS for MAPF}

Recall from Section~\ref{sec:formulation} that each agent $i$ has a start vertex $s_i\in\mathcal{V}$ and a goal $g_i\in\mathcal{V}$. 
We may treat joint MAPF states $\mathbf{s}=(s^1,\dots,s^n)$ and define a tree-based search over feasible joint actions $\mathbf{a}=(a^1,\dots,a^n)$, each of which corresponds to a proposed movement in the shared graph $\mathcal{G}$. 
At each node in the MCTS tree:
\[
(\text{State})\quad \mathbf{s}_t \;\longmapsto\; (\text{Children})\quad \{\mathbf{s}_{t+1}\}
\]
feasible next states are derived by applying valid joint actions. 
During MCTS, one typically runs four main steps:

\paragraph{1) Selection:} 
Select a path from the root to a leaf by choosing actions that balance exploitation (high-value actions) and exploration (less-visited actions). 
A common strategy is the Upper Confidence Bound for Trees (UCT):
\begin{equation}
\label{eq:uct}
a^* 
=\;
\arg \max_{a \in \mathcal{A}}
\Bigl[
Q(\mathbf{s},a) 
\;+\; 
c \sqrt{
\frac{\ln \bigl(\sum_{b}N(\mathbf{s},b)\bigr)}{N(\mathbf{s},a)}
}
\Bigr],
\end{equation}
where $Q(\mathbf{s},a)$ is the estimated action value, $N(\mathbf{s},a)$ is the visit count, and $c$ is an exploration constant.

\paragraph{2) Expansion:} 
Once a leaf node is reached, new child states are added to the tree by enumerating feasible subsequent joint actions. 
In MAPF, expansions must check collision constraints:
\[
\mathbf{s}_{\text{child}} 
\;\in\;
\Bigl\{
\mathbf{s}_{\text{leaf}} + \mathbf{a} 
\;\mid\;
\text{no collision among }\{a^1,\dots,a^n\}
\Bigr\}.
\]

\paragraph{3) Simulation:} 
From the newly expanded node, random or heuristic-based rollouts simulate the agent movements until a terminal state is reached (e.g., all agents arrive at their goals, or a collision occurs). 
A scalar payoff $R_{\text{rollout}}$ is then computed, often penalizing collisions or incomplete paths.

\paragraph{4) Backpropagation:} 
Propagate the rollout returns $R_{\text{rollout}}$ upwards to update $Q(\mathbf{s},a)$ along the selection path. 
Subsequent MCTS iterations refine these value estimates.

\subsubsection{Applications in MAPF}
MCTS-based MAPF approaches vary by how they incorporate collision penalties and domain heuristics:

\noindent
\textbf{Conflict-Avoidance with Rollout Patches.} 
\citep{skrynnik2021hybrid} employ Proximal Policy Optimization (PPO) to learn path-planning behaviors and then use MCTS expansions for collision resolution. 
The MCTS simulation step focuses on detecting conflicts quickly, effectively patching local collisions before committing to a joint action in the real environment. 
Similarly, \citep{pitanov2023monte} propose a specialized MCTS variant, guiding agents to avoid collisions via local partial plans.

\noindent
\textbf{Decentralized MCTS.} 
\citep{skrynnik2024decentralized} adapt MCTS to a multi-agent decentralized setting in a lifelong MAPF scenario, where each agent runs a local MCTS with partial observations and limited communication. 
They incorporate a rolling horizon scheme: each agent simultaneously expands possible paths for a short horizon, merges local expansions with neighbor observations, and replans as needed.

\noindent
\textbf{Multi-step Tree Search for Subgoal Allocation.} 
\citep{zhang2020learning} propose multi-step ahead tree search (MATS), which extends MCTS rollouts over multi-discrete state expansions. 
This helps in partial observability, where each agent infers likely collisions from neighbor expansions. 
The resulting MATS method yields robust collision avoidance in dense grid maps.

\subsubsection{Advantages and Limitations}
MCTS ensures systematic exploration of the joint decision tree, offering interpretable expansions of feasible MAPF states. 
However, the branching factor can be large if each agent can move freely; classical MCTS expansions may thus become computationally expensive with many agents. 
Heuristic or learned rollouts are crucial to managing complexity in large-scale MAPF.

\subsection{Supervised Learning}\label{subsec:supervised}

\begin{figure}[htb!]
    \centering
    \includegraphics[width=\linewidth]{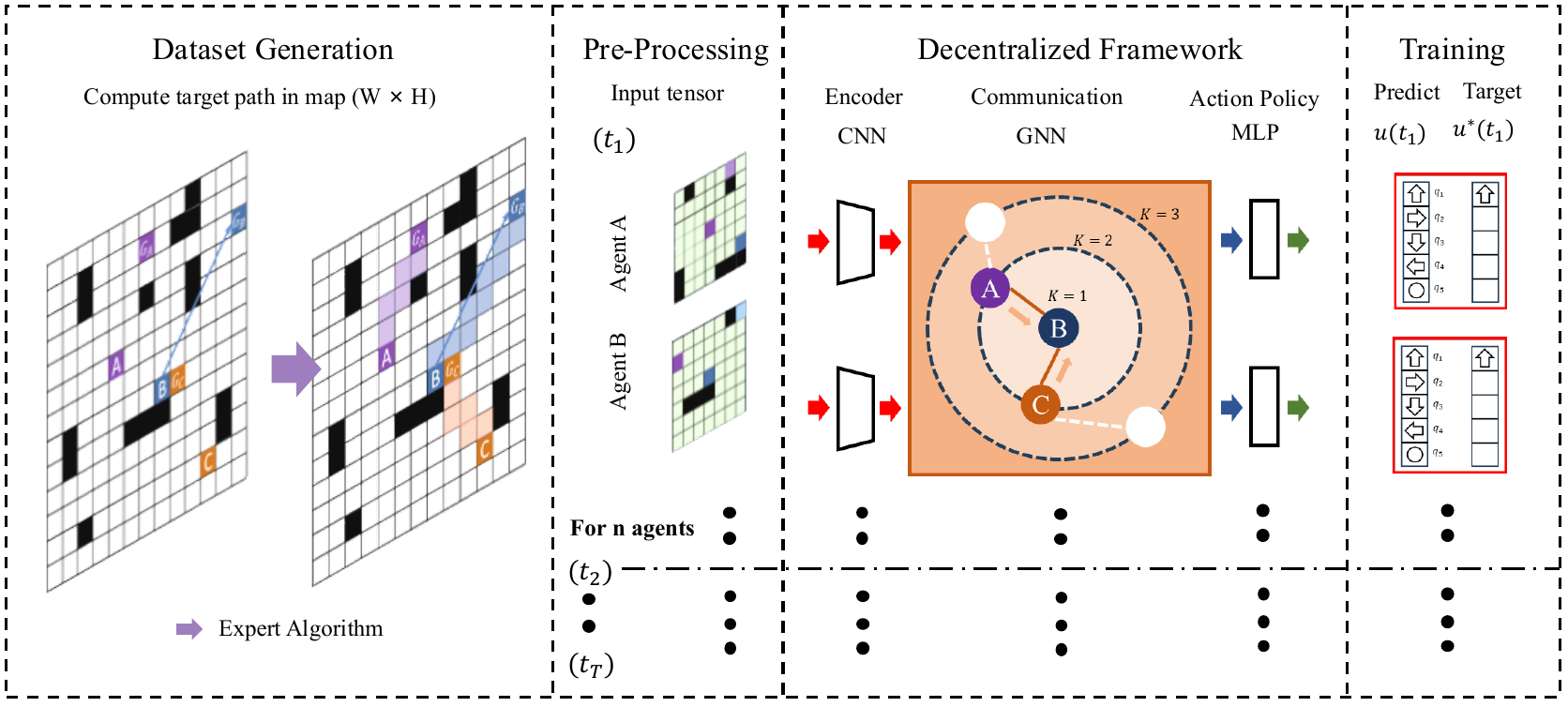}
    \caption{A supervised learning framework for solving the MAPF problem. The process includes four main stages: (1) Dataset generation – using expert algorithms to compute optimal paths on a given grid; (2) Preprocessing – extracting local observations for each agent; (3) Feature processing – applying convolutional encoders and graph neural networks for multi-hop communication, followed by MLP-based policy output; and (4) Training – using expert actions to supervise learning. This process is repeated across multiple agents and time steps.}
    \label{fig:supervised}
\end{figure}

Supervised learning (SL) leverages datasets of \emph{expert demonstrations} or \emph{labeled solutions} to train a predictive model (e.g., a neural network) that outputs recommended actions or paths for MAPF. 
In this sense, SL attempts to \emph{approximate} a collision-free solver’s behavior without performing an explicit search at inference time.
A supervised learning framework for solving the MAPF problem is shown in Figure~\ref{fig:supervised}.
This subsection covers two main SL-based approaches: \emph{assisting classical solvers} and \emph{direct solution predictions}.

\subsubsection{Mathematical Formulation of SL-based MAPF}
In supervised MAPF, we typically assume a labeled dataset $\mathcal{D}=\{(\mathbf{x}_k,\mathbf{y}_k)\}_{k=1}^M$, where:
\[
\mathbf{x}_k \;=\; \text{map features, agent positions, obstacles, \ldots}, 
\]
\[
\mathbf{y}_k \;=\; \text{expert next actions or entire agent paths from a known solver}.
\]
The learning objective is to fit model parameters $\theta$ (e.g., in a neural network) that minimize a prediction loss:
\begin{equation}\label{eq:SL-obj}
\theta^*
\;=\;
\arg \min_\theta
\sum_{k=1}^M
\mathcal{L}\bigl(f_\theta(\mathbf{x}_k),\, \mathbf{y}_k\bigr).
\end{equation}
Once trained, $f_\theta$ can produce near-instant path recommendations. 
Collision avoidance is implicit, learned from the collision-free labels $\mathbf{y}_k$.

\subsubsection{Assisting Existing Solvers}
\noindent
\textbf{Large Neighborhood Search (LNS) Guidance.} 
When integrated with LNS-based MAPF solvers (Section~\ref{sec:search}), SL can guide the selection of agent subsets to be re-optimized:
\begin{itemize}[leftmargin=*,noitemsep]
    \item \citep{huang2022anytime} train a Support Vector Machine (SVM) to rank promising subsets of agents for neighborhood destruction. 
    \item \citep{yan2024neural} propose a CNN-attention architecture to guide LNS expansions.
\end{itemize}
The new selection rule replaces hand-crafted heuristics, often yielding faster or higher-quality solutions.

\noindent
\textbf{Solver Selection.} 
\citep{zapata2024anytime} train an XGBoost classifier to choose the most suitable classical solver (e.g., CBS vs.\ priority-based methods) given input features (graph size, agent density, time limit). 
This meta-learning approach attempts to quickly pick an appropriate solver for each MAPF instance.

\subsubsection{Direct Solution Methods}
\noindent
\textbf{Imitation from CBS or A*.} 
A popular strategy is \emph{imitation learning} from classical MAPF solutions:
\begin{align}
\text{Given } \{\pi^*_i\}_{i=1}^n &\;\; \text{(expert paths from CBS or A*)}, \notag \\
\theta^* 
&=\;
\arg\min_\theta
\sum_i 
\sum_{\text{states } s_t^i} 
\mathcal{L}\Bigl(
f_\theta\bigl(\mathrm{Enc}(s_t^i)\bigr),
a_t^*\Bigr),
\label{eq:il-loss}
\end{align}
where $a^*_t$ is the expert action label at time $t$, and $\mathrm{Enc}(s_t^i)$ denotes some encoding of the agent’s local observation or map neighborhood.

Examples include:
\begin{itemize}[leftmargin=*,noitemsep]
    \item \textbf{CTRMs}~\citep{okumura2022ctrms}: 
    The authors introduce Cooperative Timed Roadmaps to focus on \emph{critical positions} for collision-free navigation, imitating CBS solutions.
    \item \textbf{MAGAT}~\citep{li2021message}: 
    A graph-attention-based model where each agent processes messages from neighbors and imitates paths computed by Enhanced Conflict-Based Search (ECBS).
    \item \textbf{CNN-GNN Integrations}~\citep{li2023multi, bignoli2021graph}: 
    These works embed 2D grid observations via CNN and then apply GNN layers to coordinate multi-robot collision avoidance, training the entire model to mimic an expert solver’s next-step guidance.
\end{itemize}

\subsubsection{Pros and Cons of SL-based MAPF}
Once trained, SL inference is typically very fast (sub-millisecond for moderate-scale networks), making it attractive for large or time-critical scenarios. 
However, generalization can degrade when new environments differ significantly from the training set. 
Moreover, pure SL does not guarantee \emph{optimality} or even feasibility in out-of-distribution cases. 
Hence, many authors propose \emph{hybrid} strategies (e.g., re-checking solutions with a classical solver or incorporating online collision checks) to ensure consistent results.

\subsection{Composite Learning Strategies}\label{subsec:composite}

\begin{figure}[htb!]
    \centering
    \includegraphics[width=0.6\linewidth]{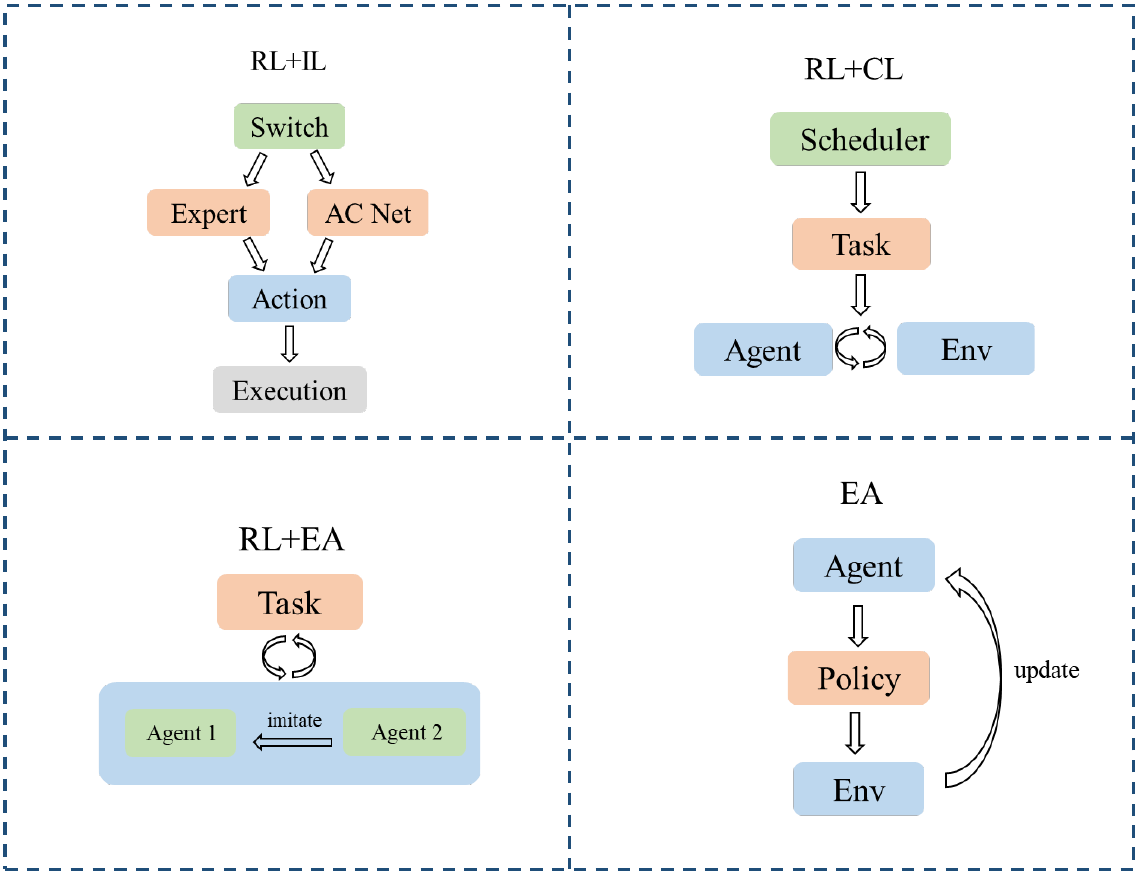}
    \caption{A set of four sub-flowcharts illustrating different hybrid approaches to solving the MAPF problem. The combinations include: (1) RL + Imitation Learning – training by switching between expert and RL policies to generate actions; (2) RL + Curriculum Learning – using a scheduler to gradually increase task difficulty; (3) RL + Evolutionary Algorithms – enabling agents to imitate better-performing peers; and (4) Pure Evolutionary Algorithm – agents execute diverse policies, replicate successful ones, and update or discard underperforming strategies.}
    \label{fig:composition}
\end{figure}

Composite learning strategies integrate multiple learning paradigms---for example, combining imitation learning with RL or leveraging progressive curriculum tasks. 
Such hybrids often seize the advantages of each technique (e.g., fast policy convergence from imitation learning, robust exploration from RL) while mitigating their individual limitations (e.g., poor generalization of imitation learning alone).
A set of four sub-flowcharts illustrating different hybrid approaches to solving the MAPF
problem is shown in Figure~\ref{fig:composition}.

\begin{table}[!htbp]
    \centering
    \caption{Composite Learning Strategies}
    \label{tab:my_label}
    \resizebox{.6\linewidth}{!}{%
    \begin{tabular}{ccccc}
    \toprule
    Models & RL & IL & CL & EA \\
    \midrule 
    PRIMAL~\citep{sartoretti2019primal} & $\checkmark$ & $\checkmark$ & \ding{55} & \ding{55} \\
    ISAC-IL~\citep{liu2024multi} & $\checkmark$ & $\checkmark$ & \ding{55} & \ding{55} \\
    C3PIL~\citep{xie2024crowd} & $\checkmark$ & $\checkmark$ & \ding{55} & \ding{55}  \\
    CACTUS~\citep{phan2024confidence} & $\checkmark$ & \ding{55} & $\checkmark$ & \ding{55} \\
    CPL~\citep{zhao2023curriculum} & $\checkmark$ & \ding{55} & $\checkmark$ & \ding{55} \\
    ~\citep{fan2022deep} & $\checkmark$ & \ding{55} & $\checkmark$ & \ding{55} \\
    HGLP~\citep{gao2023hglp} & $\checkmark$ & \ding{55} & $\checkmark$ & \ding{55} \\
    MAPPER~\citep{liu2020mapper} & $\checkmark$ & \ding{55} & $\checkmark$ & $\checkmark$ \\
    \citep{paul2022multi} & \ding{55} & \ding{55} & \ding{55} & $\checkmark$ \\
    \end{tabular}%
    }
\end{table}

\subsubsection{Imitation Learning (IL) + Reinforcement Learning (RL)}
Imitation learning in MAPF leverages expert trajectories to \emph{bootstrap} agent policies. 
Formally, agents begin by minimizing an IL loss such as \eqref{eq:il-loss}, then continue fine-tuning through RL to handle scenarios where the expert data are sparse or suboptimal:
\[
\theta_{\mathrm{combined}} 
\;=\;
\arg \min_\theta
\Bigl[
\sum_{k=1}^M \mathcal{L}\bigl(f_\theta(\mathbf{x}_k),\mathbf{y}_k\bigr) 
\;+\;
\lambda\,\mathcal{L}_{\mathrm{RL}}(\theta)
\Bigr].
\]
For example:
\begin{itemize}[leftmargin=*,noitemsep]
    \item \citep{sartoretti2019primal} combine RL and IL to train a fully decentralized policy (PRIMAL). 
    \item \citep{xie2024crowd} incorporate a Controlled Communication mechanism with IL to refine crowd-awareness among agents (C3PIL).
    \item \citep{liu2024multi} propose a multi-agent RL extension of Soft Actor-Critic (ISAC) and integrate IL to accelerate convergence (ISAC-IL).
\end{itemize}

\subsubsection{Curriculum Learning}
Curriculum Learning (CL) gradually increases the difficulty or complexity of MAPF tasks during training. 
Starting from simpler tasks (fewer agents, smaller maps) can help each agent’s policy converge to collision-free navigation strategies:
\[
\text{Train on }(\mathcal{V}_0,\mathcal{E}_0,\dots) \;\to\; 
(\mathcal{V}_1,\mathcal{E}_1,\dots) \;\to\; \ldots
\;\to\;
(\mathcal{V}_f,\mathcal{E}_f,\dots).
\]
\citep{phan2024confidence} define CACTUS, a confidence-based scheme that adaptively expands difficulty as the agent’s success rate increases. 
\citep{zhao2023curriculum} design a multi-phase approach, injecting more agents or higher obstacle density step by step. 
Empirical signals (e.g., success rate $>90\%$) trigger a progression to the next level.

\subsubsection{Evolutionary Methods}
Evolutionary Algorithms (EAs) treat agent paths or high-level policies as \emph{genomes} subject to evolutionary operators (selection, crossover, mutation). 
If $\mathbf{p}$ encodes a joint path set, the \emph{fitness function} might be:
\[
\mathrm{Fitness}(\mathbf{p})
\;=\;
-\Bigl(
\underbrace{\mathrm{SoC}(\mathbf{p})}_{\text{sum-of-costs}}
\;+\;
\alpha\,\underbrace{\mathrm{Collisions}(\mathbf{p})}_{\text{collision count}}
\Bigr),
\]
where collisions reflect the number of conflicting moves. 
In each generation, new path populations $\{\mathbf{p}^{(g+1)}\}$ evolve from $\{\mathbf{p}^{(g)}\}$:
\begin{enumerate}[leftmargin=*,noitemsep,label=(\alph*)]
    \item \emph{Selection:} retain top $K$ solutions by fitness. 
    \item \emph{Crossover:} combine partial paths from two parents. 
    \item \emph{Mutation:} randomly alter portions of the path with a collision check or local re-routing. 
\end{enumerate}
Recent examples:
\begin{itemize}[leftmargin=*,noitemsep]
    \item \citep{liu2020mapper} propose \emph{MAPPER}, combining distributed partially observable navigation and evolutionary RL for large-scale robotic tasks.
    \item \citep{paul2022multi} embed evolutionary game dynamics into multi-agent exploration, using replicator equations to update agent strategies.
\end{itemize}
Evolutionary search can handle complex and large state spaces, but runtime may grow with problem size unless carefully parallelized or hybridized with classical MAPF heuristics.

\subsection{Discussion}

The diverse learning paradigms studied in this section---MCTS, supervised methodologies, composite learning (RL+IL, curriculum), and evolutionary searches---demonstrate varied ways to fuse data-driven or iterative optimization ideas with the classical MAPF blueprint (Section~\ref{sec:formulation}). 
Table~\ref{tab:paradigm-advantages} summarizes core advantages and open challenges across these paradigms.

\begin{table}[ht!]
\centering
\caption{Key Advantages and Limitations of Non-RL Learning Paradigms in MAPF. 
While many methods yield benefits in adaptability or solution speed, each has distinct pitfalls in terms of scalability, out-of-distribution robustness, or theoretical guarantees.}
\label{tab:paradigm-advantages}
\begin{tabular}{p{2.9cm} p{4.8cm} p{4.5cm}}
\toprule
\textbf{Paradigm} & \textbf{Advantages} & \textbf{Limitations / Open Challenges}\\
\midrule
\textbf{MCTS} 
& 
\begin{itemize}[leftmargin=*,noitemsep]
\item Systematic search with built-in exploration
\item Flexible rollout policies, can incorporate domain heuristics
\end{itemize}
& 
\begin{itemize}[leftmargin=*,noitemsep]
\item Exponential branching if $n$ is large
\item Requires repeated rollouts at every decision step
\end{itemize}\\
\textbf{Supervised Learning} 
& 
\begin{itemize}[leftmargin=*,noitemsep]
\item Fast inference once trained
\item Straightforward adaptation of classical solver outputs
\end{itemize}
& 
\begin{itemize}[leftmargin=*,noitemsep]
\item No performance guarantees if test scenarios differ
\item Requires large labeled datasets, might not adapt online
\end{itemize}\\
\textbf{Composite (IL+RL, Curriculum)} 
& 
\begin{itemize}[leftmargin=*,noitemsep]
\item Synergy: IL speed + RL adaptivity
\item Progressive training can tackle large tasks
\end{itemize}
& 
\begin{itemize}[leftmargin=*,noitemsep]
\item Complex training pipeline
\item Balancing multi-objective losses can be delicate
\end{itemize}\\
\textbf{Evolutionary Methods}
& 
\begin{itemize}[leftmargin=*,noitemsep]
\item Population-based search explores diverse solutions
\item Parallel-friendly optimization
\end{itemize}
& 
\begin{itemize}[leftmargin=*,noitemsep]
\item Computation can become expensive
\item May converge slowly if mutation/crossover are not well tuned
\end{itemize}\\
\bottomrule
\end{tabular}
\end{table}

Looking forward, promising future directions include:
\begin{itemize}[leftmargin=*,noitemsep]
    \item \textbf{Hybridizing MCTS or EAs with classical solvers,} e.g., using MCTS expansions to focus on potentially conflicting sub-regions of a graph while a compilation-based solver handles the global plan.
    \item \textbf{Advanced supervision from partial solutions,} rather than full demonstration traces, possibly reducing dataset requirements.
    \item \textbf{Aggressive curriculum or meta-learning,} where tasks automatically adjust difficulty based on the agent performance, bridging offline training and real-time deployment.
    \item \textbf{Evolutionary-lifelong synergy,} combining repeated assignments (Section~\ref{subsec:lifelong}) with evolutionary population updates to continually adapt agent routes in changing environments.
\end{itemize}
These areas underscore the broader theme of \emph{integrating} learning modules into well-established MAPF frameworks (Sections~\ref{sec:search} and~\ref{sec:compilation}), aiming to achieve the best of both worlds: flexible adaptation and strong theoretical grounding.

\vspace{1em}
\noindent
In conclusion, the paradigms outlined here complement reinforcement learning methods (Section~\ref{sec:rl}) and existing classical MAPF pipelines in various ways. 
They further illustrate the growing richness of data-driven MAPF research, where advanced machine learning techniques---ranging from tree-based rollouts to supervised and evolutionary optimization---continue to push the boundaries of scalability and robustness in multi-agent path planning.

\subsection{Towards a Tighter Integration of Classical and Learning-Based Methods}

\subsubsection{Underutilized Classical Insights}

Recent years have witnessed a surge in learning-based methods for MAPF, particularly via deep multi-agent reinforcement learning or neural search heuristics. 
Yet these innovations often overlook the extensive body of work developed under the classical MAPF paradigm, which includes both search-based (e.g., Conflict-Based Search, priority-based schemes, large neighborhood search) and compilation-based (e.g., SAT, SMT, and MIP) frameworks. 
These classical methods embody decades of specialized insights into conflict resolution, combinatorial pruning, meta-agent merging, and domain-specific constraint propagation. 
While contemporary learning-based approaches sometimes adopt high-level ideas or partial code reuse from these algorithms, they rarely integrate the deeper theoretical properties that have proven indispensable for producing robust, efficient, and theoretically grounded solutions.

One critical shortfall is that classical MAPF solutions have systematically studied how to detect and handle collisions under a wide range of conditions, using sophisticated notions such as corridor or rectangle conflicts, conflict prioritization, and exhaustive ``include-exclude'' branching strategies. 
Admissible and consistent heuristics—like the pairwise dependency graph (DG) or weighted DG—substantially prune the search space by exploiting structural features unique to MAPF. 
They also enable meta-agent merges in CBS-based methods, ensuring that agents that repeatedly conflict are planned jointly, and thus reducing the risk of recurring collisions. 
However, many modern machine learning planners do not fully leverage these refined conflict-handling laws. 
They often treat collisions as uniform penalty signals in a reward function, effectively rediscovering collision dynamics ``from scratch.'' 
This leads to slow convergence during training and provides no guaranteed resolution once collisions become frequent, whereas a more direct infusion of classical conflict classification or bounding techniques could identify and prune problematic regions early, accelerating learning-based policy improvements.

Beyond collision handling, classical MAPF research has also developed systematic methods for branching and bounding solution costs. 
In conflict-based schemes, branching on cardinal conflicts first (i.e., collisions that increase the total solution cost if left unresolved) is known to boost convergence to an optimal or near-optimal solution. 
Similarly, priority-based approaches exploit partial orders on agents, reordering them dynamically when certain conflict patterns arise, so that repeated collisions do not stall planning. 
Moreover, advanced compilation-based solvers encode complex domain constraints into Boolean or linear formulations, enabling branch-and-cut to eliminate infeasible portions of the search space in a mathematically precise way. 
In principle, these bounding strategies could be embedded in a neural or reinforcement learning pipeline as ``hard'' constraints that forcibly prune unproductive states, or as differentiable approximations that guide a policy or Q-function away from infeasible actions. 
By failing to adopt such systematic bounding rules, learning-based approaches cede the algorithmic efficiency and theoretical guarantees that classical MAPF algorithms painstakingly achieve.

Classical MAPF also brings a variety of domain-specific formulations—for instance, multi-valued decision diagrams (MDDs) that restrict agent paths to only those that are cost-minimal if no collisions were present. 
These MDDs can be extended with advanced mutual-exclusion rules to prune large swaths of collision-prone paths. 
In principle, an MDD-based feasibility checker could be folded into a learning-based planner, allowing a deep policy to query whether a partial plan is viable—an operation known to dramatically reduce branching in classical MAPF. 
Equally significant is the concept of meta-agent merging, where a cluster of conflicting agents is treated as a single entity in the search, reflecting their interdependent joint moves. 
Current data-driven approaches rarely adopt such merging rules, typically treating collisions in an atomistic fashion. 
Incorporating meta-agent reasoning could enable a neural planner to reason directly about coordinated group moves, benefiting from the same conflict-abatement benefits that classical MAPF has long utilized.

A further benefit of classical MAPF is its nuance in handling unconventional constraints or real-world complexities. 
Over the years, researchers have integrated resource constraints, motion-primitive restrictions, agent heterogeneity, and multi-goal scheduling into standard MAPF formulations by carefully constructing additional constraints or objective functions. 
These domain-specific expansions are typically enforced with proven bounding mechanisms or exhaustive branching, which preserve completeness or suboptimality guarantees. 
In contrast, learning-based MAPF often tries to handle such variations by imposing a heuristic penalty or, at best, by customizing the reward function. 
While these approaches can work in principle, they are rarely accompanied by the deeper logical consistency checks that classical MAPF constraint reasoning would provide.

A related challenge is that many data-driven MAPF planners lack transparent performance bounds or completeness guarantees. 
Even suboptimal variants of classical MAPF, such as bounded suboptimal CBS or prioritized search with known approximation ratios, retain explicit guarantees on how their solutions deviate from optimum. 
Once these structures are ignored, the resulting learning-based method might converge to feasible paths in practice but offers no quantifiable measure of success under unforeseen circumstances or domain shifts. 
Although recent years have seen the emergence of ``safe RL'' paradigms that attempt to impose constraints during policy learning, the synergy between these paradigms and the proven bounding strategies from classical MAPF remains largely unexplored.

In light of these observations, there is vast potential for cross-pollination between learning-based MAPF and classic approaches. 
For instance, one could envision a ``meta-CBS'' search in which conflict branching relies on proven classical rules (identifying which conflict is cardinal, merging a group of interdependent agents), while the expansion or node-selection heuristics are guided by a learned model that predicts the most promising branches. 
Under such a framework, the rigor of collision detection and branching remains intact, but the overall search process gains speed and adaptability from the learned selection policy. 
Similarly, approximate versions of multi-valued decision diagrams might serve as a differentiable layer within a neural architecture, allowing the network to quickly rule out large classes of infeasible partial plans. 
Such hybrid strategies would incorporate the collision resolution strengths and theoretical scaffolding of classical MAPF while benefiting from the powerful function-approximation abilities of neural networks.

Ultimately, the valuable insights of classical MAPF research need not be relegated to footnotes or replaced wholesale. 
Instead, they can be adapted, approximated, or embedded into the emergent class of machine learning frameworks in a way that preserves both classical rigor and learning-based adaptability. 
This deeper fusion would not only accelerate training and improve solution quality but could also impart valuable theoretical performance bounds to systems that otherwise risk unpredictable behavior. 
Achieving this integration stands as an open challenge, but it promises a richer array of tools for future MAPF researchers seeking to move beyond the purely classical or purely data-driven paradigms.

\subsection{Underexplored Opportunities for Learning-Augmented Classical MAPF}

Researchers have increasingly recognized the potential of learning-based MAPF methods in complex or uncertain environments. 
Yet the converse direction—embedding learning within \textit{classical} MAPF solvers—remains comparatively neglected. 
Existing work on conflict-based search (CBS), priority-based search (PBS), or large neighborhood search (LNS) that incorporates machine learning has only begun to scratch the surface of what is theoretically and practically possible. 
Techniques such as learned conflict selection or agent prioritization demonstrate clear speedups and solution-quality gains, suggesting that nearly every module of a classical solver could benefit from a data-driven upgrade while retaining the solver’s transparent, systematic structure.

The most direct opportunities lie in how these solvers handle branching, pruning, and constraint propagation. 
For instance, search-based planners typically rely on heuristic expansions or handcrafted conflict-splitting rules that may be suboptimal as problem scale grows. 
In CBS, conflicts encountered at the high level must be systematically resolved by branching, yet deciding \textit{which} conflict to branch on, or \textit{how} to split constraints among child nodes, is often determined by static policies (e.g., ``choose the earliest conflict'' or ``choose the conflict with largest cost impact''). 
A learning-augmented approach could, in principle, tailor these branching decisions to the structure of actual instances. 
A trained classifier or regressor, leveraging historical search traces, might predict which specific conflict will yield the most branching overhead if left unresolved, thereby prioritizing it at the high level and reducing the overall size of the conflict tree. 
Similar logic applies to suboptimal variants of CBS, where a data-driven focal search could refine or relax bounding factors more adaptively.

Even in priority-based solvers, a machine learning model can drive dynamic reordering or grouping of agents. 
Classical PBS sets a static priority ordering for agents to follow, with collisions prompting merges of ``meta-agents.'' 
While some work has shown that learned priority assignments reduce collisions, many deeper avenues remain open. 
For example, an online learned policy could detect that certain agents—due to their goals, speeds, or frequently blocked paths—are ``high risk'' for repeated collisions, and thus reorder them to the top of the priority queue. 
Another policy might guide whether two conflicting agents should be merged into a single meta-agent at all, or if a simpler local detour would suffice. By situating these data-driven heuristics within PBS’s proven collision-avoidance framework, one can mitigate capacity or scheduling bottlenecks in real-time applications without discarding the solver’s theoretical underpinnings.

In the realm of compilation-based MAPF, learning has a potentially even greater role. 
Formulations like SAT, SMT, CSP, ASP, or MIP frequently suffer from large search trees in branching, cutting, or conflict refinement. 
Although these solvers already incorporate sophisticated general-purpose heuristics (e.g., branching rules for SAT or MIP, conflict-driven clause learning, or cutting-plane selection), they remain blind to the domain-specific structure of MAPF problems. 
A synergy with machine learning might be realized by training a specialized ``neural solver'' or ``learning-based branch-and-cut'' mechanism that recognizes which integer or Boolean variables are most critical for collision resolution, which potential cuts (e.g., collision cuts in MIP or lazy constraints in LaCAM) are likely to prune large infeasible regions, and which partial solutions are promising enough to keep exploring. 
One possible direction is to record frequent patterns of collisions or resource conflicts across MAPF instances, then train a model to inject relevant ``cuts'' or constraints earlier in the solver’s process—thereby dramatically shortening the convergence time. 
Lazy-constraint addition frameworks like LaCAM could also benefit from a learning module that predicts which constraints will actually matter, skipping many redundant checks and leaving more solver bandwidth for critical expansions.

Beyond these local improvements, a symbiosis between machine learning and classical MAPF could address high-dimensional or multi-objective variants more effectively than either approach alone. 
Industrial-scale MAPF problems often involve rich constraints (e.g., limited fuel, time windows, or capacity constraints for each agent), uncertain data streams (e.g., partial sensor data in real time), or hierarchical objectives (e.g., first minimize collisions, then total cost, then maximum lateness). 
Although classical planners can systematically incorporate these constraints once formalized, they often struggle in rapidly changing environments or when large portions of the input are noisy or approximate. 
Here, a learned module could filter or ``translate'' unstructured real-time observations—such as imperfect sensor measurements—into compact, high-level constraints that the solver can handle efficiently. 
In return, the solver’s partial solutions, conflict clauses, or explanatory branching structures may serve as ``labels'' or ``demonstrations'' to a learning algorithm, ensuring that any learned policy adheres to collision-free or resource-feasible solutions.

Despite this broad potential, current attempts at learning-augmented classical MAPF generally limit themselves to modest instance scales or rely on ad hoc ways of inserting neural components (e.g., a single learned conflict-picker for CBS). 
Much of the solver’s internal logic—such as advanced symmetry reasoning, disjoint splitting, or hierarchical branching—remains unaffected by learning, leaving vast spaces of synergy unexplored. 
In particular, there has been little effort to systematically categorize \textit{where} and \textit{how} data-driven modules might intervene. 
Such a taxonomy could map out all major stages in classical MAPF pipelines, from initial pre-processing and path generation to final conflict resolution, indicating which machine learning strategies (supervised, reinforcement, offline imitation, online adaptation) are most compatible at each stage. 
It would also help clarify the potential performance trade-offs, such as how real-time inference overhead from a neural network might compare with the time saved from fewer search-node expansions.

Ultimately, unlocking the full power of learning-augmented classical MAPF will require a tighter connection between the solver’s inherent strengths—soundness, exhaustive branching, and interpretability—and the adaptability, pattern recognition, and data-driven generalization offered by machine learning. 
Achieving this synergy calls for both deeper theoretical understanding (e.g., how best to guarantee safety or completeness when deferring key decisions to a trained module) and extensive empirical evaluation (e.g., how to ensure robust performance on varied maps or dynamic tasks). 
By moving beyond incremental or ``add-on'' usage of neural networks and embracing a more integrated, principled approach, the research community can create versatile solvers that are not only guided by data but also anchored by decades of MAPF theory, ensuring both increased efficiency within known constraints and flexible adaptation to complex real-world conditions.

\section{Experimental Settings and Comparisons}\label{sec:exp}

As introduced in Section~\ref{sec:intro}, MAPF involves devising collision-free trajectories for multiple agents within a shared environment. 
The mathematical formulation outlined in Section~\ref{sec:formulation} emphasizes representations of the environment as graphs or continuous domains, the objectives to minimize (e.g., makespan, sum-of-costs, or task completion time), and the control modes (centralized vs.\ decentralized). 
Building on those foundations, this section examines how MAPF researchers design and evaluate their experiments in practice. 
The discussion is organized into three main parts: 
(1)~\textbf{Experimental environments}, which encapsulate map types and complexity; 
(2)~\textbf{Evaluation metrics}, which capture success, safety, solution efficiency, and computation costs; 
(3)~\textbf{Scaling factors}, including the number of agents and map sizes, as well as typical baseline selections. 
Throughout, mathematical abstractions are provided to solidify the connections between problem formulations (Section~\ref{sec:formulation}) and real-world (or simulated) experiments.

\subsection{Types of Experimental Environments}
\label{subsec:env_types}

In MAPF experiments, the \emph{environment} is often modeled as $\mathcal{G}=(\mathcal{V}, \mathcal{E})$ for discrete (grid) spaces or $\Omega \subseteq \mathbb{R}^d$ for continuous domains. 
However, the structure of $\mathcal{V}$ (or $\Omega$) can vary significantly among studies. 
Table~\ref{tab:env_comparison} summarizes the most prevalent types of experimental environments. 
Where applicable, we provide an equivalent mathematical description.
Classical methods and learning-based methods essentially use the same simulation environments, therefore this section does not make a distinction between them.

\begin{table}[htbp]
\centering
\caption{Overview of MAPF Experimental Environments}
\label{tab:mapf-env}
\begin{tabular}{@{}llcc@{}}
\toprule
\textbf{Type} & \textbf{Map} & \textbf{Size} & \textbf{Preview (map in bold)}\\ \midrule
\multirow{3}{*}{City} 
    & Berlin\_1\_256      & 256 $\times$ 256 & \multirow{3}{*}{\includegraphics[width=0.09\linewidth]{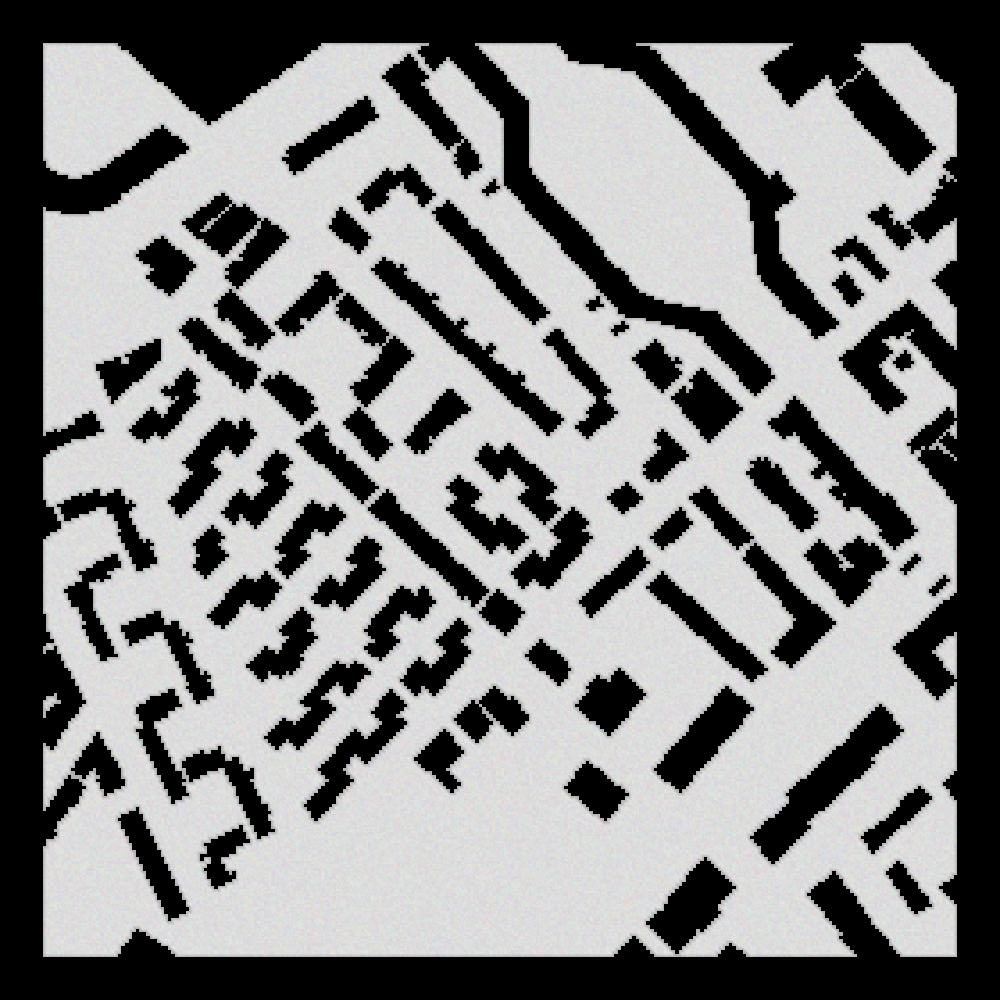}} \\
    & \textbf{Boston\_0\_256}      & 256 $\times$ 256 & \\
    & Paris\_1\_256       & 256 $\times$ 256 & \\ \midrule
\multirow{6}{*}{Dragon Age Origins} 
    & brc202d            & 481 $\times$ 530 & \multirow{6}{*}{\includegraphics[width=0.09\linewidth]{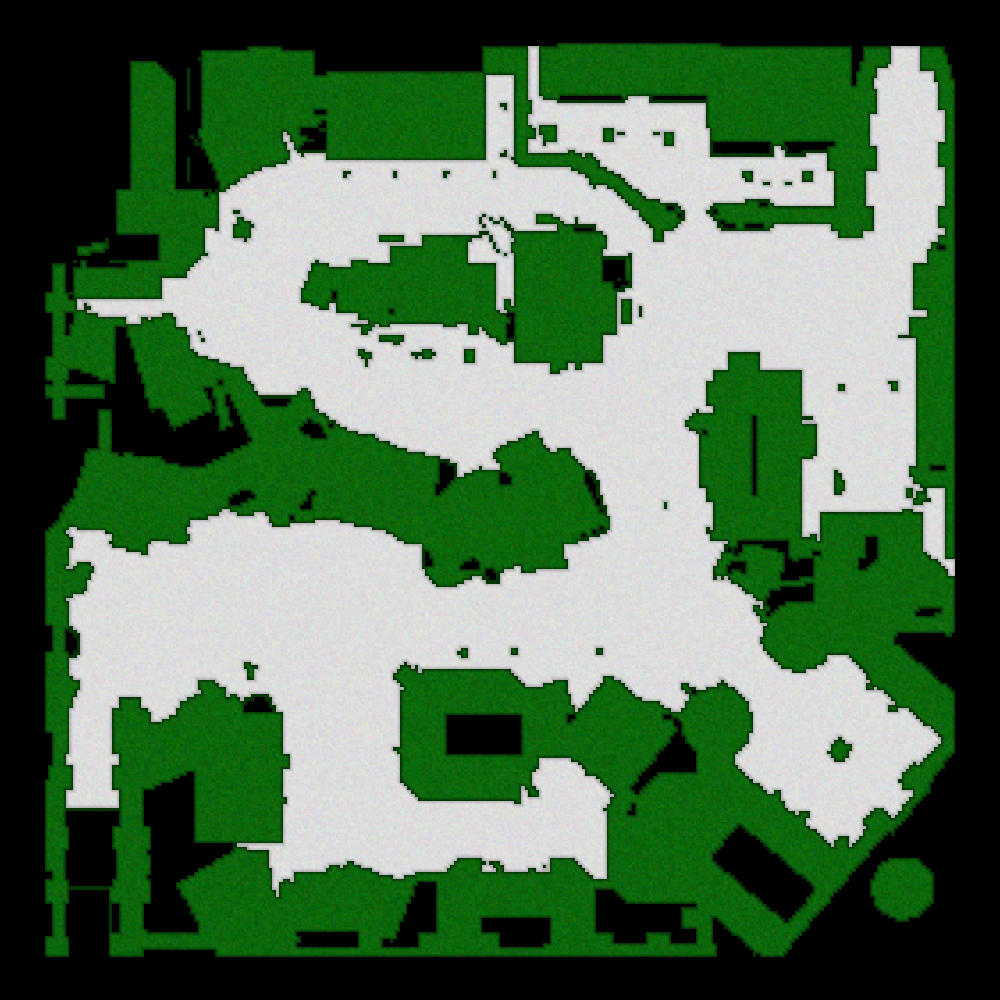}}\\
    & den312d            & 81 $\times$ 65   & \\
    & \textbf{den520d}            & 256 $\times$ 257 & \\
    & lak303d            & 194 $\times$ 194 & \\
    & orz900d            & 656 $\times$ 1491 & \\
    & ost003d            & 194 $\times$ 194 & \\ \midrule
\multirow{4}{*}{\shortstack{Dragon Age 2}} 
    & \textbf{ht\_chantry}         & 141 $\times$ 162 & \multirow{4}{*}{\includegraphics[width=0.09\linewidth]{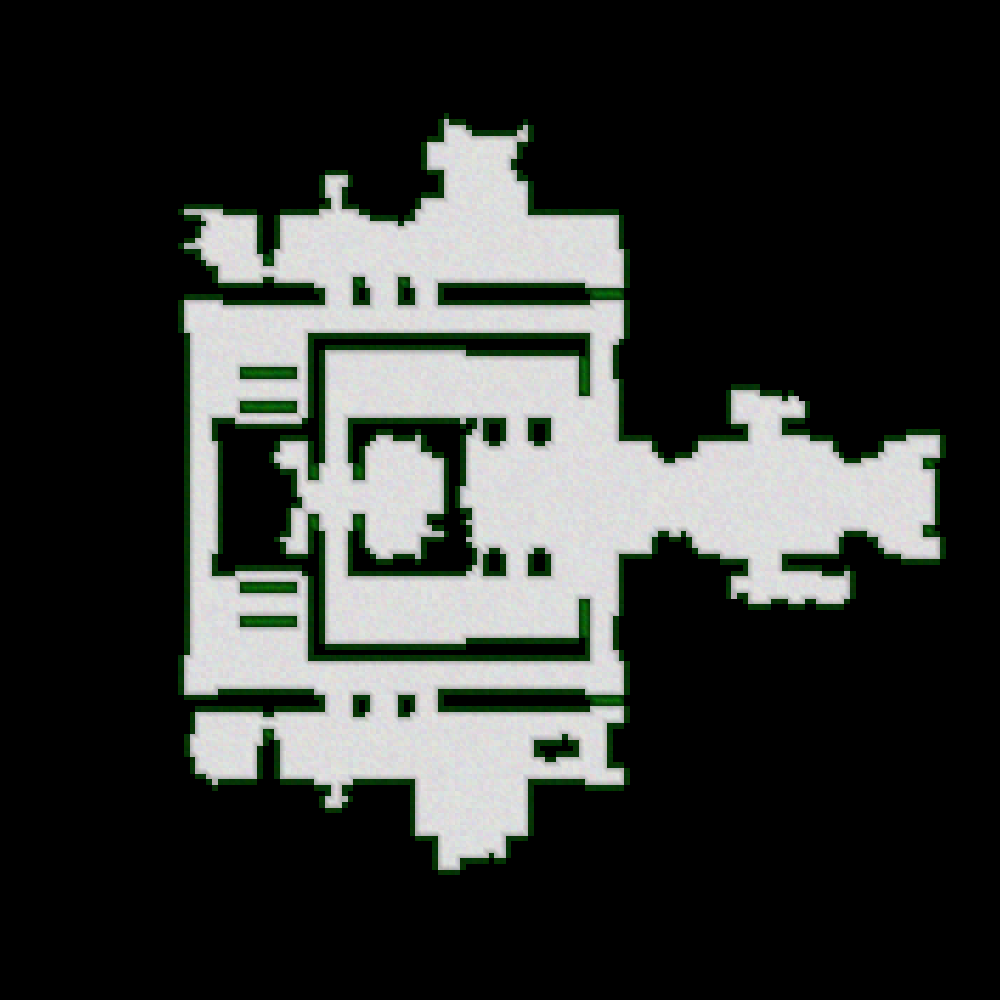}}\\
    & ht\_mansion\_n       & 270 $\times$ 133 & \\
    & w\_woundedcoast     & 578 $\times$ 642 & \\
    & lt\_gallowstemplar\_n & 180 $\times$ 251 & \\ \midrule
\multirow{4}{*}{Open} 
    & empty-8-8          & 8 $\times$ 8   & \multirow{4}{*}{\includegraphics[width=0.09\linewidth]{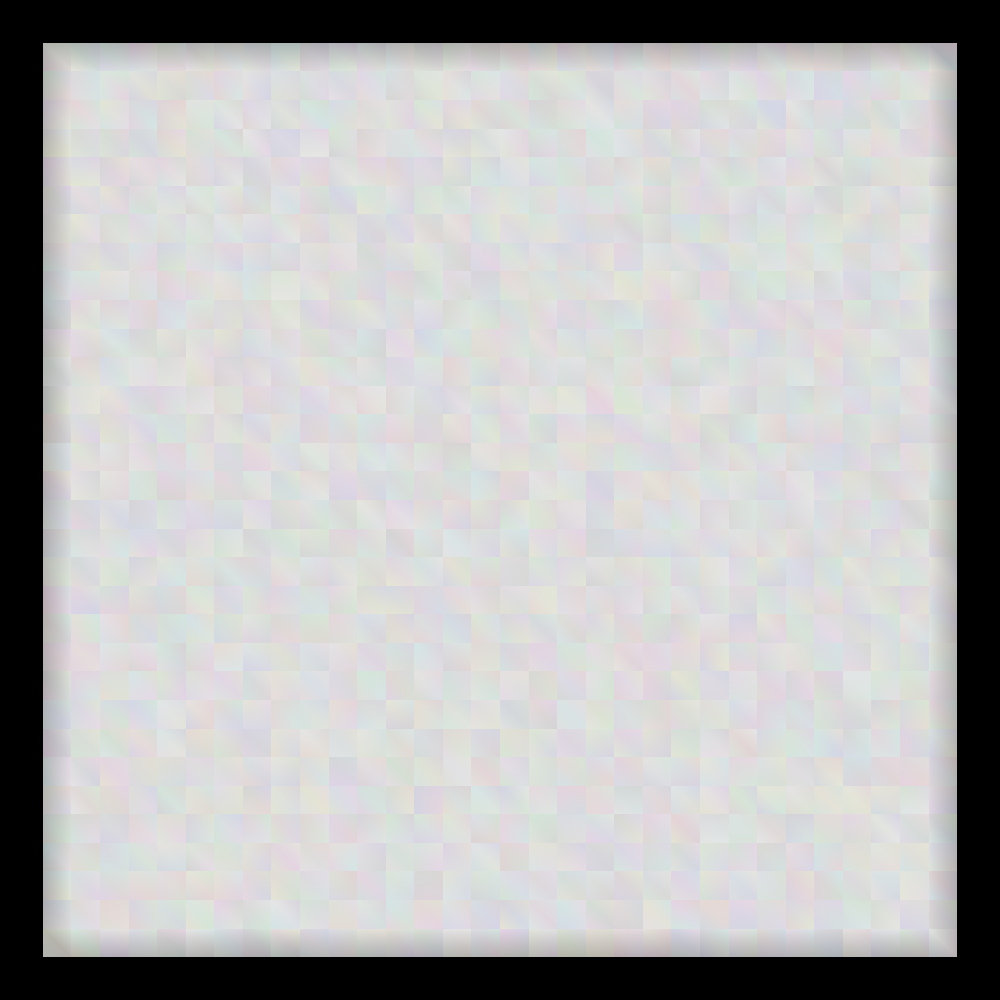}}\\
    & empty-16-16        & 16 $\times$ 16 & \\
    & \textbf{empty-32-32}        & 32 $\times$ 32 & \\
    & empty-48-48        & 48 $\times$ 48 & \\ \midrule
\multirow{4}{*}{\shortstack{Random}} 
    & random-32-32-10     & 32 $\times$ 32 & \multirow{4}{*}{\includegraphics[width=0.09\linewidth]{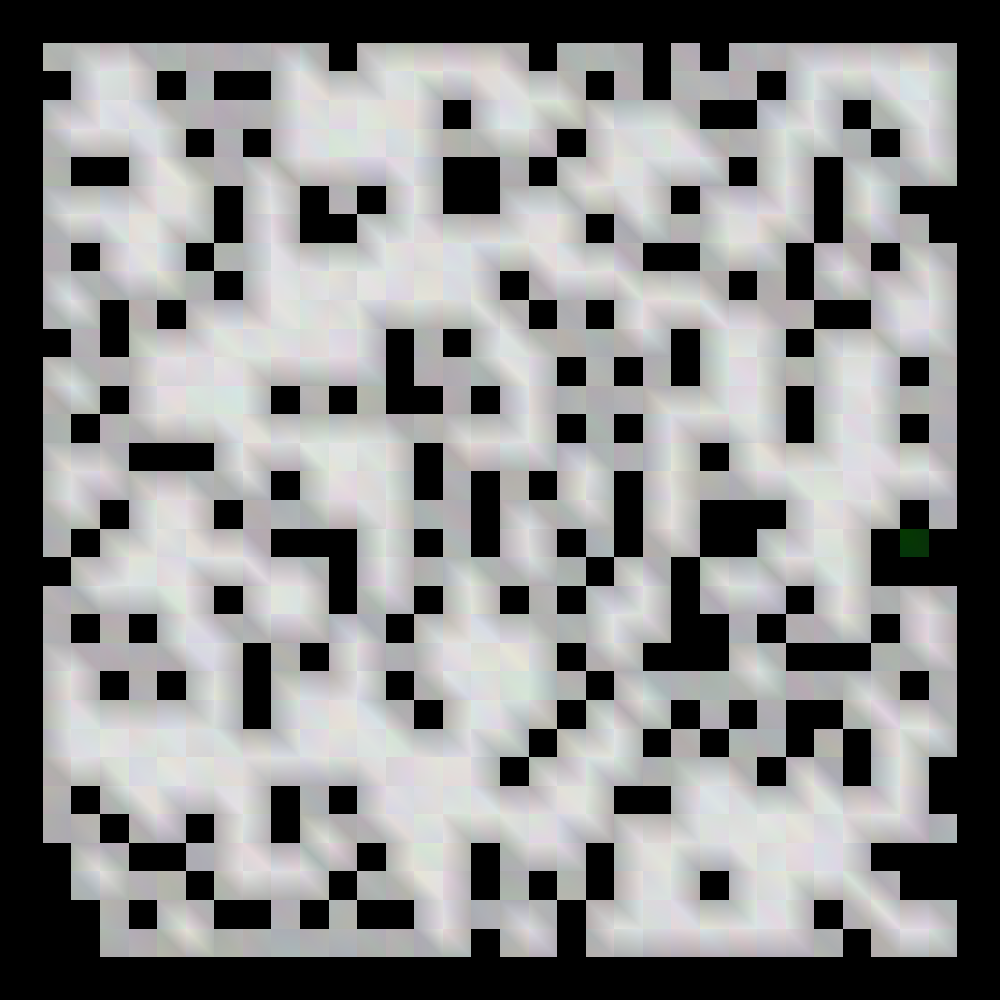}}\\
    & \textbf{random-32-32-20}     & 32 $\times$ 32 & \\
    & random-64-64-10     & 64 $\times$ 64 & \\
    & random-64-64-20     & 64 $\times$ 64 & \\ \midrule
\multirow{5}{*}{Maze} 
    & maze-32-32-2       & 32 $\times$ 32 & \multirow{5}{*}{\includegraphics[width=0.09\linewidth]{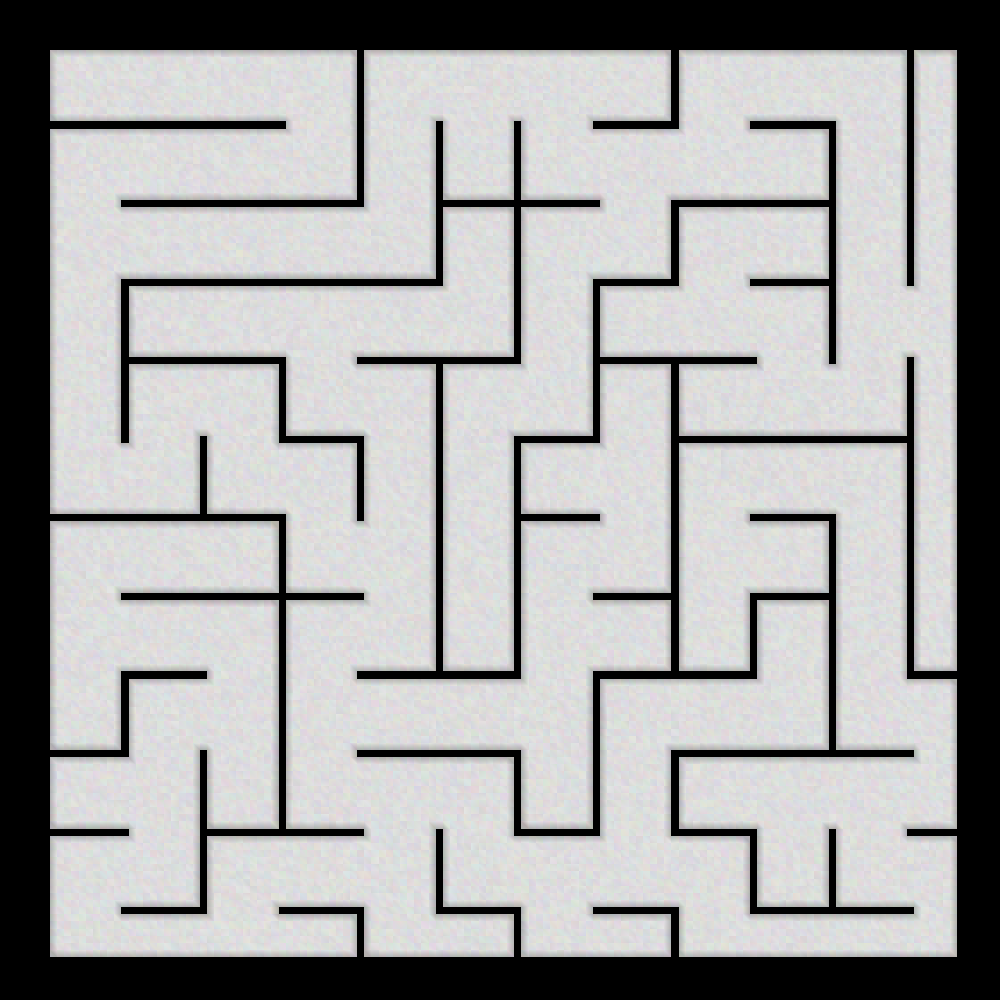}}\\
    & maze-32-32-4       & 32 $\times$ 32 & \\
    & maze-128-128-1     & 128 $\times$ 128 & \\ 
    & maze-128-128-2     & 128 $\times$ 128 & \\ 
    & \textbf{maze-128-128-10}    & 128 $\times$ 128 & \\ \midrule
\multirow{4}{*}{Room} 
    & room-32-32-4       & 32 $\times$ 32 & \multirow{4}{*}{\includegraphics[width=0.09\linewidth]{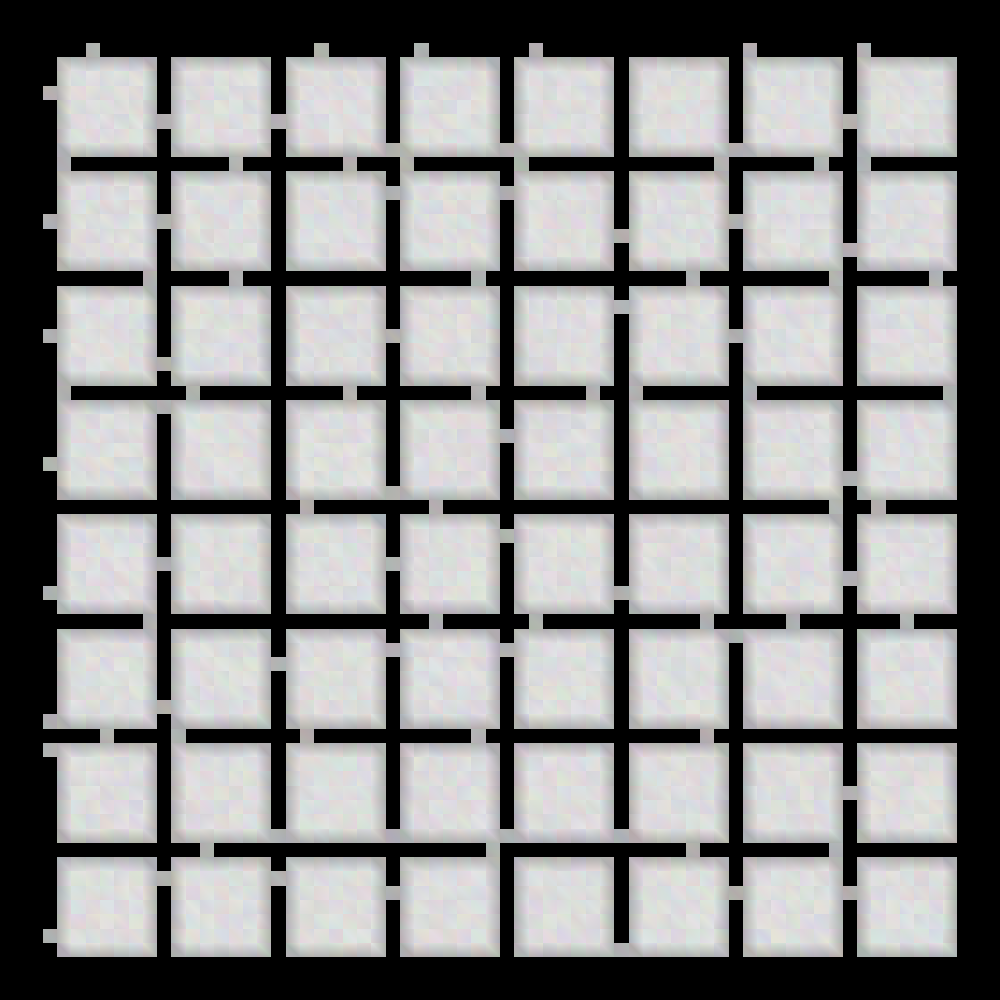}}\\
    & room-32-32-8       & 32 $\times$ 32 & \\
    & \textbf{room-64-64-8}       & 64 $\times$ 64 & \\
    & room-64-64-16      & 64 $\times$ 64 & \\ \midrule
\multirow{4}{*}{Warehouse} 
    & warehouse-10-20-10-2-1       & 161 $\times$ 63 & \multirow{4}{*}{\includegraphics[width=0.09\linewidth]{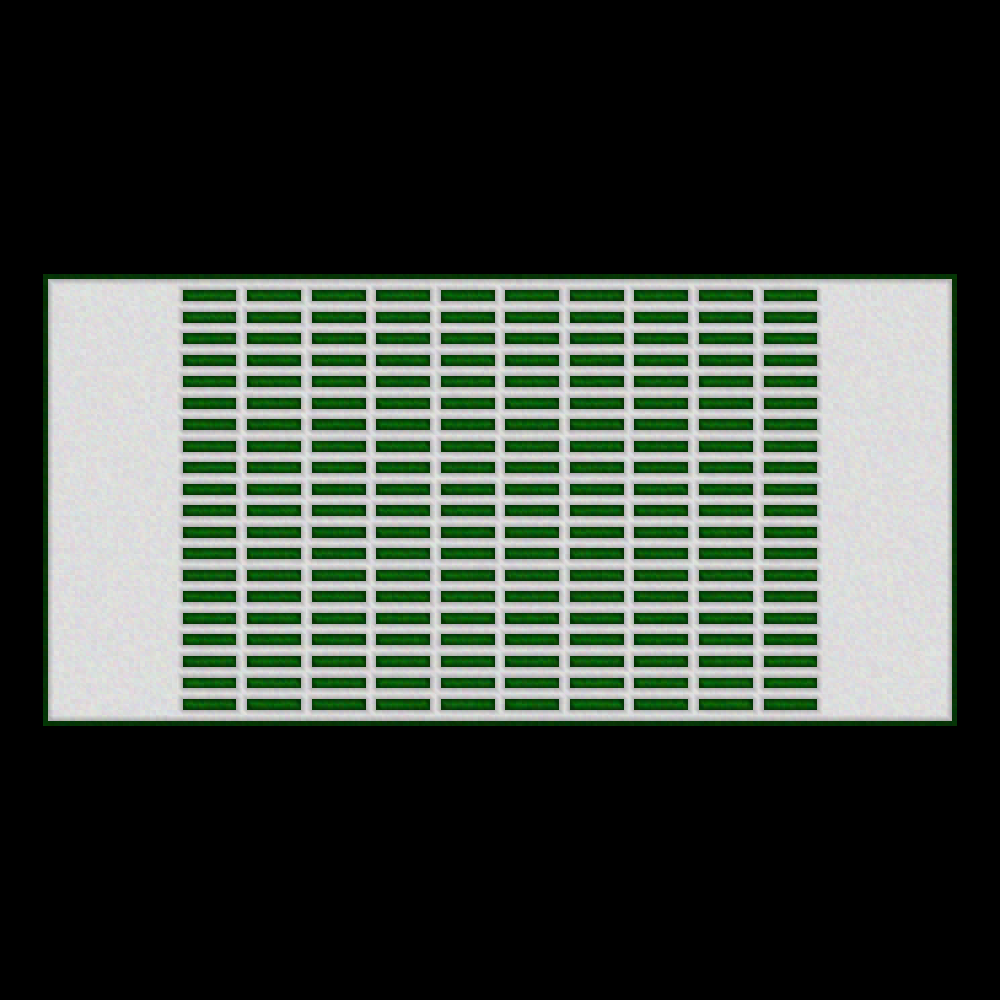}}\\
    & \textbf{warehouse-10-20-10-2-2}       & 170 $\times$ 84 & \\
    & warehouse-20-40-10-2-1       & 321 $\times$ 123 & \\
    & warehouse-20-40-10-2-2       & 340 $\times$ 164 & \\ \bottomrule
\end{tabular}%
\end{table}

Table~\ref{tab:mapf-env} catalogs the most prevalent MAPF benchmark maps, organized by environment type. 
For each environment category, i.e., \textit{City, Game (Dragon Age Origins, Dragon Age 2), Open, Random, Maze, Room}, and \textit{Warehouse}, specific map instances are listed along with their dimensions. 
The preview column provides a visual representation of a representative map from each category, offering readers a quick visual reference of the environment's structural characteristics.

These benchmark environments vary significantly in their structural complexity, ranging from empty grids (Open) to intricate city layouts and game maps. 
This diversity allows researchers to evaluate algorithm performance across different scenarios with varying levels of difficulty. 
For instance, \textit{Warehouse} simulates logistics settings with regular patterns of shelves and corridors, while \textit{Maze} presents navigation challenges with narrow passages and potential deadlocks.

\begin{sidewaystable}
\centering
\caption{Overview of Common MAPF Experimental Environments and Their Mathematical Representations}
\label{tab:env_comparison}
\renewcommand{\arraystretch}{1.2}
\resizebox{\linewidth}{!}{%
\begin{tabular}{p{0.1\linewidth} p{0.4\linewidth} p{0.5\linewidth}}
\toprule
\textbf{Environment} & 
\textbf{Mathematical Representation} &
\textbf{Key Characteristics and Challenges} \\
\midrule
\textbf{Random} 
& 
\begin{minipage}[t]{\linewidth}
\centering
$\mathcal{G}_{\text{rand}}(p)$
\end{minipage}
\vspace{0.7ex}

A random graph or grid where each vertex or edge is blocked with probability $p$.
& 
Generated via random sampling of obstacles (e.g., each grid cell is blocked w.p.\ $p$).  
Prevents overfitting to specific map structures and tests generalization.  \\
\textbf{Maze} 
& 
\begin{minipage}[t]{\linewidth}
\centering
$\mathcal{G}_{\text{maze}} = \bigl(\mathcal{V},\mathcal{E}\bigr)$ 
\newline
$\quad\,\,$ with narrow corridors or high obstacle density
\end{minipage}
& 
Walls form narrow corridors, potentially creating deadlocks. Used to assess navigation under heavy congestion or bottlenecks.  \\
\textbf{Warehouse} 
& 
\begin{minipage}[t]{\linewidth}
\centering
$\mathcal{G}_{\text{warehouse}}$ with aisles $A$ and shelves $S$
\end{minipage}
\newline
Vertices represent floor cells; edges reflect feasible movement. Shelves $S$ can be permanent obstacles or dynamic.
& 
Resembles real logistics layouts (e.g., Kiva systems). Tests algorithms for close-proximity navigation, typical for robotic automation in warehouses. \\
\textbf{City} 
& 
\begin{minipage}[t]{\linewidth}
\centering
$\mathcal{G}_{\text{city}}=(\mathcal{V},\mathcal{E})$
\end{minipage}
\newline
$\mathcal{V}$: intersections, $\mathcal{E}$: roads.
Potential constraints on flow directions or turning.
& 
Simulates urban traffic networks with intersections, roadblocks, or partial closures. Relevant for autonomous driving tasks, traffic scheduling, or robotic delivery.\\
\textbf{Game} 
& 
\begin{minipage}[t]{\linewidth}
\centering
$\mathcal{G}_{\text{game}}$ 
\end{minipage}
\newline
Extracted from game maps (e.g., \emph{MovingAI} or custom 2D/3D layouts).
& 
Provides diverse topologies (open fields, narrow corridors). Often used to benchmark pathfinding complexities encountered in virtual environments. \\
\textbf{Open} 
& 
\begin{minipage}[t]{\linewidth}
\centering
$\mathcal{G}_{\text{blank}}$ 
\end{minipage}
\newline
An empty grid or domain: $\mathcal{V}$ has no internal obstacles.
& 
Simplest environment for baseline testing of movement and interaction without obstacle complications. Useful as a fundamental performance benchmark. \\
\textbf{Dynamic} 
& 
\begin{minipage}[t]{\linewidth}
\centering
$\mathcal{G}_{t}=(\mathcal{V}_{t},\mathcal{E}_{t})$ 
\end{minipage}
\newline
Time-varying obstacle set. 
A subset $\mathcal{O}(t) \subset \mathcal{V}$ denotes occupied or blocked cells at time $t$.
& 
Obstacles or environment conditions evolve over time. Reflects real-world scenarios (e.g., moving vehicles, temporarily blocked roads). Tests adaptive, online re-planning capabilities.\\
\bottomrule
\end{tabular}%
}
\end{sidewaystable}

Table~\ref{tab:env_comparison} complements Table~\ref{tab:mapf-env} by providing a mathematical formalization of each environment type along with its key characteristics and associated challenges. 
The mathematical representations frame these environments in terms of graph theory (for discrete spaces) or geometric spaces (for continuous domains), establishing a rigorous foundation for algorithm development and analysis. 
For each environment type, the table highlights the specific navigation challenges they pose, such as congestion in narrow corridors for maze environments or dynamic obstacle avoidance in time-varying settings.

\paragraph{Remarks.} 
\emph{Random} environments are commonly used to train and test learning-based algorithms, gradually increasing size or obstacle density to evaluate scalability.  
\emph{Mazes} focus on congestion phenomena and agent coordination in highly constrained passages.  
\emph{Warehouse} and \emph{City} maps approximate real-world logistics or urban layouts, making them a standard for industrial MAPF solutions.  
\emph{Blank} maps help isolate an algorithm’s inherent path-planning efficiency without the confounding presence of obstacles.  
\emph{Dynamic} environments refine MAPF to dynamic MAPF (DMAPF) scenarios, highlighting the necessity of online or continual re-planning.

\subsection{Evaluation Metrics}
\label{subsec:eval_metrics}


Evaluating MAPF algorithms requires a comprehensive set of metrics that capture different aspects of performance, efficiency, and solution quality. 
In this section, we categorize these metrics into distinct groups and provide detailed definitions for each. 
When discussing these metrics, we consider a scenario where $n$ agents operate in an environment $\mathcal{G}$ over a time horizon $T$, with $a_i(t)$ representing the action of agent $i$ at time $t$, and $s_i(t)\in\mathcal{V}$ (or $\Omega$) denoting agent $i$'s position at time $t$.

We organize our discussion of metrics into two main categories: 
those commonly used for evaluating classic MAPF methods (Table~\ref{tab:metrics_summary_classic}) and those frequently applied to learning-based approaches (Table~\ref{tab:metrics_summary}). 
While some metrics appear in both categories, their importance and implementation often differ based on the algorithmic paradigm.

\subsubsection{Metrics for Classic Methods}

Classic MAPF methods typically prioritize completeness and optimality guarantees, with evaluation focused on solution quality, computational efficiency, and algorithmic properties. 
Table~\ref{tab:metrics_summary_classic} presents metrics commonly used for evaluating these approaches.

\begin{table}[htb!]
\centering
\caption{Classic MAPF Evaluation Metrics with Mathematical Definitions}
\label{tab:metrics_summary_classic}
\renewcommand{\arraystretch}{1.1}
\resizebox{\textwidth}{!}{%
\begin{tabular}{llp{0.65\linewidth}}
\toprule
\textbf{Category} & \textbf{Metric} & \textbf{Mathematical Definition} \\
\midrule

\multirow{3}{*}{\textbf{Success \& Failure}} 
& Success Rate (SR) & $\text{SR} = \frac{1}{n}\sum_{i=1}^n \mathbb{I}([s_i(T)=g_i])$, where $\mathbb{I}(\cdot)\in\{0,1\}$ indicates whether agent $i$ reached its goal $g_i$ by the final time $T$. \\

& Number of Instantiated Agents (NIA) & $\text{NIA} = \sum_{i=1}^n \mathbb{I}([s_i(T_d)=g_i])$, where $\mathbb{I}(\cdot)\in\{0,1\}$ indicates whether agent $i$ reached its goal $g_i$ by the deadline $T_d$. \\

& Number of On-Time Agents (NOTA) & $\text{NOTA} = \sum_{j=1}^m \mathbb{I}([t_j \leq d_j])$, where $\mathbb{I}(\cdot)\in\{0,1\}$ indicates whether task $j$ was completed at time $t_j$ before its deadline $d_j$. \\
\midrule

\multirow{7}{*}{\textbf{Solution Efficiency}} 
& Sum-of-Cost (SoC) & $\text{SoC} = \sum_{i=1}^{n}\{\mathrm{Cost}(\pi_i)\}$, where $\mathrm{Cost}(\pi_i)$ is the time until agent $i$ reaches $g_i$. \\

& Sum-of-Delay (SoD) & $\text{SoD} = \sum_{i=1}^{n}\{\mathrm{Cost}(\pi_i) - l_i\}$, where $l_i$ is the shortest path length for agent $i$. \\

& Makespan (MKSP) & $\text{MKSP} = \max_{1\le i\le n}\{\mathrm{Cost}(\pi_i)\}$ \\

& Path Cost & The average path costs per agent. \\

& Flowtime (FT) & $\text{FT} = \sum_{i=1}^n \mathrm{Cost}(\pi_i)$ \\

& Throughput (TP) & $\text{TP} = \frac{1}{T} \sum_{t=1}^{T} \sum_{i=1}^{n} \mathbb{I}([s_i(t)=g_i \wedge s_i(t-1) \neq g_i])$, average number of goals reached per timestep. \\

& Mean Execution Timesteps (MET) & $\text{MET} = \frac{1}{n} \sum_{i=1}^{n} \mathrm{Cost}(\pi_i^{exec})$, where $\mathrm{Cost}(\pi_i^{exec})$ is the actual timesteps to goal during execution. \\
\midrule

\textbf{Computation Time} 
& Runtime (RT) & Total time required by the algorithm to output a solution. \\
& Mutex Propagation Time Ratio & The average runtime ratios of mutex reasoning. \\
\midrule

\textbf{Expansions} 
& CT Node Expansions & The number of expanded CT nodes. \\
\midrule

\multirow{3}{*}{\textbf{Solution Quality}} 
& Optimality Rate & Percentage of instances where algorithm produces optimal makespans. \\
& Suboptimality Rate & $100 \times \frac{\text{found makespan} - \text{optimal makespan}}{\text{optimal makespan}}$ averaged over all solved instances. \\
& Social Welfare (SW) & $\text{SW} = \sum_{i=1}^{n} w_i(\pi_i) = \sum_{i=1}^{n}\{v_i - c_i(\pi_i)\}$, where $v_i$ is agent $i$'s value for reaching goal and $c_i(\pi_i)$ is travel cost. \\
\midrule

\multirow{2}{*}{\textbf{Resource Utilization}} 
& Station Utilization (SU) & $\text{SU} = \sum_{s=1}^{S} \sum_{t=1}^{T} \mathbb{I}([\text{station}_s \text{ is occupied at time } t])$, number of timesteps stations are occupied. \\
& Average Memory Usage (AMU) & $\text{AMU} = \frac{1}{|\mathcal{I}|} \sum_{I \in \mathcal{I}} M(I)$, where $M(I)$ is peak memory when solving instance $I$. \\
\midrule

\textbf{Conflict Analysis} 
& Conflict Distribution (CD) & $\text{CD}_\text{type} = \frac{|\mathcal{C}_\text{type}|}{|\mathcal{C}_\text{total}|}$, proportion of each conflict type among all conflicts. \\
\midrule

\textbf{Lifelong MAPF} 
& Replans & $\text{Replans} = \sum_{t=1}^{T} \mathbb{I}([\text{replan occurred at time } t])$, number of times replanning was triggered. \\

\bottomrule
\end{tabular}%
}
\end{table}

The success and failure metrics for classic methods include Success Rate (SR), which measures the proportion of agents that reach their goals, and more specialized metrics like Number of Instantiated Agents (NIA) and Number of On-Time Agents (NOTA) that account for deadline constraints. 
These metrics are particularly important for applications with strict timing requirements, such as warehouse logistics and airport operations.

Solution efficiency metrics for classic methods are often tied to optimization objectives. 
Sum-of-Cost (SoC) and Sum-of-Delay (SoD) capture the cumulative time agents spend reaching their goals, with SoD specifically measuring the excess time beyond optimal paths. 
Makespan (MKSP) evaluates the time until the last agent reaches its goal, critical for scenarios where completion time is the primary concern. 
Additional metrics like Flowtime (FT) and Throughput (TP) measure different aspects of solution efficiency.

Computational metrics for classic methods include Runtime (RT) and memory usage, which are crucial for assessing algorithmic scalability. 
Specialized metrics like CT Node Expansions provide insight into the search process of Conflict-Based Search algorithms, helping researchers understand algorithmic behavior.
Solution quality metrics such as Optimality Rate and Suboptimality Rate directly address how close solutions are to theoretical optimum, reflecting the emphasis classic methods place on optimality guarantees. Social Welfare (SW) introduces economic considerations by balancing goal values against travel costs.

The table also includes specialized metrics for resource utilization, conflict analysis, and lifelong MAPF scenarios, capturing the breadth of evaluation approaches used with classic methods.

\subsubsection{Metrics for Learning-based Methods}

Learning-based MAPF methods typically prioritize scalability, adaptability, and real-time performance, with metrics reflecting these priorities as shown in Table~\ref{tab:metrics_summary}.

\begin{sidewaystable}
\centering
\caption{Learning-based MAPF Evaluation Metrics with Mathematical Definitions}
\label{tab:metrics_summary}
\renewcommand{\arraystretch}{1.1}
\resizebox{\textwidth}{!}{%
\begin{tabular}{llp{0.65\linewidth}}
\toprule
\textbf{Category} & \textbf{Metric} & \textbf{Mathematical Definition} \\
\midrule

\multirow{3}{*}{\textbf{Success \& Failure}} 
& Success Rate (SR) & $\text{SR} = \frac{1}{n}\sum_{i=1}^n \mathbb{I}([s_i(T)=g_i])$, where $\mathbb{I}(\cdot)\in\{0,1\}$ indicates whether agent $i$ reached its goal $g_i$ by the final time $T$. \\

& Failed Agent Count (FAC) & $\text{FAC} = \sum_{i=1}^{n}\mathbb{I}[s_i(T)\neq g_i]$, number of agents that failed to reach their goals. \\

& Timeout Rate (TR) & Fraction of agents not finishing before a predefined time limit. \\
\midrule

\multirow{3}{*}{\textbf{Collision-Related}} 
& Collision Count (CC) & $\text{CC} = \sum_{t=1}^{T} \sum_{i\neq j}\mathbb{I}[s_i(t)=s_j(t)]$ for vertex collisions. Edge collisions: $\mathbb{I}[s_i(t)\leftrightarrow s_j(t+1)]$ for $i\neq j$. \\

& Collision Times Per Step (CTPS) & $\text{CTPS} = \frac{\text{CC}}{T}$, measuring collision frequency. \\

& Collision with Obstacles (CO) & Weighted count of collisions with static or dynamic obstacles. \\
\midrule

\multirow{3}{*}{\textbf{Solution Efficiency}} 
& Makespan (MK) & $\text{MK} = \max_{1\le i\le n} \{\mathrm{Cost}(\pi_i)\}$, where $\mathrm{Cost}(\pi_i)$ is the time until agent $i$ reaches $g_i$. \\

& Flowtime (FT) & $\text{FT} = \sum_{i=1}^n \mathrm{Cost}(\pi_i)$ \\

& Path Length (PL) & Sum of Euclidean or Manhattan distances traveled, aggregated across agents. \\
\midrule

\multirow{2}{*}{\textbf{Computation Time}} 
& Runtime (RT) & Total time required by the algorithm to output a solution. \\

& Per-Iteration Complexity & Time spent per planning cycle or iteration, relevant in RL-based methods and online re-planning. \\

\bottomrule
\end{tabular}%
}
\end{sidewaystable}

Success and failure metrics for learning-based methods include Success Rate (SR), similar to classic methods, but also focus on Failed Agent Count (FAC) and Timeout Rate (TR). 
The latter metrics are particularly relevant since learning-based methods often make trade-offs between optimality and completion rate.

Collision-related metrics are especially prominent for learning-based approaches, which may not provide the same collision-avoidance guarantees as classic methods. Metrics like Collision Count (CC), Collision Times Per Step (CTPS), and Collision with Obstacles (CO) quantify safety concerns, which are crucial when deploying learning-based systems in real-world environments.

Solution efficiency metrics for learning-based methods include Makespan (MK), Flowtime (FT), and Path Length (PL), similar to classic methods. 
However, these metrics are often evaluated empirically rather than with formal guarantees, reflecting the different methodological approaches.
Computation time metrics for learning-based methods include overall Runtime (RT) but emphasize Per-Iteration Complexity, which is critical for real-time applications. 
Learning-based methods frequently need to make decisions within strict time constraints, making computational efficiency per planning cycle particularly important.


\noindent
\textbf{Discussion.}
The evaluation metrics for classic and learning-based MAPF methods reflect fundamental differences in methodological priorities and application contexts. 
We can highlight several key distinctions:
\begin{itemize}[leftmargin=*]
    \item \emph{Success \& Failure} metrics are important across both paradigms, but classic methods often emphasize optimality within constraints, while learning-based methods tend to focus on robustness and completion rates in complex, uncertain environments. Classic methods typically report metrics like NIA and NOTA that incorporate strict deadlines, whereas learning-based methods more commonly report failure counts and timeout rates that acknowledge the probabilistic nature of their solutions.
    \item \emph{Collision-Related} metrics are critical for both approaches but serve different purposes. For classic methods, collisions are often theoretical constraints within the planning process and may not be explicitly measured in evaluation since many algorithms guarantee collision-free paths. In contrast, learning-based methods, which may sacrifice deterministic guarantees for scalability, must carefully quantify collision rates to assess safety implications. This explains the prominence of metrics like CC, CTPS, and CO specifically for learning-based approaches.
    \item \emph{Solution Efficiency} metrics like makespan \eqref{eq:makespan} and sum-of-costs \eqref{eq:soc} are common to both paradigms but with different emphases. Classic methods typically provide optimality bounds or guarantees with respect to these objectives, while learning-based methods report empirical performance. Additionally, classic methods often include specialized efficiency metrics (like SoD, TP, and MET) that capture nuanced aspects of performance relevant to theoretical analysis.
    \item \emph{Computation Time} represents perhaps the starkest contrast between the paradigms. Classic methods like ILP-based or SAT-based approaches prioritize finding optimal solutions even at significant computational cost, with metrics focusing on total runtime and memory usage. Learning-based methods, especially those designed for real-time deployment, emphasize per-iteration efficiency and bounded-time decision making, with metrics that reflect this real-time constraint.
    \item \emph{Algorithmic Process} metrics differ substantially between paradigms. Classic methods often report metrics specific to their algorithmic structure (e.g., CT Node Expansions, Mutex Propagation Time) to provide insight into internal processes. Learning-based methods typically focus less on algorithmic internals and more on end-to-end performance metrics relevant to deployment scenarios.
\end{itemize}

This comparison highlights how evaluation metrics reflect the fundamental trade-offs between these paradigms: classic methods prioritizing completeness, optimality, and theoretical guarantees versus learning-based approaches emphasizing scalability, adaptability, and real-time performance in complex environments. 

While the field has produced a wide array of evaluation metrics tailored to both classic and learning-based MAPF paradigms, it is important to recognize several limitations and challenges that arise from current evaluation practices. 
Notably, most published works do not report or compare results across the full spectrum of metrics outlined above. 
Instead, each study tends to select a subset of metrics that best aligns with its methodological strengths or targeted application domain. 
\textbf{This selective reporting inadvertently hampers direct comparison between algorithms and often obscures the broader landscape of algorithmic trade-offs.} 
As a result, it becomes difficult for practitioners and researchers to draw comprehensive conclusions regarding the relative merits of different approaches.

Moreover, the divergence in metric preferences between classic and learning-based MAPF communities further exacerbates this problem. 
Classic methods have traditionally emphasized theoretical guarantees and optimality-focused metrics, while learning-based approaches are more likely to report empirical performance and safety-related statistics. 
\textbf{This division creates a barrier for cross-paradigm benchmarking and restricts the transfer of insights between the two communities.} 
The lack of common ground in evaluation criteria not only impedes fair assessment but also slows progress toward integrative or hybrid MAPF solutions that could leverage the advantages of both paradigms.

To address these challenges and foster the development of the MAPF field as a whole, it would be highly beneficial for the community to \textbf{converge on a more standardized and comprehensive set of evaluation protocols}. 
Such protocols should encourage the reporting of a broader set of metrics, encompassing both theoretical guarantees and empirical behaviors, as well as those that reflect real-world constraints and safety considerations. 
Efforts to develop unified benchmark suites, open-source evaluation toolkits, and consensus on metric definitions will be instrumental in enabling more meaningful and transparent comparisons between algorithms. 
In turn, this will facilitate clearer identification of research gaps, accelerate the adoption of MAPF solutions in practice, and promote the cross-fertilization of ideas between the classic and learning-based research communities.

\subsection{Environment Scale and Baseline Selections}
\label{subsec:scale_baselines}

\subsubsection{Scaling in Map Size and Agent Number}

The relationship between map size and agent population represents one of the most critical factors affecting MAPF problem difficulty. 
As these parameters increase, the computational complexity grows exponentially, creating significant challenges for both classical and learning-based approaches.
In formal terms, consider a grid-based MAPF instance with map dimensions $H \times W$ and agent set ${1,\dots,n}$. 
Each experimental configuration can be represented as a triple $(H,W,n)$. 
Given a map $\mathcal{G}$ of size $H\times W$, the effective state space expands to $\mathcal{V}^n \equiv (H\times W)^n$ for the possible permutations of agent positions. 
This state space grows combinatorially as either the map dimensions $(H,W)$ or the agent count $n$ increases, representing a fundamental computational challenge in MAPF.

\begin{figure}[htb!]
\centering
\includegraphics[width=\linewidth]{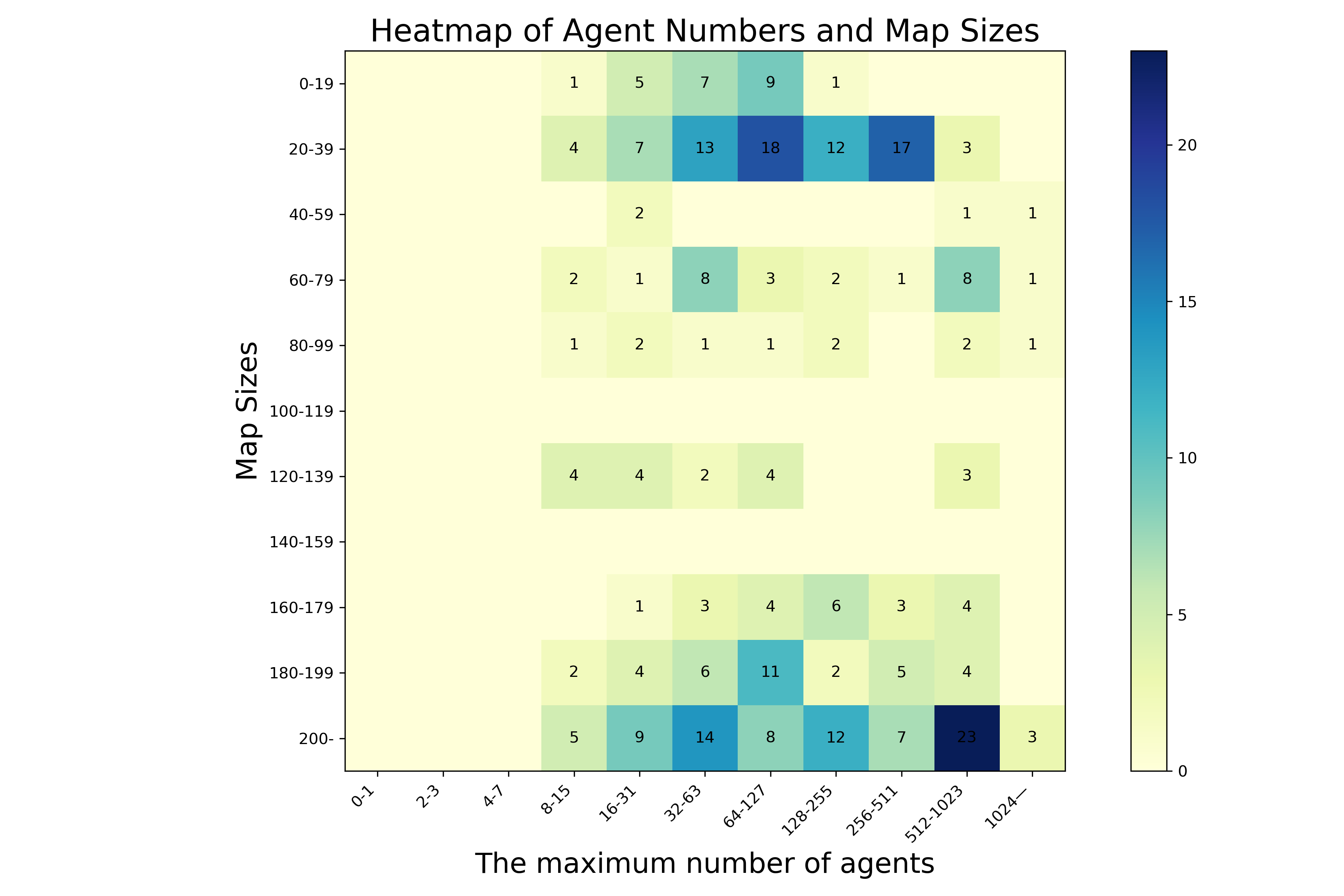}
\caption{Representative heatmap (schematic) depicting experimental configurations of map size vs.\ agent number for classical methods. 
The horizontal axis denotes the agent population $n$, 
while the vertical axis denotes the map. 
The color intensity indicates the frequency of usage in surveyed papers.}
\label{fig:classic_map_agent}
\end{figure}

\begin{figure}[htb!]
\centering
\includegraphics[width=\linewidth]{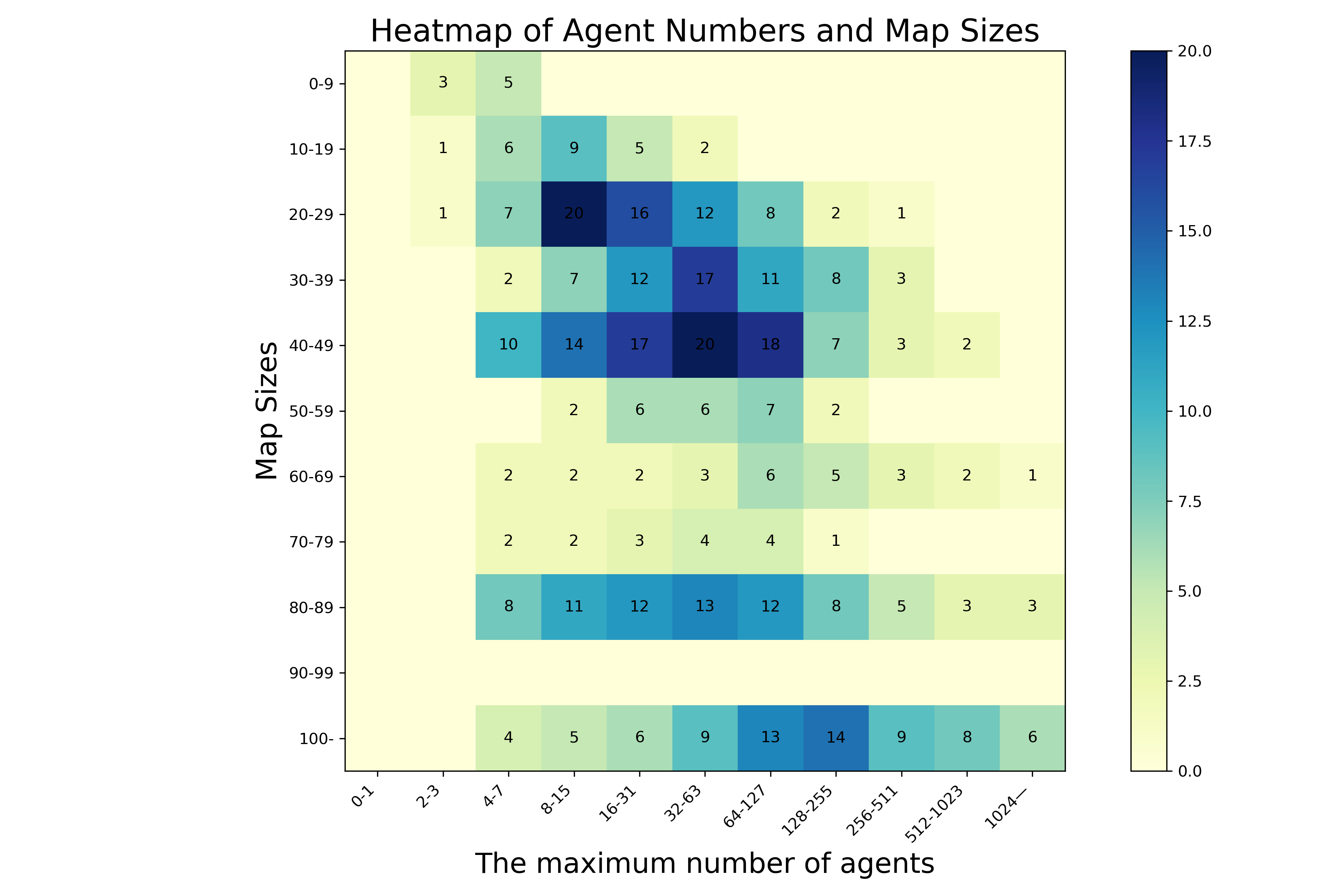}
\caption{Representative heatmap (schematic) depicting experimental configurations of map size vs.\ agent number for learning-based methods. 
The horizontal axis denotes the agent population $n$, 
while the vertical axis denotes the map dimension (e.g., $H\times W$). 
The color intensity indicates the frequency of usage in surveyed papers.}
\label{fig:map_agent}
\end{figure}

Figures~\ref{fig:classic_map_agent} and~\ref{fig:map_agent} visualize the landscape of experimental configurations used across the MAPF literature. 
These heatmaps represent the frequency with which particular $(H\times W,n)$ pairs appear in research studies, with color intensity indicating usage frequency. 
Based on our survey, we can categorize MAPF problem instances into four distinct scales:

\begin{itemize}[leftmargin=*]
\item \textbf{Small-Scale} (e.g., $H,W<10$ or $n<10$): These configurations serve primarily as algorithmic testbeds, enabling quick proofs-of-concept, debugging sessions, and theoretical validations. While limited in practical application, they allow researchers to isolate algorithmic behaviors without prohibitive computational costs.

\item \textbf{Medium-Scale} ($10\leq H,W\leq 50$ or $10 \leq n \leq64$): Representing the most common configurations in academic benchmarks, these instances strike a balance between computational feasibility and real-world relevance. Many standard MAPF benchmarks like warehouse, game, and city maps fall within this category, making it the principal testing ground for comparing algorithmic approaches.

\item \textbf{Large-Scale} ($H,W>50\times50$ or$64 \leq n \leq512$): As industrial applications gain prominence, this scale has become increasingly important. Configurations in this range test algorithms under conditions approximating real-world deployment scenarios, such as warehouse automation systems with hundreds of robots \citep{friedrich2024scalable}. At this scale, many optimal algorithms become impractical, shifting focus toward bounded-suboptimal or incomplete methods.

\item \textbf{Very Large-Scale} ($n>512$ or $H,W >100\times100$): The frontier of MAPF research, these configurations push algorithms to their limits and reveal fundamental scalability bottlenecks. At this scale, decentralized approaches, hierarchical methods, and learning-based solutions often become necessary, as centralized optimal planning becomes computationally intractable.
\end{itemize}

These two figures show that, classic methods demonstrate a clear dichotomy in the size of maps they tackle. 
On one hand, a significant portion of classic works target medium-scale problems, often with map side lengths in the range of 20 to 50. 
These settings are generally more tractable and allow for detailed benchmarking of algorithmic improvements. 
On the other hand, several studies explicitly stress-test the scalability of classic methods on very large-scale problems, with map side lengths reaching approximately 200. 
This dual focus demonstrates the longstanding tradition in the MAPF community to both refine algorithmic efficiency for moderate-sized environments and push the boundaries of scalability.
With respect to agent count, classic methods commonly address scenarios with 64 to 1024 agents. 
The distribution across this range is relatively even, and more than half of the surveyed works handle settings categorized as large-scale (hundreds of agents) or very large-scale (over a thousand agents). 
This breadth underscores the maturity and robustness of classic algorithms in managing high agent densities, a critical requirement for real-world deployments such as warehouse automation or traffic management.

In contrast, learning-based methods, which leverage deep reinforcement learning, imitation learning, or other data-driven techniques, also display polarization in the map sizes considered. 
Most current research focuses on medium-scale maps (side length 10--50) and, to a lesser extent, large-scale maps (side length 80--100). 
However, the number of agents handled by these methods is generally much lower, typically ranging from 8 to 256 agents. 
The vast majority of studies remain within the bounds of large-scale or smaller problem instances. 
This concentration on relatively modest scales suggests that, despite strong recent advances, the application of learning-based approaches to truly large and dense MAPF scenarios remains limited in practice.

It is particularly noteworthy that \textbf{learning-based methods, despite their purported advantage of scalability, are predominantly evaluated on smaller problem instances compared to classic methods}. 
This observation is somewhat counterintuitive, given that a central claim of learning-based MAPF research is the potential for improved scalability through parallelism, generalization, and efficient policy learning. 
Several factors may contribute to this gap. 
Training deep models on very large maps with hundreds or thousands of agents is computationally intensive and may be hampered by sample inefficiency or instability. 
The community has gravitated towards established benchmarks that emphasize smaller or medium-sized problems, possibly due to the high cost of data generation and verification in larger settings. 
Moreover, current neural architectures may struggle with long-range coordination and global conflict resolution in massive environments, necessitating further innovation. 
There is also a lack of standardized metrics for large-scale MAPF that are both fair and informative across classic and learning-based methods.

To more fully realize the promise of learning-based MAPF methods and to facilitate a fair comparison with classic approaches, future research should focus on several directions. 
First, there is a need to scale up benchmark problems and adopt new benchmarks that reflect the scale and density of real-world applications, with larger maps and agent populations. 
Second, algorithmic innovation is required, particularly in the design of scalable neural architectures---such as graph neural networks, hierarchical policies, or multi-stage training---that can effectively manage global coordination among thousands of agents. 
Third, the development and investigation of hybrid methods that combine classic and learning-based techniques may leverage the strengths of both paradigms and enable more efficient handling of large, complex environments. 
Additionally, more efficient training strategies, such as curriculum learning, transfer learning, and distributed training, are essential to mitigate the computational overhead of scaling up learning-based solutions. 
It is also important to establish comprehensive and standardized evaluation practices that span a broad range of problem scales, ensuring meaningful and reproducible comparisons across different methodologies. 
Finally, theoretical analysis of the scalability and generalization limits of learning-based approaches would provide valuable insights into their potential advantages and limitations compared to classic techniques.

\subsubsection{Typical Baselines}

In this section, we systematically analyze the selection and distribution of baselines in the MAPF literature, drawing insights from both classical and learning-based approaches.

\begin{figure}[htb!]
\centering
\includegraphics[width=\linewidth]{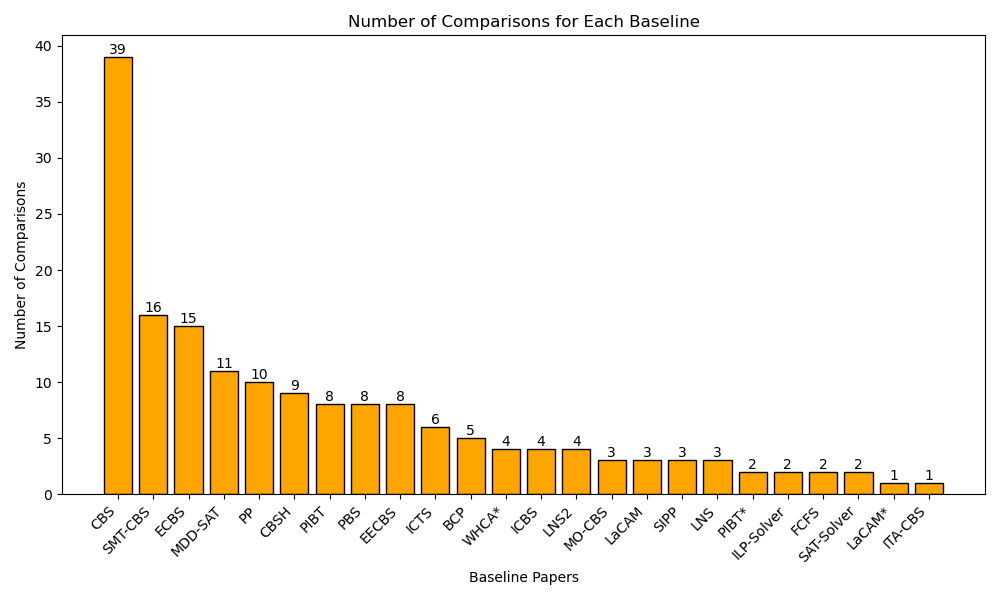}
\caption{Histogram (schematic) illustrating the frequency of classical baseline algorithms used in MAPF experiments. 
Horizontal axis: baseline names or abbreviations. 
Vertical axis: number of publications that adopt each baseline for comparison.}
\label{fig:classical_baseline_compare}
\end{figure}

\paragraph{Baselines in Classical Methods.}
A statistical overview of classical baselines is shown in Figure~\ref{fig:classical_baseline_compare}. 
This histogram demonstrates the frequency with which different classical algorithms are adopted as comparison baselines in MAPF experimental studies. 
Notably, Conflict-Based Search (CBS) overwhelmingly dominates as the baseline of choice, being cited in more publications than all other methods by a factor of two to four. 
Other CBS variants—such as SMT-CBS and ECBS—as well as compilation-based solvers like MDD-SAT and PP, follow at a considerable distance. 
The distribution exhibits a pronounced long-tail characteristic: 
a small number of methods are routinely compared, while the majority are only sporadically included. 
It is also worth noting that, in classical MAPF research, learning-based methods are rarely considered as baselines, reflecting a clear methodological divide in the literature.

\begin{figure}[htb!]
\centering
\includegraphics[width=\linewidth]{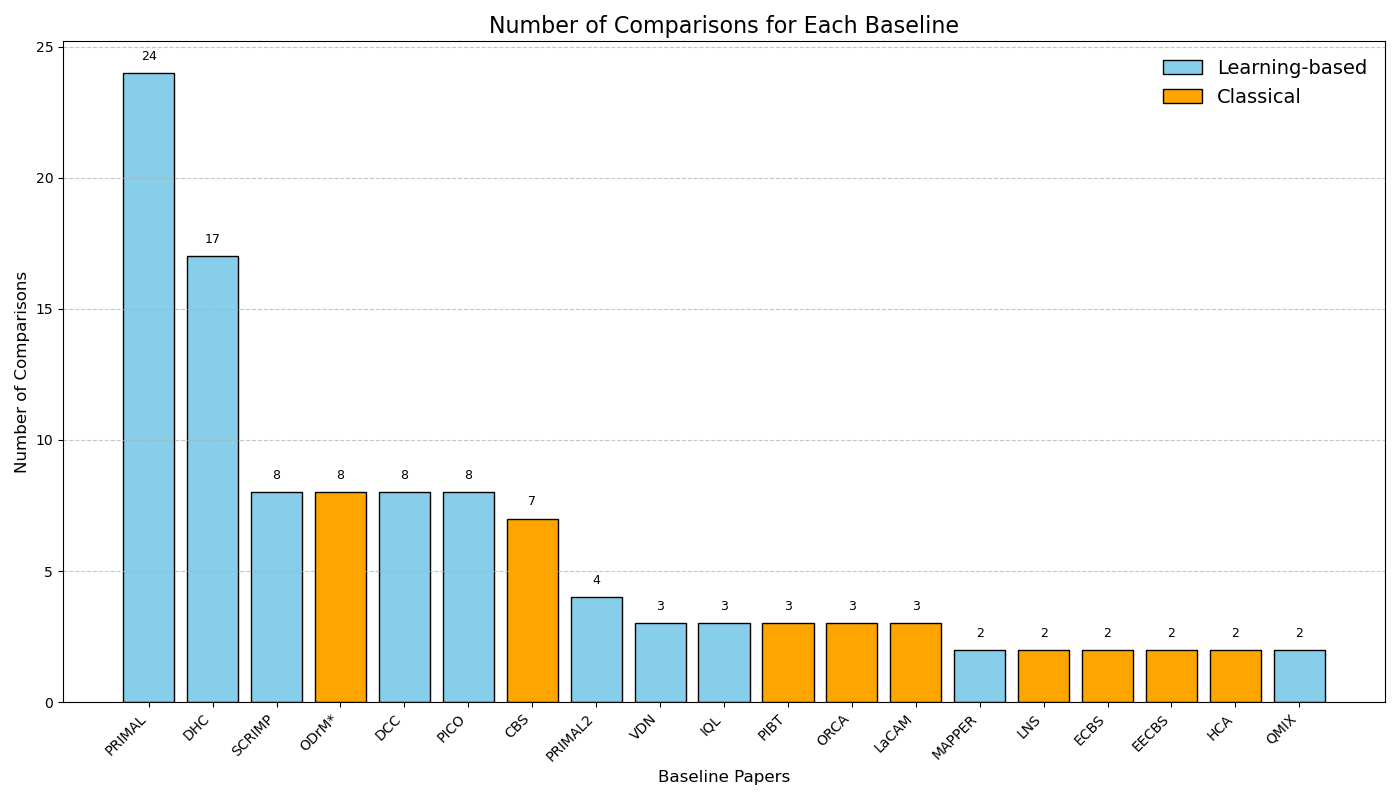}
\caption{Histogram (schematic) illustrating the frequency of various baseline algorithms used in MAPF experiments. 
Horizontal axis: baseline names or abbreviations (e.g., CBS, PRIMAL, DHC). 
Vertical axis: number of publications that adopt each baseline for comparison.}
\label{fig:baseline_compare}
\end{figure}

\paragraph{Baselines in Learning-based Methods.}
Figure~\ref{fig:baseline_compare} extends this analysis to studies proposing learning-based or hybrid MAPF solutions. 
Here, the long-tail pattern persists, with a few dominant methods—specifically PRIMAL and DHC—being featured in most comparisons. 
Methods such as SCRIMP, DCC, and PICO also appear with moderate frequency. 
Importantly, the set of baselines in this category spans both learning-based and classical approaches, with ODrM* and CBS remaining the most frequently adopted classical algorithms. 
Among learning-based baselines, PRIMAL stands out as the method of choice, while other highly-cited baselines are predominantly communication-based or reinforcement learning (RL) algorithms. 
This trend underlines a growing recognition of the value of cross-paradigm comparisons within the MAPF community.

\begin{sidewaystable}
\centering
\caption{Representative Baselines Commonly Used in MAPF Experiments}
\label{tab:baseline_list}
\renewcommand{\arraystretch}{1.2}
\resizebox{\linewidth}{!}{%
\begin{tabular}{p{0.2\linewidth} p{0.6\linewidth} p{0.2\linewidth}}
\toprule
\textbf{Method} & \textbf{Brief Description} & \textbf{Example References} \\
\midrule
\textbf{CBS (Conflict-Based Search)} 
& 
Optimal search-based approach that resolves collisions by branching on conflict constraints. 
& 
\citep{sharon2015conflict}, \citep{stern2019multi}. \\

\textbf{DHC (Deep Heuristic Conflict Resolution)} 
& 
Integrates an RL-based heuristic with a conflict-based search or priority-based scheme. 
& 
\citep{sartoretti2019primal}, \citep{alkazzi2024comprehensive}. \\

\textbf{PRIMAL} 
& 
Hierarchical RL framework for one-shot or lifelong MAPF, focusing on partial observability. Often used in warehouse-like settings. 
& 
\citep{sartoretti2019primal}, \citep{damani2021primal}. \\

\textbf{ILP/SAT Solvers} 
& 
Compilation-based methods that translate MAPF to integer linear programs or SAT formulas, then leverage generic solvers. 
& 
\citep{surynek2016efficient}, \citep{surynek2022problem}. \\

\textbf{Independent RL} 
& 
Each agent learns a single-agent policy (e.g., DQN or PPO) ignoring multi-agent coordination except via collisions. 
& 
\citep{qiu2020multi}. \\

\textbf{MADDPG/ CTDE} 
& 
MARL with centralized training and decentralized execution, common for continuous or partially observable tasks. 
& 
\citep{Pham2023OptimizingCM}, \citep{ma2021learning}. \\
\bottomrule
\end{tabular}%
}
\end{sidewaystable}

Table~\ref{tab:baseline_list} provides a structured summary of representative baselines commonly used in MAPF experiments, including brief descriptions and example references.

\paragraph{Discussion and Analysis.}
A closer examination of these baseline selection patterns reveals several important phenomena. 
Firstly, the overwhelming preference for CBS and its variants in classical MAPF research underscores its status as the de facto standard for both optimality and interpretability. 
However, this reliance can obscure the diversity of problem characteristics, particularly in scenarios where CBS's search-based paradigm may not scale efficiently.
In contrast, learning-based and hybrid methods, such as PRIMAL and DHC, have established themselves as essential baselines in recent literature, especially for large-scale, partially observable, or dynamically changing environments. 
The prevalence of these methods illuminates the community’s shift towards tackling real-world complexities that challenge traditional solvers.

Interestingly, \textbf{the long-tail pattern in baseline selection suggests a lack of standardization and highlights the heterogeneity of experimental settings in MAPF research.} 
This phenomenon is even more pronounced in learning-based studies, where the choice of baseline is influenced by the target scenario as well as by the computational resources available for large-scale training and evaluation.
Moreover, the comparative analysis between independent RL and centralized training/decentralized execution (CTDE) paradigms—exemplified by MADDPG—reflects an ongoing exploration of trade-offs between scalability, coordination complexity, and solution quality. 
While independent RL is simple and scalable, it often underperforms in highly-coupled MAPF settings. 
Conversely, CTDE approaches can better capture agent interactions but at the cost of increased training complexity.

In practical terms, most publications select two to three baselines to highlight either 
(i) the performance gap in large-scale or online scenarios—where classical solvers may fail due to computational bottlenecks, or 
(ii) the near-optimality of learning-based methods on small- to medium-sized benchmarks, where exact solutions are feasible and serve as a gold standard.
Despite these advances, two key limitations persist. 
First, the dichotomy between classical and learning-based baselines can hinder fair and comprehensive evaluation, especially as hybrid methods become more prevalent. 
Second, the long-tail distribution of baseline usage complicates cross-paper comparison and meta-analysis, as different studies often report results against disjoint sets of baselines.

To address these issues, future research should strive for more systematic benchmarking protocols, including the adoption of diverse and representative baseline sets across both classical and learning-based paradigms. 
It is essential to encourage the inclusion of strong learning-based baselines in classical research, and vice versa, to foster cross-paradigm insight. 
Furthermore, the development of standardized evaluation environments and open-source baseline implementations will enhance reproducibility and fairness in MAPF benchmarking, ultimately accelerating progress in the field.

\subsection{Beyond Static MAPF: Dynamic Environment}
\label{subsec:dynamic_mapf}


Despite significant progress in classical MAPF algorithms, the extension of these approaches to dynamic environments remains an open and relatively underexplored research direction. 
In dynamic MAPF, the environment itself evolves over time, with obstacles appearing or disappearing as agents move or wait. 
This setting more accurately reflects real-world applications, such as warehouse robotics and urban mobility, where exogenous changes in the map can occur unpredictably and may substantially impact agent coordination.

To systematically assess the robustness of classical MAPF methods under such environmental shifts, we leveraged the POGEMA~\citep{skrynnik2024pogema} benchmark\footnote{\url{https://github.com/CognitiveAISystems/pogema-benchmark}.}, which facilitates controlled experimentation with varying proportions of dynamic obstacles. 
Here, a dynamic environment is defined as one in which a certain fraction of obstacles—relative to the total—may appear or disappear as agents interact with the map. 
This experimental design enables the evaluation of MAPF solvers across a spectrum from static to highly dynamic scenarios.

\begin{figure}[htb!]
\centering
\includegraphics[width=.5\linewidth]{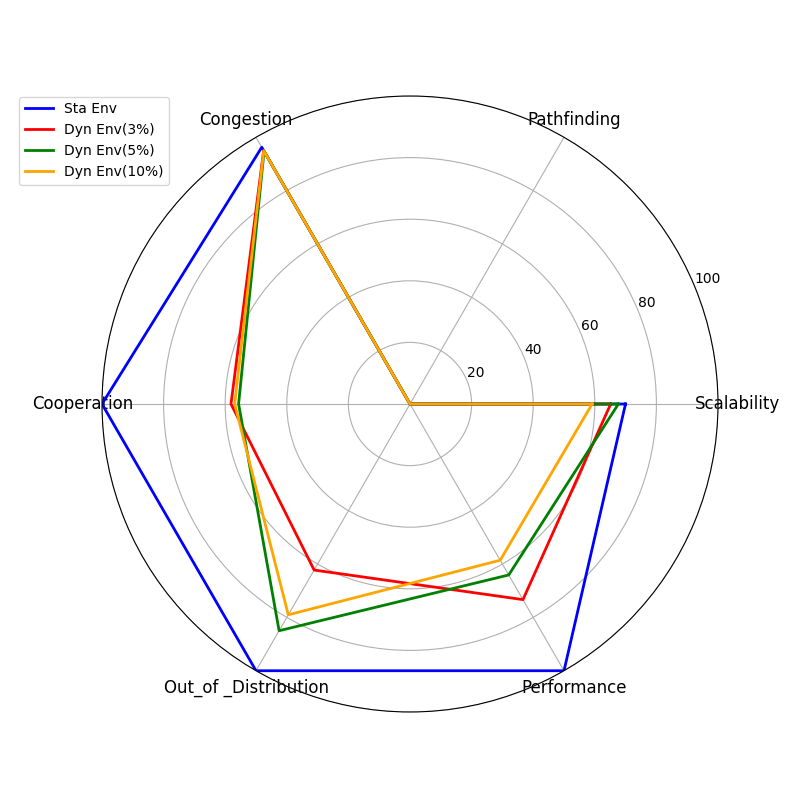}
\caption{Results of the evaluation of DCC algorithm in static and dynamic envirnments}
\label{Distribution_Schematic}
\end{figure}

Our empirical analysis, summarized in Figure~\ref{Distribution_Schematic}, compares the performance of the DCC algorithm across static and dynamic conditions using a diverse set of benchmark metrics. 
These metrics include: overall performance on random maps, generalization to out-of-distribution (MovingAI-tiles) maps, scalability with respect to agent count, cooperation efficiency on dense puzzle maps, congestion minimization, and single-agent pathfinding optimality. 

The results indicate that classical MAPF algorithms experience a noticeable decline in overall performance as the proportion of dynamic obstacles increases. 
Specifically, both out-of-distribution generalization and cooperation metrics degrade in more dynamic settings, highlighting the increased difficulty of maintaining effective coordination when the environment is non-stationary. 
In contrast, the scalability and congestion metrics remain relatively stable, suggesting that these aspects are less sensitive to environmental dynamics, at least under the tested conditions. 
Notably, the pathfinding metric remains at zero across all settings, reflecting the inability of the current algorithm to find optimal solutions even in the single-agent case under dynamic constraints.

These findings underscore the urgent need for principled algorithmic advances in dynamic MAPF. 
Future research should prioritize the development of solvers that can adapt in real time to environmental changes, possibly by integrating online learning, predictive modeling of obstacle dynamics, or robust planning under uncertainty. 
Furthermore, new benchmarks and evaluation protocols—such as those enabled by POGEMA—will be essential for tracking progress and establishing reliable baselines in this challenging domain. 
Ultimately, bridging the gap between static and dynamic MAPF remains a critical step toward deploying multi-agent planning systems in complex, real-world environments.

\section{Future Work}\label{sec:future}

As MAPF continues to evolve in both classical and learning-based paradigms, several promising research directions are emerging to address the increasing complexity, dynamism, and heterogeneity of real-world applications. 
In this section, we outline key avenues for future investigation, ranging from mixed-motive and generative MAPF to language-grounded planning, collision-informed frameworks, large-scale agent coordination, neural solver integration, formal verification, and adaptation to dynamic environments. 
Each subsection discusses open challenges and potential methodologies, aiming to provide a roadmap for advancing the theoretical and practical boundaries of MAPF research.

\subsection{Mixed-Motive MAPF}

Traditional MAPF formulations typically presume that all agents adhere to a single global objective or, at minimum, have no direct incentives to stray from the centrally prescribed collision-free plan. 
This assumption, though compatible with single-organization settings, is less valid when agents belong to separate stakeholders or hold private goals and constraints~\citep{friedrich2024scalable,he2024social}. 
A promising line of inquiry thus lies in mechanism design~\citep{kollock1998social} and information design~\citep{dughmi2016algorithmic}, whereby the MAPF engine not only imposes combinatorial collision-avoidance rules but further orchestrates how agents reveal their preferences, how resource usage is priced or compensated, and how incentives unfold under partial information or strategic misreporting. 
In such scenarios, agents can, for instance, exaggerate the cost of diverting around congested corridors or withhold crucial timing details about resource availability. 
Disentangling this strategic complexity involves carefully crafting the ``rules of the game'' (mechanism design) alongside the ``rules of communication'' (information design) so that rational agents ultimately converge upon collision-free, cost-effective routes.

On the mechanism design side, one can embed proven approaches such as Vickrey–Clarke–Groves into classical methods like conflict-based or priority-based search, effectively transforming each conflict resolution or priority assignment into a small ``auction,'' where agents bid or signal valuations of potential paths. 
This process relies on transfer payments or side payments to disincentivize blocking maneuvers or the withholding of crucial route information. 
However, layering mechanism design onto conflict-splitting or meta-agent merging poses novel algorithmic questions: 
each branching step could redistribute payoffs to reflect the combinatorial implications of agent conflicts, and the complexity of running repeated auctions in large MAPF instances may grow quickly unless the solver intelligently prunes or aggregates conflict sets. 
Bridging domain-specific heuristics (e.g., corridor constraints, target symmetries) with the incentive rules thus becomes critical for tractable deployment.

In parallel, information design tactics can address the reality that agents might not wish to share complete data on their start times, fuel costs, or future tasks. 
Revealing too much can expose sensitive business operations, but revealing too little can lead to wasted capacity or collisions in heavily trafficked areas. 
One avenue is to devise partial revelation policies, whereby high-level conflict signals (such as approximate arrival times or corridor usage) are disclosed, while private valuations or exact cost structures remain hidden. 
Such partial information can be channeled through multi-agent reinforcement learning modules, which learn how to integrate coarse signals into bidding or negotiation schemes without requiring every detail of an agent’s internal state. 
This is particularly relevant in dynamic or partially observable environments, where continuous-time updates or abrupt schedule changes make full disclosure impossible or undesirable.

Learning components can further refine these mechanism- and information-aware protocols by adapting them to empirical agent behaviors in large-scale or rapidly changing contexts. 
A reinforcement learner, for instance, might observe trends in how certain conflict auctions are repeatedly won or lost and adjust cost functions or penalty structures so that no single agent has disproportionate power to stall or block traffic. 
Equilibrium-based MARL techniques can incorporate game-theoretic analyses—ensuring that collision-free outcomes remain stable against unilateral deviations—while exploiting data-driven training signals to handle highly non-stationary conditions~\citep{yang2020learning,lin2023information}. 
The interplay between classical branching heuristics, incentive-compatible pricing, and partial information sharing thus presents a rich field for new theoretical models and practical implementations. 
Ultimately, these research aim to produce MAPF solvers that not only rule out collisions but also account for strategic considerations and incomplete information, enabling robust and efficient multi-agent coordination where full cooperation cannot be taken for granted.

\subsection{Generative MAPF}

Generative modeling offers a fundamentally different perspective on MAPF compared to classical search-based algorithms. 
Instead of navigating a collision-free solution space incrementally, one trains a diffusion-based or flow-based model to sample entire spatio-temporal agent trajectories simultaneously~\citep{liang2024multi,shaoul2025multirobot,andreychuk2025mapf}. 
Diffusion models, for instance, learn to reverse a progressive noise-injection process, thus mapping random noise to structured outputs~\citep{ho2020denoising};
in a MAPF context, each output would encode a multi-agent solution capturing both temporal dependencies (when agents move) and spatial constraints (which routes they follow). 
While collisions may not be strictly eliminated at generation time, the training process can steer the learned distribution toward more feasible configurations by incorporating approximate collision penalties or domain-aligned regularizations. 
Upon deployment, a batch of trajectories is sampled in parallel and can be quickly post-processed by classical MAPF methods such as conflict-based or acceptance–rejection checks. 
This two-stage pipeline rapidly explores a diverse space of solutions while preserving the guarantees of collision resolution.

Normalizing flows provide a complementary approach, in which an invertible transform maps a simple base distribution to the manifold of agent paths~\citep{papamakarios2021normalizing}.
By carefully designing coupling layers to partition or reorder spatial and temporal coordinates, one can embed approximate collision-avoidance hints directly in the architecture. 
Although enforcing strict feasibility within these flows is non-trivial due to highly non-linear collision constraints, such biases can reduce the burden on subsequent classical search or local repair. 
Moreover, controlling the likelihood of a sample allows the solver to prioritize those paths the model deems more probable under the learned distribution, potentially improving efficiency when large agent teams are present.

Another avenue of research concerns how these generative processes can adapt in real time. 
In-context learning~\citep{xie2022an} and meta-learning~\citep{finn2017model} mechanisms suggest that once a generative model has learned a broad ``prior'' over multi-agent trajectories (possibly from diverse environments or partial solver outputs), it can refine its sampling behavior when presented with a small set of example configurations from the current scenario. 
Rather than retraining from scratch, the model conditions on these examples to better accommodate new obstacles, agent dynamics, or unexpected start–goal patterns. 
If collisions arise even after adaptation, domain knowledge from classic MAPF remains essential for final adjustments like corridor conflict splitting or priority-based local repairs. 
This synergy underscores how generative modeling should not stand alone but instead integrate with core MAPF heuristics, ensuring robust feasibility checks and the ability to fine-tune solutions for complex domains.

Future investigations must probe deeper into how to embed domain constraints directly into the generative process, thereby reducing the reliance on ad hoc collision penalties and post-hoc rejigs. 
One possibility is to incorporate differentiable approximations of collision detection—enabling gradient-based updates that actively push sampled paths away from high-conflict regions—or to engineer flow-based layers that inherently separate agent trajectories across time steps~\citep{christopher2024constrained}. 
Another research direction would explore advanced diffusion variants~\citep{de2021diffusion}, such as diffusion probabilistic fields~\citep{zhuang2023diffusion}, in which spatio-temporal dependencies are represented more flexibly and can be more directly aligned with grid- or graph-based MAPF formulations. 
Further, collecting training datasets that reflect diverse conflict scenarios—rather than a narrow subset of feasible solutions—will enhance the model’s capacity to handle previously unseen, highly congested conditions. 
Ultimately, generative methods for MAPF promise to balance global exploration with established collision-avoidance paradigms, providing a rich ground for algorithmic breakthroughs that blend the creativity of deep generative sampling with the reliability of classical solver principles.

\subsection{Language-Grounded MAPF}

The emergence of large language models presents novel opportunities to significantly broaden the methodological landscape of MAPF planning, both in classical and learning-based paradigms. 
From the perspective of classical algorithms, many established methods rely heavily on carefully tailored heuristics or branching strategies. 
Instead of manually designing these components, one may employ an LLM-driven meta-search process such as a FunSearch-like framework~\citep{romera2024mathematical,liu2024evolution} that synthesizes and refines new heuristics through evolutionary computation and language-guided reasoning. 
The key idea is to let the LLM propose candidate heuristic formulas or conflict-splitting schemes based on descriptive natural-language prompts, then automatically evaluate and evolve them within the MAPF setting, continually searching for high-quality or domain-specific strategies that outperform static hand-designed approaches.

In a learning-based context, LLMs can serve more directly as a policy generator by leveraging their built-in commonsense and reasoning abilities to propose agent moves or path expansions~\citep{zeng2024perceive,chen2024solving,atasever2025multiagent,seo2025llmdr}. 
Combined with ``tool-using'' functionalities, the LLM could offload subproblems—such as temporarily resolving tight collisions or revalidating partial solutions—to external classical solvers. 
In this way, the language model orchestrates a hybrid workflow where parts of the MAPF pipeline remain driven by exact or bounded-optimal algorithms, yet the overall agent behavior is flexibly guided by the LLM’s adaptive priors, effectively lowering the barrier to tackling new or unstructured constraints. 
Extending this further, an agentic workflow approach would allow the LLM itself to manage parallel requests, re-planning triggers, or conflict merges and splits as circumstances evolve in real time. 
By respecting key MAPF constraints while drawing on the LLM’s substantial capacity for semantic understanding and multi-step reasoning, the framework might scale seamlessly to more complex scenarios and dynamic constraints than purely algorithmic or purely data-driven methods could feasibly address.

The transformative potential of these LLM-integrated approaches lies in their ability to expand the boundaries of MAPF. 
With advanced language understanding, it becomes feasible to incorporate ambiguous or high-level directives (such as safety zones, time windows, or ethical guidelines) into a single integrated solver pipeline~\citep{10903304}. 
LLMs also invite us to rethink the granularity and scope of MAPF: 
Instead of limiting research to discretized, grid-centric formulations, a language-grounded system could harmonize classical domain representations with higher-level, continuously updated knowledge about tasks, agent roles, and mission objectives, stepping beyond short-horizon collision avoidance toward richer multi-agent orchestration. 
By trusting in the surprising range of LLM capabilities and experimenting with nontrivial orchestration workflows, researchers and practitioners alike can seek to solve more ambitious MAPF tasks that were previously deemed too complex to formalize or too large to handle through conventional techniques alone. 
Ultimately, rather than producing standalone solutions, the richness of LLM-based MAPF encourages a more systematic synergy, emphasizing the interplay between evolutionary search, language-driven policy design, exact verification, and multi-agent adaptation to create a new generation of MAPF solvers unconstrained by traditional boundaries.

\subsection{Collision-Informed MAPF}

A promising research avenue lies in reconciling the discrete foundations of classical MAPF methods with the continuous safety and collision-avoidance constraints prevalent in real-world robotics. 
One approach is to incorporate geometric barrier functions, partial differential equations, or other collision-informed constraints directly into the neural architecture so that collision-free behavior is maintained throughout training and inference~\citep{raissi2019physics,liu2024physics}. 
By encoding these constraints into the underlying policy or trajectory generator, it becomes feasible to guarantee adherence to physical safety margins and dynamic feasibility conditions without requiring explicit post-processing steps. 
This integration can be particularly powerful when paired with classic search-based planners (such as CBS or LNS), where the discrete expansions would co-exist with learned, collision-aware modules that efficiently navigate continuous dynamics or partial observability. 

A second direction is to leverage specialized neural architectures that exhibit symmetry or equivariance properties consistent with MAPF tasks. 
Many multi-agent grid-based environments enforce uniform motion rules and relative positioning constraints that could be captured by group-equivariant neural networks or graph neural networks~\citep{satorras2021n,gerken2023geometric}. 
These architectures can be further tailored by incorporating insights from domain decomposition and agent grouping used in priority- or constraint-based algorithms. 
They would thereby preserve the interpretability of classical methods and simultaneously harness data-driven generalization. 
An additional refinement would involve neural architecture search (NAS) to automatically discover task-specific architectures that effectively balance expressive power and computational overhead, ensuring scalability to large agent counts and continuous-time maneuvers~\citep{zoph2017neural,elsken2019neural}.

Furthermore, there is growing interest in formally verifying learned neural modules so as to provide correctness guarantees under a wide range of deployment scenarios. 
If the policy or planner can be represented symbolically, formal methods could systematically check whether collision-free paths are produced for all valid agent configurations and environment states within specified bounds~\citep{sun2019formal,ehlers2017formal,corsi2021formal}. 
Although bridging neural networks with rigorous verification poses nontrivial algorithmic and computational hurdles, recent progress in combining solver-based techniques (e.g., SMT, MIP) with neural certificate generation offers a glimpse into how collision-free properties can be mathematically assured~\citep{katz2017reluplex,eleftheriadis2022neural}. 
Such integrative solutions would not only raise confidence in the deployment of learned MAPF systems but also illuminate potential failure modes that can guide subsequent network architecture revisions.

\subsection{Many-Agent Pathfinding} 

As agent teams grow to hundreds or even thousands of individuals, contemporary MAPF solutions struggle to balance real-time responsiveness with global collision guarantees. 
Traditional centralized algorithms excel in principled conflict resolution but often exhibit superlinear computation times as the agent count rises. 
Purely data-driven approaches, while adaptable to large heterogeneous environments, risk overlooking worst-case collisions if not carefully constrained. 
A productive research direction is thus to integrate classic multi-level MAPF formulations with newly emerging theoretical and learning-based insights, providing a unifying framework capable of scaling to ultra-large agent populations.

One promising avenue is to adopt a hierarchical decomposition approach at multiple levels of granularity, dividing the environment into manageable regions connected by well-defined interfaces~\citep{zhang2021hierarchical,lee2021parallel}. 
While classical corridor reasoning or disjoint splitting can handle local collisions effectively, significant open questions remain in how to maintain global consistency across inter-region boundaries. 
This challenge can be tackled by refining high-level conflict-detection schemes or priority-based synchronization protocols, which ensure that independently computed subsolutions do not generate collisions when merged. 
Crucially, learning-based modules can be incorporated to predict congestion patterns or to re-partition regions on the fly, especially under dynamic task assignments. 
In this scheme, each layer could exploit classical completeness and optimality checks, but rely on data-driven components for decisions such as how to group agents, when to initiate re-routing, or which regions can be safely skipped during intermediate conflict detection. 
Successfully merging these components requires developing robust meta-planning algorithms that preserve collision-free guarantees yet adapt to shifting agent distributions or high-density bottlenecks at runtime.

Another prospective direction involves leveraging the mean-field or other large-population theories to characterize coordination as an aggregate phenomenon, especially when detailed one-to-one interactions become computationally intractable~\citep{yang2018mean,park2024mean}. 
While mean-field approximations can mitigate dimensionality issues by focusing on ensemble behavior, a key research hurdle is to ensure the resulting macroscopic analysis remains relevant to discrete collision checks and agent-level path constraints. 
Coupling such theoretical frameworks with localized collision resolution demands creative approximations or hybrid models that reconcile continuous-density abstractions with discrete pathfinding. 
For instance, machine learning could aid in periodically mapping dense agent flows onto collision-aware graph abstractions, yielding a fluid-to-discrete handover that maintains tractable solution spaces. 
These research efforts could invoke domain-crossing methods from network flow theory, computational physics, or distributed control, thereby pushing MAPF toward a unified methodology that simultaneously handles micro-level path feasibility and macro-level congestion effects.

Beyond algorithmic advances, there is considerable room for cross-disciplinary systems integration and real-world validations. 
In large-scale robotics or warehouse logistics, where thousands of mobile units operate simultaneously, the interplay of hardware limits and safety requirements adds layers of complexity. 
Investigating parallel computing paradigms or distributed ledger technologies may uncover efficient ways to coordinate agent subgroups and encode conflict resolution constraints at scale. 
Leveraging cloud-based or high-performance computing resources might enable near real-time solutions for ultra-large MAPF instances, while novel concurrency control strategies can address synchronization overheads in a distributed environment. 
Integrating these systems-level solutions with hierarchical or mean-field formulations holds the potential to move from purely theoretical frameworks toward reliable, large-scale industrial deployments.

\subsection{Neural MAPF Solver} 

A promising extension of current compilation-based methods for MAPF lies in the integration of neural solvers that harness recent progress in deep learning for combinatorial optimization. 
Classical approaches that convert MAPF into SAT, SMT, CSP, ASP, or MIP formulations typically rely on general-purpose solvers with hardcoded branching heuristics and conflict resolution strategies. 
While such solvers have matured significantly, they often lack adaptability to domain-specific structures in MAPF, such as dynamic collision-avoidance patterns or the distinct bottlenecks introduced by dense agent populations. 
Neural solvers, in contrast, can learn search policies, branching rules, or cut-generation heuristics tailored to these very properties, thus offering a new level of flexibility and problem-awareness~\citep{sun2018learning,chen2022learning}.

One key research direction is the design of differentiable solver components that incorporate learned embeddings of MAPF constraints~\citep{ryu2019plug,sun2021scalable}. 
By embedding agents’ spatiotemporal interactions and collision constraints into deep neural architectures, one could enable backpropagation of solution-quality signals through the solver’s decision pipeline. 
This approach might draw inspiration from graph neural networks or Transformer-based models, where the nodes represent time-indexed agent states and the edges capture conflict relationships. 
Iterative message-passing schemes could then identify tight collision sets or promising feasible transitions, guiding the solver toward near-optimal solutions more quickly than generic combinatorial methods.

Another direction is the development of neural heuristics for column generation or conflict-driven clause learning. 
In a MIP setting, candidate path ``columns'' that fail to meaningfully reduce conflicts might be pruned early by a learned ranking function, allowing the solver to focus on highly efficacious routes. 
In a SAT or SMT framework, learned conflict-clause generation could allow the solver to incrementally add only the most potent constraints, rather than enumerating a large swath of collisions indiscriminately. 
Such targeted addition would reduce solver overhead and produce solutions that reflect context-aware collision avoidance. 
These techniques may benefit from large-scale pretraining on diverse MAPF scenarios, enabling models to learn broad priors about collision structure and thematically consistent path flows.

A further challenge in neural solver integration is accommodating the idiosyncrasies of MAPF, such as highly heterogeneous agent goals, domain-specific resource constraints, and multi-objective cost structures. 
Where standard neural combinatorial optimization often focuses on uniform problem setups, MAPF tasks can vary significantly in complexity and could include partial observability or on-the-fly environment changes. 
Building robust neural solvers that adapt to these variations requires layering explicit MAPF constraints over trainable solver components~\citep{bengio2021machine,mazyavkina2021reinforcement}. 
Coupling domain knowledge—like joint collision checks and known bottleneck patterns—with learned strategies for branching or constraint propagation offers a fertile research path. 
This synergy, if pursued systematically, has the potential to move beyond single-paper demonstrations and establish a coherent body of work on neural solver design for MAPF.

Finally, an important strand of future research concerns theoretical validation and practical deployment~\citep{martin2024learning}. 
Neural solvers can reduce runtime while discovering high-quality solutions that classical solvers might overlook. 
However, guaranteeing completeness or bounding potential suboptimality can be challenging when the solver’s decision logic is partially determined by learned modules. 
Techniques derived from explainable AI and formal verification may be adapted to certify the correctness of neural ``subroutines,'' ensuring that they do not violate MAPF’s fundamental collision-avoidance requirements. 
This interplay of data-driven guidance and rigorous solver-level checks embodies a broader ``AI for Science'' ambition, wherein machine learning and symbolic search unite to tackle complex multi-agent coordination problems with unprecedented speed and reliability. 
By actively addressing these verification and validation questions, neural solvers for compilation-based MAPF could achieve both strong practical performance and robust theoretical assurances.

\subsection{Automated MAPF Proving}

A promising avenue for advancing both classical and learning-based MAPF methods is the integration of automated theorem proving (ATP) frameworks, such as Lean4, with state-of-the-art machine learning techniques~\citep{moura2021lean,wang2024theoremllama}. 
Classical MAPF solvers, which strive for optimality or completeness guarantees through explicit search, stand to benefit from formal verification strategies that ensure the correctness of algorithmic steps and collision-avoidance proofs. 
In particular, encoding these algorithms in Lean4 could help verify key properties of conflict resolution, priority assignments, or subproblem-specific constraints, thus yielding robust correctness certificates and facilitating the discovery of overlooked corner cases. 
Such certified correctness has the potential to substantially accelerate the development cycle for classical methods by enabling rapid prototyping, rigorous debugging, and targeted refinements through machine-assisted proofs.

At the same time, learning-based approaches for MAPF, including deep neural planners, Large Neighborhood Search (LNS) heuristics guided by data-driven models, or reinforcement learning policies with partial observability, may similarly benefit from an ATP perspective~\citep{zhang2025understanding,li2025formalization}. 
Although most learning pipelines offer no hard completeness or optimality guarantees, it is conceivable to couple them with Lean4 reasoning modules that analyze candidate policies to ensure adherence to baseline collision-free constraints or to produce counterexamples when violations occur. 
This integration would require extending the expressiveness of Lean4 libraries to capture the specific temporal and spatial properties of multi-agent coordination, as well as devising pragmatic methods for encapsulating learned heuristics as logical axioms or hypotheses. 
By codifying the interactions between data-driven modules and the underlying MAPF rules, one could systematically identify when learned models are falling outside valid solution spaces.

Such a synthesis could also open a path for hybrid verification strategies, where a learning-based or classical solver generates candidate solutions that are then incrementally refined through symbolic reasoning. 
For instance, if a partial solution passes initial semantic checks in Lean4 but triggers unresolved proof obligations, a learning-based method might attempt localized corrections, or the classical solver might apply constraint-splitting techniques guided by the theorem prover’s feedback. 
This bidirectional workflow could form a self-correcting loop, offering stronger completeness assurances while still harnessing the scalability of machine-driven exploration. 
Looking beyond near-term implementations, one might further explore how these formal verification pipelines can incorporate emerging advances in automated proof search, including LLM-based theorem provers or interactive proof systems that accept high-level domain insights from MAPF practitioners. 
By intertwining automated theorem proving, classical MAPF theory, and learning-based heuristics within a single ecosystem, future research can aspire to create rigorously verified solvers that remain adaptable to the multifaceted and ever-evolving challenges of real-world multi-agent coordination.

\section{Related Work}\label{sec:related}

Owing to the importance of MAPF in robotics, logistics, and numerous other sectors, the research community has produced an extensive array of surveys over the last few years. 
On the classical side of MAPF, several works have established foundational terminology, formulations, and categorizations, covering search-based, rule-based, and compilation-based algorithms for grid-based and other graph formulations~\citep{stern2019multi,stern2019mapfsurvey,wu2023review,ma2022graph,gao2024review,yang2023path,salzman2020research,tjiharjadi2022systematic,zhou2023research}. 
These surveys focus on the central challenge of achieving collision-free paths for multiple agents, providing overviews of optimal and suboptimal pathways such as conflict-based search (CBS), reduction-based formulations (SAT, MILP, CSP), priority-based heuristics, and decoupling techniques. 
They also highlight real-world applications, enumerating key issues like scalability, dynamic constraints, and conflicting objectives.

A second category of surveys has emerged with the explicit aim of incorporating machine learning into MAPF. 
This line of work examines RL, imitation learning, and other learning paradigms geared toward approximate or decentralized controllers~\citep{chung2024learning,alkazzi2024comprehensive,yakovlev2022planning,yang2023path,zhou2023research}. 
While such surveys effectively showcase the flexibility and adaptability of data-driven approaches, they often concentrate on a narrower set of techniques—primarily modern deep RL or policy-based algorithms—and typically offer less extensive coverage of classical techniques for large-scale problems. 
Problem compilation surveys form a separate, more specialized niche. 
These works discuss how MAPF can be encoded into Boolean satisfiability, integer linear programs, constraint satisfaction, or SMT~\citep{surynek2022problem}, underscoring how such transformations leverage efficient solver to achieve high solution quality.

More recent surveys have begun to merge perspectives by discussing beyond-classical scenarios, such as handling heterogeneous agents, tasks with temporal or resource constraints, and real-time or partially observable systems~\citep{tjiharjadi2022systematic,ma2022graph,gao2024review,yang2023path}. 
Some of these reviews expand the scope to include multi-agent pickup and delivery, continuous-space motion planning, and swarm robotics~\citep{salzman2020research,yakovlev2022planning}, providing valuable insights on how MAPF can be adapted to more complex or realistic domains. 
Nonetheless, in many of these works, discussions of learning-based methods are presented independently from the traditional MAPF literature, offering limited guidance on how the two paradigms might reinforce each other.

In contrast to prior surveys, our paper offers a more unified treatment of MAPF. 
First, we bridge classical and learning-based approaches in a single cohesive framework, systematically examining the strengths and weaknesses of search-based and compilation-based methods alongside a spectrum of data-driven algorithms (including RL, imitation learning, evolutionary techniques, and large language model approaches). 
Second, we link the theoretical underpinnings of classical MAPF—e.g., guarantees on optimality and completeness—with the adaptivity afforded by learning-based approaches in uncertain and dynamic environments. 
This integration, which remains underexplored in previous reviews, underscores opportunities for synergy, such as replacing classical heuristic modules with learned policies or combining classical back-ends with representation learning for improved scalability. 
Third, unlike earlier works that frame empirical analyses in domain-specific terms, we provide a comparative study of experimental design across both classical and learning-based methods, shedding light on how the choice of map type, agent population, performance metrics, and baseline comparisons significantly affects reported outcomes.

Overall, we position our survey as an effort to build on and extend existing literature. 
We not only synthesize mature methods from decades of MAPF research but also draw attention to recent data-driven trends and novel hybrid solutions. 
In doing so, we aim to inspire fresh lines of inquiry that promote a holistic view of MAPF. 
By consolidating diverse approaches under a single umbrella and emphasizing rigorous benchmarking and real-world applicability, our survey advances the field beyond established boundaries, thereby offering practical guidance to both academic researchers and industrial practitioners.

\section{Conclusion}

This survey has provided a comprehensive examination of Multi-Agent Path Finding, systematically bridging the traditionally separate domains of classical algorithmic approaches and modern learning-based methods. 
Through our analysis, several key insights have emerged that shape our understanding of the current state and future trajectory of MAPF research.

\noindent\textbf{Synthesis of Key Findings.}
Our investigation reveals that MAPF has evolved far beyond its origins as a purely theoretical problem. 
Classical methods, exemplified by Conflict-Based Search and its variants, have achieved remarkable scalability—routinely handling thousands of agents while maintaining optimality guarantees. 
These approaches embody decades of algorithmic refinements, from sophisticated heuristics and symmetry reasoning to mutex propagation and disjoint splitting. 
Compilation-based methods have similarly matured, leveraging powerful general-purpose solvers to transform MAPF into well-studied formalisms. 
However, these classical approaches often struggle with real-time constraints, dynamic environments, and partial observability—precisely the scenarios where learning-based methods excel.

The emergence of data-driven approaches has introduced new possibilities for adaptive, robust multi-agent coordination. 
Reinforcement learning methods, particularly those employing multi-agent frameworks with communication protocols, have demonstrated the ability to discover emergent coordination strategies that classical planners might overlook. 
Yet our analysis also reveals a surprising paradox: 
despite claims of superior scalability, learning-based methods are predominantly evaluated on smaller problem instances than their classical counterparts, rarely exceeding a few hundred agents or modest grid sizes.

\noindent\textbf{The Promise of Integration.}
Perhaps the most significant contribution of this survey is highlighting the vast, underexplored potential for synergy between classical and learning paradigms. 
Learning-augmented classical solvers represent a particularly promising direction, where data-driven components enhance specific modules (conflict selection, node prioritization, neighborhood destruction) while preserving the theoretical scaffolding that ensures completeness and bounded suboptimality. 
This hybrid approach exemplifies a broader principle: 
rather than viewing classical and learning-based methods as competing paradigms, the field benefits most when they are seen as complementary tools in a unified toolkit.

\noindent\textbf{Methodological Implications.}
Our systematic analysis of experimental practices reveals critical gaps in current evaluation methodologies. 
The lack of standardized benchmarks, inconsistent metric reporting, and limited cross-paradigm comparisons hinder progress and make it difficult to assess the true capabilities of different approaches. 
We advocate for comprehensive evaluation protocols that span both theoretical guarantees and empirical performance, encompassing diverse environment types, varying agent densities, and realistic dynamic constraints. 
Only through such rigorous benchmarking can the community make informed decisions about which methods to deploy in specific application contexts.

\noindent\textbf{Future Outlook.} 
Looking ahead, MAPF research stands at an inflection point. 
The integration of large language models, generative approaches, and neural solver architectures promises to expand the boundaries of what is computationally feasible. 
Mixed-motive settings that incorporate game-theoretic considerations reflect the reality of multi-stakeholder systems. 
Collision-informed frameworks that embed safety constraints directly into neural architectures address critical deployment concerns. 
As MAPF applications extend to thousands or even millions of agents in smart cities and large-scale logistics networks, hierarchical and mean-field approaches will become essential.

\noindent\textbf{Final Remarks.} 
The journey from theoretical foundations to practical deployment of MAPF solutions exemplifies the evolution of AI research more broadly. 
What began as a discrete optimization problem on simple grids has expanded to encompass continuous domains, dynamic environments, heterogeneous agents, and complex real-world constraints. 
By providing this comprehensive survey that unifies classical and learning-based perspectives, we hope to accelerate progress toward robust, scalable, and intelligent multi-agent coordination systems. 
The future of MAPF lies not in choosing between classical rigor and learning-based flexibility, but in creatively combining their strengths to address the increasingly complex challenges of autonomous multi-agent systems in our interconnected world.


\bibliography{jair}

\begin{thebibliography}{207}
\providecommand{\natexlab}[1]{#1}
\providecommand{\url}[1]{\texttt{#1}}
\expandafter\ifx\csname urlstyle\endcsname\relax
  \providecommand{\doi}[1]{doi: #1}\else
  \providecommand{\doi}{doi: \begingroup \urlstyle{rm}\Url}\fi

\bibitem[Ach{\'a} et~al.(2021)Ach{\'a}, L{\'o}pez, Hagedorn, and Baier]{acha2021new}
Roberto~As{\'\i}n Ach{\'a}, Rodrigo L{\'o}pez, Sebasti{\'a}n Hagedorn, and Jorge~A Baier.
\newblock A new boolean encoding for {MAPF} and its performance with {ASP} and {MaxSAT} solvers.
\newblock In \emph{SoCS}, 2021.

\bibitem[Alkazzi \& Okumura(2024)Alkazzi and Okumura]{alkazzi2024comprehensive}
Jean-Marc Alkazzi and Keisuke Okumura.
\newblock A comprehensive review on leveraging machine learning for multi-agent path finding.
\newblock \emph{IEEE Access}, 12:\penalty0 57390--57409, 2024.

\bibitem[Alkazzi et~al.(2022)Alkazzi, Rizk, Salomon, and Makhoul]{alkazzi2022mapfaster}
Jean-Marc Alkazzi, Anthony Rizk, Michel Salomon, and Abdallah Makhoul.
\newblock Mapfaster: A faster and simpler take on multi-agent path finding algorithm selection.
\newblock In \emph{2022 IEEE/RSJ International Conference on Intelligent Robots and Systems (IROS)}, pp.\  10088--10093. IEEE, 2022.

\bibitem[Almadhoun et~al.(2019)Almadhoun, Taha, Seneviratne, and Zweiri]{almadhoun2019survey}
Randa Almadhoun, Tarek Taha, Lakmal Seneviratne, and Yahya Zweiri.
\newblock A survey on multi-robot coverage path planning for model reconstruction and mapping.
\newblock \emph{SN Applied Sciences}, 1:\penalty0 1--24, 2019.

\bibitem[Andreychuk et~al.(2022)Andreychuk, Yakovlev, Surynek, Atzmon, and Stern]{andreychuk2022multi}
Anton Andreychuk, Konstantin Yakovlev, Pavel Surynek, Dor Atzmon, and Roni Stern.
\newblock Multi-agent pathfinding with continuous time.
\newblock \emph{Artificial Intelligence}, 305:\penalty0 103662, 2022.

\bibitem[Andreychuk et~al.(2025)Andreychuk, Yakovlev, Panov, and Skrynnik]{andreychuk2025mapf}
Anton Andreychuk, Konstantin Yakovlev, Aleksandr Panov, and Alexey Skrynnik.
\newblock Mapf-gpt: Imitation learning for multi-agent pathfinding at scale.
\newblock In \emph{AAAI}, 2025.

\bibitem[Atasever et~al.(2025)Atasever, Kulkarni, Li, Hong, and Deshmukh]{atasever2025multiagent}
Merve Atasever, Mihir~Nitin Kulkarni, Qingpei Li, Matthew Hong, and Jyotirmoy~V. Deshmukh.
\newblock Multi-agent path finding via decision transformer and {LLM} collaboration, 2025.
\newblock URL \url{https://openreview.net/forum?id=Mvn48u0ehO}.

\bibitem[Barer et~al.(2014)Barer, Sharon, Stern, and Felner]{barer2014suboptimal}
Max Barer, Guni Sharon, Roni Stern, and Ariel Felner.
\newblock Suboptimal variants of the conflict-based search algorithm for the multi-agent pathfinding problem.
\newblock In \emph{SoCS}, 2014.

\bibitem[Bart{\'a}k \& {\v{S}}vancara(2019)Bart{\'a}k and {\v{S}}vancara]{bartak2019sat}
Roman Bart{\'a}k and Ji{\v{r}}{\'\i} {\v{S}}vancara.
\newblock On {SAT}-based approaches for multi-agent path finding with the sum-of-costs objective.
\newblock In \emph{SoCS}, 2019.

\bibitem[Bengio et~al.(2021)Bengio, Lodi, and Prouvost]{bengio2021machine}
Yoshua Bengio, Andrea Lodi, and Antoine Prouvost.
\newblock Machine learning for combinatorial optimization: a methodological tour d’horizon.
\newblock \emph{European Journal of Operational Research}, 290\penalty0 (2):\penalty0 405--421, 2021.

\bibitem[Bignoli(2021)]{bignoli2021graph}
Andrea Bignoli.
\newblock Graph attentional neural network in multi-agent pickup and delivery problems.
\newblock \emph{Unpublished Manuscript}, 2021.
\newblock URL \url{https://www.politesi.polimi.it/bitstream/10589/191682/4/Tesi_Bignoli.pdf}.
\newblock Accessed: 2024-12-03.

\bibitem[Boyarski et~al.(2015{\natexlab{a}})Boyarski, Felner, Sharon, and Stern]{boyarski2015don}
Eli Boyarski, Ariel Felner, Guni Sharon, and Roni Stern.
\newblock Don't split, try to work it out: Bypassing conflicts in multi-agent pathfinding.
\newblock In \emph{ICAPS}, 2015{\natexlab{a}}.

\bibitem[Boyarski et~al.(2015{\natexlab{b}})Boyarski, Felner, Stern, Sharon, Betzalel, Tolpin, and Shimony]{boyarski2015icbs}
Eli Boyarski, Ariel Felner, Roni Stern, Guni Sharon, Oded Betzalel, David Tolpin, and Eyal Shimony.
\newblock Icbs: The improved conflict-based search algorithm for multi-agent pathfinding.
\newblock In \emph{SoCS}, 2015{\natexlab{b}}.

\bibitem[Boyarski et~al.(2022)Boyarski, Chan, Atzmon, Felner, and Koenig]{boyarski2022merging}
Eli Boyarski, Shao-Hung Chan, Dor Atzmon, Ariel Felner, and Sven Koenig.
\newblock On merging agents in multi-agent pathfinding algorithms.
\newblock In \emph{SoCS}, 2022.

\bibitem[{\v{C}}apek \& Surynek(2021){\v{C}}apek and Surynek]{vcapek2021dpll}
Martin {\v{C}}apek and Pavel Surynek.
\newblock {DPLL(MAPF)}: an integration of multi-agent path finding and {SAT} solving technologies.
\newblock In \emph{SoCS}, 2021.

\bibitem[Chan et~al.(2022)Chan, Li, Gange, Harabor, Stuckey, and Koenig]{chan2022flex}
Shao-Hung Chan, Jiaoyang Li, Graeme Gange, Daniel Harabor, Peter~J Stuckey, and Sven Koenig.
\newblock Flex distribution for bounded-suboptimal multi-agent path finding.
\newblock In \emph{AAAI}, 2022.

\bibitem[Chen et~al.(2023{\natexlab{a}})Chen, Wang, Zhang, Ding, Zhong, Pu, and Zhang]{chen2023towards}
Feng Chen, Chenghe Wang, Fuxiang Zhang, Hao Ding, Qiaoyong Zhong, Shiliang Pu, and Zongzhang Zhang.
\newblock Towards deployment-efficient and collision-free multi-agent path finding (student abstract).
\newblock In \emph{AAAI}, 2023{\natexlab{a}}.

\bibitem[Chen et~al.(2023{\natexlab{b}})Chen, Wang, Miao, Mo, Feng, Zhou, and Wang]{chen2023transformer}
Lin Chen, Yaonan Wang, Zhiqiang Miao, Yang Mo, Mingtao Feng, Zhen Zhou, and Hesheng Wang.
\newblock Transformer-based imitative reinforcement learning for multirobot path planning.
\newblock \emph{IEEE Transactions on Industrial Informatics}, 19\penalty0 (10):\penalty0 10233--10243, 2023{\natexlab{b}}.

\bibitem[Chen et~al.(2022{\natexlab{a}})Chen, Chen, Chen, Heaton, Liu, Wang, and Yin]{chen2022learning}
Tianlong Chen, Xiaohan Chen, Wuyang Chen, Howard Heaton, Jialin Liu, Zhangyang Wang, and Wotao Yin.
\newblock Learning to optimize: A primer and a benchmark.
\newblock \emph{Journal of Machine Learning Research}, 23\penalty0 (189):\penalty0 1--59, 2022{\natexlab{a}}.

\bibitem[Chen et~al.(2024)Chen, Koenig, and Dilkina]{chen2024solving}
Weizhe Chen, Sven Koenig, and Bistra Dilkina.
\newblock Why solving multi-agent path finding with large language model has not succeeded yet.
\newblock \emph{arXiv preprint arXiv:2401.03630}, 2024.

\bibitem[Chen et~al.(2022{\natexlab{b}})Chen, Li, Harabor, Stuckey, and Koenig]{chen2022multi}
Zhe Chen, Jiaoyang Li, Daniel Harabor, Peter~J Stuckey, and Sven Koenig.
\newblock Multi-train path finding revisited.
\newblock In \emph{SoCS}, 2022{\natexlab{b}}.

\bibitem[Christopher et~al.(2024)Christopher, Baek, and Fioretto]{christopher2024constrained}
Jacob~K Christopher, Stephen Baek, and Nando Fioretto.
\newblock Constrained synthesis with projected diffusion models.
\newblock In \emph{NeurIPS}, 2024.

\bibitem[Chung et~al.(2024)Chung, Fayyad, Younes, and Najjaran]{chung2024learning}
Jaehoon Chung, Jamil Fayyad, Younes~Al Younes, and Homayoun Najjaran.
\newblock Learning team-based navigation: a review of deep reinforcement learning techniques for multi-agent pathfinding.
\newblock \emph{Artificial Intelligence Review}, 57\penalty0 (2):\penalty0 41, 2024.

\bibitem[Corsi et~al.(2021)Corsi, Marchesini, and Farinelli]{corsi2021formal}
Davide Corsi, Enrico Marchesini, and Alessandro Farinelli.
\newblock Formal verification of neural networks for safety-critical tasks in deep reinforcement learning.
\newblock In \emph{UAI}. PMLR, 2021.

\bibitem[Damani et~al.(2021)Damani, Luo, Wenzel, and Sartoretti]{damani2021primal}
Mehul Damani, Zhiyao Luo, Emerson Wenzel, and Guillaume Sartoretti.
\newblock Primal$\_2$: Pathfinding via reinforcement and imitation multi-agent learning-lifelong.
\newblock \emph{IEEE Robotics and Automation Letters}, 6\penalty0 (2):\penalty0 2666--2673, 2021.

\bibitem[De~Bortoli et~al.(2021)De~Bortoli, Thornton, Heng, and Doucet]{de2021diffusion}
Valentin De~Bortoli, James Thornton, Jeremy Heng, and Arnaud Doucet.
\newblock Diffusion schr{\"o}dinger bridge with applications to score-based generative modeling.
\newblock In \emph{NeurIPS}, 2021.

\bibitem[Dresner \& Stone(2004)Dresner and Stone]{dresner2004multiagent}
Kurt Dresner and Peter Stone.
\newblock Multiagent traffic management: A reservation-based intersection control mechanism.
\newblock In \emph{Autonomous Agents and Multiagent Systems, International Joint Conference on}, volume~3, pp.\  530--537. Citeseer, 2004.

\bibitem[Dughmi \& Xu(2016)Dughmi and Xu]{dughmi2016algorithmic}
Shaddin Dughmi and Haifeng Xu.
\newblock Algorithmic bayesian persuasion.
\newblock In \emph{STOC}, 2016.

\bibitem[Ehlers(2017)]{ehlers2017formal}
Ruediger Ehlers.
\newblock Formal verification of piece-wise linear feed-forward neural networks.
\newblock In \emph{ATVA 2017}, 2017.

\bibitem[Eleftheriadis et~al.(2022)Eleftheriadis, Kekatos, Katsaros, and Tripakis]{eleftheriadis2022neural}
Charis Eleftheriadis, Nikolaos Kekatos, Panagiotis Katsaros, and Stavros Tripakis.
\newblock On neural network equivalence checking using smt solvers.
\newblock In \emph{FORMATS}, 2022.

\bibitem[Elsken et~al.(2019)Elsken, Metzen, and Hutter]{elsken2019neural}
Thomas Elsken, Jan~Hendrik Metzen, and Frank Hutter.
\newblock Neural architecture search: A survey.
\newblock \emph{Journal of Machine Learning Research}, 20\penalty0 (55):\penalty0 1--21, 2019.

\bibitem[Erdmann \& Lozano-Perez(1987)Erdmann and Lozano-Perez]{erdmann1987multiple}
Michael Erdmann and Tomas Lozano-Perez.
\newblock On multiple moving objects.
\newblock \emph{Algorithmica}, 2:\penalty0 477--521, 1987.

\bibitem[Fan et~al.(2022)Fan, Wu, Liao, Cao, Guo, Sartoretti, and Wu]{fan2022deep}
Mingfeng Fan, Yaoxin Wu, Tianjun Liao, Zhiguang Cao, Hongliang Guo, Guillaume Sartoretti, and Guohua Wu.
\newblock Deep reinforcement learning for {UAV} routing in the presence of multiple charging stations.
\newblock \emph{IEEE Transactions on Vehicular Technology}, 72\penalty0 (5):\penalty0 5732--5746, 2022.

\bibitem[Fan et~al.(2020)Fan, Long, Liu, and Pan]{fan2020distributed}
Tingxiang Fan, Pinxin Long, Wenxi Liu, and Jia Pan.
\newblock Distributed multi-robot collision avoidance via deep reinforcement learning for navigation in complex scenarios.
\newblock \emph{The International Journal of Robotics Research}, 39\penalty0 (7):\penalty0 856--892, 2020.

\bibitem[Felner et~al.(2012)Felner, Goldenberg, Sharon, Stern, Beja, Sturtevant, Schaeffer, and Holte]{felner2012partial}
Ariel Felner, Meir Goldenberg, Guni Sharon, Roni Stern, Tal Beja, Nathan Sturtevant, Jonathan Schaeffer, and Robert Holte.
\newblock Partial-expansion a* with selective node generation.
\newblock In \emph{Proceedings of the AAAI Conference on Artificial Intelligence}, volume~26, pp.\  471--477, 2012.

\bibitem[Felner et~al.(2018)Felner, Li, Boyarski, Ma, Cohen, Kumar, and Koenig]{felner2018adding}
Ariel Felner, Jiaoyang Li, Eli Boyarski, Hang Ma, Liron Cohen, TK~Satish Kumar, and Sven Koenig.
\newblock Adding heuristics to conflict-based search for multi-agent path finding.
\newblock In \emph{ICAPS}, 2018.

\bibitem[Feng et~al.(2024)Feng, Li, and Lu]{feng2024multi}
Yi~Feng, Cheng Li, and Yun Lu.
\newblock Multi-agent path finding in dynamic environment guided by historical experience.
\newblock In \emph{CCC}, 2024.

\bibitem[Ferner et~al.(2013)Ferner, Wagner, and Choset]{ferner2013odrm}
Cornelia Ferner, Glenn Wagner, and Howie Choset.
\newblock Odrm* optimal multirobot path planning in low dimensional search spaces.
\newblock In \emph{2013 IEEE international conference on robotics and automation}, pp.\  3854--3859. IEEE, 2013.

\bibitem[Finn et~al.(2017)Finn, Abbeel, and Levine]{finn2017model}
Chelsea Finn, Pieter Abbeel, and Sergey Levine.
\newblock Model-agnostic meta-learning for fast adaptation of deep networks.
\newblock In \emph{ICML}, 2017.

\bibitem[Friedrich et~al.(2024)Friedrich, Zhang, Curry, Dierks, McAleer, Li, Sandholm, and Seuken]{friedrich2024scalable}
Paul Friedrich, Yulun Zhang, Michael Curry, Ludwig Dierks, Stephen McAleer, Jiaoyang Li, Tuomas Sandholm, and Sven Seuken.
\newblock Scalable mechanism design for multi-agent path finding.
\newblock \emph{arXiv preprint arXiv:2401.17044}, 2024.

\bibitem[Gange et~al.(2019)Gange, Harabor, and Stuckey]{gange2019lazy}
Graeme Gange, Daniel Harabor, and Peter~J Stuckey.
\newblock Lazy cbs: implicit conflict-based search using lazy clause generation.
\newblock In \emph{Proceedings of the international conference on automated planning and scheduling}, volume~29, pp.\  155--162, 2019.

\bibitem[Gao et~al.(2023)Gao, Ye, Li, and Li]{gao2023hglp}
Jianqi Gao, Zhaohui Ye, Xinyi Li, and Yanjie Li.
\newblock {HGLP}: Hierarchical solver for combined task assignment and path finding.
\newblock In \emph{CCDC}, 2023.

\bibitem[Gao et~al.(2024{\natexlab{a}})Gao, Li, Li, Yan, Lin, and Wu]{gao2024review}
Jianqi Gao, Yanjie Li, Xinyi Li, Kejian Yan, Ke~Lin, and Xinyu Wu.
\newblock A review of graph-based multi-agent pathfinding solvers: From classical to beyond classical.
\newblock \emph{Knowledge-Based Systems}, 283:\penalty0 111121, 2024{\natexlab{a}}.

\bibitem[Gao et~al.(2024{\natexlab{b}})Gao, Li, Ye, and Wu]{gao2024pce}
Jianqi Gao, Yanjie Li, Zhaohui Ye, and Xinyu Wu.
\newblock {PCE}: Multi-agent path finding via priority-aware communication \& experience learning.
\newblock \emph{IEEE Transactions on Intelligent Vehicles}, 2024{\natexlab{b}}.

\bibitem[Gerken et~al.(2023)Gerken, Aronsson, Carlsson, Linander, Ohlsson, Petersson, and Persson]{gerken2023geometric}
Jan~E Gerken, Jimmy Aronsson, Oscar Carlsson, Hampus Linander, Fredrik Ohlsson, Christoffer Petersson, and Daniel Persson.
\newblock Geometric deep learning and equivariant neural networks.
\newblock \emph{Artificial Intelligence Review}, 56\penalty0 (12):\penalty0 14605--14662, 2023.

\bibitem[G{\'o}mez et~al.(2020)G{\'o}mez, Hern{\'a}ndez, and Baier]{gomez2020solving}
Rodrigo~N G{\'o}mez, Carlos Hern{\'a}ndez, and Jorge~A Baier.
\newblock Solving sum-of-costs multi-agent pathfinding with answer-set programming.
\newblock In \emph{AAAI}, 2020.

\bibitem[G{\'o}mez et~al.(2021)G{\'o}mez, Hern{\'a}ndez, and Baier]{gomez2021compact}
Rodrigo~N G{\'o}mez, Carlos Hern{\'a}ndez, and Jorge~A Baier.
\newblock A compact answer set programming encoding of multi-agent pathfinding.
\newblock \emph{IEEE Access}, 9:\penalty0 26886--26901, 2021.

\bibitem[Guan et~al.(2022)Guan, Gao, Zhao, Yang, Deng, and Lam]{guan2022ab}
Huifeng Guan, Yuan Gao, Min Zhao, Yong Yang, Fuqin Deng, and Tin~Lun Lam.
\newblock Ab-mapper: Attention and bicnet based multi-agent path planning for dynamic environment.
\newblock In \emph{IROS}, 2022.

\bibitem[He et~al.(2024{\natexlab{a}})He, Duhan, Tulsyan, Kim, and Sartoretti]{he2024social}
Chengyang He, Tanishq Duhan, Parth Tulsyan, Patrick Kim, and Guillaume Sartoretti.
\newblock Social behavior as a key to learning-based multi-agent pathfinding dilemmas.
\newblock \emph{arXiv preprint arXiv:2408.03063}, 2024{\natexlab{a}}.

\bibitem[He et~al.(2024{\natexlab{b}})He, Yang, Duhan, Wang, and Sartoretti]{he2024alpha}
Chengyang He, Tianze Yang, Tanishq Duhan, Yutong Wang, and Guillaume Sartoretti.
\newblock Alpha: Attention-based long-horizon pathfinding in highly-structured areas.
\newblock In \emph{ICRA}, 2024{\natexlab{b}}.

\bibitem[Ho et~al.(2020{\natexlab{a}})Ho, Geraldes, Gon{\c{c}}alves, Rigault, Sportich, Kubo, Cavazza, and Prendinger]{ho2020decentralized}
Florence Ho, R{\'u}ben Geraldes, Artur Gon{\c{c}}alves, Bastien Rigault, Benjamin Sportich, Daisuke Kubo, Marc Cavazza, and Helmut Prendinger.
\newblock Decentralized multi-agent path finding for uav traffic management.
\newblock \emph{IEEE Transactions on Intelligent Transportation Systems}, 23\penalty0 (2):\penalty0 997--1008, 2020{\natexlab{a}}.

\bibitem[Ho et~al.(2020{\natexlab{b}})Ho, Jain, and Abbeel]{ho2020denoising}
Jonathan Ho, Ajay Jain, and Pieter Abbeel.
\newblock Denoising diffusion probabilistic models.
\newblock In \emph{NeurIPS}, 2020{\natexlab{b}}.

\bibitem[Huang et~al.(2021{\natexlab{a}})Huang, Dilkina, and Koenig]{huang2021learning-aamas}
Taoan Huang, Bistra Dilkina, and Sven Koenig.
\newblock Learning node-selection strategies in bounded suboptimal conflict-based search for multi-agent path finding.
\newblock In \emph{AAMAS}, 2021{\natexlab{a}}.

\bibitem[Huang et~al.(2021{\natexlab{b}})Huang, Koenig, and Dilkina]{huang2021learning}
Taoan Huang, Sven Koenig, and Bistra Dilkina.
\newblock Learning to resolve conflicts for multi-agent path finding with conflict-based search.
\newblock In \emph{AAAI}, 2021{\natexlab{b}}.

\bibitem[Huang et~al.(2022)Huang, Li, Koenig, and Dilkina]{huang2022anytime}
Taoan Huang, Jiaoyang Li, Sven Koenig, and Bistra Dilkina.
\newblock Anytime multi-agent path finding via machine learning-guided large neighborhood search.
\newblock In \emph{AAAI}, 2022.

\bibitem[Katz et~al.(2017)Katz, Barrett, Dill, Julian, and Kochenderfer]{katz2017reluplex}
Guy Katz, Clark Barrett, David~L Dill, Kyle Julian, and Mykel~J Kochenderfer.
\newblock Reluplex: An efficient smt solver for verifying deep neural networks.
\newblock In \emph{CAV}, 2017.

\bibitem[Kollock(1998)]{kollock1998social}
Peter Kollock.
\newblock Social dilemmas: The anatomy of cooperation.
\newblock \emph{Annual review of sociology}, 24\penalty0 (1):\penalty0 183--214, 1998.

\bibitem[Kong et~al.(2024)Kong, Zhou, Li, and Wang]{kong2024multi}
Xiaoran Kong, Yatong Zhou, Zhe Li, and Shaohai Wang.
\newblock Multi-{UAV} simultaneous target assignment and path planning based on deep reinforcement learning in dynamic multiple obstacles environments.
\newblock \emph{Frontiers in Neurorobotics}, 17:\penalty0 1302898, 2024.

\bibitem[Lam et~al.(2022)Lam, Le~Bodic, Harabor, and Stuckey]{lam2022branch}
Edward Lam, Pierre Le~Bodic, Daniel Harabor, and Peter~J Stuckey.
\newblock Branch-and-cut-and-price for multi-agent path finding.
\newblock \emph{Computers \& Operations Research}, 144:\penalty0 105809, 2022.

\bibitem[Lam et~al.(2023)Lam, Harabor, Stuckey, and Li]{lam2023exact}
Edward Lam, Daniel~D Harabor, Peter~J Stuckey, and Jiaoyang Li.
\newblock Exact anytime multi-agent path finding using branch-and-cut-and-price and large neighborhood search.
\newblock In \emph{ICAPS}, 2023.

\bibitem[Lee et~al.(2021)Lee, Motes, Morales, and Amato]{lee2021parallel}
Hannah Lee, James Motes, Marco Morales, and Nancy~M Amato.
\newblock Parallel hierarchical composition conflict-based search for optimal multi-agent pathfinding.
\newblock \emph{IEEE Robotics and Automation Letters}, 6\penalty0 (4):\penalty0 7001--7008, 2021.

\bibitem[Leet et~al.(2024)Leet, Morris, and Pradeep]{leet2024safe}
Christopher~J Leet, Robert Morris, and Priyank Pradeep.
\newblock Safe, efficient, and fair utm airspace management.
\newblock In \emph{DASC}, 2024.

\bibitem[Li et~al.(2025)Li, Xu, Sun, Zhou, and Wen]{li2025formalization}
Chenyi Li, Shengyang Xu, Chumin Sun, Li~Zhou, and Zaiwen Wen.
\newblock Formalization of optimality conditions for smooth constrained optimization problems.
\newblock \emph{arXiv preprint arXiv:2503.18821}, 2025.

\bibitem[Li et~al.(2019{\natexlab{a}})Li, Felner, Boyarski, Ma, and Koenig]{li2019improved}
Jiaoyang Li, Ariel Felner, Eli Boyarski, Hang Ma, and Sven Koenig.
\newblock Improved heuristics for multi-agent path finding with conflict-based search.
\newblock In \emph{IJCAI}, 2019{\natexlab{a}}.

\bibitem[Li et~al.(2019{\natexlab{b}})Li, Harabor, Stuckey, Felner, Ma, and Koenig]{li2019disjoint}
Jiaoyang Li, Daniel Harabor, Peter~J Stuckey, Ariel Felner, Hang Ma, and Sven Koenig.
\newblock Disjoint splitting for multi-agent path finding with conflict-based search.
\newblock In \emph{ICAPS}, 2019{\natexlab{b}}.

\bibitem[Li et~al.(2019{\natexlab{c}})Li, Harabor, Stuckey, Ma, and Koenig]{li2019symmetry}
Jiaoyang Li, Daniel Harabor, Peter~J Stuckey, Hang Ma, and Sven Koenig.
\newblock Symmetry-breaking constraints for grid-based multi-agent path finding.
\newblock In \emph{AAAI}, 2019{\natexlab{c}}.

\bibitem[Li et~al.(2020)Li, Gange, Harabor, Stuckey, Ma, and Koenig]{li2020new}
Jiaoyang Li, Graeme Gange, Daniel Harabor, Peter~J Stuckey, Hang Ma, and Sven Koenig.
\newblock New techniques for pairwise symmetry breaking in multi-agent path finding.
\newblock In \emph{ICAPS}, 2020.

\bibitem[Li et~al.(2021{\natexlab{a}})Li, Chen, Harabor, Stuckey, and Koenig]{li2021anytime}
Jiaoyang Li, Zhe Chen, Daniel Harabor, Peter~J Stuckey, and Sven Koenig.
\newblock Anytime multi-agent path finding via large neighborhood search.
\newblock In \emph{IJCAI}, 2021{\natexlab{a}}.

\bibitem[Li et~al.(2021{\natexlab{b}})Li, Chen, Zheng, Chan, Harabor, Stuckey, Ma, and Koenig]{li2021scalable}
Jiaoyang Li, Zhe Chen, Yi~Zheng, Shao-Hung Chan, Daniel Harabor, Peter~J Stuckey, Hang Ma, and Sven Koenig.
\newblock Scalable rail planning and replanning: Winning the 2020 flatland challenge.
\newblock In \emph{ICAPS}, 2021{\natexlab{b}}.

\bibitem[Li et~al.(2021{\natexlab{c}})Li, Harabor, Stuckey, Ma, Gange, and Koenig]{li2021pairwise}
Jiaoyang Li, Daniel Harabor, Peter~J Stuckey, Hang Ma, Graeme Gange, and Sven Koenig.
\newblock Pairwise symmetry reasoning for multi-agent path finding search.
\newblock \emph{Artificial Intelligence}, 301:\penalty0 103574, 2021{\natexlab{c}}.

\bibitem[Li et~al.(2021{\natexlab{d}})Li, Ruml, and Koenig]{li2021eecbs}
Jiaoyang Li, Wheeler Ruml, and Sven Koenig.
\newblock {EECBS}: A bounded-suboptimal search for multi-agent path finding.
\newblock In \emph{AAAI}, 2021{\natexlab{d}}.

\bibitem[Li et~al.(2022{\natexlab{a}})Li, Chen, Harabor, Stuckey, and Koenig]{li2022mapf}
Jiaoyang Li, Zhe Chen, Daniel Harabor, Peter~J Stuckey, and Sven Koenig.
\newblock Mapf-lns2: Fast repairing for multi-agent path finding via large neighborhood search.
\newblock In \emph{AAAI}, 2022{\natexlab{a}}.

\bibitem[Li et~al.(2021{\natexlab{e}})Li, Lin, Liu, and Prorok]{li2021message}
Qingbiao Li, Weizhe Lin, Zhe Liu, and Amanda Prorok.
\newblock Message-aware graph attention networks for large-scale multi-robot path planning.
\newblock \emph{IEEE Robotics and Automation Letters}, 6\penalty0 (3):\penalty0 5533--5540, 2021{\natexlab{e}}.

\bibitem[Li et~al.(2022{\natexlab{b}})Li, Chen, Jin, Tan, Zha, and Wang]{li2022multi}
Wenhao Li, Hongjun Chen, Bo~Jin, Wenzhe Tan, Hongyuan Zha, and Xiangfeng Wang.
\newblock Multi-agent path finding with prioritized communication learning.
\newblock In \emph{ICRA}, 2022{\natexlab{b}}.

\bibitem[Li et~al.(2023)Li, Gao, and Li]{li2023multi}
Xinyi Li, Jianqi Gao, and Yanjie Li.
\newblock Multi-agent path finding based on graph neural network.
\newblock In \emph{CCC}, 2023.

\bibitem[Liang et~al.(2024)Liang, Christopher, Koenig, and Fioretto]{liang2024multi}
Jinhao Liang, Jacob~K Christopher, Sven Koenig, and Ferdinando Fioretto.
\newblock Multi-agent path finding in continuous spaces with projected diffusion models.
\newblock \emph{arXiv preprint arXiv:2412.17993}, 2024.

\bibitem[Lin \& Ma(2023)Lin and Ma]{lin2023sacha}
Qiushi Lin and Hang Ma.
\newblock {SACHA}: Soft actor-critic with heuristic-based attention for partially observable multi-agent path finding.
\newblock \emph{IEEE Robotics and Automation Letters}, 8\penalty0 (8):\penalty0 5100--5107, 2023.

\bibitem[Lin et~al.(2023)Lin, Li, Zha, and Wang]{lin2023information}
Yue Lin, Wenhao Li, Hongyuan Zha, and Baoxiang Wang.
\newblock Information design in multi-agent reinforcement learning.
\newblock In \emph{NeurIPS}, 2023.

\bibitem[Liu et~al.(2024{\natexlab{a}})Liu, Xialiang, Yuan, Lin, Luo, Wang, Lu, and Zhang]{liu2024evolution}
Fei Liu, Tong Xialiang, Mingxuan Yuan, Xi~Lin, Fu~Luo, Zhenkun Wang, Zhichao Lu, and Qingfu Zhang.
\newblock Evolution of heuristics: Towards efficient automatic algorithm design using large language model.
\newblock In \emph{ICML}, 2024{\natexlab{a}}.

\bibitem[Liu et~al.(2024{\natexlab{b}})Liu, Gao, Zhu, Qiao, Chen, Guo, and Li]{liu2024multi-agent}
Qi~Liu, Jianqi Gao, Dongjie Zhu, Zhongjian Qiao, Pengbin Chen, Jingxiang Guo, and Yanjie Li.
\newblock Multi-agent target assignment and path finding for intelligent warehouse: A cooperative multi-agent deep reinforcement learning perspective.
\newblock \emph{arXiv:2408.13750}, 2024{\natexlab{b}}.

\bibitem[Liu et~al.(2024{\natexlab{c}})Liu, Wang, and Liu]{liu2024multi}
Xi~Liu, Zhonghua Wang, and Shun Liu.
\newblock Multi robot path planning based on reinforcement learning.
\newblock In \emph{IMCEC}, 2024{\natexlab{c}}.

\bibitem[Liu et~al.(2024{\natexlab{d}})Liu, Ni, and Qureshi]{liu2024physics}
Yuchen Liu, Ruiqi Ni, and Ahmed~H Qureshi.
\newblock Physics-informed neural mapping and motion planning in unknown environments.
\newblock \emph{arXiv preprint arXiv:2410.09883}, 2024{\natexlab{d}}.

\bibitem[Liu et~al.(2020)Liu, Chen, Zhou, Koushik, Hebert, and Zhao]{liu2020mapper}
Zuxin Liu, Baiming Chen, Hongyi Zhou, Guru Koushik, Martial Hebert, and Ding Zhao.
\newblock Mapper: Multi-agent path planning with evolutionary reinforcement learning in mixed dynamic environments.
\newblock In \emph{IROS}, 2020.

\bibitem[Ma(2022)]{ma2022graph}
Hang Ma.
\newblock Graph-based multi-robot path finding and planning.
\newblock \emph{Current Robotics Reports}, 3\penalty0 (3):\penalty0 77--84, 2022.

\bibitem[Ma et~al.(2017{\natexlab{a}})Ma, Li, Kumar, and Koenig]{ma2017lifelong}
Hang Ma, Jiaoyang Li, TK~Kumar, and Sven Koenig.
\newblock Lifelong multi-agent path finding for online pickup and delivery tasks.
\newblock \emph{arXiv:1705.10868}, 2017{\natexlab{a}}.

\bibitem[Ma et~al.(2017{\natexlab{b}})Ma, Yang, Cohen, Kumar, and Koenig]{ma2017feasibility}
Hang Ma, Jingxing Yang, Liron Cohen, TK~Kumar, and Sven Koenig.
\newblock Feasibility study: Moving non-homogeneous teams in congested video game environments.
\newblock In \emph{AIIDE}, 2017{\natexlab{b}}.

\bibitem[Ma et~al.(2019)Ma, Harabor, Stuckey, Li, and Koenig]{ma2019searching}
Hang Ma, Daniel Harabor, Peter~J Stuckey, Jiaoyang Li, and Sven Koenig.
\newblock Searching with consistent prioritization for multi-agent path finding.
\newblock In \emph{AAAI}, 2019.

\bibitem[Ma et~al.(2021{\natexlab{a}})Ma, Luo, and Ma]{ma2021distributed}
Ziyuan Ma, Yudong Luo, and Hang Ma.
\newblock Distributed heuristic multi-agent path finding with communication.
\newblock In \emph{ICRA}, 2021{\natexlab{a}}.

\bibitem[Ma et~al.(2021{\natexlab{b}})Ma, Luo, and Pan]{ma2021learning}
Ziyuan Ma, Yudong Luo, and Jia Pan.
\newblock Learning selective communication for multi-agent path finding.
\newblock \emph{IEEE Robotics and Automation Letters}, 7\penalty0 (2):\penalty0 1455--1462, 2021{\natexlab{b}}.

\bibitem[Martin \& Furieri(2024)Martin and Furieri]{martin2024learning}
Andrea Martin and Luca Furieri.
\newblock Learning to optimize with convergence guarantees using nonlinear system theory.
\newblock \emph{IEEE Control Systems Letters}, 2024.

\bibitem[Matsui(2023)]{matsui2023investigation}
Toshihiro Matsui.
\newblock Investigation of integrating solution techniques for lifelong {MAPD} problem considering endpoints.
\newblock In \emph{PAAMS}, 2023.

\bibitem[Mazyavkina et~al.(2021)Mazyavkina, Sviridov, Ivanov, and Burnaev]{mazyavkina2021reinforcement}
Nina Mazyavkina, Sergey Sviridov, Sergei Ivanov, and Evgeny Burnaev.
\newblock Reinforcement learning for combinatorial optimization: A survey.
\newblock \emph{Computers \& Operations Research}, 134:\penalty0 105400, 2021.

\bibitem[Mohanty et~al.(2020)Mohanty, Nygren, Laurent, Schneider, Scheller, Bhattacharya, Watson, Egli, Eichenberger, Baumberger, et~al.]{mohanty2020flatland}
Sharada Mohanty, Erik Nygren, Florian Laurent, Manuel Schneider, Christian Scheller, Nilabha Bhattacharya, Jeremy Watson, Adrian Egli, Christian Eichenberger, Christian Baumberger, et~al.
\newblock Flatland-{RL}: Multi-agent reinforcement learning on trains.
\newblock \emph{arXiv:2012.05893}, 2020.

\bibitem[Morris et~al.(2016)Morris, Pasareanu, Luckow, Malik, Ma, Kumar, and Koenig]{morris2016planning}
Robert Morris, Corina~S Pasareanu, Kasper Luckow, Waqar Malik, Hang Ma, TK~Satish Kumar, and Sven Koenig.
\newblock Planning, scheduling and monitoring for airport surface operations.
\newblock In \emph{Workshops at the AAAI}, 2016.

\bibitem[Moura \& Ullrich(2021)Moura and Ullrich]{moura2021lean}
Leonardo~de Moura and Sebastian Ullrich.
\newblock The lean 4 theorem prover and programming language.
\newblock In \emph{CADE}. Springer, 2021.

\bibitem[Okoso et~al.(2019)Okoso, Otaki, and Nishi]{okoso2019multi}
Ayano Okoso, Keisuke Otaki, and Tomoki Nishi.
\newblock Multi-agent path finding with priority for cooperative automated valet parking.
\newblock In \emph{ITSC}, 2019.

\bibitem[Okoso et~al.(2022)Okoso, Otaki, Koide, and Nishi]{okoso2022high}
Ayano Okoso, Keisuke Otaki, Satoshi Koide, and Tomoki Nishi.
\newblock High density automated valet parking via multi-agent path finding.
\newblock In \emph{ITSC}, 2022.

\bibitem[Okumura(2023{\natexlab{a}})]{okumura2023improving}
Keisuke Okumura.
\newblock Improving lacam for scalable eventually optimal multi-agent pathfinding.
\newblock \emph{arXiv:2305.03632}, 2023{\natexlab{a}}.

\bibitem[Okumura(2023{\natexlab{b}})]{okumura2023lacam}
Keisuke Okumura.
\newblock {LaCAM}: Search-based algorithm for quick multi-agent pathfinding.
\newblock In \emph{AAAI}, 2023{\natexlab{b}}.

\bibitem[Okumura(2024)]{okumura2024engineering}
Keisuke Okumura.
\newblock Engineering {LaCAM*}: Towards real-time, large-scale, and near-optimal multi-agent pathfinding.
\newblock In \emph{AAMAS}, 2024.

\bibitem[Okumura \& Tixeuil(2023)Okumura and Tixeuil]{okumura2023fault}
Keisuke Okumura and S{\'e}bastien Tixeuil.
\newblock Fault-tolerant offline multi-agent path planning.
\newblock In \emph{AAAI}, 2023.

\bibitem[Okumura et~al.(2019)Okumura, Tamura, and D{\'e}fago]{okumura2019winpibt}
Keisuke Okumura, Yasumasa Tamura, and Xavier D{\'e}fago.
\newblock winpibt: Extended prioritized algorithm for iterative multi-agent path finding.
\newblock \emph{arXiv preprint arXiv:1905.10149}, 2019.

\bibitem[Okumura et~al.(2021{\natexlab{a}})Okumura, Bonnet, Tamura, and D{\'e}fago]{okumura2021offline}
Keisuke Okumura, Fran{\c{c}}ois Bonnet, Yasumasa Tamura, and Xavier D{\'e}fago.
\newblock Offline time-independent multi-agent path planning.
\newblock \emph{arXiv:2105.07132}, 2021{\natexlab{a}}.

\bibitem[Okumura et~al.(2021{\natexlab{b}})Okumura, Tamura, and D{\'e}fago]{okumura2021iterative}
Keisuke Okumura, Yasumasa Tamura, and Xavier D{\'e}fago.
\newblock Iterative refinement for real-time multi-robot path planning.
\newblock In \emph{IROS}, 2021{\natexlab{b}}.

\bibitem[Okumura et~al.(2021{\natexlab{c}})Okumura, Tamura, and D{\'e}fago]{okumura2021time}
Keisuke Okumura, Yasumasa Tamura, and Xavier D{\'e}fago.
\newblock Time-independent planning for multiple moving agents.
\newblock In \emph{AAAI}, 2021{\natexlab{c}}.

\bibitem[Okumura et~al.(2022{\natexlab{a}})Okumura, Machida, D{\'e}fago, and Tamura]{okumura2022priority}
Keisuke Okumura, Manao Machida, Xavier D{\'e}fago, and Yasumasa Tamura.
\newblock Priority inheritance with backtracking for iterative multi-agent path finding.
\newblock \emph{Artificial Intelligence}, 310:\penalty0 103752, 2022{\natexlab{a}}.

\bibitem[Okumura et~al.(2022{\natexlab{b}})Okumura, Yonetani, Nishimura, and Kanezaki]{okumura2022ctrms}
Keisuke Okumura, Ryo Yonetani, Mai Nishimura, and Asako Kanezaki.
\newblock {CTRMs}: Learning to construct cooperative timed roadmaps for multi-agent path planning in continuous spaces.
\newblock \emph{arXiv:2201.09467}, 2022{\natexlab{b}}.

\bibitem[Ou et~al.(2024)Ou, Luo, Xu, Feng, and Zhao]{ou2024reinforcement}
Wen Ou, Biao Luo, Xiaodong Xu, Yu~Feng, and Yuqian Zhao.
\newblock Reinforcement learned multi--agent cooperative navigation in hybrid environment with relational graph learning.
\newblock \emph{IEEE Transactions on Artificial Intelligence}, 6\penalty0 (1):\penalty0 25--36, 2024.

\bibitem[Paolo et~al.(2024)Paolo, Gonzalez-Billandon, and K{\'e}gl]{paoloposition}
Giuseppe Paolo, Jonas Gonzalez-Billandon, and Bal{\'a}zs K{\'e}gl.
\newblock Position: A call for embodied ai.
\newblock In \emph{ICML}, 2024.

\bibitem[Papamakarios et~al.(2021)Papamakarios, Nalisnick, Rezende, Mohamed, and Lakshminarayanan]{papamakarios2021normalizing}
George Papamakarios, Eric Nalisnick, Danilo~Jimenez Rezende, Shakir Mohamed, and Balaji Lakshminarayanan.
\newblock Normalizing flows for probabilistic modeling and inference.
\newblock \emph{Journal of Machine Learning Research}, 22\penalty0 (57):\penalty0 1--64, 2021.

\bibitem[Park et~al.(2024)Park, Kim, and Alaa]{park2024mean}
Sungwoo Park, Dongjun Kim, and Ahmed Alaa.
\newblock Mean-field chaos diffusion models.
\newblock In \emph{ICML}, 2024.

\bibitem[Paul \& Deshmukh(2022)Paul and Deshmukh]{paul2022multi}
Sheryl Paul and Jyotirmoy~V Deshmukh.
\newblock Multi agent path finding using evolutionary game theory.
\newblock \emph{arXiv:2212.02010}, 2022.

\bibitem[Pham \& Bera(2024)Pham and Bera]{Pham2023OptimizingCM}
Phu Pham and Aniket Bera.
\newblock Optimizing crowd-aware multi-agent path finding through local communication with graph neural networks.
\newblock In \emph{IROS}, 2024.

\bibitem[Phan et~al.(2024{\natexlab{a}})Phan, Driscoll, Romberg, and Koenig]{phan2024confidence}
Thomy Phan, Joseph Driscoll, Justin Romberg, and Sven Koenig.
\newblock Confidence-based curriculum learning for multi-agent path finding.
\newblock \emph{arXiv preprint arXiv:2401.05860}, 2024{\natexlab{a}}.

\bibitem[Phan et~al.(2024{\natexlab{b}})Phan, Huang, Dilkina, and Koenig]{phan2024adaptive}
Thomy Phan, Taoan Huang, Bistra Dilkina, and Sven Koenig.
\newblock Adaptive anytime multi-agent path finding using bandit-based large neighborhood search.
\newblock In \emph{AAAI}, 2024{\natexlab{b}}.

\bibitem[Phillips \& Likhachev(2011)Phillips and Likhachev]{phillips2011sipp}
Mike Phillips and Maxim Likhachev.
\newblock Sipp: Safe interval path planning for dynamic environments.
\newblock In \emph{2011 IEEE international conference on robotics and automation}, pp.\  5628--5635. IEEE, 2011.

\bibitem[Pitanov et~al.(2023)Pitanov, Skrynnik, Andreychuk, Yakovlev, and Panov]{pitanov2023monte}
Yelisey Pitanov, Alexey Skrynnik, Anton Andreychuk, Konstantin Yakovlev, and Aleksandr Panov.
\newblock Monte-carlo tree search for multi-agent pathfinding: Preliminary results.
\newblock In \emph{HAIS}, 2023.

\bibitem[Pyke \& Stark(2021)Pyke and Stark]{pyke2021dynamic}
Lewis~M Pyke and Craig~R Stark.
\newblock Dynamic pathfinding for a swarm intelligence based uav control model using particle swarm optimisation.
\newblock \emph{Frontiers in Applied Mathematics and Statistics}, 7:\penalty0 744955, 2021.

\bibitem[Qiu(2020)]{qiu2020multi}
Hongda Qiu.
\newblock Multi-agent navigation based on deep reinforcement learning and traditional pathfinding algorithm.
\newblock \emph{arXiv:2012.09134}, 2020.

\bibitem[Rahmani \& Pelechano(2020)Rahmani and Pelechano]{rahmani2020multi}
Vahid Rahmani and Nuria Pelechano.
\newblock Multi-agent parallel hierarchical path finding in navigation meshes (ma-hna*).
\newblock \emph{Computers \& Graphics}, 86:\penalty0 1--14, 2020.

\bibitem[Raissi et~al.(2019)Raissi, Perdikaris, and Karniadakis]{raissi2019physics}
Maziar Raissi, Paris Perdikaris, and George~E Karniadakis.
\newblock Physics-informed neural networks: A deep learning framework for solving forward and inverse problems involving nonlinear partial differential equations.
\newblock \emph{Journal of Computational physics}, 378:\penalty0 686--707, 2019.

\bibitem[Rashid et~al.(2020)Rashid, Samvelyan, De~Witt, Farquhar, Foerster, and Whiteson]{rashid2020monotonic}
Tabish Rashid, Mikayel Samvelyan, Christian~Schroeder De~Witt, Gregory Farquhar, Jakob Foerster, and Shimon Whiteson.
\newblock Monotonic value function factorisation for deep multi-agent reinforcement learning.
\newblock \emph{Journal of Machine Learning Research}, 21\penalty0 (178):\penalty0 1--51, 2020.

\bibitem[Ren et~al.(2021)Ren, Sathiyanarayanan, Ewing, Senbaslar, and Ayanian]{ren2021mapfast}
Jingyao Ren, Vikraman Sathiyanarayanan, Eric Ewing, Baskin Senbaslar, and Nora Ayanian.
\newblock Mapfast: A deep algorithm selector for multi agent path finding using shortest path embeddings.
\newblock \emph{arXiv preprint arXiv:2102.12461}, 2021.

\bibitem[Ren et~al.(2024)Ren, Cai, and Wang]{ren2024multi}
Zhongqiang Ren, Yilin Cai, and Hesheng Wang.
\newblock Multi-agent teamwise cooperative path finding and traffic intersection coordination.
\newblock In \emph{IROS}, 2024.

\bibitem[Romera-Paredes et~al.(2024)Romera-Paredes, Barekatain, Novikov, Balog, Kumar, Dupont, Ruiz, Ellenberg, Wang, Fawzi, et~al.]{romera2024mathematical}
Bernardino Romera-Paredes, Mohammadamin Barekatain, Alexander Novikov, Matej Balog, M~Pawan Kumar, Emilien Dupont, Francisco~JR Ruiz, Jordan~S Ellenberg, Pengming Wang, Omar Fawzi, et~al.
\newblock Mathematical discoveries from program search with large language models.
\newblock \emph{Nature}, 625\penalty0 (7995):\penalty0 468--475, 2024.

\bibitem[Ryu et~al.(2019)Ryu, Liu, Wang, Chen, Wang, and Yin]{ryu2019plug}
Ernest Ryu, Jialin Liu, Sicheng Wang, Xiaohan Chen, Zhangyang Wang, and Wotao Yin.
\newblock Plug-and-play methods provably converge with properly trained denoisers.
\newblock In \emph{ICML}, 2019.

\bibitem[Salzman \& Stern(2020)Salzman and Stern]{salzman2020research}
Oren Salzman and Roni Stern.
\newblock Research challenges and opportunities in multi-agent path finding and multi-agent pickup and delivery problems.
\newblock In \emph{AAMAS}, 2020.

\bibitem[Sartoretti et~al.(2019)Sartoretti, Kerr, Shi, Wagner, Kumar, Koenig, and Choset]{sartoretti2019primal}
Guillaume Sartoretti, Justin Kerr, Yunfei Shi, Glenn Wagner, TK~Satish Kumar, Sven Koenig, and Howie Choset.
\newblock {PRIMAL}: Pathfinding via reinforcement and imitation multi-agent learning.
\newblock \emph{IEEE Robotics and Automation Letters}, 4\penalty0 (3):\penalty0 2378--2385, 2019.

\bibitem[Satorras et~al.(2021)Satorras, Hoogeboom, and Welling]{satorras2021n}
V{\i}ctor~Garcia Satorras, Emiel Hoogeboom, and Max Welling.
\newblock E (n) equivariant graph neural networks.
\newblock In \emph{ICML}, 2021.

\bibitem[Seo et~al.(2025)Seo, Kim, Shin, and Suh]{seo2025llmdr}
Seungbae Seo, Junghwan Kim, Minjeong Shin, and Bongwon Suh.
\newblock Llmdr: Llm-driven deadlock detection and resolution in multi-agent pathfinding.
\newblock \emph{arXiv preprint arXiv:2503.00717}, 2025.

\bibitem[Shaoul et~al.(2025)Shaoul, Mishani, Vats, Li, and Likhachev]{shaoul2025multirobot}
Yorai Shaoul, Itamar Mishani, Shivam Vats, Jiaoyang Li, and Maxim Likhachev.
\newblock Multi-robot motion planning with diffusion models.
\newblock In \emph{ICLR}, 2025.

\bibitem[Sharon et~al.(2011)Sharon, Stern, Goldenberg, and Felner]{sharon2011pruning}
Guni Sharon, Roni Stern, Meir Goldenberg, and Ariel Felner.
\newblock Pruning techniques for the increasing cost tree search for optimal multi-agent pathfinding.
\newblock In \emph{Proceedings of the International Symposium on Combinatorial Search}, volume~2, pp.\  150--157, 2011.

\bibitem[Sharon et~al.(2015)Sharon, Stern, Felner, and Sturtevant]{sharon2015conflict}
Guni Sharon, Roni Stern, Ariel Felner, and Nathan~R Sturtevant.
\newblock Conflict-based search for optimal multi-agent pathfinding.
\newblock \emph{Artificial intelligence}, 219:\penalty0 40--66, 2015.

\bibitem[Shrestha et~al.(2021)Shrestha, Bajracharya, and Kim]{shrestha20216g}
Rakesh Shrestha, Rojeena Bajracharya, and Shiho Kim.
\newblock 6g enabled unmanned aerial vehicle traffic management: A perspective.
\newblock \emph{IEEE Access}, 9:\penalty0 91119--91136, 2021.

\bibitem[Sigurdson et~al.(2019)Sigurdson, Bulitko, Koenig, Hernandez, and Yeoh]{sigurdson2019automatic}
Devon Sigurdson, Vadim Bulitko, Sven Koenig, Carlos Hernandez, and William Yeoh.
\newblock Automatic algorithm selection in multi-agent pathfinding.
\newblock \emph{arXiv preprint arXiv:1906.03992}, 2019.

\bibitem[Silver(2005)]{silver2005cooperative}
David Silver.
\newblock Cooperative pathfinding.
\newblock In \emph{Proceedings of the aaai conference on artificial intelligence and interactive digital entertainment}, volume~1, pp.\  117--122, 2005.

\bibitem[Skrynnik et~al.(2021)Skrynnik, Yakovleva, Davydov, Yakovlev, and Panov]{skrynnik2021hybrid}
Alexey Skrynnik, Alexandra Yakovleva, Vasilii Davydov, Konstantin Yakovlev, and Aleksandr~I Panov.
\newblock Hybrid policy learning for multi-agent pathfinding.
\newblock \emph{IEEE Access}, 9:\penalty0 126034--126047, 2021.

\bibitem[Skrynnik et~al.(2024{\natexlab{a}})Skrynnik, Andreychuk, Borzilov, Chernyavskiy, Yakovlev, and Panov]{skrynnik2024pogema}
Alexey Skrynnik, Anton Andreychuk, Anatolii Borzilov, Alexander Chernyavskiy, Konstantin Yakovlev, and Aleksandr Panov.
\newblock {POGEMA}: A benchmark platform for cooperative multi-agent navigation.
\newblock \emph{arXiv:2407.14931}, 2024{\natexlab{a}}.

\bibitem[Skrynnik et~al.(2024{\natexlab{b}})Skrynnik, Andreychuk, Nesterova, Yakovlev, and Panov]{skrynnik2024learn}
Alexey Skrynnik, Anton Andreychuk, Maria Nesterova, Konstantin Yakovlev, and Aleksandr Panov.
\newblock Learn to follow: Decentralized lifelong multi-agent pathfinding via planning and learning.
\newblock In \emph{AAAI}, 2024{\natexlab{b}}.

\bibitem[Skrynnik et~al.(2024{\natexlab{c}})Skrynnik, Andreychuk, Yakovlev, and Panov]{skrynnik2024decentralized}
Alexey Skrynnik, Anton Andreychuk, Konstantin Yakovlev, and Aleksandr Panov.
\newblock Decentralized monte carlo tree search for partially observable multi-agent pathfinding.
\newblock In \emph{AAAI}, 2024{\natexlab{c}}.

\bibitem[Song et~al.(2023)Song, Zhang, and Cheng]{song2023helsa}
Zhaoyi Song, Rongqing Zhang, and Xiang Cheng.
\newblock Helsa: Hierarchical reinforcement learning with spatiotemporal abstraction for large-scale multi-agent path finding.
\newblock In \emph{IROS}, 2023.

\bibitem[Standley(2010)]{standley2010finding}
Trevor Standley.
\newblock Finding optimal solutions to cooperative pathfinding problems.
\newblock In \emph{Proceedings of the AAAI conference on artificial intelligence}, volume~24, pp.\  173--178, 2010.

\bibitem[Stern(2019)]{stern2019mapfsurvey}
Roni Stern.
\newblock Multi-agent path finding--an overview.
\newblock \emph{Artificial Intelligence: 5th RAAI Summer School, Tutorial Lectures}, pp.\  96--115, 2019.

\bibitem[Stern et~al.(2019)Stern, Sturtevant, Felner, Koenig, Ma, Walker, Li, Atzmon, Cohen, Kumar, et~al.]{stern2019multi}
Roni Stern, Nathan Sturtevant, Ariel Felner, Sven Koenig, Hang Ma, Thayne Walker, Jiaoyang Li, Dor Atzmon, Liron Cohen, TK~Kumar, et~al.
\newblock Multi-agent pathfinding: Definitions, variants, and benchmarks.
\newblock In \emph{SoCS}, 2019.

\bibitem[Sun et~al.(2018)Sun, Chen, Shi, Hong, Fu, and Sidiropoulos]{sun2018learning}
Haoran Sun, Xiangyi Chen, Qingjiang Shi, Mingyi Hong, Xiao Fu, and Nicholas~D Sidiropoulos.
\newblock Learning to optimize: Training deep neural networks for interference management.
\newblock \emph{IEEE Transactions on Signal Processing}, 66\penalty0 (20):\penalty0 5438--5453, 2018.

\bibitem[Sun et~al.(2019)Sun, Khedr, and Shoukry]{sun2019formal}
Xiaowu Sun, Haitham Khedr, and Yasser Shoukry.
\newblock Formal verification of neural network controlled autonomous systems.
\newblock In \emph{HSCC}, 2019.

\bibitem[Sun et~al.(2021)Sun, Wu, Xu, Wohlberg, and Kamilov]{sun2021scalable}
Yu~Sun, Zihui Wu, Xiaojian Xu, Brendt Wohlberg, and Ulugbek~S Kamilov.
\newblock Scalable plug-and-play admm with convergence guarantees.
\newblock \emph{IEEE Transactions on Computational Imaging}, 7:\penalty0 849--863, 2021.

\bibitem[Sunehag et~al.(2018)Sunehag, Lever, Gruslys, Czarnecki, Zambaldi, Jaderberg, Lanctot, Sonnerat, Leibo, Tuyls, et~al.]{sunehag2018value}
Peter Sunehag, Guy Lever, Audrunas Gruslys, Wojciech~Marian Czarnecki, Vinicius Zambaldi, Max Jaderberg, Marc Lanctot, Nicolas Sonnerat, Joel~Z Leibo, Karl Tuyls, et~al.
\newblock Value-decomposition networks for cooperative multi-agent learning based on team reward.
\newblock In \emph{AAMAS}, 2018.

\bibitem[Surynek(2019{\natexlab{a}})]{surynek2019conflict}
Pavel Surynek.
\newblock Conflict handling framework in generalized multi-agent path finding: Advantages and shortcomings of satisfiability modulo approach.
\newblock In \emph{ICAART}, 2019{\natexlab{a}}.

\bibitem[Surynek(2019{\natexlab{b}})]{surynek2019multi}
Pavel Surynek.
\newblock Multi-agent path finding with continuous time viewed through satisfiability modulo theories (smt).
\newblock \emph{arXiv preprint arXiv:1903.09820}, 2019{\natexlab{b}}.

\bibitem[Surynek(2019{\natexlab{c}})]{surynek2019multi2}
Pavel Surynek.
\newblock Multi-agent path finding with continuous time and geometric agents viewed through satisfiability modulo theories ({SMT}).
\newblock In \emph{SoCS}, 2019{\natexlab{c}}.

\bibitem[Surynek(2019{\natexlab{d}})]{surynek2019tour}
Pavel Surynek.
\newblock On the tour towards {DPLL(MAPF)} and beyond.
\newblock \emph{arXiv:1907.07631}, 2019{\natexlab{d}}.

\bibitem[Surynek(2020{\natexlab{a}})]{surynek2020continuous}
Pavel Surynek.
\newblock Continuous multi-agent path finding via satisfiability modulo theories ({SMT}).
\newblock In \emph{ICAART}, 2020{\natexlab{a}}.

\bibitem[Surynek(2020{\natexlab{b}})]{surynek2020multi}
Pavel Surynek.
\newblock Multi-agent path finding modulo theory with continuous movements and the sum of costs objective.
\newblock In \emph{KI}, 2020{\natexlab{b}}.

\bibitem[Surynek(2021{\natexlab{a}})]{surynek2021conceptual}
Pavel Surynek.
\newblock Conceptual comparison of compilation-based solvers for multi-agent path finding: {MIP} vs. {SAT}.
\newblock In \emph{SoCS}, 2021{\natexlab{a}}.

\bibitem[Surynek(2021{\natexlab{b}})]{surynek2021sum}
Pavel Surynek.
\newblock Sum of costs optimal multi-agent path finding with continuous time via satisfiability modulo theories.
\newblock In \emph{SoCS}, 2021{\natexlab{b}}.

\bibitem[Surynek(2022)]{surynek2022problem}
Pavel Surynek.
\newblock Problem compilation for multi-agent path finding: A survey.
\newblock In \emph{IJCAI}, 2022.

\bibitem[Surynek et~al.(2016)Surynek, Felner, Stern, and Boyarski]{surynek2016efficient}
Pavel Surynek, Ariel Felner, Roni Stern, and Eli Boyarski.
\newblock Efficient sat approach to multi-agent path finding under the sum of costs objective.
\newblock In \emph{ECAI}. IOS Press, 2016.

\bibitem[Surynek et~al.(2017)Surynek, {\v{S}}vancara, Felner, and Boyarski]{surynek2017integration}
Pavel Surynek, Ji{\v{r}}{\'\i} {\v{S}}vancara, Ariel Felner, and Eli Boyarski.
\newblock Integration of independence detection into sat-based optimal multi-agent path finding-a novel sat-based optimal {MAPF} solver.
\newblock In \emph{ICAART}, 2017.

\bibitem[Surynek et~al.(2018)Surynek, Felner, Stern, and Boyarski]{surynek2018sub}
Pavel Surynek, Ariel Felner, Roni Stern, and Eli Boyarski.
\newblock Sub-optimal sat-based approach to multi-agent path-finding problem.
\newblock In \emph{SoCS}, 2018.

\bibitem[Surynek et~al.(2021)Surynek, Li, Zhang, Satish~Kumar, and Koenig]{surynek2021mutex}
Pavel Surynek, Jiaoyang Li, Han Zhang, TK~Satish~Kumar, and Sven Koenig.
\newblock Mutex propagation for {SAT}-based multi-agent path finding.
\newblock In \emph{PRIMA}, 2021.

\bibitem[Surynek et~al.(2022)Surynek, Stern, Boyarski, and Felner]{surynek2022migrating}
Pavel Surynek, Roni Stern, Eli Boyarski, and Ariel Felner.
\newblock Migrating techniques from search-based multi-agent path finding solvers to sat-based approach.
\newblock \emph{Journal of Artificial Intelligence Research}, 73:\penalty0 553--618, 2022.

\bibitem[Tan et~al.(2024)Tan, Luo, Li, and Ma]{tan2024benchmarking}
Jiaqi Tan, Yudong Luo, Jiaoyang Li, and Hang Ma.
\newblock Benchmarking large neighborhood search for multi-agent path finding.
\newblock \emph{arXiv:2407.09451}, 2024.

\bibitem[Tang et~al.(2024{\natexlab{a}})Tang, Berto, and Park]{tang2024ensembling}
Huijie Tang, Federico Berto, and Jinkyoo Park.
\newblock Ensembling prioritized hybrid policies for multi-agent pathfinding.
\newblock \emph{arXiv:2403.07559}, 2024{\natexlab{a}}.

\bibitem[Tang et~al.(2024{\natexlab{b}})Tang, Mao, and Ma]{tang2024large}
Jingtao Tang, Zining Mao, and Hang Ma.
\newblock Large-scale multi-robot coverage path planning on grids with path deconfliction.
\newblock \emph{arXiv:2411.01707}, 2024{\natexlab{b}}.

\bibitem[Tang et~al.(2024{\natexlab{c}})Tang, Koenig, and Li]{tang2024ita}
Yimin Tang, Sven Koenig, and Jiaoyang Li.
\newblock Ita-ecbs: A bounded-suboptimal algorithm for combined target-assignment and path-finding problem.
\newblock In \emph{SoCS}, 2024{\natexlab{c}}.

\bibitem[Tao et~al.(2024)Tao, Kang, Dong, Zhang, Ye, Chen, and Zheng]{tao2024poaql}
Lesong Tao, Miao Kang, Jinpeng Dong, Songyi Zhang, Ke~Ye, Shitao Chen, and Nanning Zheng.
\newblock {POAQL}: A partially observable altruistic q-learning method for cooperative multi-agent reinforcement learning.
\newblock In \emph{ICRA}, 2024.

\bibitem[Teng et~al.(2017)Teng, Lau, and Kumar]{teng2017coordinating}
Teck-Hou Teng, Hoong~Chuin Lau, and Akshat Kumar.
\newblock Coordinating vessel traffic to improve safety and efficiency.
\newblock In \emph{AAMAS}, 2017.

\bibitem[Tjiharjadi et~al.(2022)Tjiharjadi, Razali, and Sulaiman]{tjiharjadi2022systematic}
Semuil Tjiharjadi, Sazalinsyah Razali, and Hamzah~Asyrani Sulaiman.
\newblock A systematic literature review of multi-agent pathfinding for maze research.
\newblock \emph{Journal of Advances in Information Technology}, 13\penalty0 (4), 2022.

\bibitem[van Knippenberg et~al.(2021)van Knippenberg, Holenderski, and Menkovski]{van2021time}
Marijn van Knippenberg, Mike Holenderski, and Vlado Menkovski.
\newblock Time-constrained multi-agent path finding in non-lattice graphs with deep reinforcement learning.
\newblock In \emph{ACML}, 2021.

\bibitem[Varambally et~al.(2022)Varambally, Li, and Koenig]{varambally2022mapf}
Sumanth Varambally, Jiaoyang Li, and Sven Koenig.
\newblock Which mapf model works best for automated warehousing?
\newblock In \emph{SoCS}, 2022.

\bibitem[von~der Burg et~al.(2024)von~der Burg, Kamphof, Soomers, and Sharpanskykh]{von2024towards}
Malte von~der Burg, Jorick Kamphof, Joost Soomers, and Alexei Sharpanskykh.
\newblock Towards autonomous airport surface movement operations using hierarchical multi-agent planning.
\newblock \emph{Available at SSRN 4916874}, 2024.

\bibitem[Wagner \& Choset(2011)Wagner and Choset]{wagner2011m}
Glenn Wagner and Howie Choset.
\newblock M*: A complete multirobot path planning algorithm with performance bounds.
\newblock In \emph{2011 IEEE/RSJ international conference on intelligent robots and systems}, pp.\  3260--3267. IEEE, 2011.

\bibitem[Wang et~al.(2020)Wang, Liu, Li, and Prorok]{wang2020mobile}
Binyu Wang, Zhe Liu, Qingbiao Li, and Amanda Prorok.
\newblock Mobile robot path planning in dynamic environments through globally guided reinforcement learning.
\newblock \emph{IEEE Robotics and Automation Letters}, 5\penalty0 (4):\penalty0 6932--6939, 2020.

\bibitem[Wang et~al.(2024{\natexlab{a}})Wang, Diller, and Han]{10903304}
Chenyang Wang, Jonathan Diller, and Qi~Han.
\newblock Llm for generating simulation inputs to evaluate path planning algorithms.
\newblock In \emph{ICMLA}, 2024{\natexlab{a}}.

\bibitem[Wang et~al.(2019)Wang, Li, Ma, Koenig, and Kumar]{wang2019new}
Jiangxing Wang, Jiaoyang Li, Hang Ma, Sven Koenig, and S~Kumar.
\newblock A new constraint satisfaction perspective on multi-agent path finding: Preliminary results.
\newblock In \emph{AAMAS}, 2019.

\bibitem[Wang et~al.(2024{\natexlab{b}})Wang, Veerapaneni, Wu, Li, and Likhachev]{wang2024mapf}
Qian Wang, Rishi Veerapaneni, Yu~Wu, Jiaoyang Li, and Maxim Likhachev.
\newblock Mapf in 3d warehouses: Dataset and analysis.
\newblock In \emph{ICAPS}, 2024{\natexlab{b}}.

\bibitem[Wang et~al.(2024{\natexlab{c}})Wang, Zhang, Jia, Pan, Diao, Pi, and Zhang]{wang2024theoremllama}
Ruida Wang, Jipeng Zhang, Yizhen Jia, Rui Pan, Shizhe Diao, Renjie Pi, and Tong Zhang.
\newblock Theoremllama: Transforming general-purpose llms into lean4 experts.
\newblock In \emph{EMNLP}, 2024{\natexlab{c}}.

\bibitem[Wang et~al.(2023)Wang, Bulitko, Huang, Koenig, and Stern]{wang2023synthesizing}
Shuwei Wang, Vadim Bulitko, Taoan Huang, Sven Koenig, and Roni Stern.
\newblock Synthesizing priority planning formulae for multi-agent pathfinding.
\newblock In \emph{AIIDE}, 2023.

\bibitem[Wang et~al.(2024{\natexlab{d}})Wang, Duhan, Li, and Sartoretti]{wang2024lns2+}
Yutong Wang, Tanishq Duhan, Jiaoyang Li, and Guillaume Sartoretti.
\newblock {LNS2+ RL}: Combining multi-agent reinforcement learning with large neighborhood search in multi-agent path finding.
\newblock \emph{arXiv:2405.17794}, 2024{\natexlab{d}}.

\bibitem[Wu et~al.(2023)Wu, Yan, Hasan, and Jamaluddin]{wu2023review}
Mengdie Wu, Wenyao Yan, Haslin Hasan, and Robiatul~A’Dawiah Jamaluddin.
\newblock A review of multi-agent path finding algorithms.
\newblock In \emph{ISCTech}, 2023.

\bibitem[Xie et~al.(2024{\natexlab{a}})Xie, Zhang, Ouyang, Yang, Dong, Shi, and Jin]{xie2024improved}
Jing Xie, Yongjun Zhang, Qianying Ouyang, Huanhuan Yang, Fang Dong, Dianxi Shi, and Songchang Jin.
\newblock Improved communication and collision-avoidance in dynamic multi-agent path finding.
\newblock In \emph{IJCNN}, 2024{\natexlab{a}}.

\bibitem[Xie et~al.(2024{\natexlab{b}})Xie, Zhang, Yang, Ouyang, Dong, Guo, Jin, and Shi]{xie2024crowd}
Jing Xie, Yongjun Zhang, Huanhuan Yang, Qianying Ouyang, Fang Dong, Xinyu Guo, Songchang Jin, and Dianxi Shi.
\newblock Crowd perception communication-based multi-agent path finding with imitation learning.
\newblock \emph{IEEE Robotics and Automation Letters}, 9\penalty0 (10):\penalty0 8929--8936, 2024{\natexlab{b}}.

\bibitem[Xie et~al.(2022)Xie, Raghunathan, Liang, and Ma]{xie2022an}
Sang~Michael Xie, Aditi Raghunathan, Percy Liang, and Tengyu Ma.
\newblock An explanation of in-context learning as implicit bayesian inference.
\newblock In \emph{ICLR}, 2022.

\bibitem[Yakovlev et~al.(2022)Yakovlev, Andreychuk, Skrynnik, and Panov]{yakovlev2022planning}
KS~Yakovlev, AA~Andreychuk, Aleksei~Aleksandrovich Skrynnik, and Aleksandr~Igorevich Panov.
\newblock Planning and learning in multi-agent path finding.
\newblock \emph{Doklady Mathematics}, 106\penalty0 (Suppl 1):\penalty0 S79--S84, 2022.

\bibitem[Yan \& Wu(2024)Yan and Wu]{yan2024neural}
Zhongxia Yan and Cathy Wu.
\newblock Neural neighborhood search for multi-agent path finding.
\newblock In \emph{ICLR}, 2024.

\bibitem[Yang et~al.(2020)Yang, Li, Farajtabar, Sunehag, Hughes, and Zha]{yang2020learning}
Jiachen Yang, Ang Li, Mehrdad Farajtabar, Peter Sunehag, Edward Hughes, and Hongyuan Zha.
\newblock Learning to incentivize other learning agents.
\newblock In \emph{NeurIPS}, 2020.

\bibitem[Yang et~al.(2023)Yang, Li, Qian, Quan, Miao, Liu, Hu, and Memetimin]{yang2023path}
Liwei Yang, Ping Li, Song Qian, He~Quan, Jinchao Miao, Mengqi Liu, Yanpei Hu, and Erexidin Memetimin.
\newblock Path planning technique for mobile robots: A review.
\newblock \emph{Machines}, 11\penalty0 (10):\penalty0 980, 2023.

\bibitem[Yang et~al.(2018)Yang, Luo, Li, Zhou, Zhang, and Wang]{yang2018mean}
Yaodong Yang, Rui Luo, Minne Li, Ming Zhou, Weinan Zhang, and Jun Wang.
\newblock Mean field multi-agent reinforcement learning.
\newblock In \emph{ICML}, 2018.

\bibitem[Yang et~al.(2024)Yang, Fan, He, Wang, Huang, and Sartoretti]{yang2024attention}
Yibin Yang, Mingfeng Fan, Chengyang He, Jianqiang Wang, Heye Huang, and Guillaume Sartoretti.
\newblock Attention-based priority learning for limited time multi-agent path finding.
\newblock In \emph{AAMAS}, 2024.

\bibitem[Yao et~al.(2024)Yao, Shen, Xu, Chen, and Nie]{yao2024accelerating}
Zhenyu Yao, Han Shen, Yan Xu, Yu~Chen, and Jing Nie.
\newblock Accelerating multi-agent path finding based on roadmap construction in non-grid environment.
\newblock In \emph{ASCC}, 2024.

\bibitem[Ye et~al.(2022)Ye, Li, Guo, Gao, and Fu]{ye2022multi}
Zhaohui Ye, Yanjie Li, Ronghao Guo, Jianqi Gao, and Wen Fu.
\newblock Multi-agent pathfinding with communication reinforcement learning and deadlock detection.
\newblock In \emph{ICIRA}, 2022.

\bibitem[Yin et~al.(2024)Yin, Wang, Yan, Xiang, Cai, and Wei]{yin2024deep}
Heqing Yin, Chang Wang, Chao Yan, Xiaojia Xiang, Boliang Cai, and Changyun Wei.
\newblock Deep reinforcement learning with multi-critic {TD3} for decentralized multi-robot path planning.
\newblock \emph{IEEE Transactions on Cognitive and Developmental Systems}, 16\penalty0 (4):\penalty0 1233--1247, 2024.

\bibitem[Yu et~al.(2023)Yu, Li, Gao, and Prorok]{yu2023accelerating}
Chenning Yu, Qingbiao Li, Sicun Gao, and Amanda Prorok.
\newblock Accelerating multi-agent planning using graph transformers with bounded suboptimality.
\newblock In \emph{ICRA}, 2023.

\bibitem[Zapata et~al.(2024)Zapata, Godoy, and As{\'\i}n-Ach{\'a}]{zapata2024anytime}
Angelo Zapata, Julio Godoy, and Roberto As{\'\i}n-Ach{\'a}.
\newblock Anytime automatic algorithm selection for the multi-agent path finding problem.
\newblock \emph{IEEE Access}, 12:\penalty0 62177--62188, 2024.

\bibitem[Zeng et~al.(2024)Zeng, Yang, Dong, Du, Zheng, Xu, and Li]{zeng2024perceive}
Qingbin Zeng, Qinglong Yang, Shunan Dong, Heming Du, Liang Zheng, Fengli Xu, and Yong Li.
\newblock Perceive, reflect, and plan: Designing llm agent for goal-directed city navigation without instructions.
\newblock \emph{arXiv preprint arXiv:2408.04168}, 2024.

\bibitem[Zhang et~al.(2021)Zhang, Yao, Liu, Li, Terr, Chan, Kumar, and Koenig]{zhang2021hierarchical}
Han Zhang, Mingze Yao, Ziang Liu, Jiaoyang Li, Lucas Terr, Shao-Hung Chan, TK~Satish Kumar, and Sven Koenig.
\newblock A hierarchical approach to multi-agent path finding.
\newblock In \emph{SoCS}, 2021.

\bibitem[Zhang et~al.(2022{\natexlab{a}})Zhang, Li, Surynek, Kumar, and Koenig]{zhang2022multi}
Han Zhang, Jiaoyang Li, Pavel Surynek, TK~Satish Kumar, and Sven Koenig.
\newblock Multi-agent path finding with mutex propagation.
\newblock \emph{Artificial Intelligence}, 311:\penalty0 103766, 2022{\natexlab{a}}.

\bibitem[Zhang et~al.(2022{\natexlab{b}})Zhang, Li, Huang, Koenig, and Dilkina]{zhang2022learning}
Shuyang Zhang, Jiaoyang Li, Taoan Huang, Sven Koenig, and Bistra Dilkina.
\newblock Learning a priority ordering for prioritized planning in multi-agent path finding.
\newblock In \emph{SoCS}, 2022{\natexlab{b}}.

\bibitem[Zhang et~al.(2025)Zhang, Lyu, Li, Duan, Zhan, Liu, Cheng, and Li]{zhang2025understanding}
Xiangteng Zhang, Yao Lyu, Shengbo~Eben Li, Jingliang Duan, Guojian Zhan, Chang Liu, Bo~Cheng, and Keqiang Li.
\newblock Understanding connection between pmp and hjb equations from the perspective of hamilton dynamics.
\newblock \emph{IEEE Transactions on Artificial Intelligence}, 2025.

\bibitem[Zhang et~al.(2020)Zhang, Qian, Yao, Hu, and Xu]{zhang2020learning}
Yi~Zhang, Yu~Qian, Yichen Yao, Haoyuan Hu, and Yinghui Xu.
\newblock Learning to cooperate: Application of deep reinforcement learning for online agv path finding.
\newblock In \emph{AAMAS}, 2020.

\bibitem[Zhao et~al.(2023)Zhao, Zhuang, Huang, and Liu]{zhao2023curriculum}
Cheng Zhao, Liansheng Zhuang, Yihong Huang, and Haonan Liu.
\newblock Curriculum learning based multi-agent path finding for complex environments.
\newblock In \emph{IJCNN}, 2023.

\bibitem[Zhiyao \& Sartoretti(2020)Zhiyao and Sartoretti]{zhiyao2020deep}
Luo Zhiyao and Guillaume Sartoretti.
\newblock Deep reinforcement learning based multiagent pathfinding.
\newblock \emph{Technical Report}, 2020.

\bibitem[Zhou et~al.(2024)Zhou, Pang, and Li]{zhou2024dhaa}
Dongming Zhou, Zhengbin Pang, and Wei Li.
\newblock {DHAA}: Distributed heuristic action aware multi-agent path finding in high density scene.
\newblock \emph{Multimedia Tools and Applications}, pp.\  1--19, 2024.

\bibitem[Zhou(2023)]{zhou2023research}
Ruining Zhou.
\newblock Research of the methods on multi-agent path finding.
\newblock \emph{Highlights in Science, Engineering and Technology}, 39:\penalty0 1131--1140, 2023.

\bibitem[Zhuang et~al.(2023)Zhuang, Abnar, Gu, Schwing, Susskind, and Bautista]{zhuang2023diffusion}
Peiye Zhuang, Samira Abnar, Jiatao Gu, Alex Schwing, Joshua~M Susskind, and Miguel~Angel Bautista.
\newblock Diffusion probabilistic fields.
\newblock In \emph{ICLR}, 2023.

\bibitem[Zoph \& Le(2017)Zoph and Le]{zoph2017neural}
Barret Zoph and Quoc Le.
\newblock Neural architecture search with reinforcement learning.
\newblock In \emph{ICLR}, 2017.

\end{thebibliography}
\bibliographystyle{colm2025_conference}


\end{document}